\documentclass[nohyperref]{article}

\usepackage{microtype}
\usepackage{graphicx}
\usepackage{subfigure}
\usepackage{booktabs} 

\usepackage{hyperref}

\usepackage[accepted]{icml2022}

\usepackage{amsmath}
\usepackage{amssymb}
\usepackage{mathtools}
\usepackage{amsthm}

\usepackage[capitalize,noabbrev]{cleveref}

\theoremstyle{plain}

\theoremstyle{definition}

\theoremstyle{remark}

\usepackage[textsize=tiny]{todonotes}
\usepackage{multirow}
\usepackage[T1]{fontenc}

\icmltitlerunning{Cliff Diving: Exploring Reward Surfaces in Reinforcement Learning Environments}

\begin{document}

\twocolumn[
\icmltitle{Cliff Diving: Exploring Reward Surfaces in Reinforcement Learning Environments}



\icmlsetsymbol{equal}{*}

\begin{icmlauthorlist}
\icmlauthor{Ryan Sullivan}{equal,swarmlabs,umd}
\icmlauthor{J K Terry}{equal,swarmlabs,umd}
\icmlauthor{Benjamin Black}{equal,swarmlabs,umd}
\icmlauthor{John P.~Dickerson}{umd}
\end{icmlauthorlist}

\icmlaffiliation{umd}{Department of Computer Science, University of Maryland, College Park}
\icmlaffiliation{swarmlabs}{Swarm Labs}

\icmlcorrespondingauthor{Ryan Sullivan}{rsulli@umd.edu}

\icmlkeywords{Machine Learning, ICML}

\vskip 0.3in
]



\printAffiliationsAndNotice{\icmlEqualContribution} 

\begin{abstract}
Visualizing optimization landscapes has led to many fundamental insights in numeric optimization, and novel improvements to optimization techniques. However, visualizations of the objective that reinforcement learning optimizes (the ``reward surface'') have only ever been generated for a small number of narrow contexts. This work presents reward surfaces and related visualizations of 27 of the most widely used reinforcement learning environments in Gym for the first time. We also explore reward surfaces in the policy gradient direction and show for the first time that many popular reinforcement learning environments have frequent ``cliffs'' (sudden large drops in expected return). We demonstrate that A2C often ``dives off'' these cliffs into low reward regions of the parameter space while PPO avoids them, confirming a popular intuition for PPO's improved performance over previous methods. We additionally introduce a highly extensible library that allows researchers to easily generate these visualizations in the future. Our findings provide new intuition to explain the successes and failures of modern RL methods, and our visualizations concretely characterize several failure modes of reinforcement learning agents in novel ways.
\end{abstract}

\section{Introduction}

Reinforcement learning typically attempts to optimize the expected discounted return of an agent's policy. Policy gradient methods represent the policy with a neural network, and learn this policy by approximating the gradient of the reinforcement learning objective over policy network parameters. This means that the benefits and challenges of deep learning apply to policy gradient methods. However, unlike most deep learning tasks, reinforcement learning is notoriously unstable, and agents often experience sharp drops in reward during training. Studying the reward surface and how RL algorithms interact with it is critical to understanding the successes and failures of deep reinforcement learning. 

\begin{figure}[t]
\centering
\begin{tabular}{c}
    \includegraphics[scale=0.52]{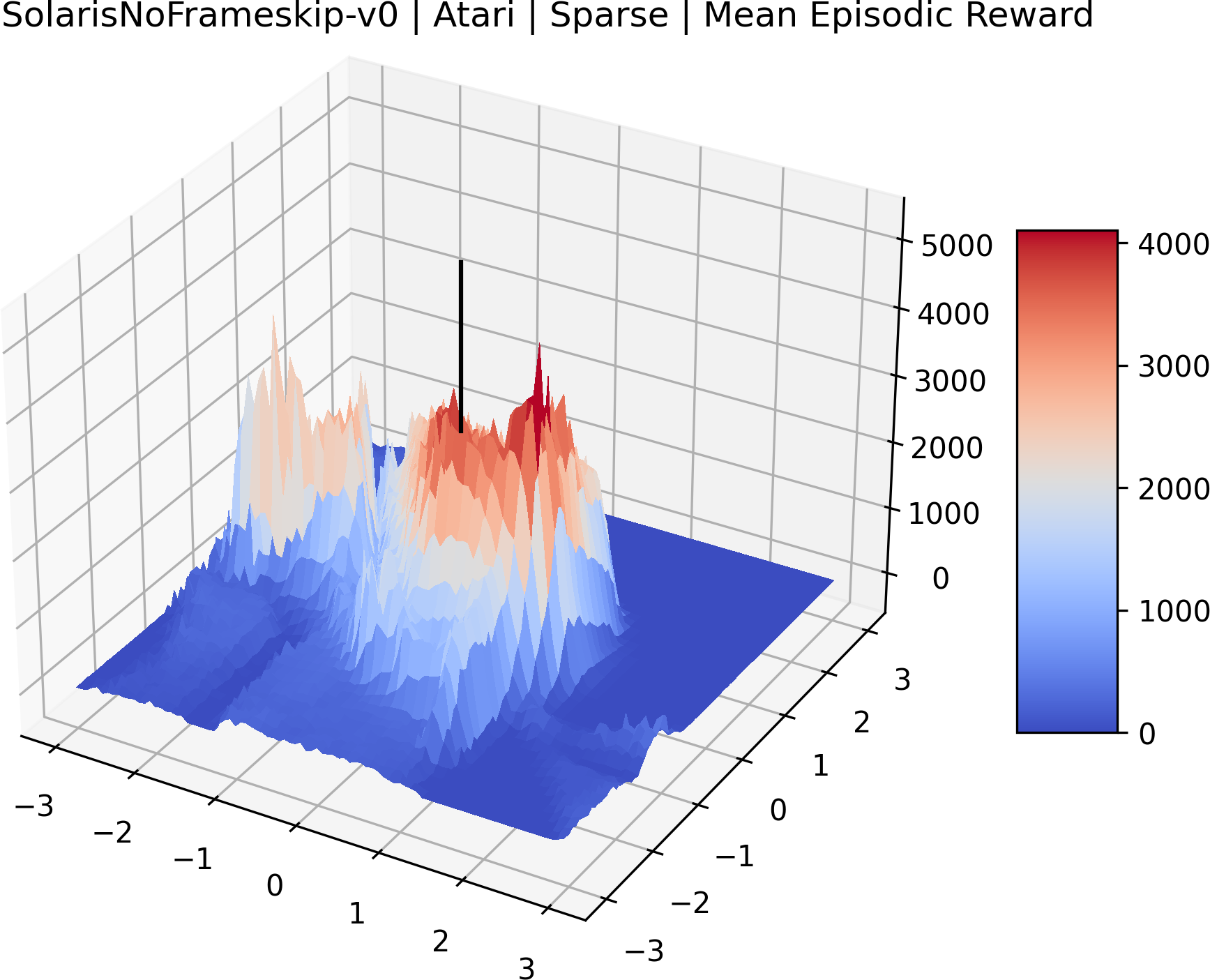} \\
\end{tabular}
\caption{Reward surface for the Solaris Atari environment.}
\label{fig:sample_solaris}
\end{figure}

A ``reward surface'' is the high dimensional surface of the reinforcement learning objective (the cumulative expected reward when following a given policy in an environment) over the policy network parameter space. Reward surfaces have been used to study specific questions in limited contexts \citep{Ilyas2020A} but have never been generated for a wide variety of environments. Loss landscapes have been used in computer vision to understand the effect that residual connections have on computer vision tasks \citep{li2017visualizing}, and we expect that visualizations of reward surfaces could be similarly valuable for understanding techniques in reinforcement learning. As a first step toward this goal, we produce visualizations of the reward surfaces for 27 of the most popular reinforcement learning environments, and identify new patterns in these surfaces for the first time. We see common traits of environments that are generally perceived to be "easy" for reinforcement learning to solve, and visual explanations of failure modes in sparse reward environments.

We additionally conduct a series of novel visualizations of the reward surface, finding evidence of steep ``cliffs'' in reward surfaces plotted in the policy gradient direction of numerous environments. These are directions where the returns are constant or increase for a short distance, then drop sharply. Reinforcement learning notoriously suffers from instability, and these cliffs may explain the sudden performance drops that agents experience during training. Our plots offer conclusive visual evidence that these cliff exist, as well as methods to study them further. We show that the specific cliffs present in our visualizations impact learning performance in some situations, and by comparing the performance of PPO \citep{schulman2017proximal} and A2C \citep{mnih2016asynchronous} on these cliffs, we provide an explanation for PPO's improved efficacy over previous methods.

Finally, to accelerate future research on reward surfaces and the impact of cliffs in reinforcement learning, we release the code for this paper as a comprehensive, modular, and easy to use library for researchers to plot reward surfaces and other visualizations used in this paper.

\section{Background and Related Work}

\subsection{Loss Landscapes}
\citet{li2017visualizing} developed a filter-normalization, a technique for plotting 3d visualizations of loss landscapes, the surface generated by a loss function over neural network parameters. They were able to demonstrate that the sharpness of loss landscapes plotted using filter-normalized random directions correlated with neural network generalization error. They create their plots by choosing perturbations $d$ of trained parameters that give an informative view of the neural network's local behavior. However, uniform random perturbations are known to be misleading in neural network analysis, because neural networks with ReLU activations have scale invariant weights \citep{li2017visualizing}. To mitigate this problem, they propose filter-normalized random directions. They represent the neural network as a vector $\theta$ indexed by layer $i$ and filter (not filter weight) $j$.\footnote{Note that this method also works for fully connected layers, which are equivalent to a convolutional layer with a 1x1 output feature map.} Then, they sample a random Gaussian direction $d$, and scale each filter of this random direction to match the magnitude of the neural network parameters in the corresponding filter, by applying the following formula. $$ d_{i,j} = \frac{d_{i,j}}{\| d_{i,j}\|} \| \theta_{i,j} \| $$ \citet{li2017visualizing} used this method to demonstrate that skip connections can limit the increase in non-convexity as network depth increases. Their work was originally applied to image classification networks, but we adapt these techniques to visualize reward surfaces for policy networks.

\subsection{Reinforcement Learning}
Deep reinforcement learning aims to optimize a policy $\pi$ to maximize the expected return over neural network parameters $\theta$. This objective can be written as $J(\pi_\theta)=E_{\tau\sim\pi_{\theta}}R(\tau)$ where $R(\tau)=\sum_{t=0}^{n}\gamma^{t}r_{t}$. Here $\tau$ is a trajectory, $r_t$ is the reward at time step $t$, and $\gamma$ is the discount factor and we sum the rewards across the entire episode of $n$ time steps. \citet{10.5555/3398761.3398871} showed that the discounted policy gradient is not the gradient over the surface of average discounted rewards, and is in fact not the gradient of any surface. To avoid this issue and make interpretation easier, we plot the undiscounted reward surface where $\gamma = 1$.

\subsection{Policy Gradient Methods}
Policy gradient methods estimate the gradient of the policy network and use gradient ascent to increase the probability of selecting actions that lead to high rewards. The gradient of the objective $J$ is $\nabla_\theta J(\pi_\theta) = E_{\tau \sim \pi_\theta} \Big[ \sum^{T}_{t=0} \nabla_\theta \log \pi_\theta(a_t | s_t) A^{\pi_\theta}(s_t | a_t) \Big]$ where $A^{\pi_\theta}(s_t|a_t)$ is the advantage function for the current policy $\pi_\theta$. 

\subsection{Reward surfaces}
The ``reward surface'' is the reinforcement learning objective function $J(\pi_\theta)$. Reward surfaces were first visualized by \citet{Ilyas2020A} to characterize problems with policy gradient estimates. The authors plotted a policy gradient estimate vs a uniform random direction, showing via visually striking examples that low sample estimates of the policy gradient rarely guide the policy in a better direction than a random direction. Note that this work did not make use of filter-normalized directions.

\citet{bekci2020visualizing} then used loss landscapes from \citet{li2017visualizing} to study Soft Actor-Critic agents \citep{haarnoja2018soft} trained on an inventory optimization task. They visualize the impact of policy stochasticity and action smoothing on the curvature of the loss landscapes for 4 MuJoCo environments.

Later, \citet{ota2021training} used loss landscapes from \citet{li2017visualizing} directly to compare shallow neural networks for reinforcement learning to deeper neural networks, showing that deep neural networks perform poorly because their loss landscape has more complex curvature. They used this visual insight to develop methods that can train deeper networks for reinforcement learning tasks. This work plots the loss function of SAC agents which includes an entropy-regularization term \citep{haarnoja2018soft}, while we directly plot the reinforcement learning objective.

This paper is different from each of these previous works in the breadth of environments explored, and in that we are the first to use filter-normalized directions from \citet{li2017visualizing} to visualize reward surfaces rather than loss landscapes. We additionally introduce novel experiments and results inspired by the findings in our initial reward surface plots.

\subsection{Proximal Policy Optimization}
Proximal Policy Optimization (PPO) \citep{schulman2017proximal} is a derivative method of Trust Region Policy Optimization (TRPO) \citep{schulman2015trust} intended to be easier to implement and more hyperparameter invariant. Over the past 4 years, PPO has become a ``default'' method for many deep RL practitioners and has been used in numerous high profile works \citep{berner2019dota, Mirhoseini2021AGP}. TRPO and PPO claim to offer enhanced empirical performance over previous methods by preventing dramatic changes in the policy via trust regions and ratio clipping respectively. Ratio clipping is conceptually a useful heuristic, however we are not aware of any work demonstrating why it results in empirically good performance across so many domains.

Instability during training is common in deep reinforcement learning, and agents often experience large drops in reward that can take long to recover from. One intuition for the utility of PPO's gradient and ratio clipping that has been idly discussed in the community is that they prevent agents from taking gradient steps that result in this collapsing performance \citep{hui_2018}. More precisely, the intuition is that by preventing large changes to the policy, it avoids gradient steps that move the policy into regions of the parameter space with poor rewards and uninformative gradients. To the best of our knowledge, this intuition was first described in a widely circulated medium article \citet{hui_2018} released roughly 1 year after the original PPO paper. We are aware of no prior work in the academic literature which directly support this intuition, and the TRPO and PPO papers do not directly allude to it. We discover evidence of these cliffs in plots of the policy gradient direction, and perform experiments to confirm that they negatively impact learning. As a result of its empirically good performance PPO has become the de-facto policy gradient method, so we chose it as the agent in our reward surface experiments.

\section{Environment Selection}
In exploring these reward surfaces, we sought to cover many widely used benchmark environments. We generated plots for all ``Classic Control`` and ``MuJoCo'' environments in Gym \citep{brockman2016openai} and for many popular Atari environments from the Arcade Learning Environment \citep{bellemare2013arcade}. Video games with graphical observations and discrete actions are a common benchmark for the field, and Atari games are the most popular games for this purpose. The MuJoCo environments are high quality physics simulations used for prominent robotics work with continuous actions and vector observations. The Classic Control games represent some of the early toy environments in reinforcement learning, and continue to be used as a standard set of "easy" environments. We generate surfaces for all classic control and MuJoCo environments, but we only generate surfaces for a representative selection of 12 Atari environments instead of all 59 Atari environments in Gym for the sake of brevity. To make sure we explore diverse reward schemes, we specifically picked six sparse reward environments (Montezuma's Revenge, Pitfall!, Solaris, Private Eye, Freeway, Venture), three popular dense reward environments (Bank Heist, Q*Bert, Ms. Pac-Man), and three popular easy exploration environments (Breakout, Pong and Space Invaders), according to the standard taxonomy by \citet{bellemare2016unifying}.

\section{Initial Explorations of Reward Surfaces}
\begin{figure*}[ht]
\centering
\begin{tabular}{cc}
 \includegraphics[width=0.4\linewidth]{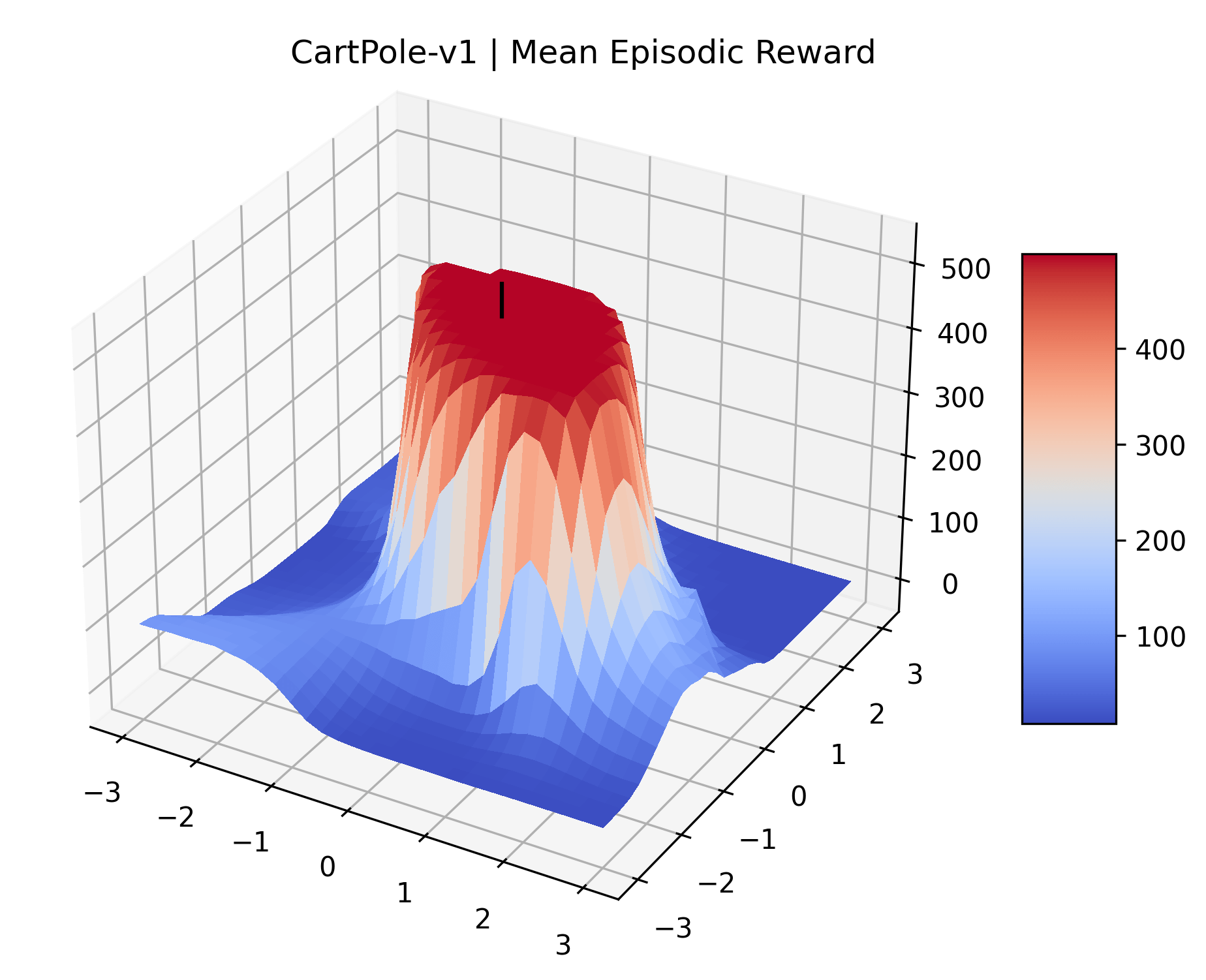} &
 \includegraphics[width=0.4\linewidth]{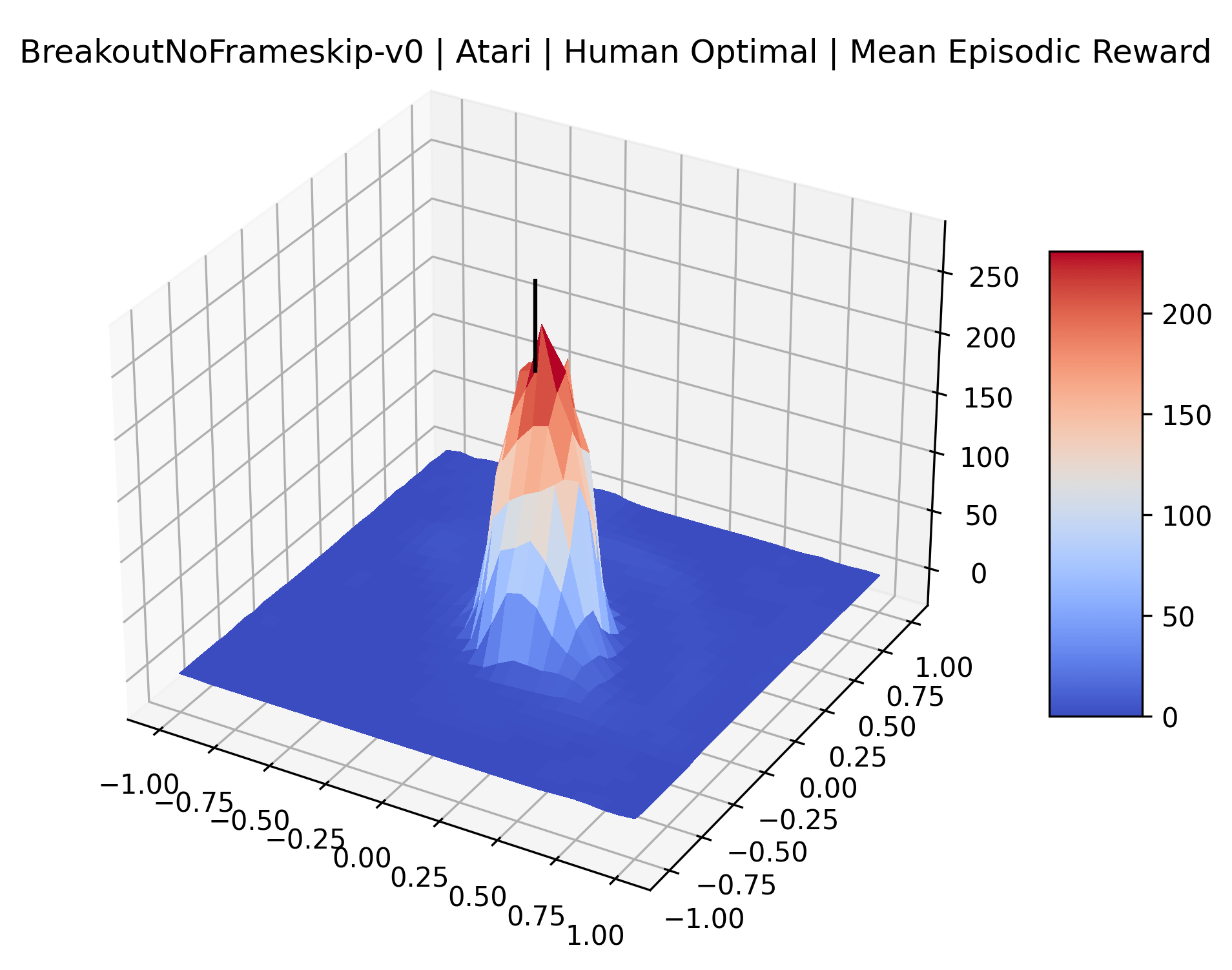} \\
 \includegraphics[width=0.4\linewidth]{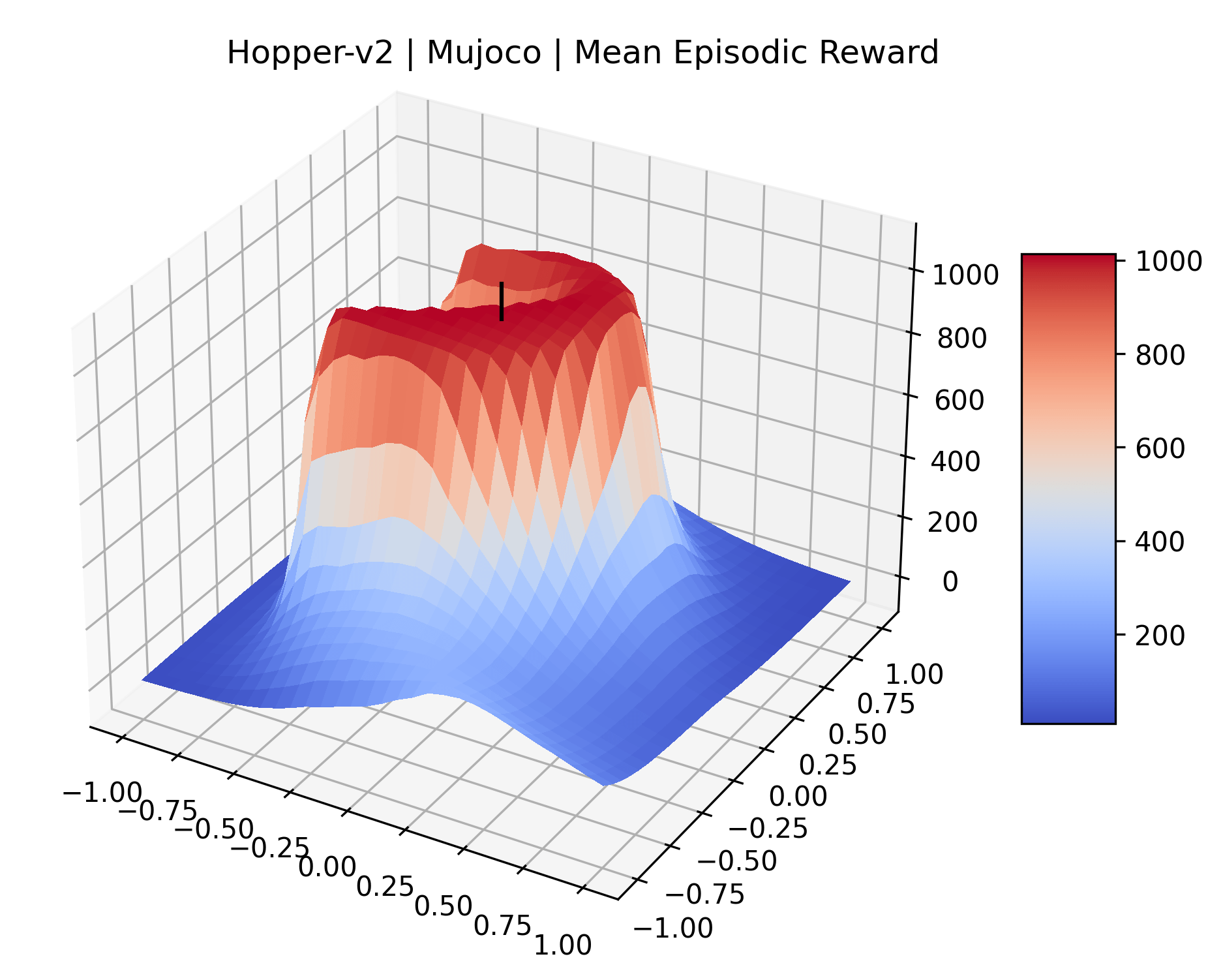} &
 \includegraphics[width=0.4\linewidth]{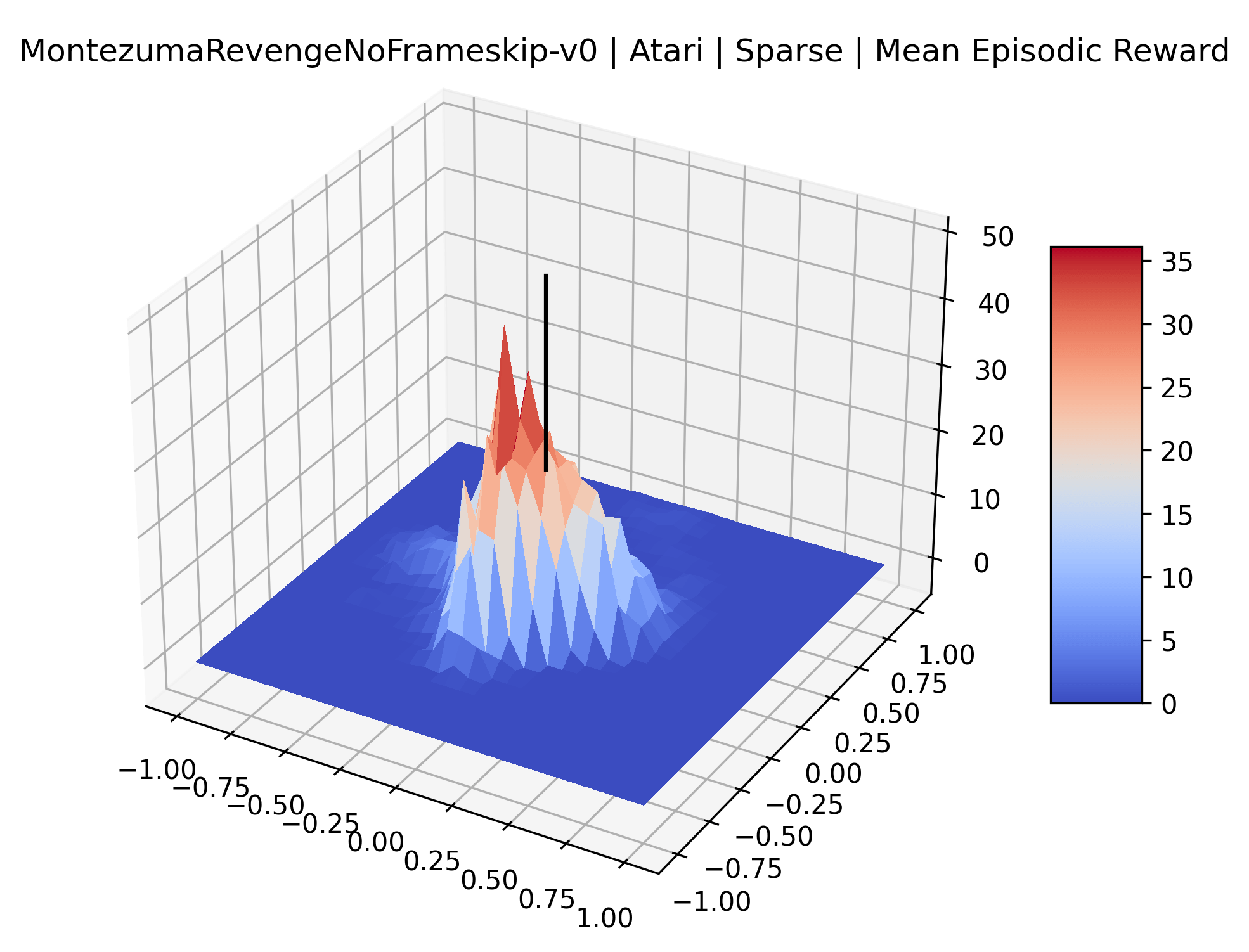} \\
\end{tabular}
\caption{Reward surfaces for CartPole-v1, Breakout, Hopper-v2, Montezuma's Revenge}
\label{fig:example_rewardsurface_table}
\end{figure*}

\subsection{Methodology}
\label{subsection:rand-dir}
In this work, we adapt the techniques from \citet{li2017visualizing} to visualize the reward surfaces of PPO agents in reinforcement learning environments. As in that paper, we deal with the high-dimensional domain of the reward surface by focusing our analysis around points in the policy space visited during training. Given a training checkpoint with parameters $\theta$, we are interested in understanding the local surface of rewards generated by the policy represented by parameters $\theta + d$ for small perturbations $d$.

To visualize this local region in 3 dimensions we plot the empirical expected episodic return on the z axis against independently sampled filter-normalized random directions on the x and y axes.\footnote{Note that because this is a high-dimensional space, these directions are orthogonal with high probability.} The plots are additionally scaled manually to highlight features of interest, so note the marks on the axes which indicate those manual scales. The scale, resolution, and number of environment steps used in each environment are listed in \autoref{fig:options_table}.

A reward surface is dependent on the chosen learning algorithm, network architecture, hyperparameters, and random seed, so for these experiments we chose to plot the reward surface of PPO agents using the tuned hyperparameters found in RL Zoo 3 \citep{rl-zoo3}. However, reward surfaces for a given environment are extremely visually similar across these variables, which we discuss in \autoref{sec:reproducibility}. To understand what challenges RL algorithms face towards the end of training after sufficient exploration has occurred, we chose the best checkpoint during training, evaluated on 25 episodes, with a preference for later checkpoints when evaluations showed equal rewards. The best checkpoint was typically found during the last 10\% of training.

\subsection{Results}
A sample of the visualizations of the reward surfaces can be seen in \autoref{fig:example_rewardsurface_table}. The full set of reward surface plots can be found in \autoref{appendix:reward_surfaces}. We additionally do a small number of experiments on network architecture, which we include in the appendix as a curiosity \autoref{appendix:network_architecture}. In the following sections we present our key findings and discuss the validity of our reward surface visualizations.

\subsubsection{Findings in Plots}
\begin{figure*}[ht]
\centering
\begin{tabular}{ccc}
 \includegraphics[width=0.3\linewidth]{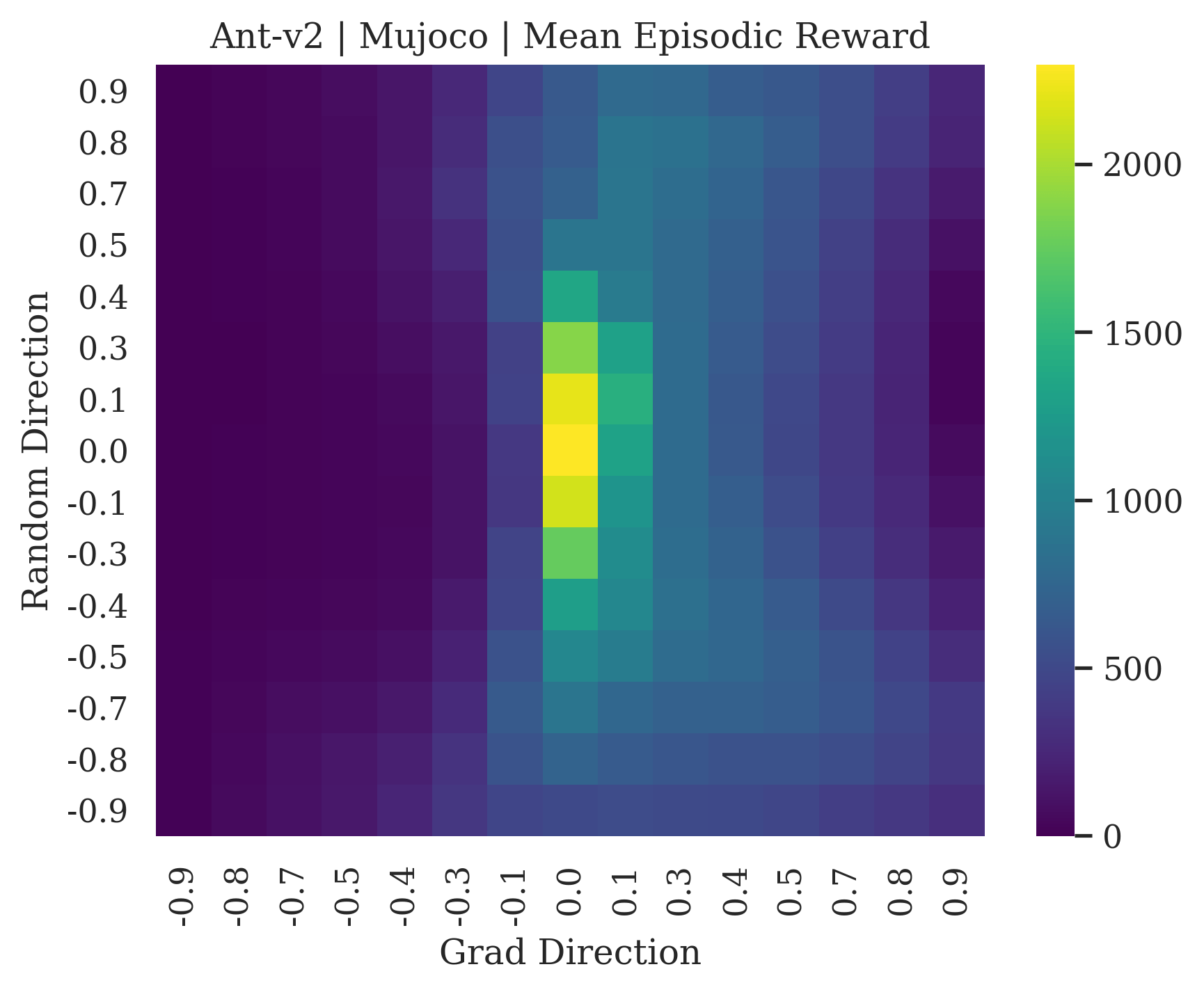} &
 \includegraphics[width=0.3\linewidth]{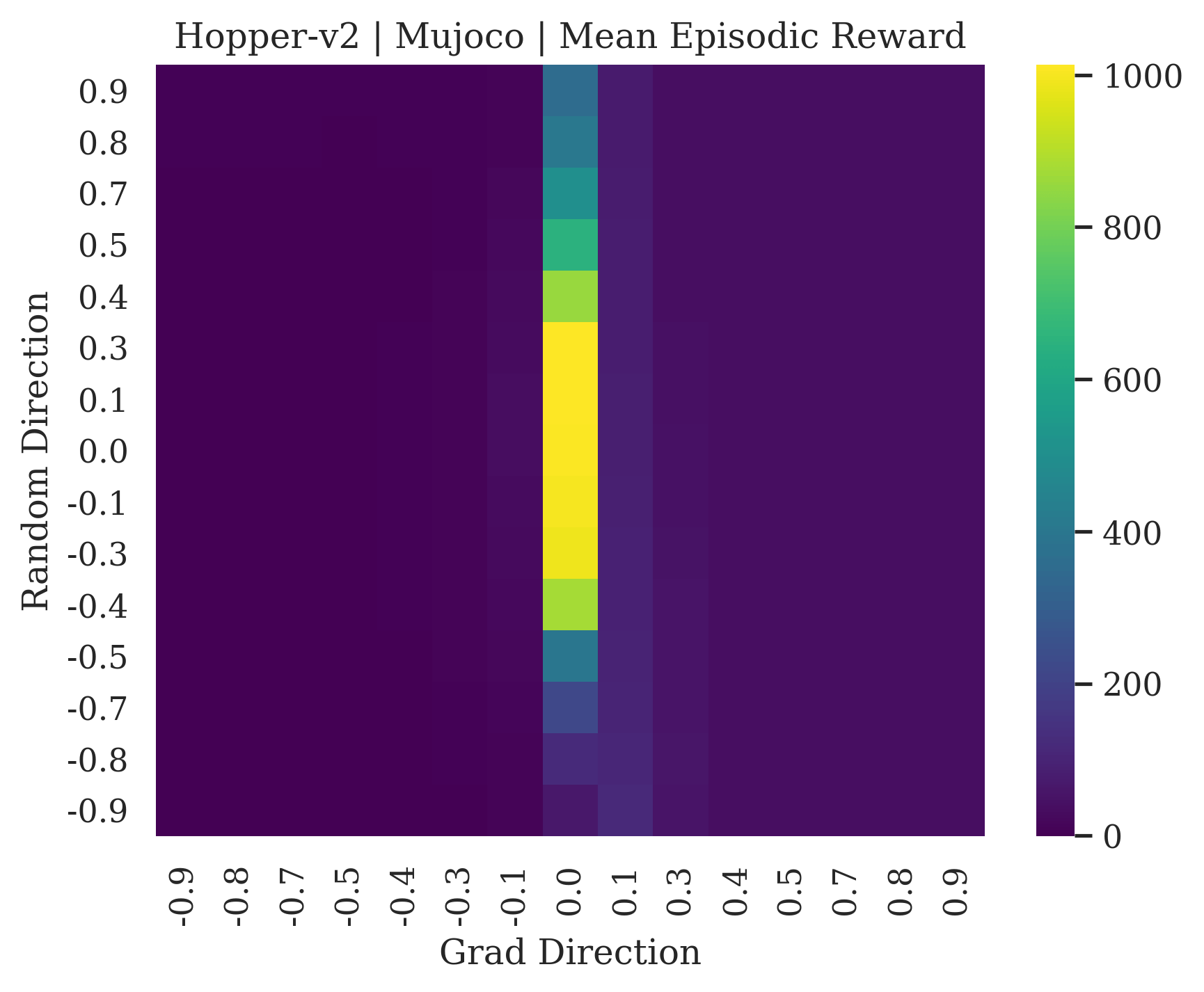} &
 \includegraphics[width=0.3\linewidth]{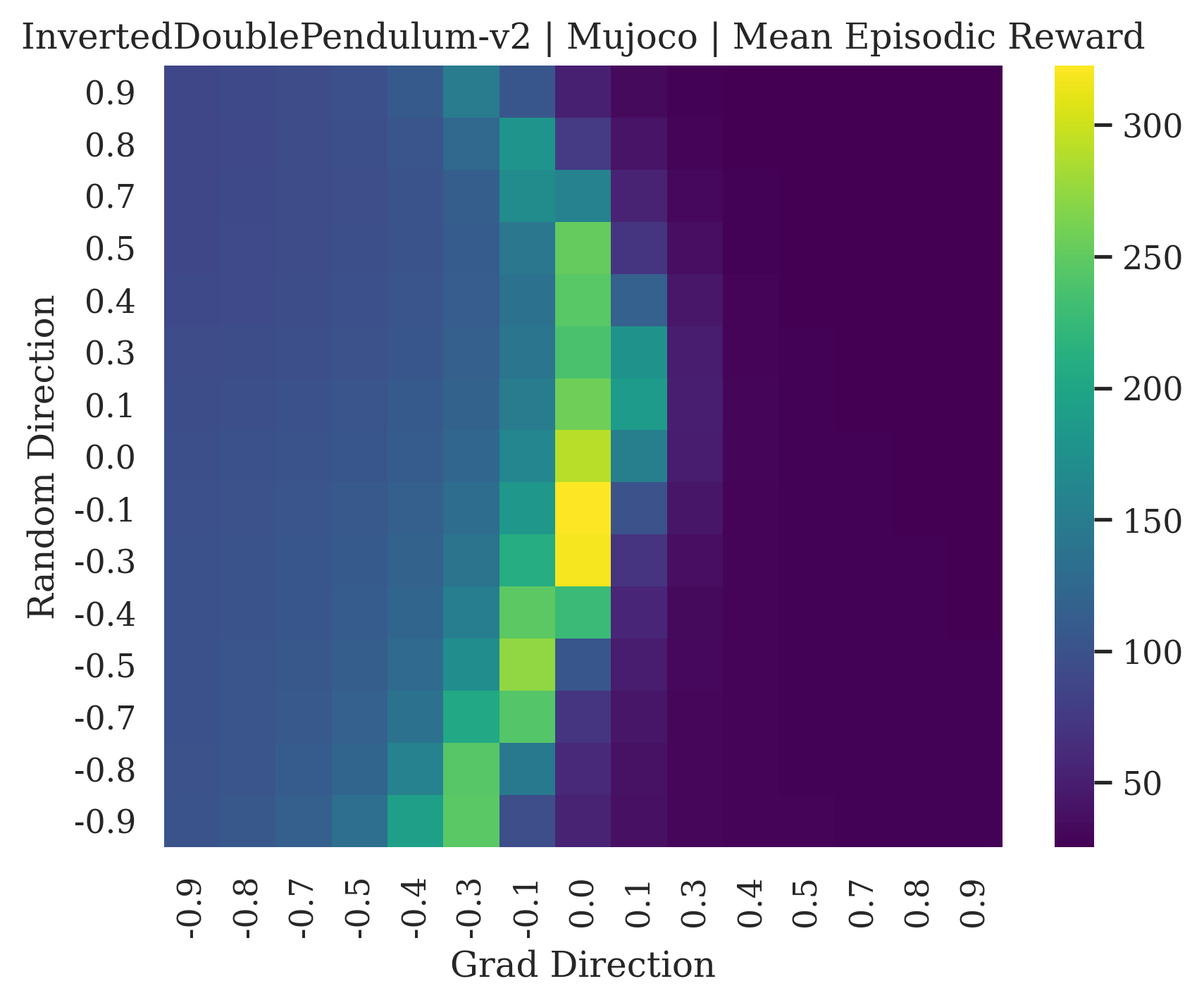} \\
\end{tabular}
\caption{Gradient heat maps for Ant, Hopper, and InvertedDoublePendulum. The heat map for Hopper and InvertedDoublePendulum shows a cliff-like gradient direction which falls off sharply compared to the filter-normalized direction.}
\label{fig:heatmap_table}
\end{figure*}

One obvious observation is that the size and shape of the maximizers present in the reward surfaces roughly correlates with the perceived difficulty of the environment. Though there is some recent work attempting to quantify the difficulty of RL environments, it is still a challenging and unsolved problem \citep{Oller2020AnalyzingRL, furuta2021pic}. However, researchers have used these benchmarks for years, and we now have some intuition for identifying environments that are easier for modern methods to solve. The Classic Control environments were designed as simple toy environments, and modern methods easily find strong policies for them. The Atari environments have been taxonomized by \citet{bellemare2016unifying} to indicate their relative difficulty. Breakout, Pong, and Space Invaders are listed as "Easy Exploration" games. Among the "Hard Exploration" games Bank Heist, Ms. Pacman, and Q*bert have dense rewards, while Freeway, Montezuma's Revenge, Pitall!, Private Eye, Solaris, and Venture have a more challenging sparse reward structure. The MuJoCo environments have not been officially categorized according to difficulty to our knowledge. 

Using these rough categories as a guide, we see that the Classic Control environments shown in \autoref{fig:classiccontrol_rewardsurface_table}, certainly the easiest set that we study, have large maximizers that often span the entire domain of the plot. In the Atari environments shown in \autoref{fig:atari_rewardsurface_table}, we see that the "Human Optimal" (Easy Exploration) and "Dense Reward" (Hard Exploration) environments have large smooth maximizers relative the the chaotic landscapes seen in the "Sparse Reward" environments. This finding complements those of \citet{li2017visualizing} who found that the sharpness of loss landscapes using filter-normalized directions correlated with generalization error. This observation also raises the possibility of future work creating a metric based on reward surfaces to gauge the difficulty of RL environments.

We also observe some interesting characteristics of individual environments. In \autoref{fig:classiccontrol_rewardsurface_table}, MountainCarContinuous has a strangely spiky reward surface for a Classic Control environment. This may be a result of the uniquely low training time required to solve it. Using the hyperparameters from RL Zoo 3 we train this agent for only 20,000 steps, while the next lowest training time in RL Zoo 3 is 200,000 timesteps. Looking at \autoref{fig:mujoco_rewardsurface_table}, Humanoid Standup has fairly spiky reward surface, suggesting that it's reward function is sensitive to small perturbations in the policy. While many maximizers in these plots appear as a single hill or peak in an otherwise flat region, some of the surfaces have uniquely identifiable curvature. Hopper has a large semi-circular curve in its reward surface. Cartpole, MountainCar, and InvertedPendulum have large plateaus at their peak reward.

The sparse reward Atari environments in \autoref{fig:atari_rewardsurface_table} are particularly interesting to examine. Note that each point in the sparse Atari reward surfaces were evaluated with either 1 or 2 million environment steps, to limit the standard error for each estimate. We see that Freeway's reward surface has large flat regions of varying heights surrounding a single smooth maximizer. Montezuma's Revenge and Venture have short noisy maximizers. The agents for the sparse environments (except Freeway) perform poorly, so note that even the maximizers in these plots represent weak policies. As a result of this, we see that the reward surfaces for Montezuma's Revenge, Solaris, and Venture show nearby maximizers that the agent was unable to find during training. Private Eye has a large region of zero reward and a far away maximum with much higher rewards. Finally, as its name suggests, the reward surface for Pitfall! is mostly flat and marred by several deep pits. 

The idea that sparse rewards structures lead to flat regions in a reward surface is intuitive -- in environments where rewards are issued solely in infrequent goal states, large yet precise policy improvements may be required to experience variations in rewards. Despite the intuitive nature of this observation, we're unaware of this being visually documented in the literature. We also find that in regions where the reward surfaces are not flat, they are often extremely noisy. This is supported by plots of the surface of standard deviations at each point, which we include in \autoref{appendix:standard_deviation}. We see in these plots that the standard deviation is often significantly larger than the average reward in the reward surfaces for Montezuma's Revenge, Private Eye, and Venture, unlike any of the other environments we study. These plots seem to highlight different failure modes of sparse environments -- either the surface is flat when the agent experiences no variation in rewards, or the surface is spiky when rewards are sparse and noisy.

\subsubsection{Plot Reproducibility and Standard Error}
\label{sec:reproducibility}
To demonstrate the consistency of these experiments across multiple random seeds, we repeated our reward surface plots 18 times for Acrobot, HalfCheetah, Breakout, and Montezuma's Revenge. For each trial, we trained and evaluated a new agent with a new random seed. We can see from the plots in \autoref{appendix:reproducibility} that the reward surfaces are extremely visually similar for a particular environment, indicating that training appears to converge to visually similar regions of the reward landscape, and that the characteristics of these plots are consistent across multiple seeds. 

We evaluated for at least 200,000 time steps (1 or 2 million steps for the sparse reward Atari games) at each point to ensure that the standard error for these estimates is small. We record the standard error of each plot in \autoref{appendix:standard_error}.

\section{Discovering Cliffs in the Gradient Direction}
\label{subsection:explore-grad}

\begin{figure*}[ht]
\centering
\begin{tabular}{ccc}
 \includegraphics[width=0.3\linewidth]{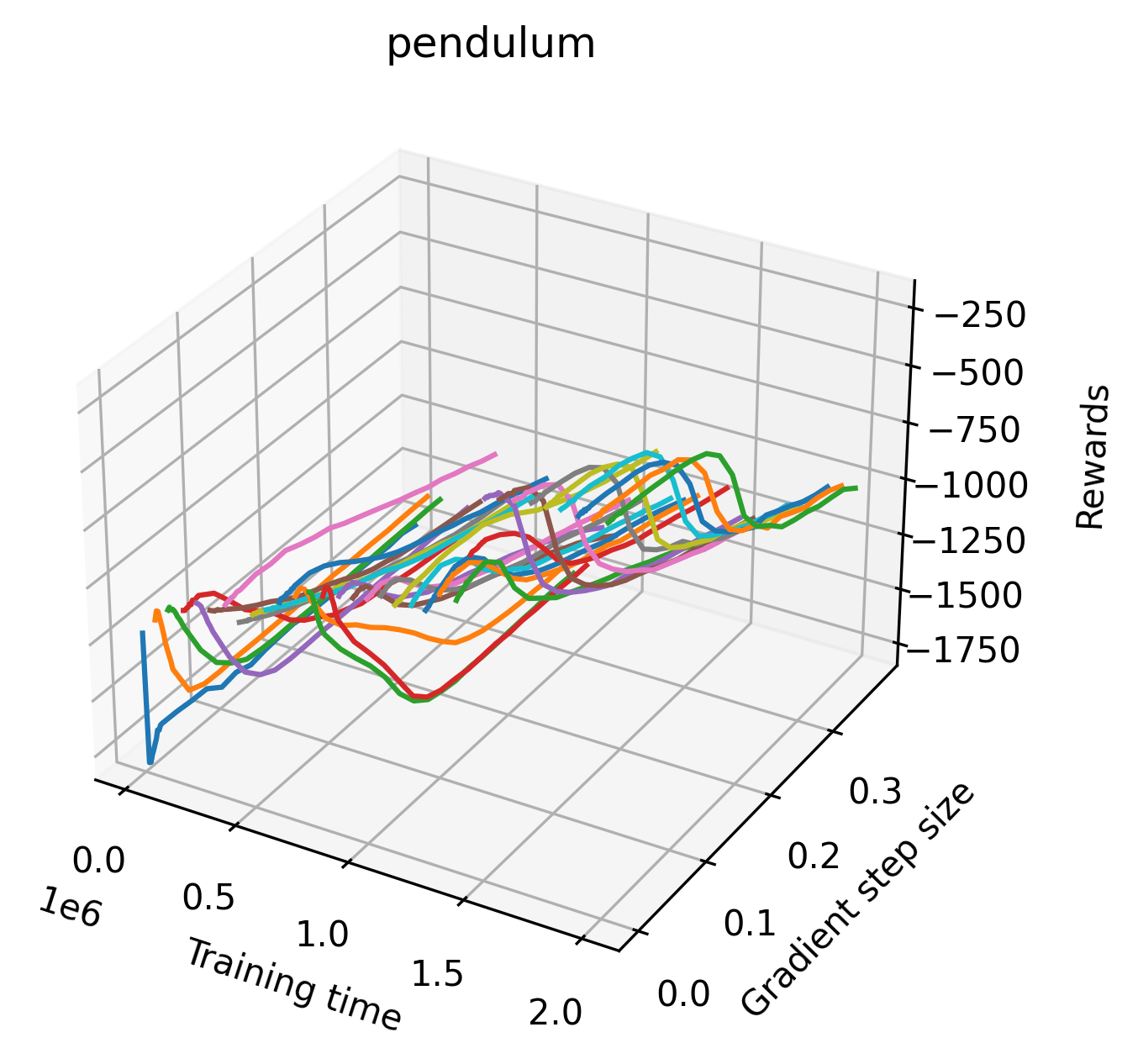} &
 \includegraphics[width=0.3\linewidth]{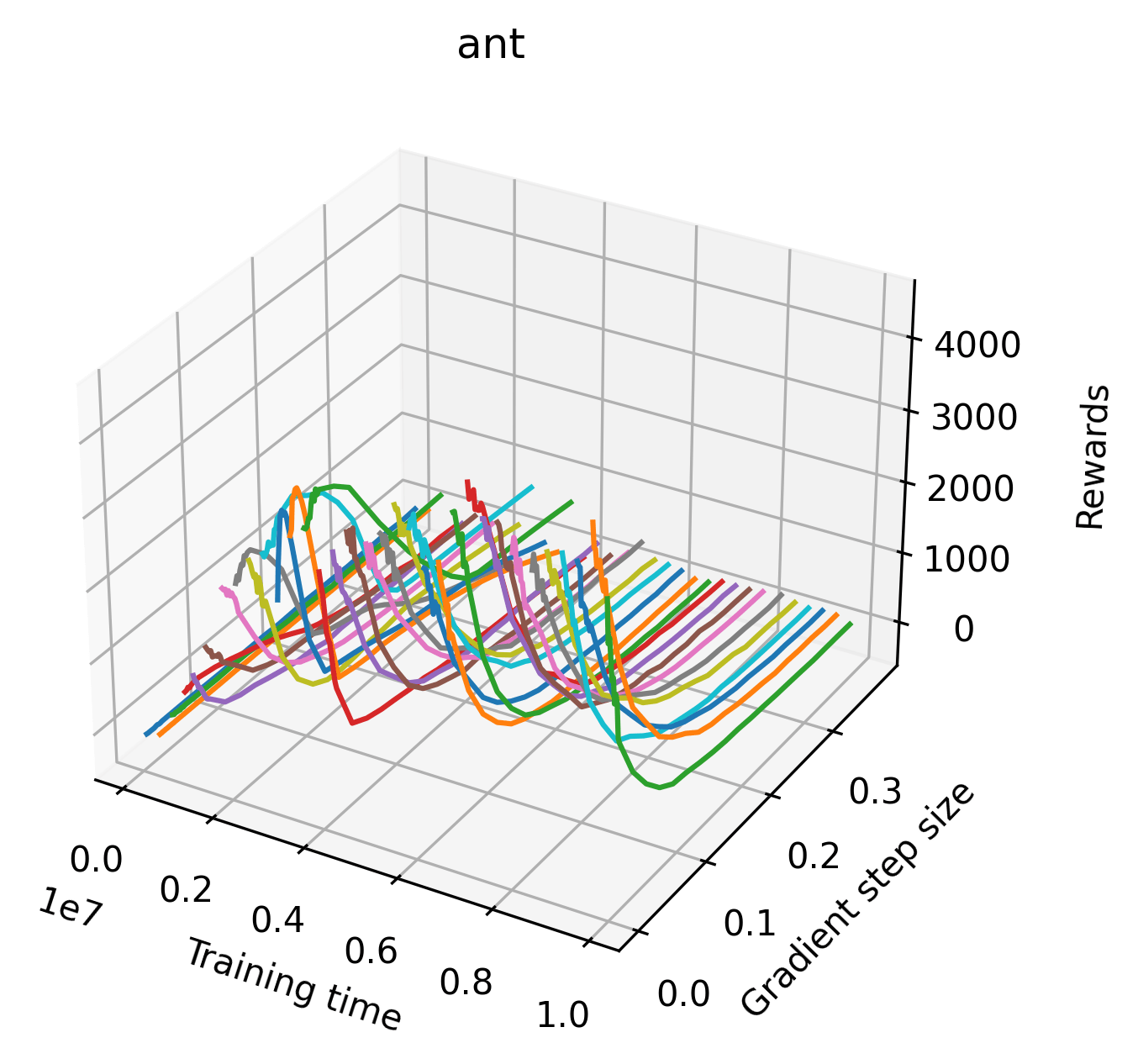} &
 \includegraphics[width=0.3\linewidth]{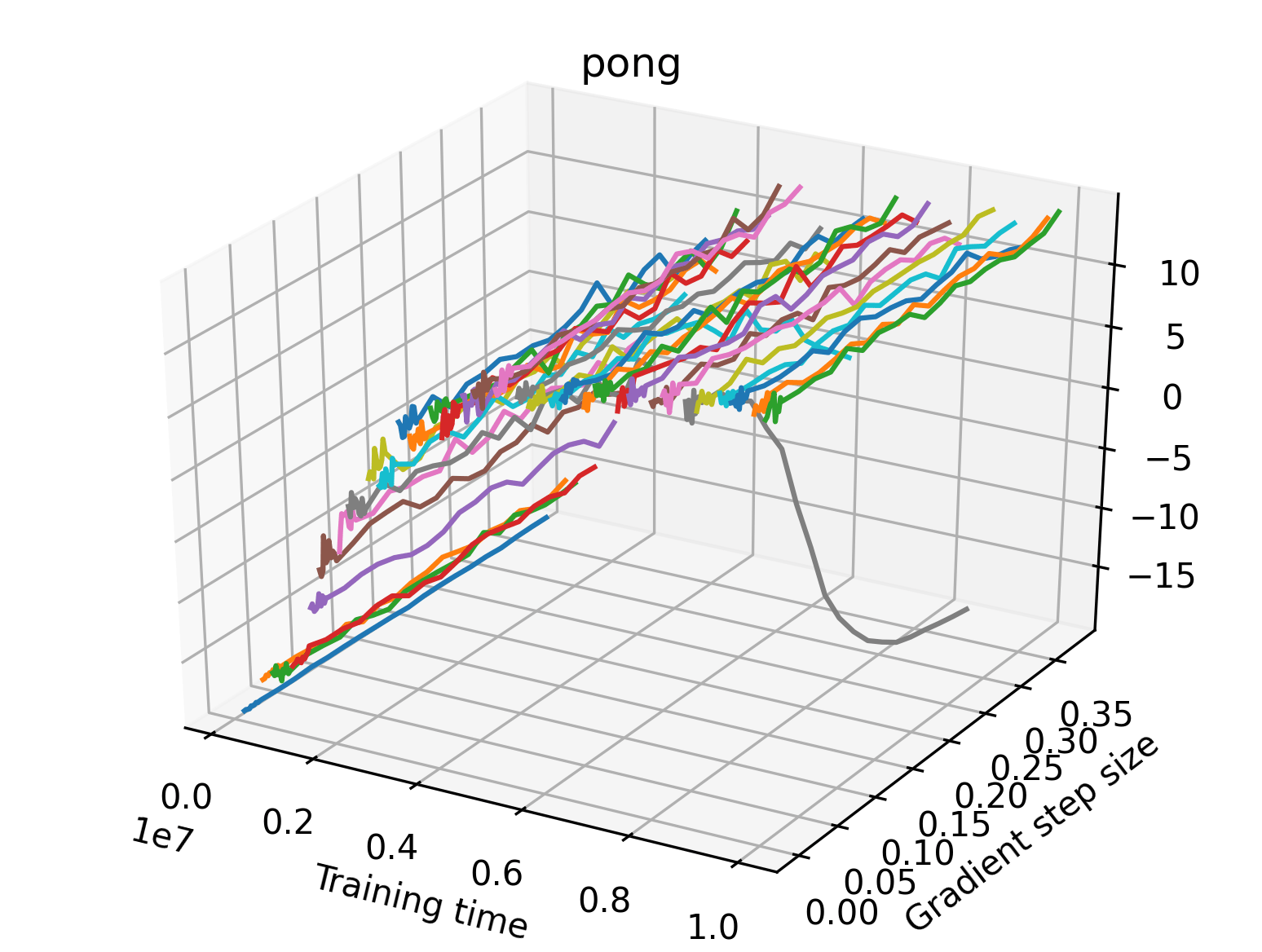} \\
\end{tabular}
\caption{Gradient line search plots for Pendulum, Ant, and Pong. The line plot for Ant shows several checkpoints that exhibit cliff-like properties.}
\label{fig:lineplot_table}
\end{figure*}

Filter normalized random directions provide a broad sense of the local reward landscape, and are useful for analysis near the end of training, but they do not represent the direction of training because the sampled random directions are likely orthogonal to the gradient direction. To better understand the optimization characteristics of these environments, we performed similar experiments using the policy gradient direction. We use a high quality estimate of the policy gradient computed over 1,000,000 environment steps as in \citet{Ilyas2020A}. In many of these plots, we find evidence of ``cliffs'' in the reward surface. These are regions of the reward surface in which rewards sharply decrease after a short distance. We discuss the difficulty of precisely defining a sharp decrease in \autoref{appendix:limitations} and explain the heuristic criteria used to identify cliffs in \autoref{appendix:cliff-selection}.
    
One difficulty of plotting the gradient direction is that the gradient magnitudes vary drastically for different environments and at different points during training \citep{mccandlish2018empirical}. Additionally, any maximum in a reward surface can be made to look like a sharp cliff by using a large enough gradient scale, or like a large plateau by using a smaller gradient scale. To provide a fair comparison of the gradient direction's sharpness, we normalize the gradient direction by dividing each component by it's L2 norm.

\subsection{Gradient Directions vs. Filter-Normalized Random Directions}

We show the differences between filter normalized random directions and the gradient direction by creating plots of the reward surface using the gradient direction on the x axis and a random filter normalized direction on the y axis. Due to the frequent sharp changes in reward that we observe in these plots, 3d surface plots can become partially obscured by peaks, so we choose to plot these surfaces as heat maps instead. A sample of the heat maps can be seen in \autoref{fig:heatmap_table} and the full set of heat maps for Classic Control and MuJoCo environments can be found in \autoref{appendix:heat_maps}.

\subsubsection{Preliminary Observations}
The most striking observation is that rewards in the gradient direction often change much more rapidly than the rewards in random directions. We see a sample of these gradient heat maps in \autoref{fig:heatmap_table}. In the gradient heat maps for Hopper and InvertedDoublePendulum, rewards in the gradient direction seem to form cliffs and drop off rapidly after a short distance. The plot for InvertedDoublePendulum is particularly interesting. It is possible to argue that the different normalization schemes that we use for random and gradient directions make these plots falsely appear to have cliffs. However, the rewards for InvertedDoublePendulum drop much more quickly in the gradient direction than in the negative gradient direction. Due to these potential concerns about normalization, in the next section we directly visualize the gradient directions across multiple training checkpoints for each environment to find more convincing evidence of cliffs.

\subsection{Visualizing Rewards in the Gradient Direction During Training}

\subsubsection{Methodology}
In order to study the gradient direction's reward surface over the course of training, we plot a 1-dimensional projection of the rewards along the gradient direction for a series of checkpoints taken at uniform training step increments. A sample of these can be seen in \autoref{fig:lineplot_table} and the full set of plots can be found in \autoref{appendix:line_plots} Since the training checkpoints are relatively far apart from one another, the plot is somewhat discontinuous. However, we selected uniformly distributed checkpoints across the entire span of training, so they should be representative of all points visited during training. We sample 20 points along a distance of 0.4 in the normalized gradient direction, and another 10 points between the first and second sample point. This results in a single high resolution segment at the start of each plot that allows us to more easily identify cliffs early in the plots. The Atari environments Montezuma's Revenge, Pitfall!, Private Eye, and Solaris were plotted with 1,000,000 environment steps per sample point to limit evaluation noise, while the remaining environments used 200,000 time steps. 

\subsubsection{Observations}
We find that many of the same observations here as we do in the original reward surfaces. The gradient directions for easy, dense reward environments tend to point toward better rewards, and sparse reward environments often have flat trajectories in the gradient direction. Some of the sparse environments also have extremely noisy gradient directions, with wide swings in reward. This typically occurs in environments where the agent performs poorly. For example, we see in Freeway, a sparse reward environment where our agent finds a nearly optimal policy, that the gradient line plot looks very similar to that of dense reward Atari environments like Pong. However, we also note some unique properties of the gradient direction. In some plots, for example in Pong, we see ``cliffs'' in the reward surface where the reward briefly increases, then sharply decreases. We find that these cliffs occur occasionally in almost every environment.

\section{Cliffs Impact Policy Gradient Training}

\begin{table*}[ht]
\centering
\begin{tabular}{|c|c|c|cc|}
\hline
 & & & Cliff & Non-Cliff \\ \hline
 \multirow{8}{*}{N steps = 128}
    & \multirow{2}{*}{LR = 0.000001} & PPO & 0.03\% & 0.03\% \\ & & A2C & -0.3\% & 0.2\% \\ \cline{2-5}
    & \multirow{2}{*}{LR = 0.0001} & PPO & -0.01\% & -0.03\% \\ & & A2C & -4\% & 2.3\% \\ \cline{2-5}
    & \multirow{2}{*}{LR = 0.01} & PPO & 0.0\% & 0.4\% \\ & & A2C & -0.3\% & 2.0\% \\ \cline{2-5}
    & \multirow{2}{*}{LR = 0.5} & PPO & 0.0\% & 0.0\% \\ & & A2C & -8.5\% & 1.6\% \\ \hline
 \multirow{8}{*}{N steps = 2048} 
    & \multirow{2}{*}{LR = 0.000001} & PPO & -0.1\% & -0.4\% \\ & & A2C & -0.5\% & 0.2\% \\ \cline{2-5}
    & \multirow{2}{*}{LR = 0.0001} & PPO & -0.2\% & -0.1\% \\ & & A2C & -7.0\% & 4.4\% \\ \cline{2-5}
    & \multirow{2}{*}{LR = 0.01} & PPO & 0.1\% & 0.1\% \\ & & A2C & -3.9\% & 2.9\% \\ \cline{2-5}
    & \multirow{2}{*}{LR = 0.5} & PPO & 0.0\% & 0.3\% \\ & & A2C & -3.6\% & -0.3\% \\ \hline
\end{tabular}
\caption{Table of A2C and PPO's average percent change in reward after taking a few gradient steps on cliff and non-cliff checkpoints for various sets of hyperparameters. These results are averaged among 10 trials each evaluated for 1000 episodes.}
\label{fig:a2c_v_ppo_table}
\end{table*}

We are clearly able to see cliffs in our line plots, but we needed to confirm that these cliff-like gradient directions are not simply a visualization artifact and that they affect the performance of agents. We hypothesize that methods which approximate the policy gradient will occasionally step too far and fall off of these cliffs, thereby performing worse on cliff-like checkpoints than normal checkpoints. This hypothesis follows from the intuition that motivated PPO and TRPO. To test our hypothesis, we use the line plots to identify trained checkpoints where the true policy gradient points towards a cliff, and compare the performance of A2C on these cliffs versus less cliff-like checkpoints. We find that A2C performs significantly worse on cliff checkpoints than non-cliff checkpoints. We further hypothesize that PPO, which uses ratio clipping to avoid significant changes to its policy and better hyperparameters, will perform better than A2C on cliff checkpoints. We run the same experiment using PPO, and compare the results in  \autoref{fig:a2c_v_ppo_table}.

\subsection{Methodology}
\label{sec:cliff_methodology}

To evaluate performance on cliffs relative to baseline performance on standard checkpoints, we first select 12 cliff and 12 non-cliff checkpoints from our gradient line search data using a heuristic explained in \autoref{appendix:cliff-selection}.

For both A2C and PPO we evaluate the percent change in reward that results from training for 2048 environment steps at a particular checkpoint. For each checkpoint, we perform 10 trials in which we evaluate the starting performance of the agent over 1000 episodes, the agent takes a few gradient steps, and then we evaluate the resulting policy over 1000 episodes as well. We calculate the percent change in reward, and average this value over every checkpoint in each set. This produces an average change in reward for the set of cliff checkpoints, and the set of non-cliff checkpoints, which we list in \autoref{fig:a2c_v_ppo_table}.

We also want to ensure that the method takes a step in the direction that we are studying, and steps far enough to reach the cliff. As such, we try increasing both the learning rate (LR) and the number of steps per parallel environment per training update (N steps). The remaining hyperparameters are the optimal hyperparameters from RL Zoo 3 \citep{rl-zoo3} as in previous experiments. We use these hyperparameters to validate the existence of cliffs and demonstrate that cliffs can have a negative effect on training, whether or not it commonly occurs in practice.

\subsection{Results}


We see the results of our experiments with several hyperparameter sets in \autoref{fig:a2c_v_ppo_table}. We find that on cliff checkpoints, A2C's gradient step consistently results in a decrease in rewards, while on non-cliff checkpoints it increases expected return. On the other hand, PPO sees nearly the same percent change in performance across cliff and non-cliff checkpoints. From these results, we confirm that the cliffs present in our gradient line searches can have a real effect on optimization, and are not simply a visualization artifact. We also show that for all tested hyperparameters, PPO is affected by cliffs less than A2C. Only one of our experiments shows a minor decrease in rewards for PPO, while all of them show a larger decrease for A2C. \citet{Engstrom2020Implementation} found that PPO's performance could largely be attributed to hyperparameter improvements and implementation tricks, so we leave a thorough investigation of the exact components that cause PPO to be less affected by cliffs as future work.

\section{Library}
We developed an extensive software library for plotting the reward surfaces of reinforcement learning agents to produce this work and encourage future research using these visualizations. The library includes code for training agents using all of the options available in Stable Baselines 3 \citep{stable-baselines3} and hyperparameters from RL Zoo 3 \citep{rl-zoo3}. We provide algorithms for estimating the gradient and hessian of policy networks along with code for evaluating the rewards or discounted returns of trained agents. The entire code base supports the use of arbitrary directions for investigation, and specifically provides tools for using filter-normalized and policy gradient directions. We include routines for creating 3d plots, line plots, heat maps, and gifs of reward surfaces. Finally, all experiments in the library are parallelized across multiple environments, and scripts are included for generating reward surfaces on SLURM clusters. The library is well organized and documented, and it can be found at \href{https://github.com/RyanNavillus/reward-surfaces}{https://github.com/RyanNavillus/reward-surfaces}.

\section{Discussion}
In this work we introduce valuable new methods for studying deep reinforcement learning, and use them to discover new results about the optimization characteristics of popular RL environments. Reward surfaces provide a useful overview of the reward structure of an environment. Loss landscapes have already been used in debugging tools for computer vision tasks \citep{bain2021lossplot}, and we hope that reward surfaces could be similarly useful in debugging RL systems. In particular, the reward surfaces for sparse environments allow us to see large regions of flat rewards, and extreme evaluation noise at individual points. Gradient methods cannot optimize flat surfaces, so solutions to this problem are constrained to either modifying the reward structure (e.g. curiosity or bonus-based exploration methods \citep{pathak2017curiosity, burda2018exploration}) or condensing the action space such that simple exploration methods are tractable (e.g. DIAYN and related work \citep{DBLP:conf/iclr/EysenbachGIL19, Sharma2020Dynamics-Aware}). 

All this suggests many interesting opportunities for future work, either visualizing the effects of bonus-based exploration methods on the reward surface, or quantification of reward sparsity. Additionally, our gradient line searches visualize optimization characteristics of the environment, and could allow us to select gradient methods more suited to reinforcement learning. We see a few worthwhile research directions using these techniques.

Our gradient line searches show evidence of cliffs in most popular RL environments. Interestingly, although these cliffs appear in almost every environment occasionally, the most extreme examples occur in relatively easy environments like CartPole and Inverted Double Pendulum, while the harder Atari environments are characterized by mostly flat and noisy gradient directions. Despite the apparent difference, these noisy spikes can also be considered small cliffs on the scale of an individual gradient step. Our experiments demonstrate that the extreme cliffs can have an impact on training over a few gradient steps, but we suspect that smaller cliffs have a slower, degrading effect over the course of training. Future work could attempt to directly study the degree to which cliffs affect training in practice.

Our experiments comparing PPO and A2C provide an empirical justification for why PPO is so effective. We find that PPO performs significantly better than A2C on checkpoints with steep cliffs in the gradient direction's reward surface. A deeper understanding of why existing methods work will allow us to develop stronger algorithms in the future. Previous work has focused on the hyperparameters and algorithmic advances that contribute to PPO's strong performance \citep{Engstrom2020Implementation}, but we believe that the specific situations where PPO outperforms previous methods warrants further investigation.

Finally, this work focuses on policy networks, but the tools that we introduce could be applied to study Q networks or value networks. It would also be interesting to see reward surfaces for multiagent algorithms, such as independent learning or parameter sharing~\citep{10.1007/978-3-319-71682-4_5}.

\section{Conclusion}
This work is the first to use filter-normalized directions to visualize reward surfaces for a large collection of popular reinforcement learning environments. We are also the first to find visual evidence of the cliffs that inspired TRPO and PPO, and we perform experiments demonstrating their negative impact on policy gradient methods. This offers new potential insights into why deep RL works, and why reinforcement learning is seemingly so challenging when compared to other areas of deep learning. To accelerate future works in this field we created an extensive, well-documented library for plotting reward surfaces. We thoroughly outline limitations this work has in \autoref{appendix:limitations}. We hope that this work inspires future research on the specific optimization challenges that reinforcement learning faces, and that it enables new studies using reward surfaces.

\section{Acknowledgements}

We would like to thank Joseph Suarez, Costa Huang, and our reviewers for their helpful comments and writing suggestions.


\bibliography{main}
\bibliographystyle{icml2022}

\newpage
\appendix
\onecolumn
\newcommand\surfacescale{0.31\linewidth}

\section{All Reward Surfaces}
\label{appendix:reward_surfaces}

\subsection{Classic Control}
\begin{figure*}[!ht]
\centering
\begin{tabular}{ccc}
 \includegraphics[width=\surfacescale]{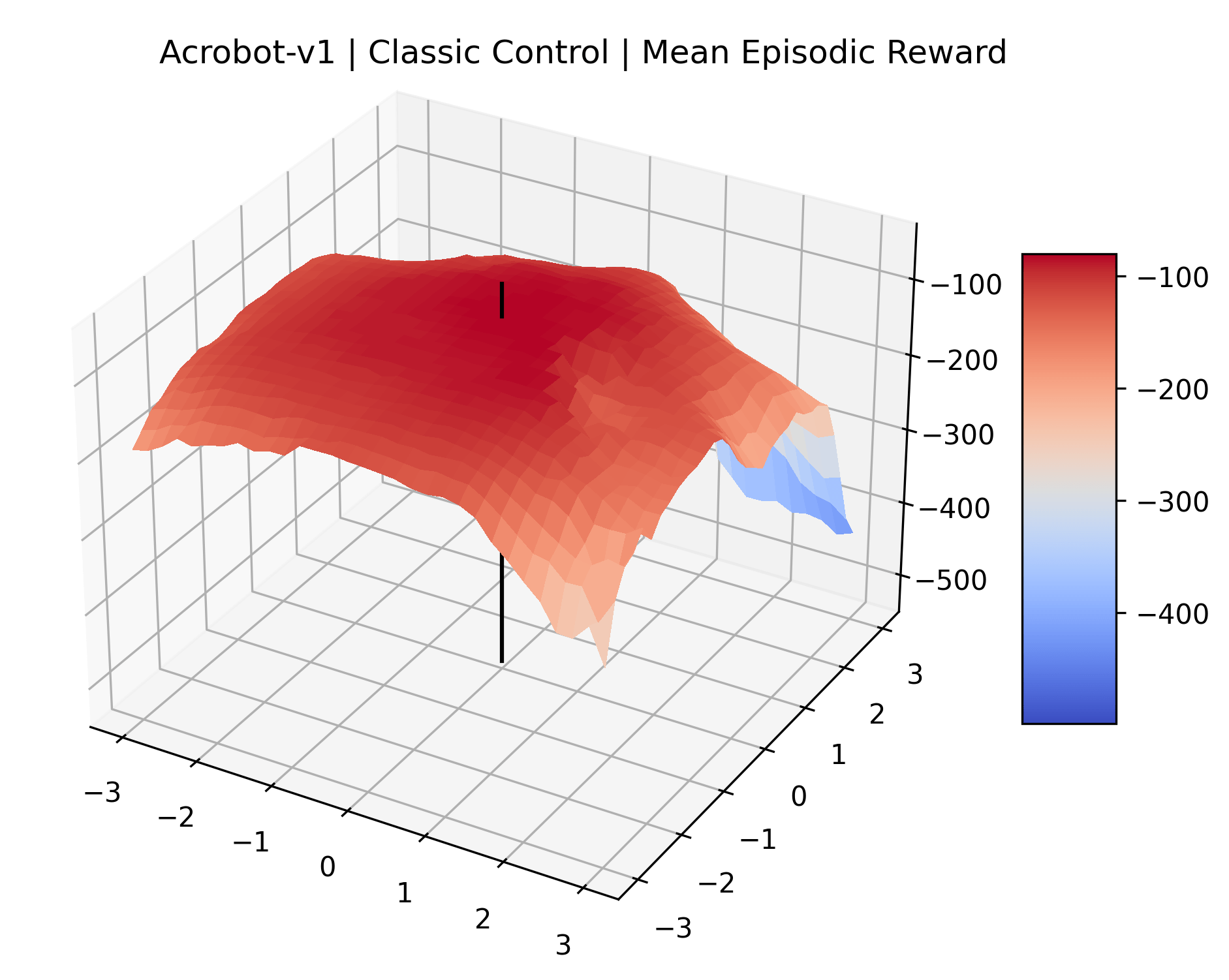} &
 \includegraphics[width=\surfacescale]{./plots/cartpole3x3_episoderewards_3dsurface.png} &
 \includegraphics[width=\surfacescale]{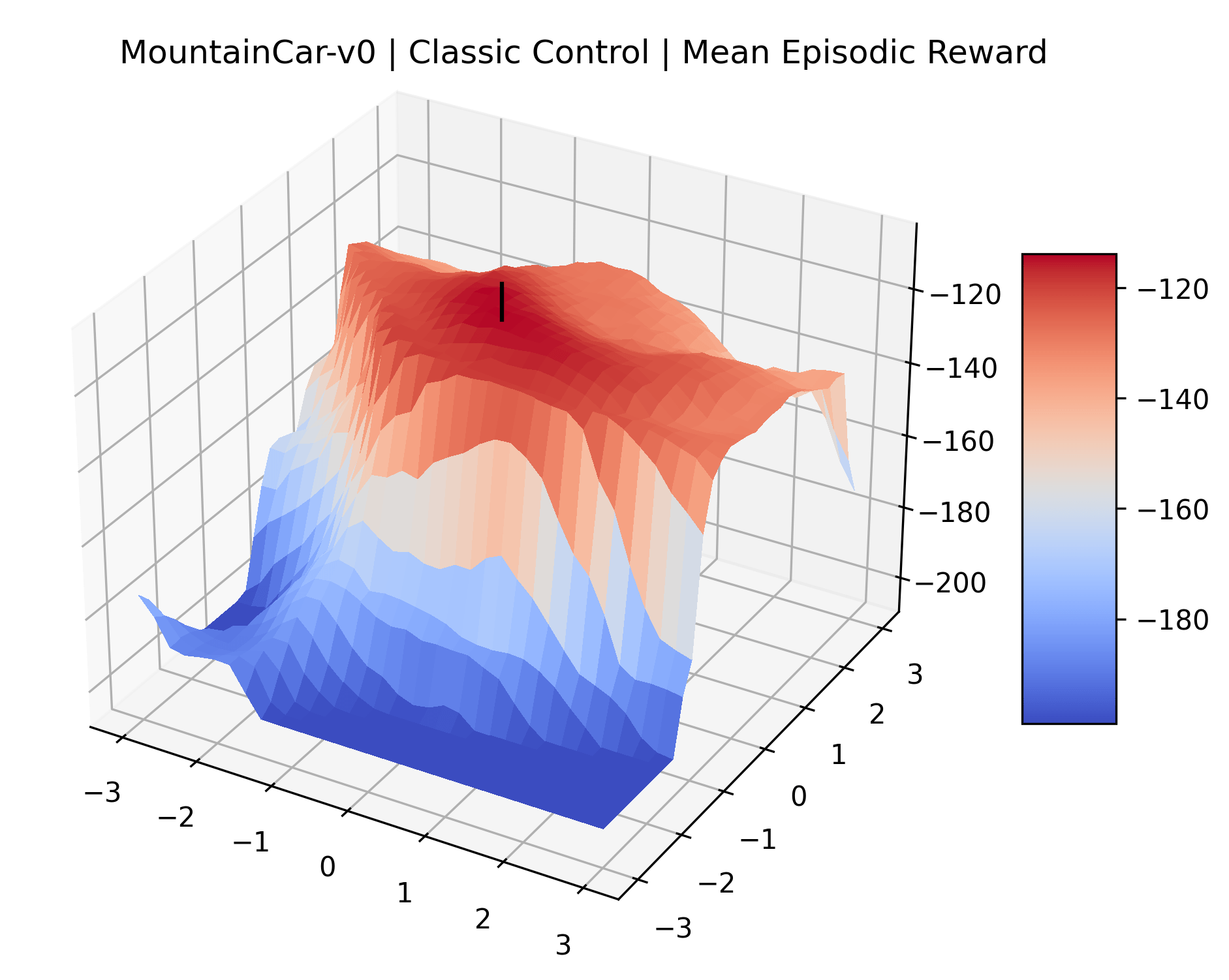} \\
\end{tabular}
\begin{tabular}{cc}
 \includegraphics[width=\surfacescale]{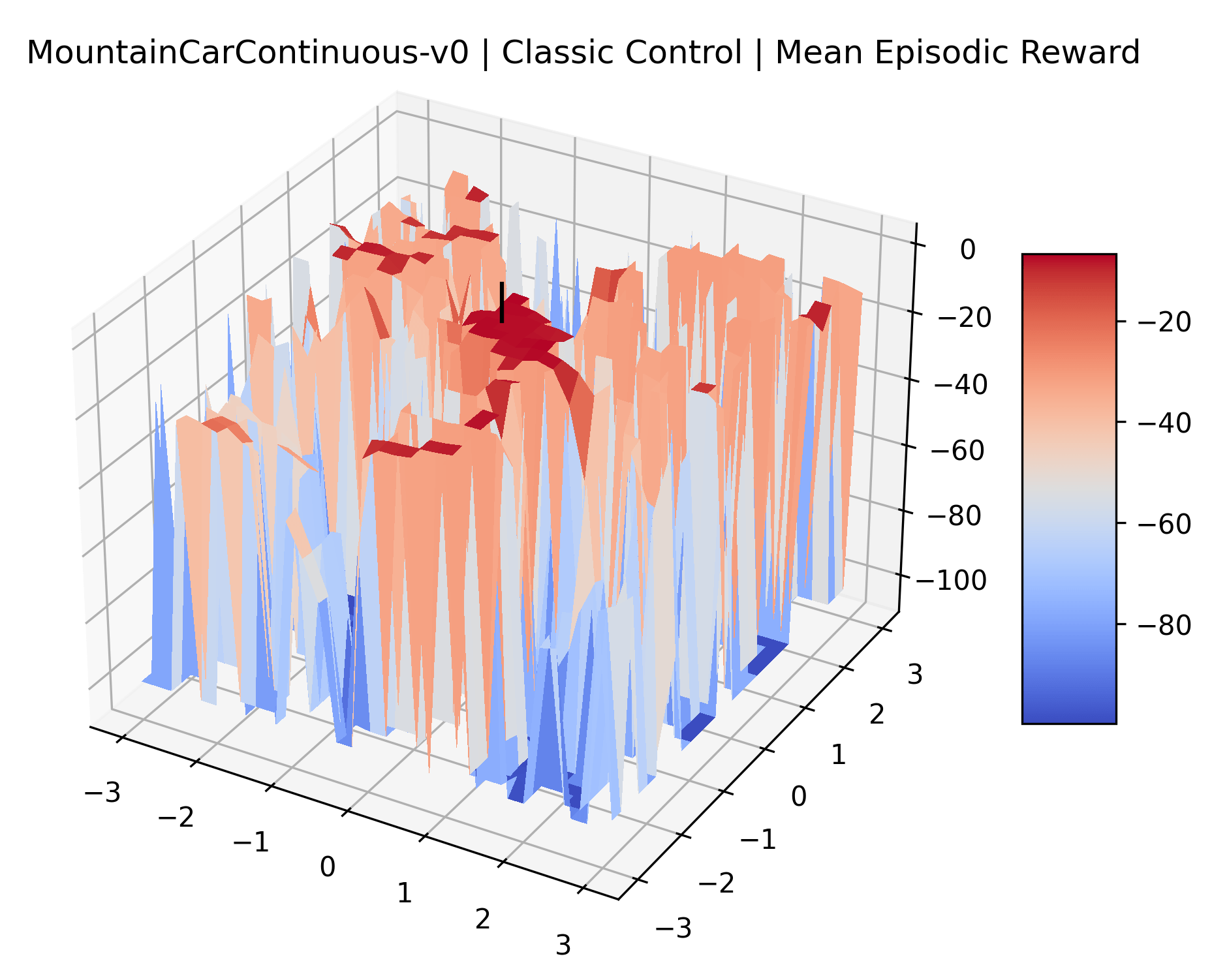} &
 \includegraphics[width=\surfacescale]{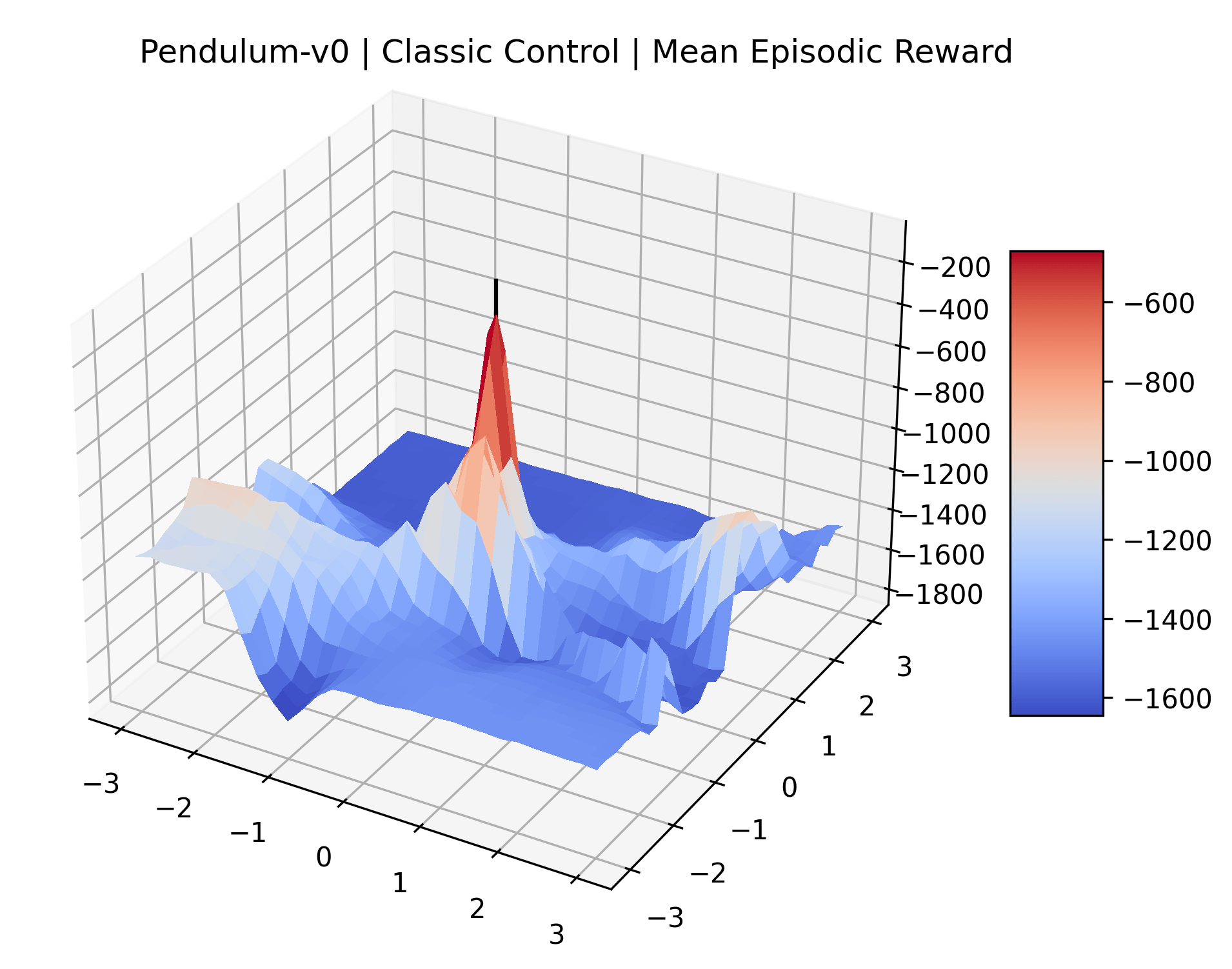} \\
\end{tabular}
\caption{Reward surfaces for 5 Classic Control environments.}
\label{fig:classiccontrol_rewardsurface_table}
\end{figure*}
\pagebreak

\subsection{MuJoCo}
\begin{figure*}[!ht]
\centering
\begin{tabular}{ccc}
 \includegraphics[width=\surfacescale]{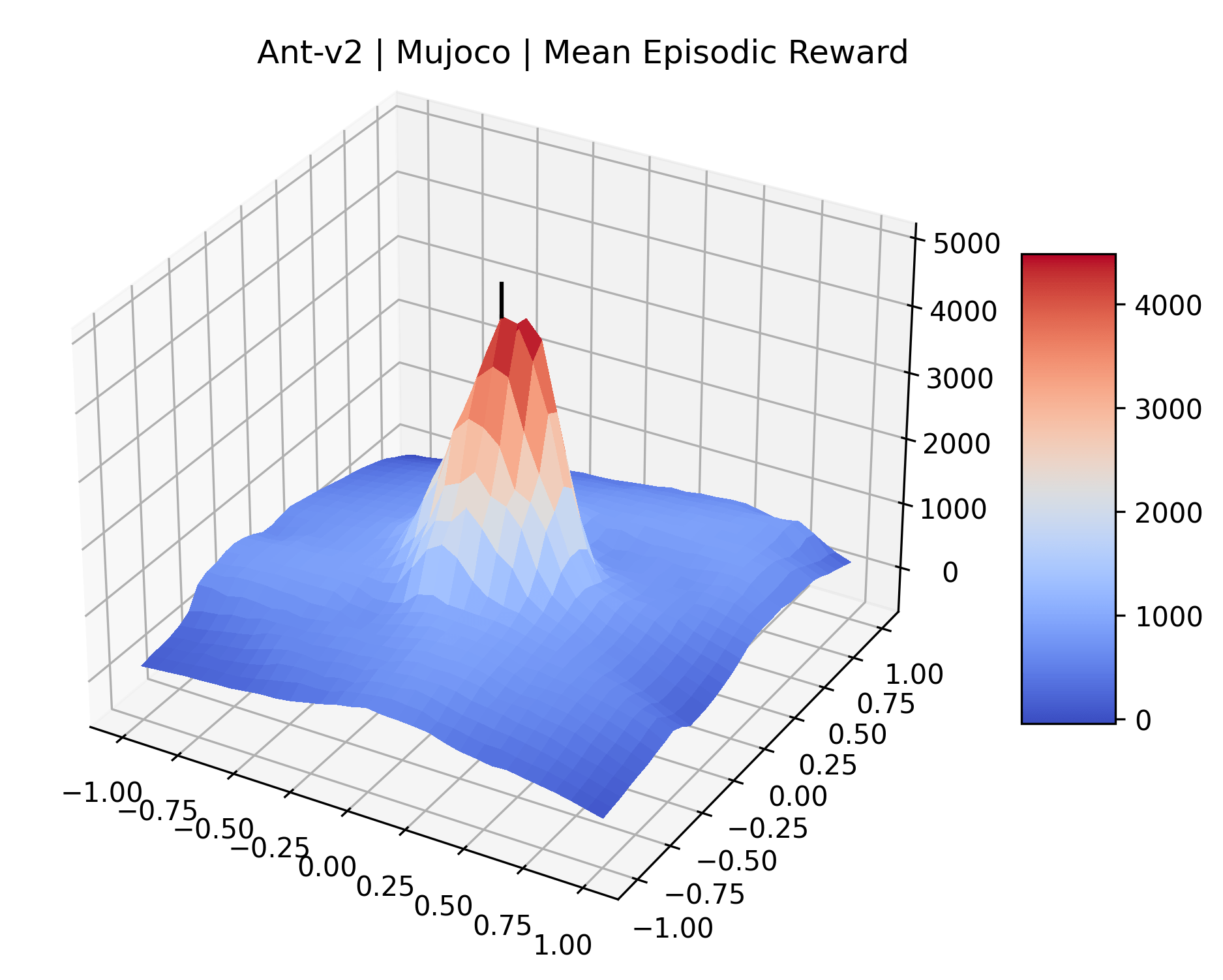} &
 \includegraphics[width=\surfacescale]{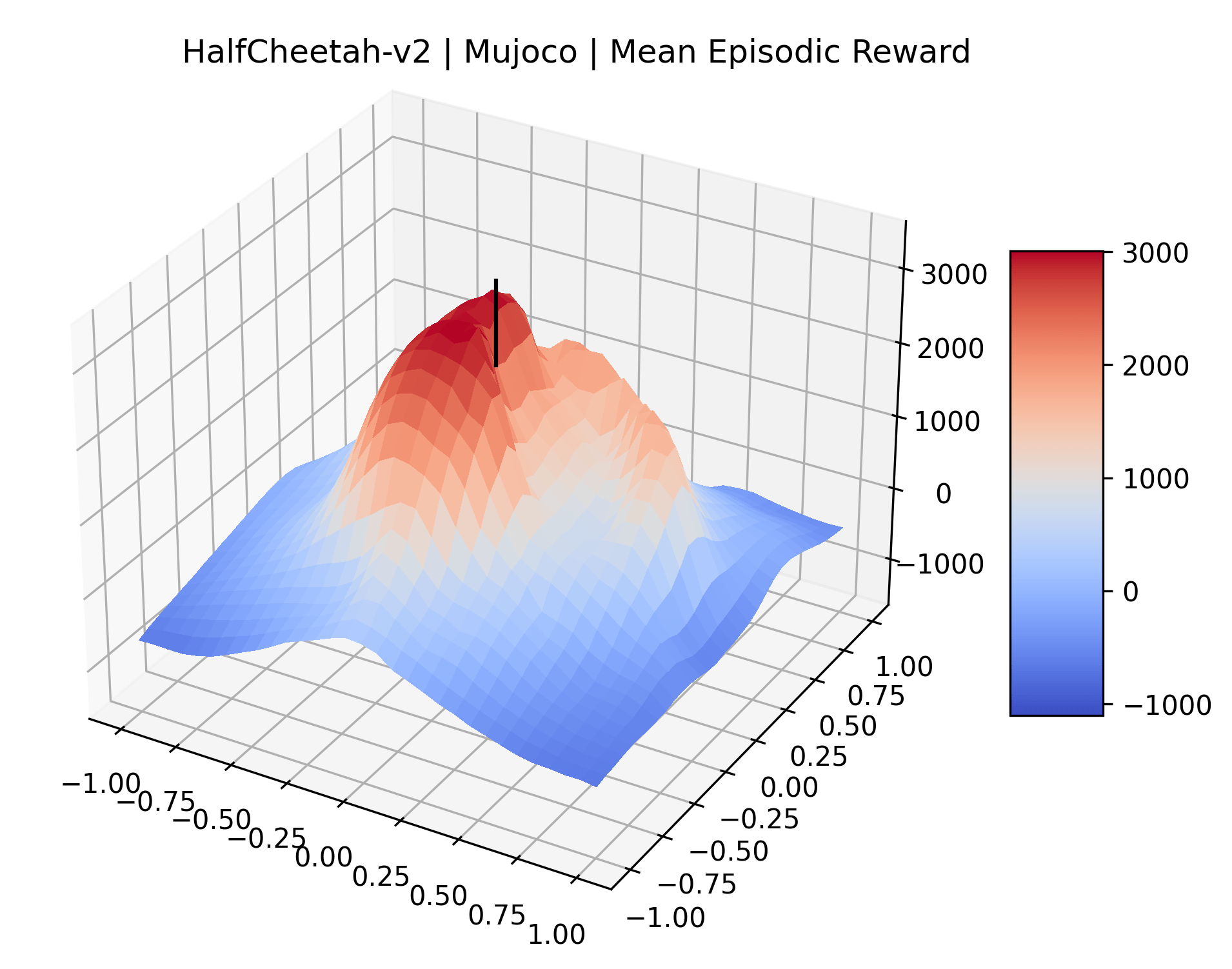} &
 \includegraphics[width=\surfacescale]{./plots/hopper_episoderewards_3dsurface.png} \\
 \includegraphics[width=\surfacescale]{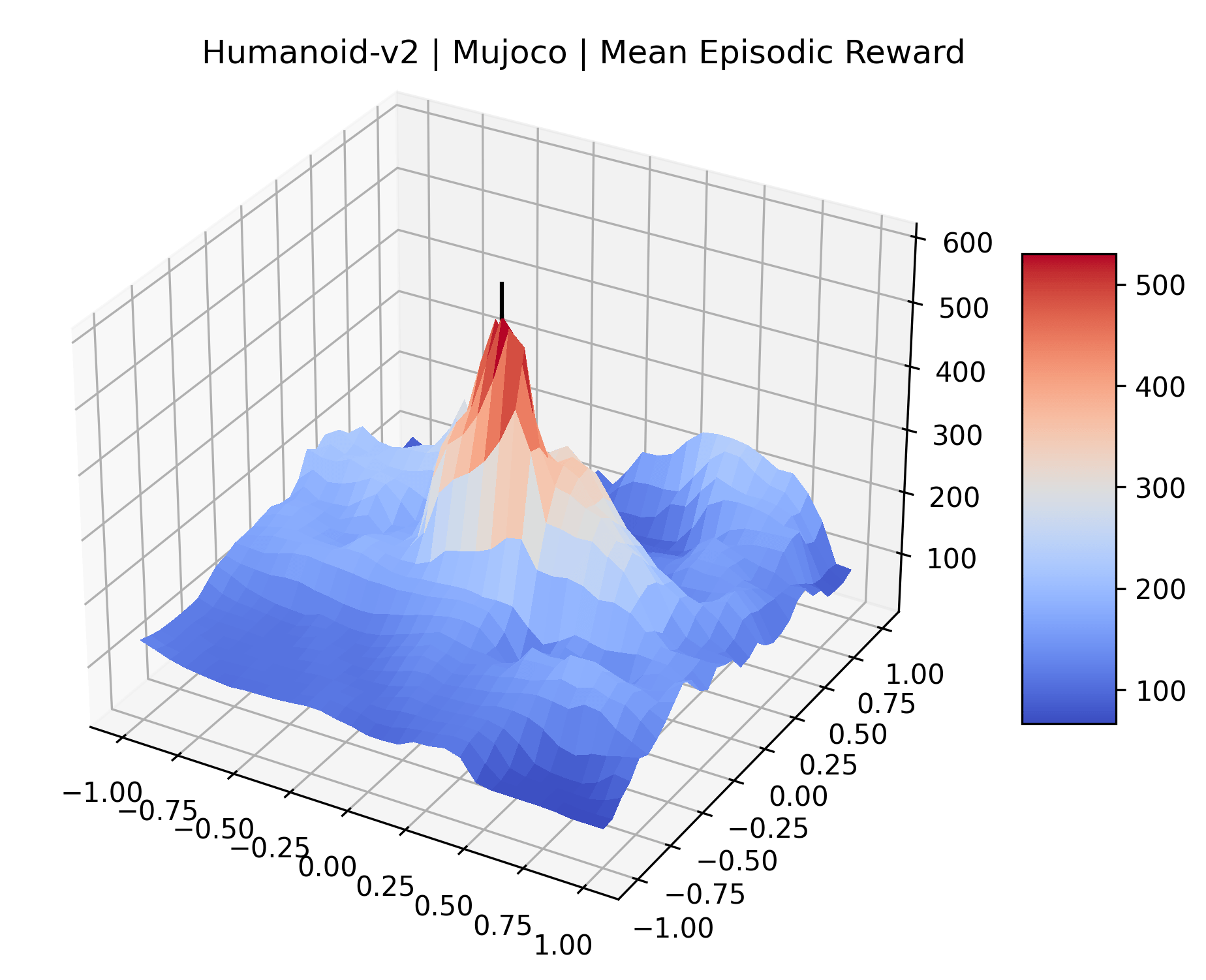} &
 \includegraphics[width=\surfacescale]{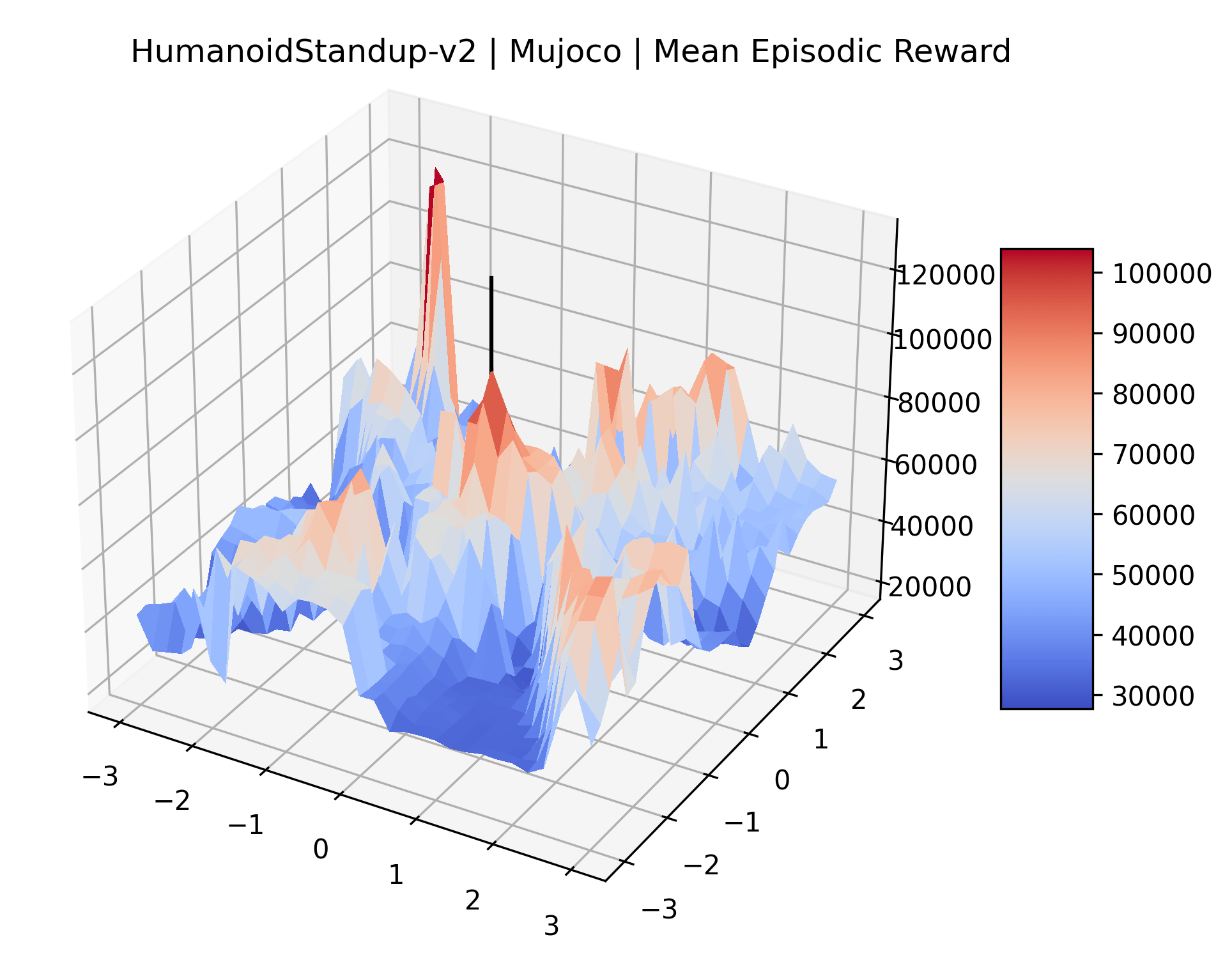} &
 \includegraphics[width=\surfacescale]{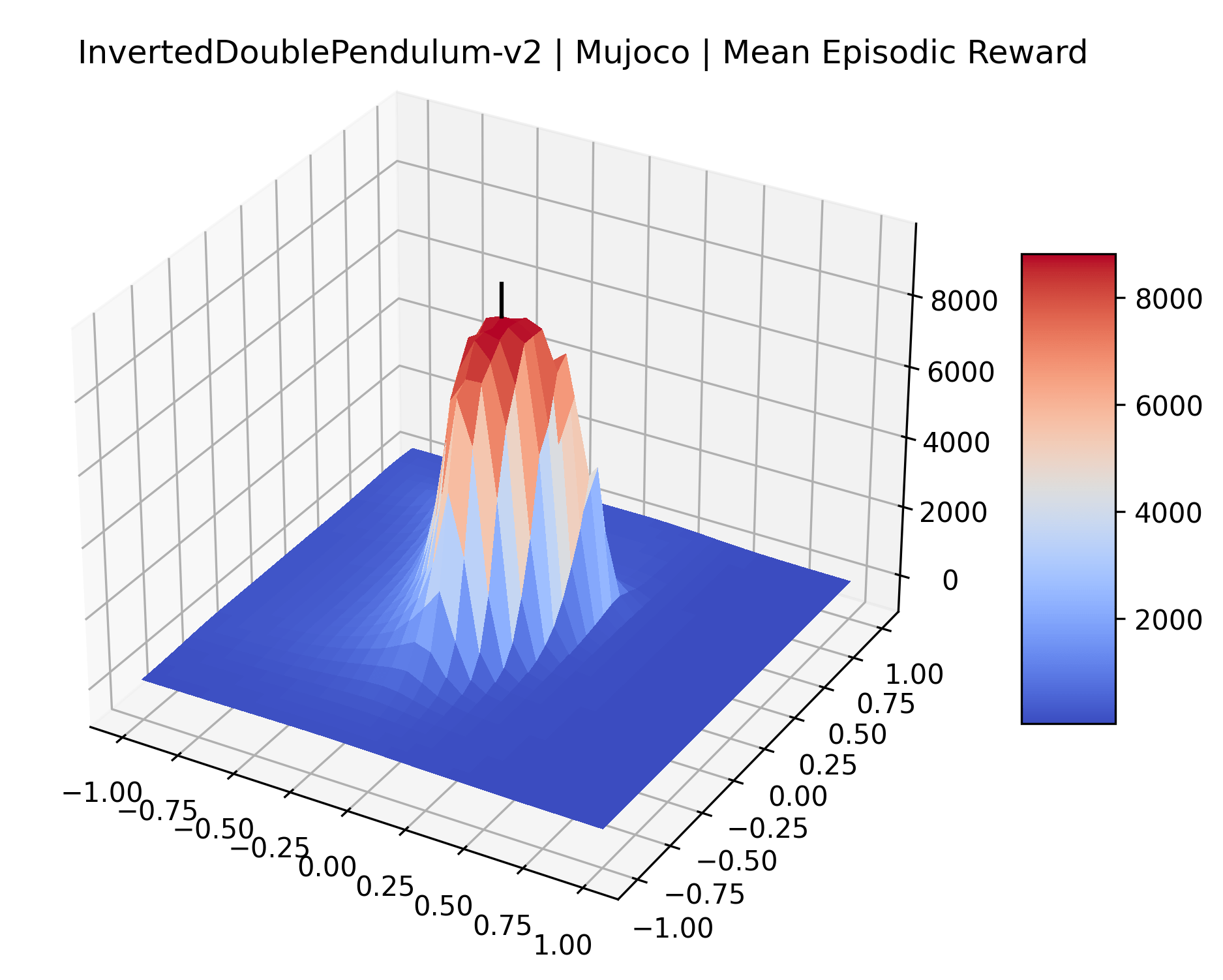} \\
 \includegraphics[width=\surfacescale]{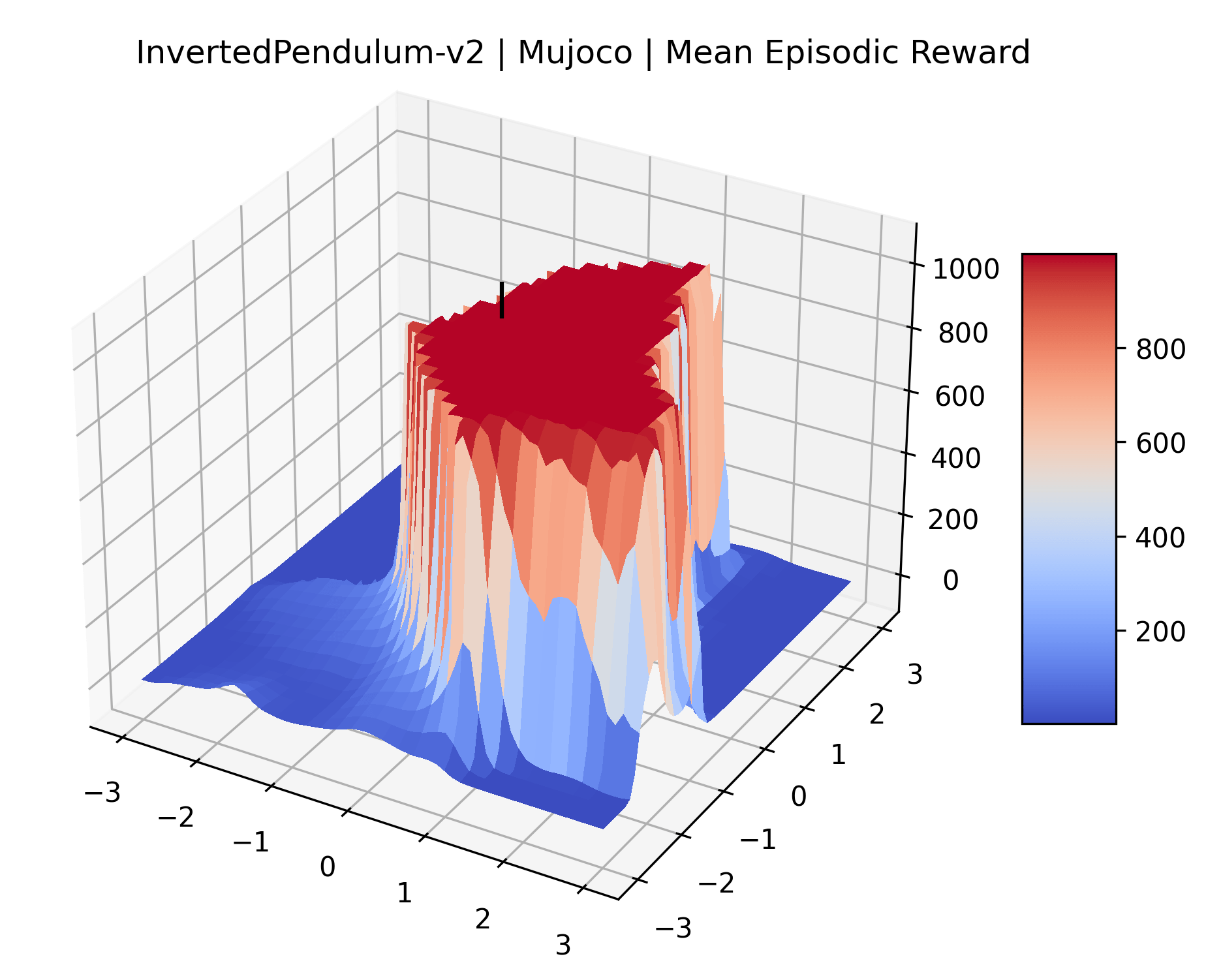} &
 \includegraphics[width=\surfacescale]{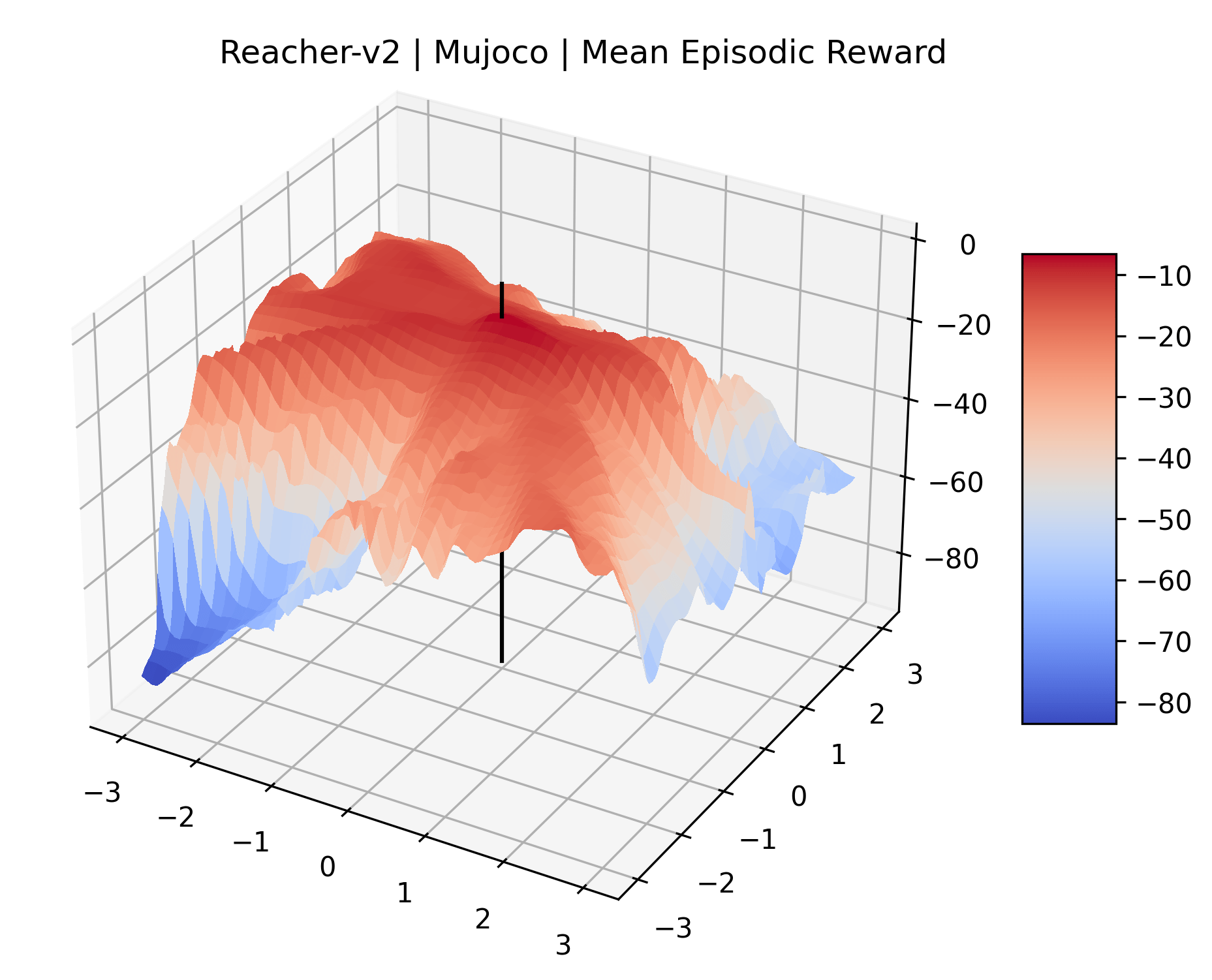} &
 \includegraphics[width=\surfacescale]{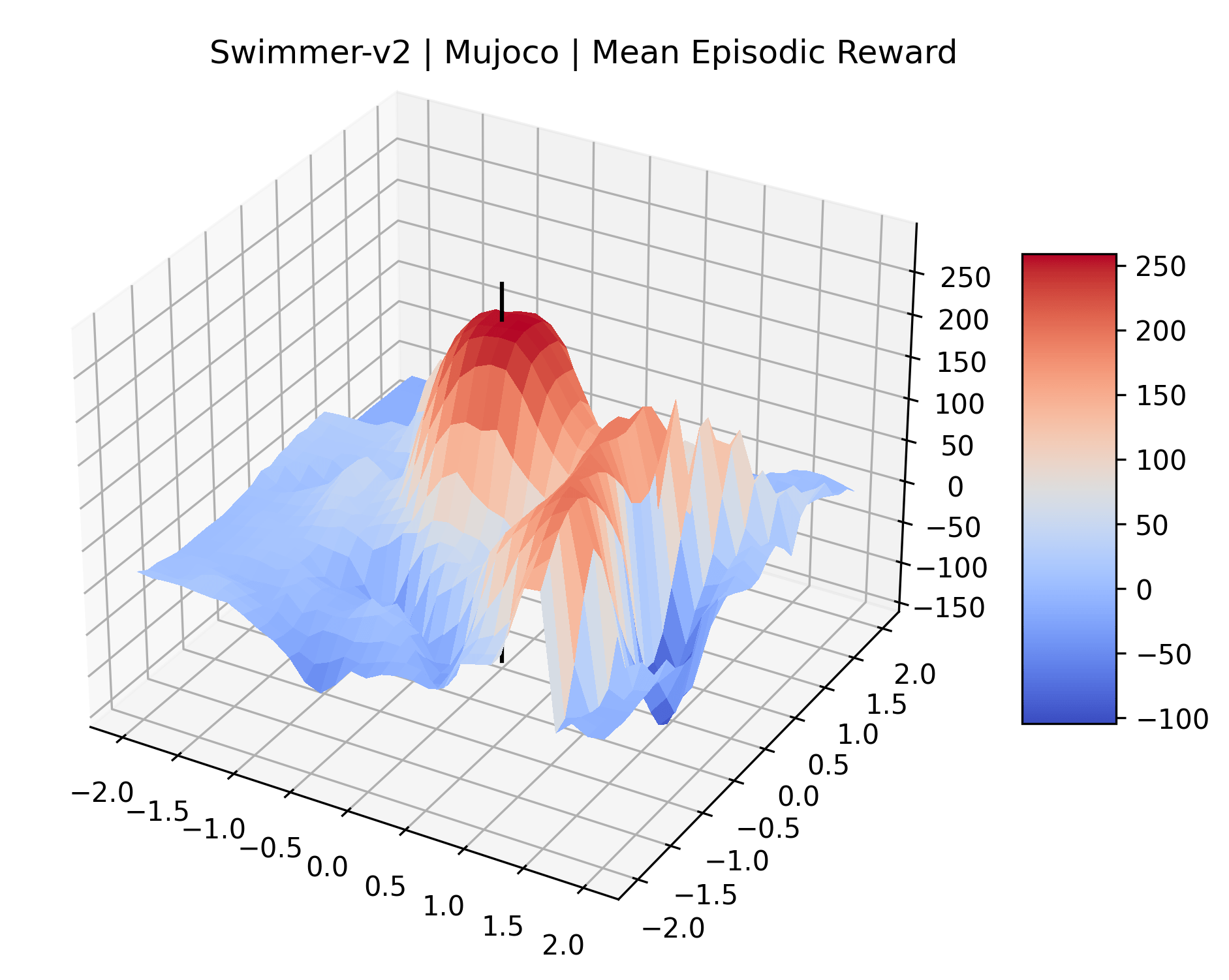} \\
 & \includegraphics[width=\surfacescale]{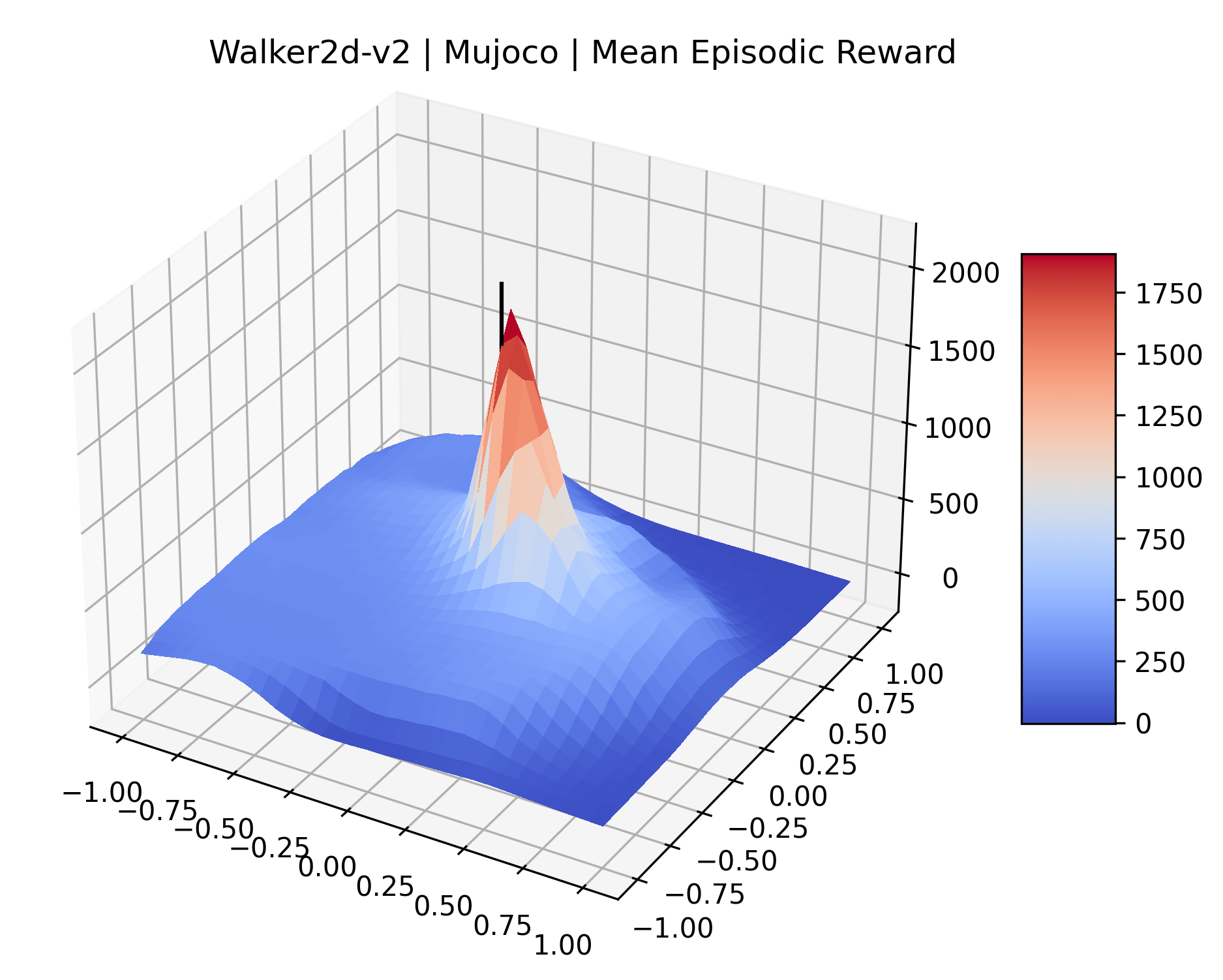} & \\
\end{tabular}
\caption{Reward surfaces for 10 MuJoCo environments.}
\label{fig:mujoco_rewardsurface_table}
\end{figure*}
\pagebreak

\subsection{Atari}
\begin{figure*}[!ht]
\centering
\begin{tabular}{ccc}
 \includegraphics[width=\surfacescale]{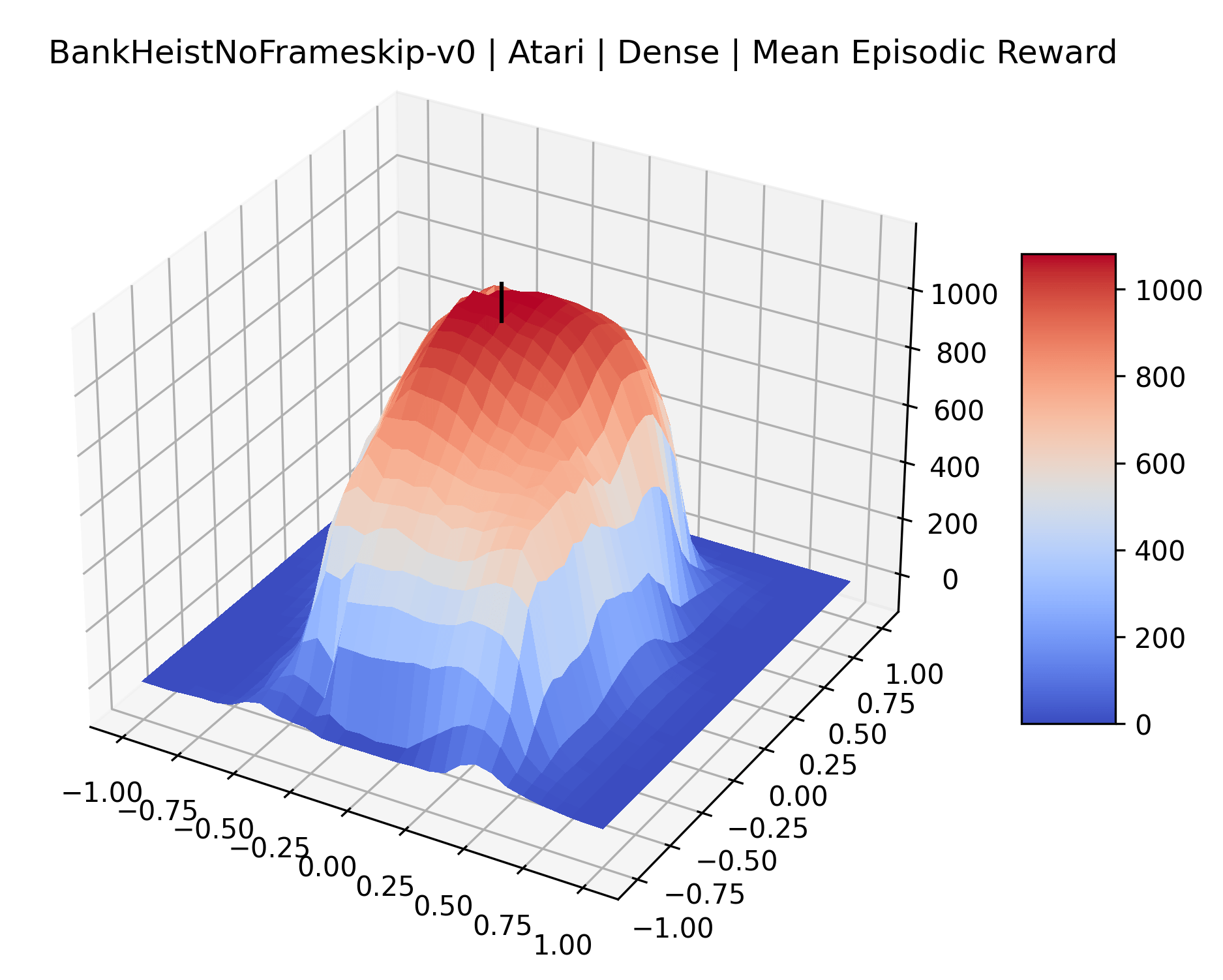} &
 \includegraphics[width=\surfacescale]{./plots/breakout_episoderewards_3dsurface.png} &
 \includegraphics[width=\surfacescale]{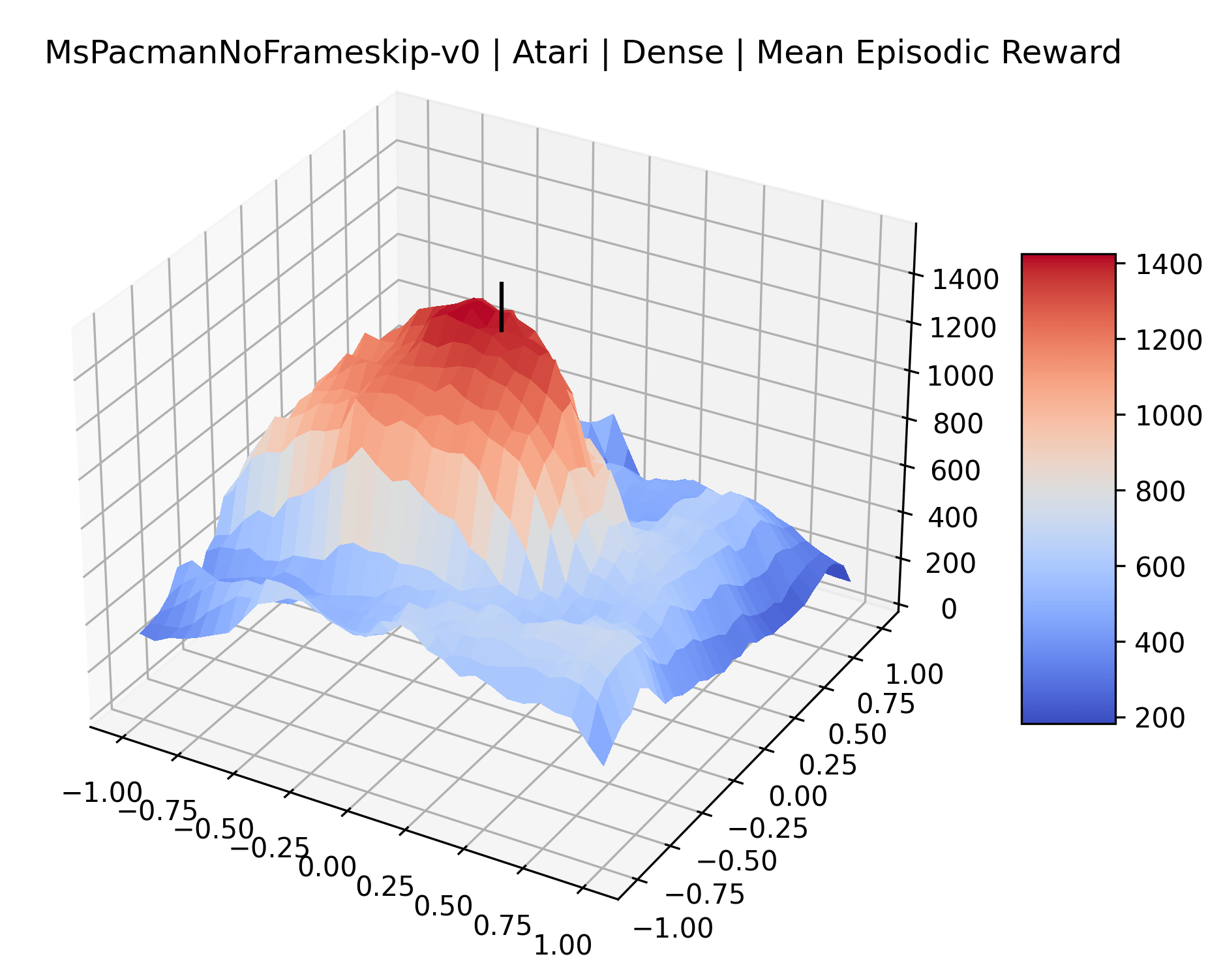} \\
 \includegraphics[width=\surfacescale]{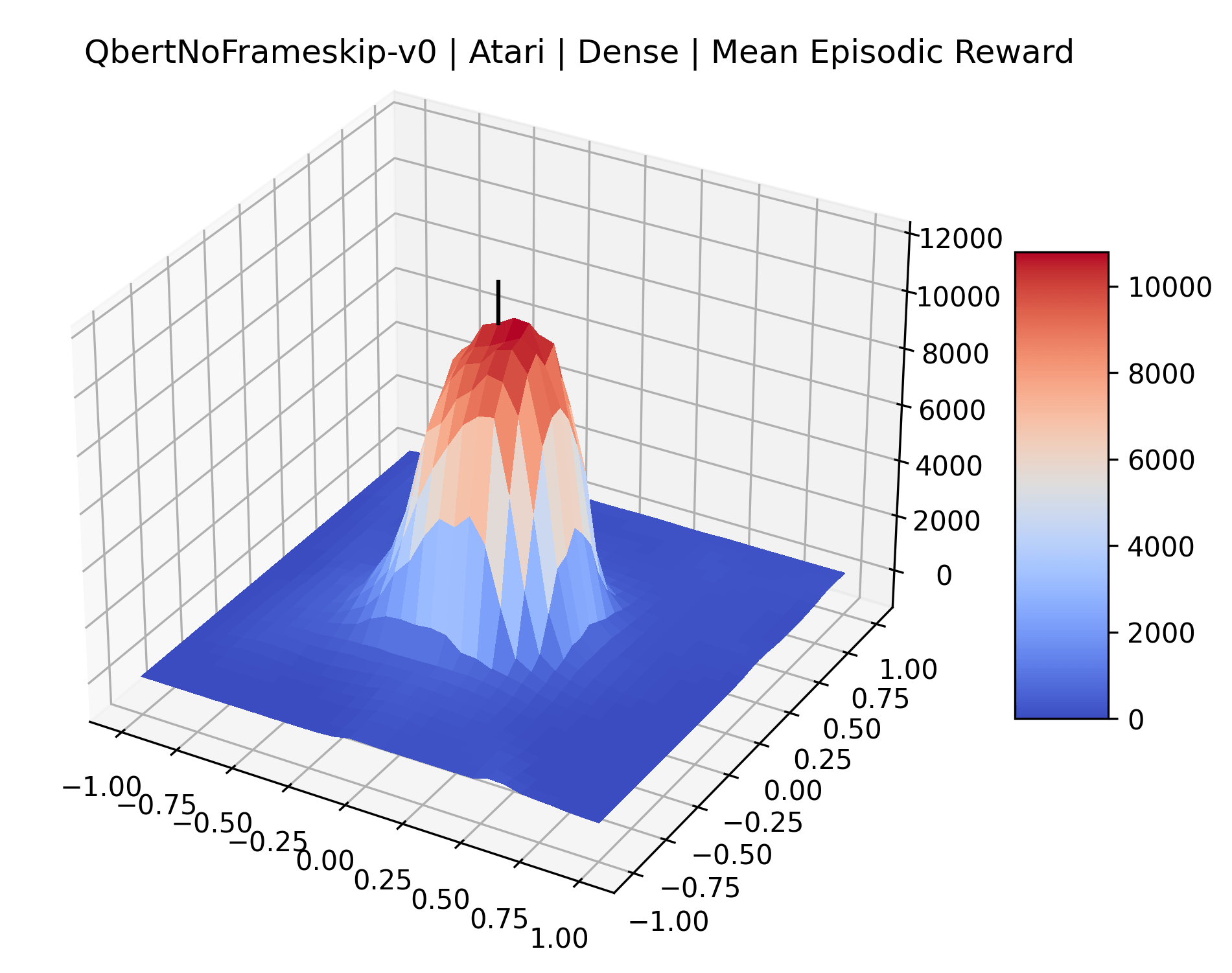} &
 \includegraphics[width=\surfacescale]{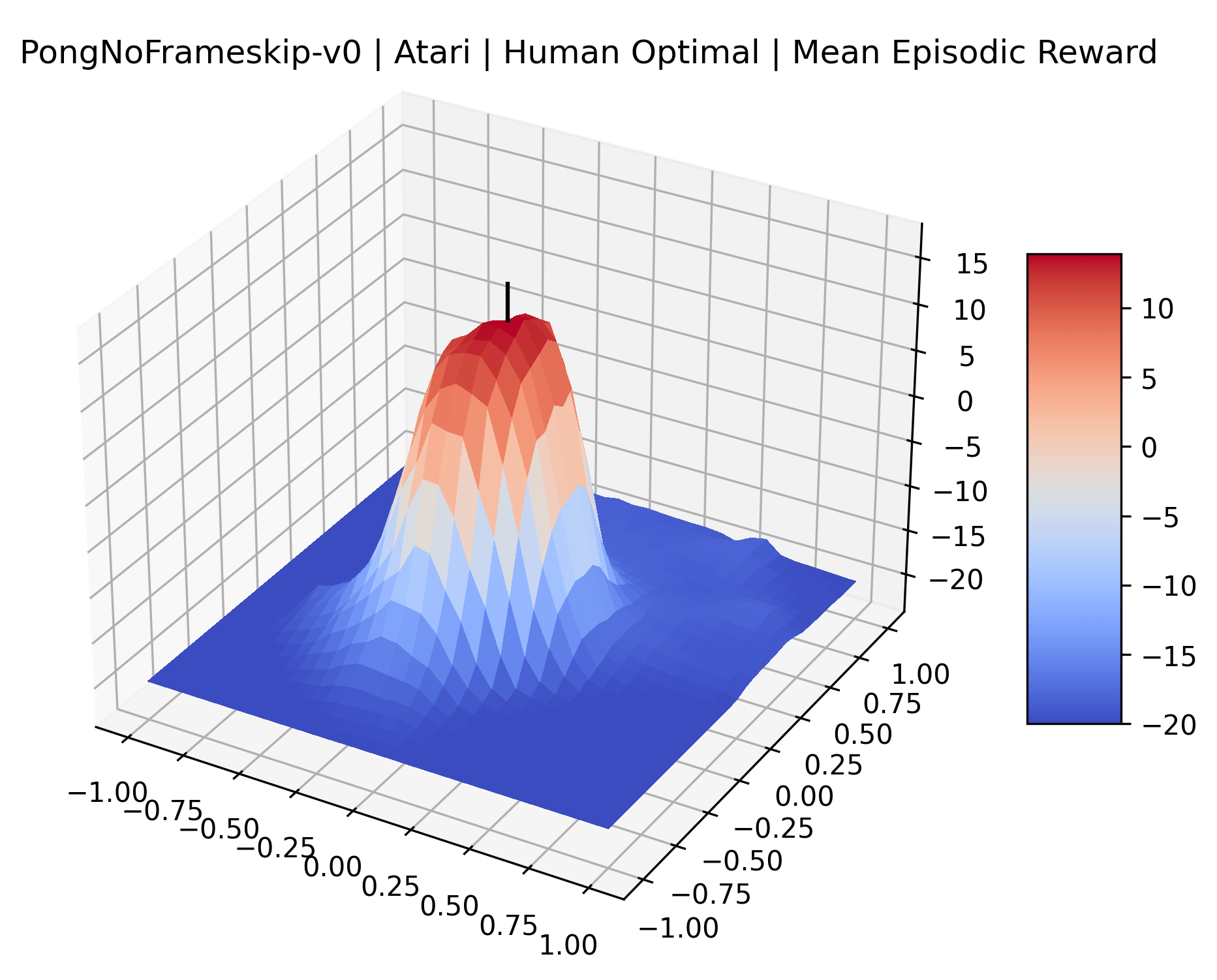} &
 \includegraphics[width=\surfacescale]{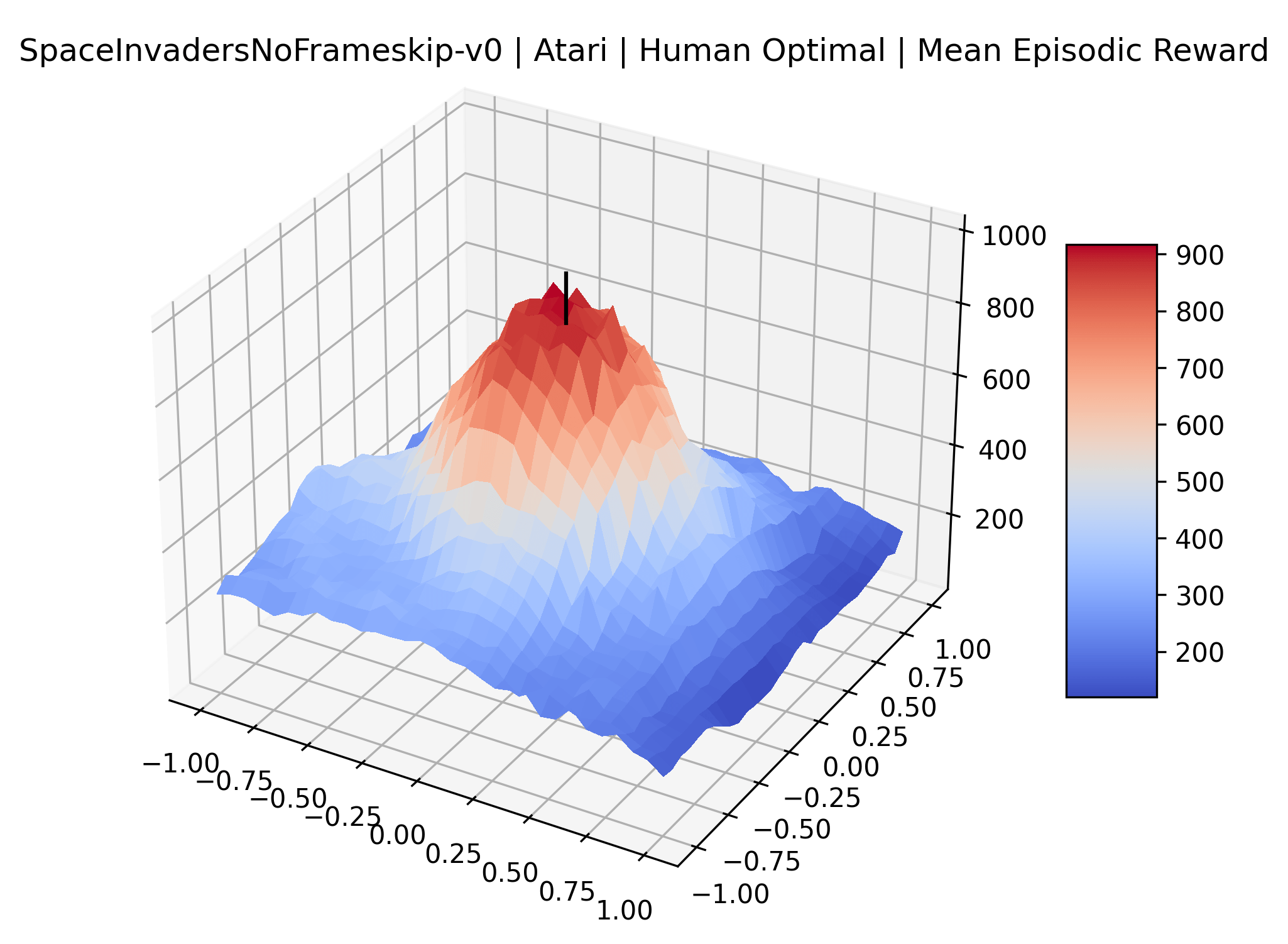} \\
 \includegraphics[width=\surfacescale]{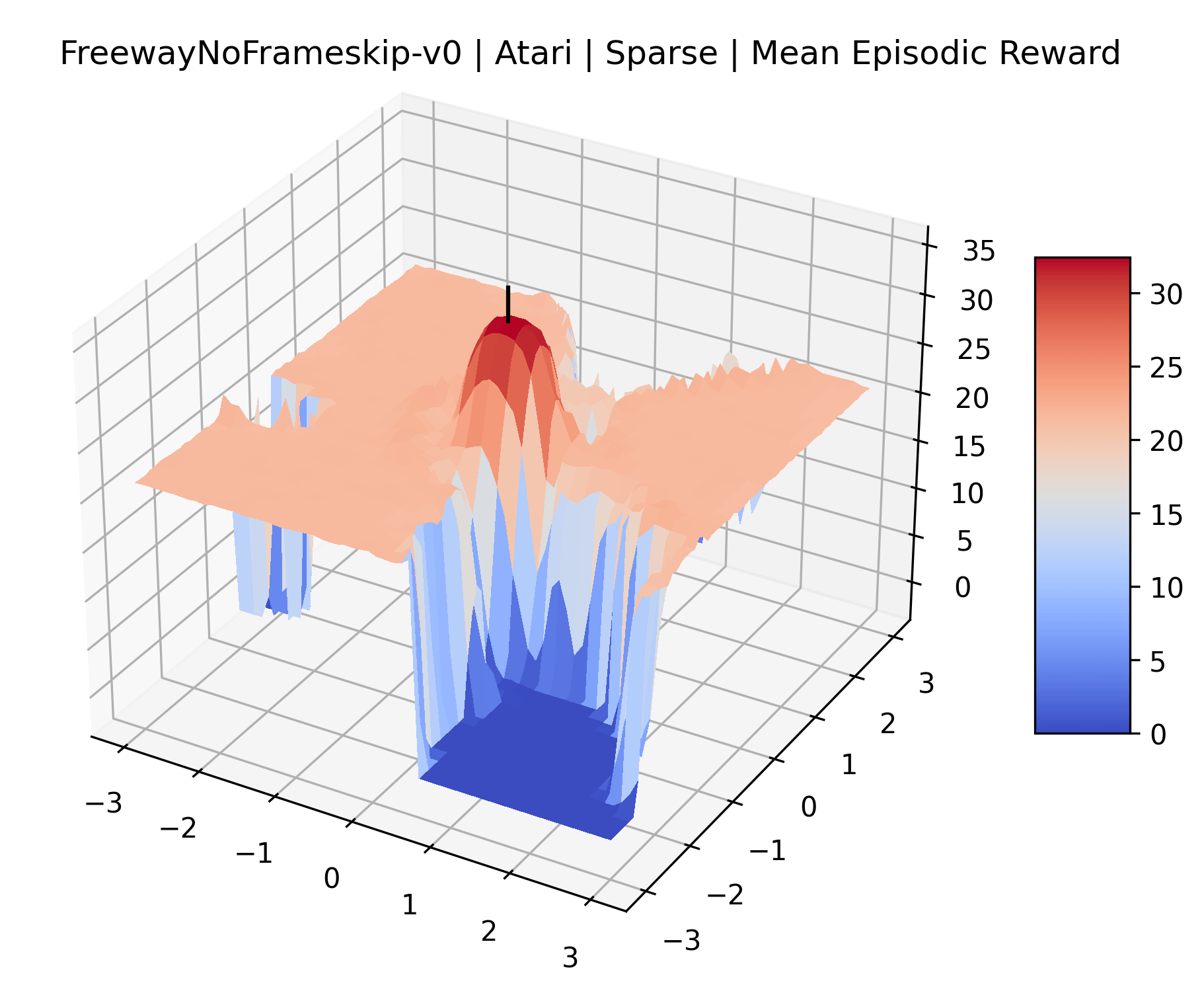} &
 \includegraphics[width=\surfacescale]{./plots/montezuma_episoderewards_3dsurface.png} &
 \includegraphics[width=\surfacescale]{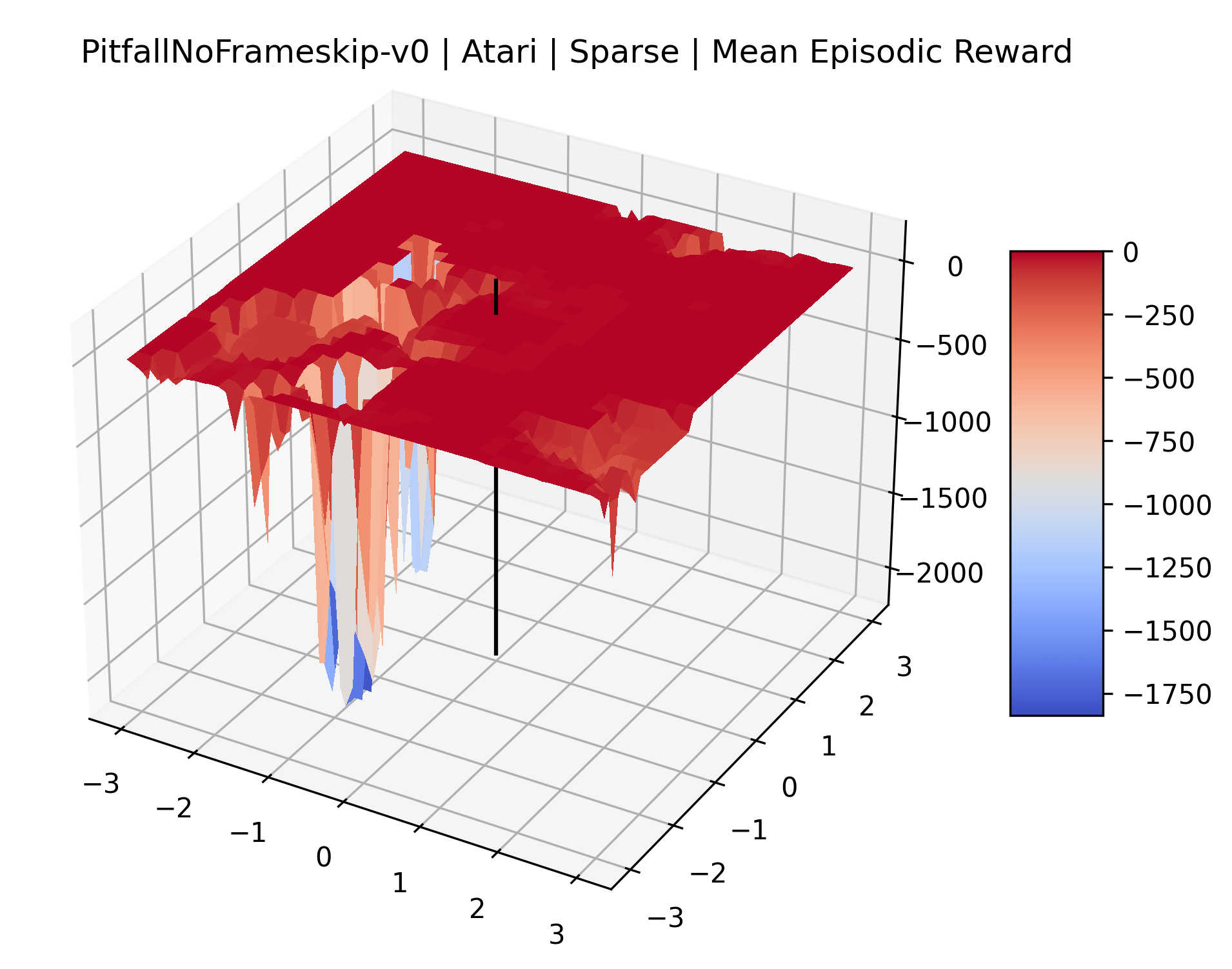} \\ \includegraphics[width=\surfacescale]{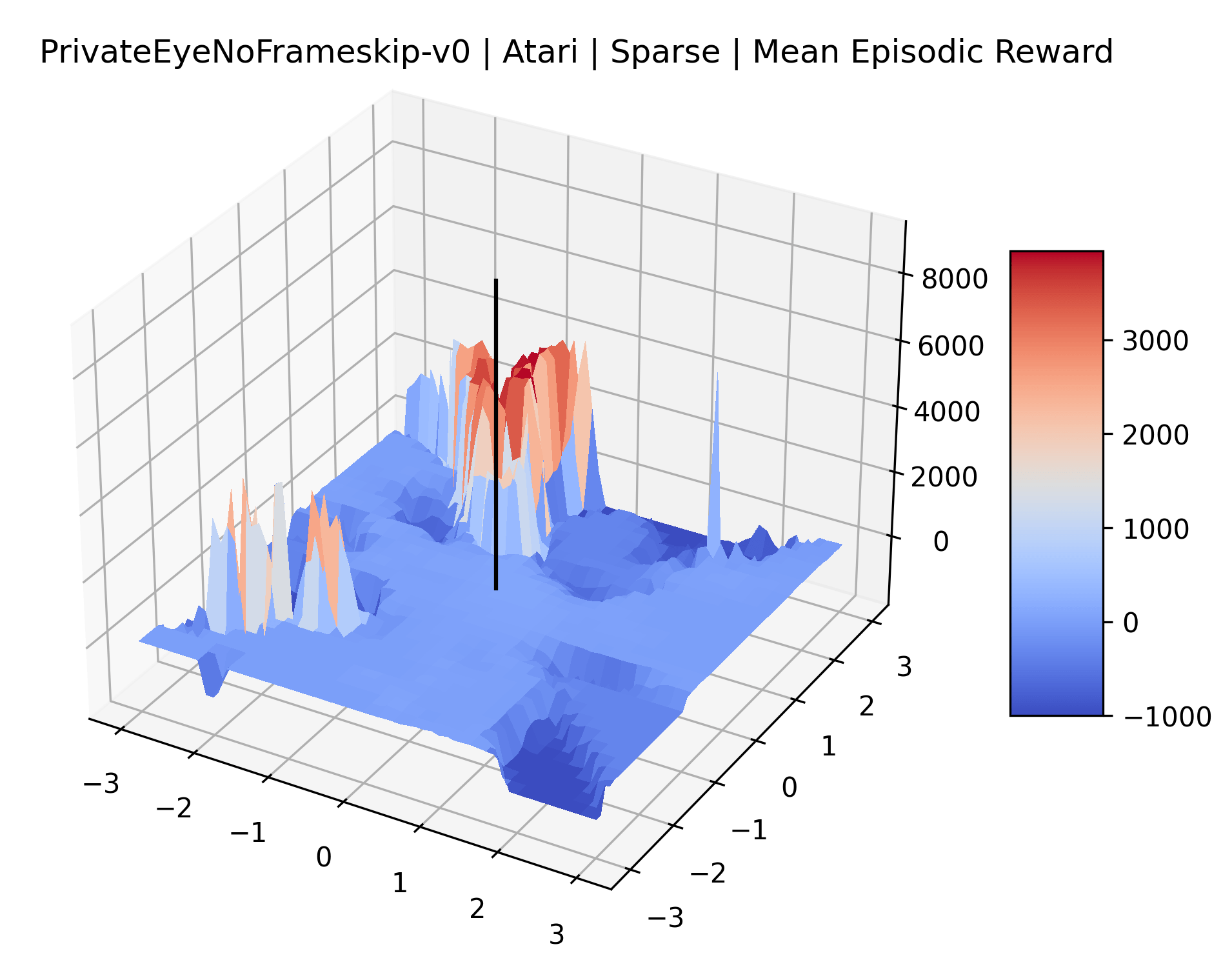} &
 \includegraphics[width=\surfacescale]{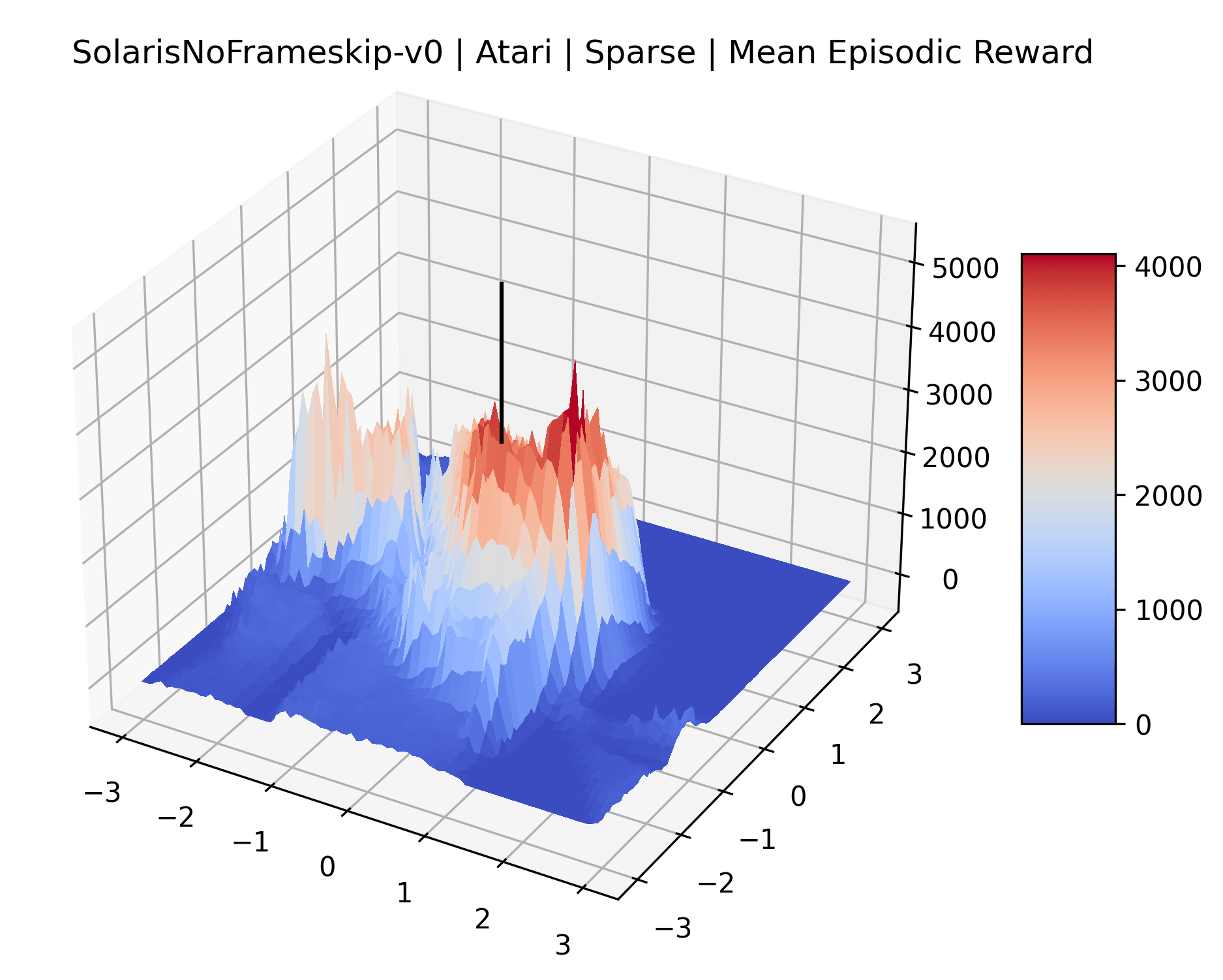} &
 \includegraphics[width=\surfacescale]{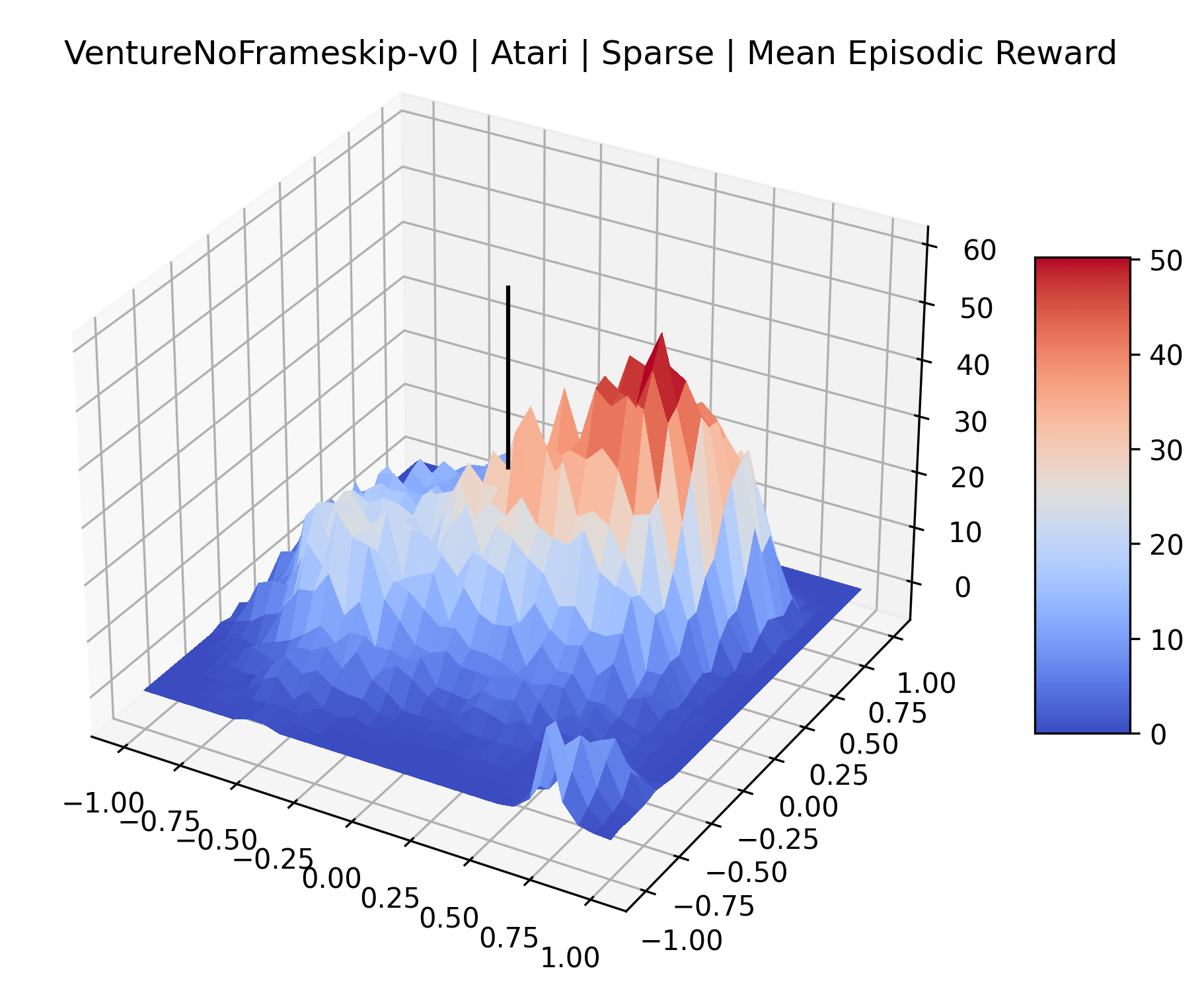} \\
\end{tabular}
\caption{Reward surfaces for 12 Atari environments.}
\label{fig:atari_rewardsurface_table}
\end{figure*}
\pagebreak

\subsection{Reward Surface Plot Options}
\begin{table*}[ht]
\centering
\begin{tabular}{|c|c|c|c|}
 \hline
 Environment & Grid Range & Grid Samples & Environment Steps \\
 \hline
 Acrobot & [-3, 3] & 31 x 31 & 200,000 \\
 CartPole & [-3, 3] & 31 x 31 & 200,000 \\
 MountainCar & [-3, 3] & 31 x 31 & 200,000 \\
 MountainCarContinuous & [-3, 3] & 31 x 31 & 200,000 \\
 Pendulum & [-3, 3] & 31 x 31 & 200,000 \\
 \hline
 Ant & [-1, 1] & 31 x 31 & 200,000 \\
 HalfCheetah & [-1, 1] & 31 x 31 & 200,000 \\
 Hopper & [-1, 1] & 31 x 31 & 200,000 \\
 Humanoid & [-1, 1] & 31 x 31 & 500,000 \\
 HumanoidStandup & [-3, 3] & 61 x 61 & 500,000 \\
 InvertedDoublePendulum & [-1, 1] & 31 x 31 & 200,000 \\
 InvertedPendulum & [-3, 3] & 61 x 61 & 200,000 \\
 Reacher & [-3, 3] & 91 x 91 & 200,000 \\
 Swimmer & [-2, 2] & 31 x 31 & 500,000 \\
 Walker2d & [-1, 1] & 31 x 31 & 200,000 \\
 \hline
 Breakout & [-1, 1] & 31 x 31 & 200,000 \\
 Pong & [-1, 1] & 31 x 31 & 200,000 \\
 SpaceInvaders & [-1, 1] & 31 x 31 & 200,000 \\
 BankHeist & [-1, 1] & 31 x 31 & 200,000 \\
 MsPacman & [-1, 1] & 31 x 31 & 200,000 \\
 Q*bert & [-1, 1] & 31 x 31 & 200,000 \\
 Freeway & [-3, 3] & 61 x 61 & 1,000,000 \\
 Montezuma's Revenge & [-1, 1] & 31 x 31 & 200,000 \\
 Pitfall! & [-3, 3] & 61 x 61 & 1,000,000 \\
 Private Eye & [-3, 3] & 61 x 61 & 1,000,000 \\
 Solaris & [-3, 3] & 91 x 91 & 2,000,000 \\
 Venture & [-1, 1] & 31 x 31 & 200,000 \\
 \hline

\end{tabular}
\caption{Settings used to generate reward surfaces for each environment. These settings were manually chosen to highlight interesting features of the reward surface and to reduce standard error in the plots.}
\label{fig:options_table}
\end{table*}
\pagebreak

\section{Standard Deviation Plots}
\label{appendix:standard_deviation}

\subsection{Classic Control}
\begin{figure*}[!ht]
\centering
\begin{tabular}{ccc}
 \includegraphics[width=\surfacescale]{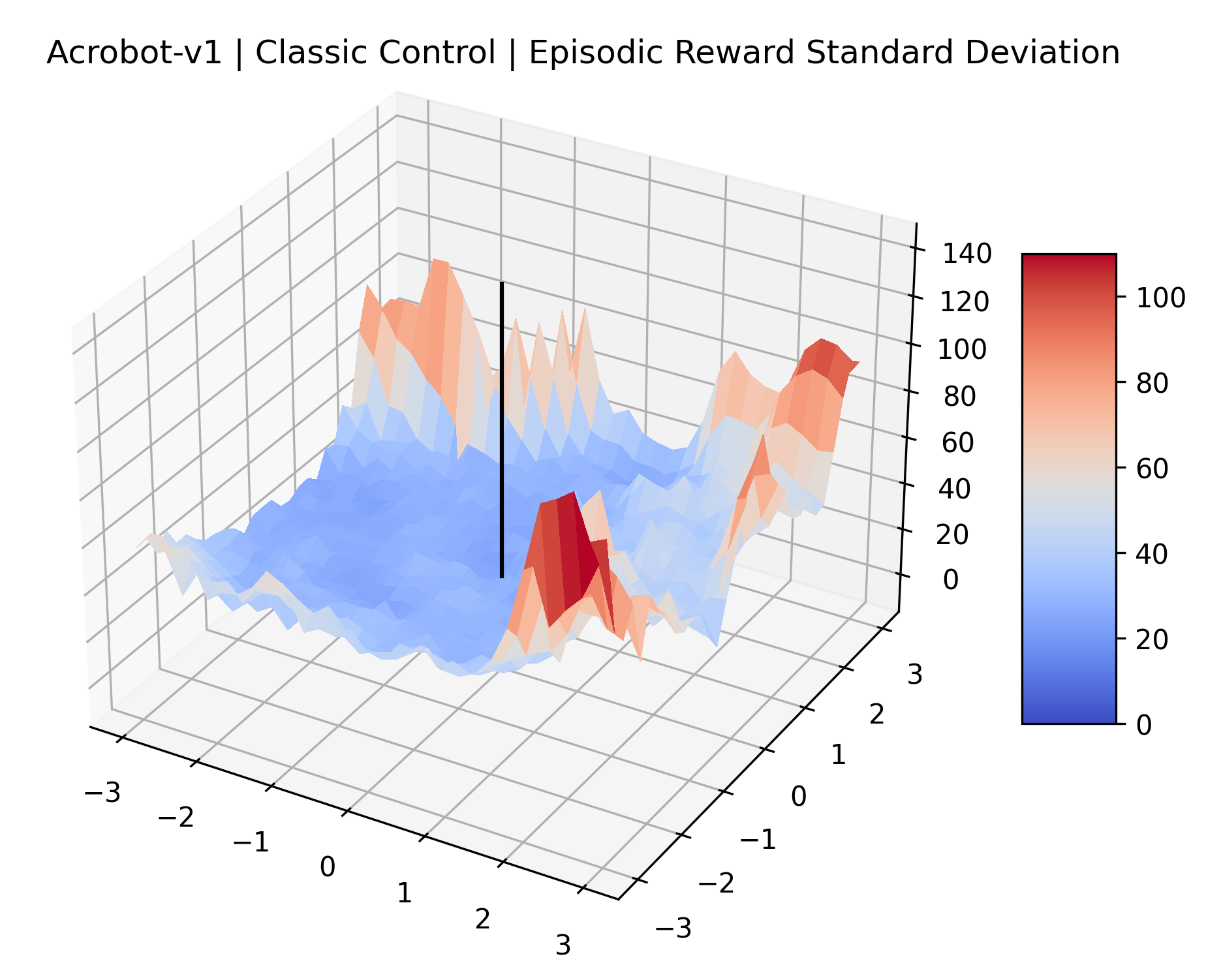} &
 \includegraphics[width=\surfacescale]{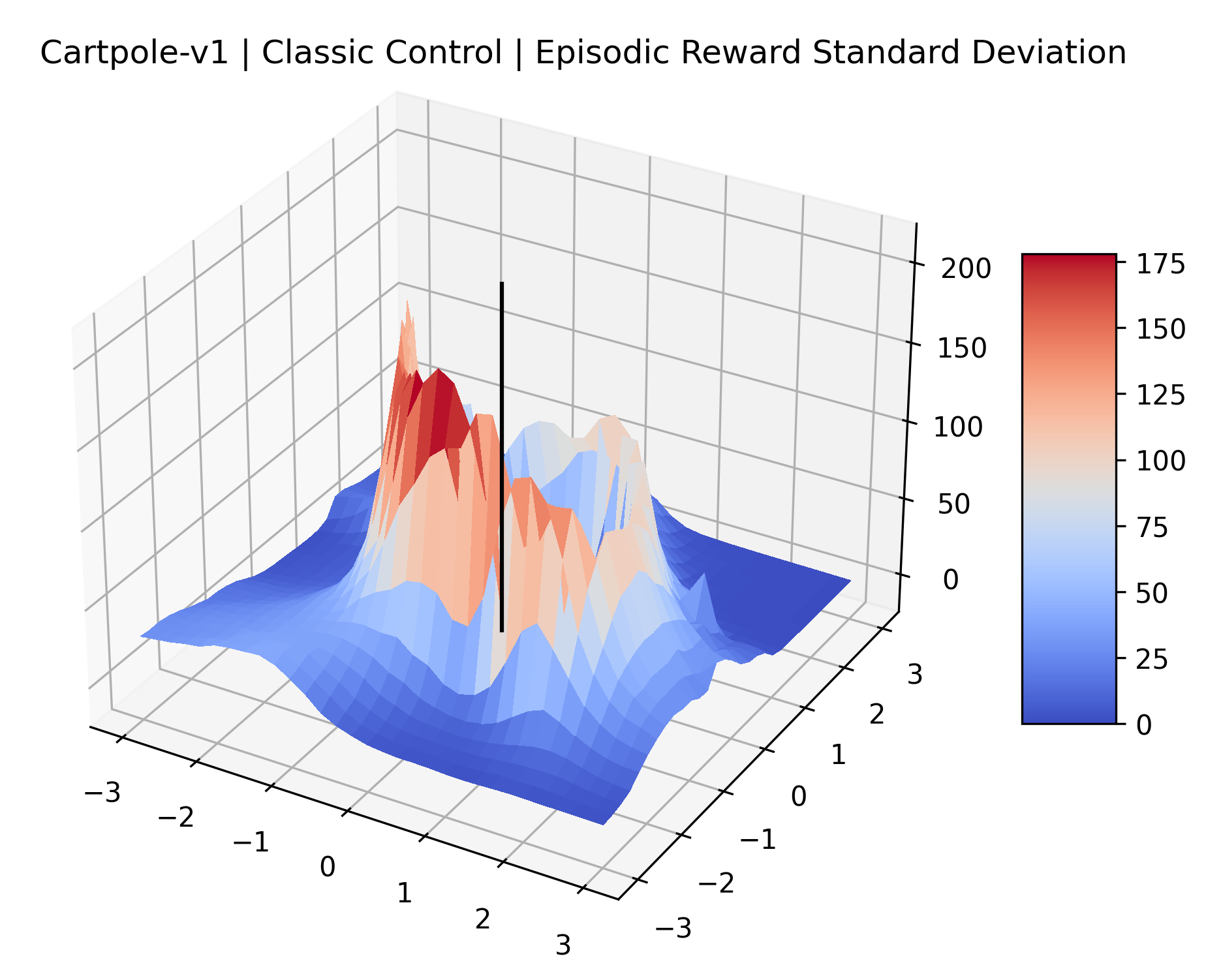} &
 \includegraphics[width=\surfacescale]{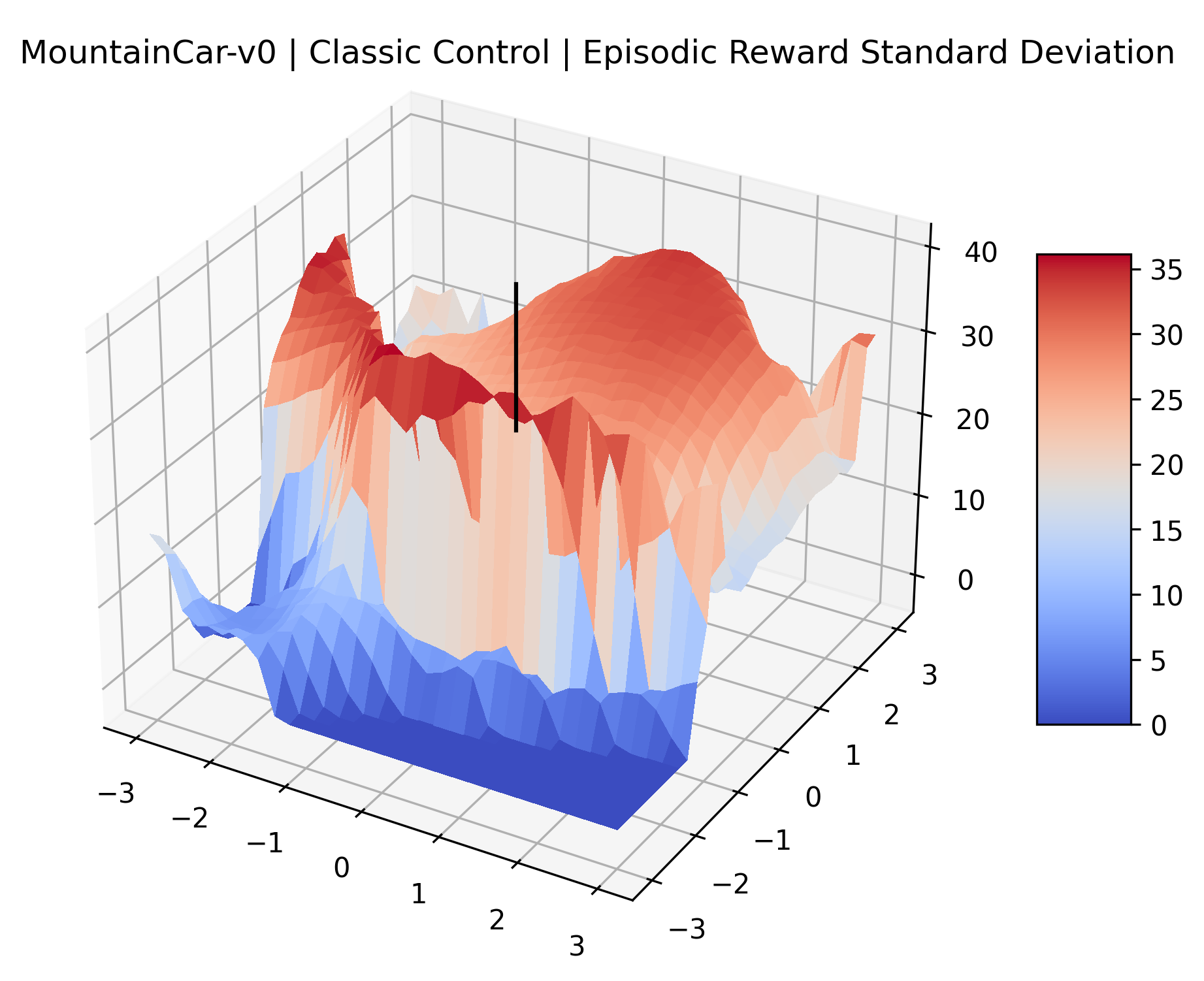} \\
\end{tabular}
\begin{tabular}{cc}
 \includegraphics[width=\surfacescale]{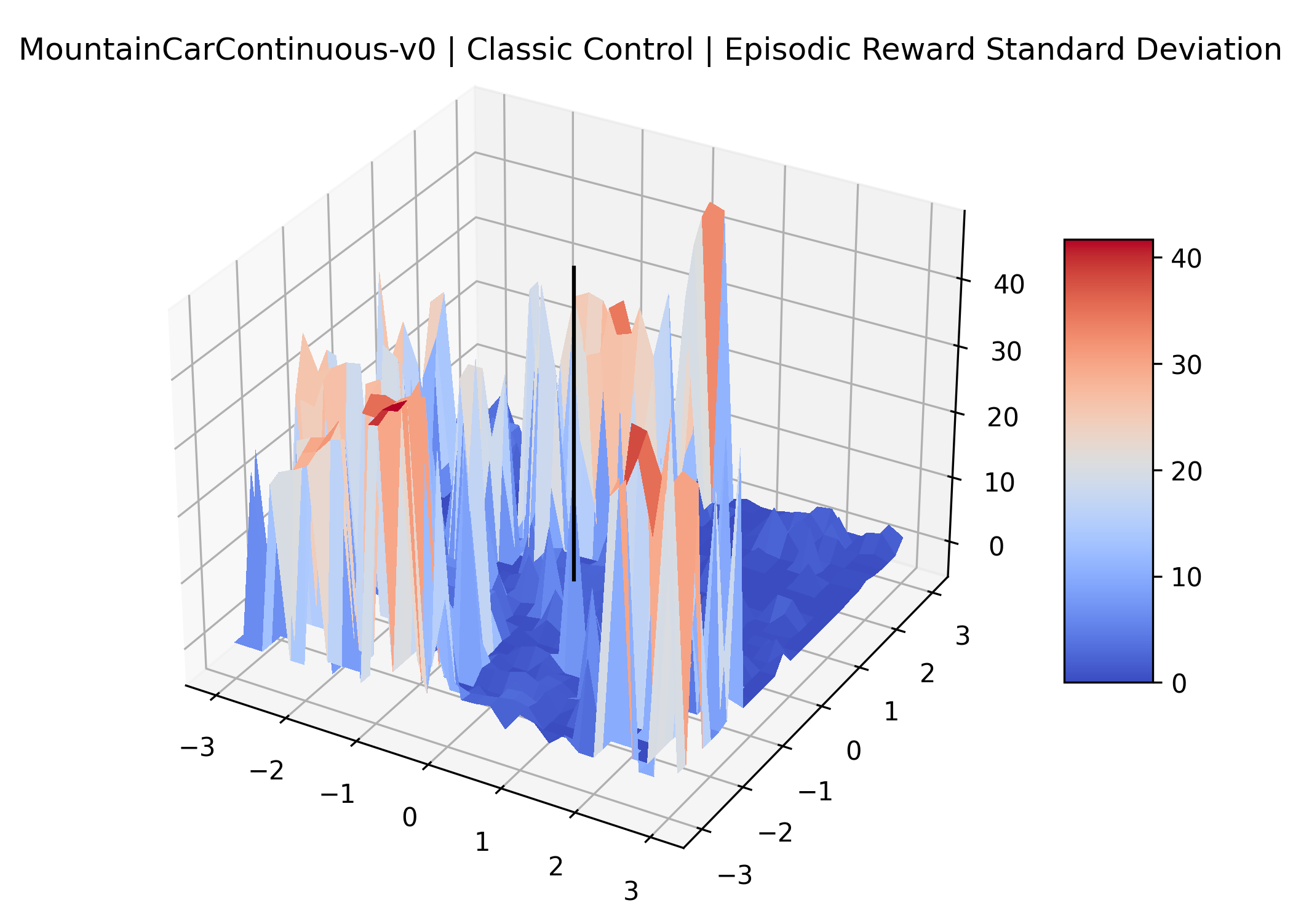} &
 \includegraphics[width=\surfacescale]{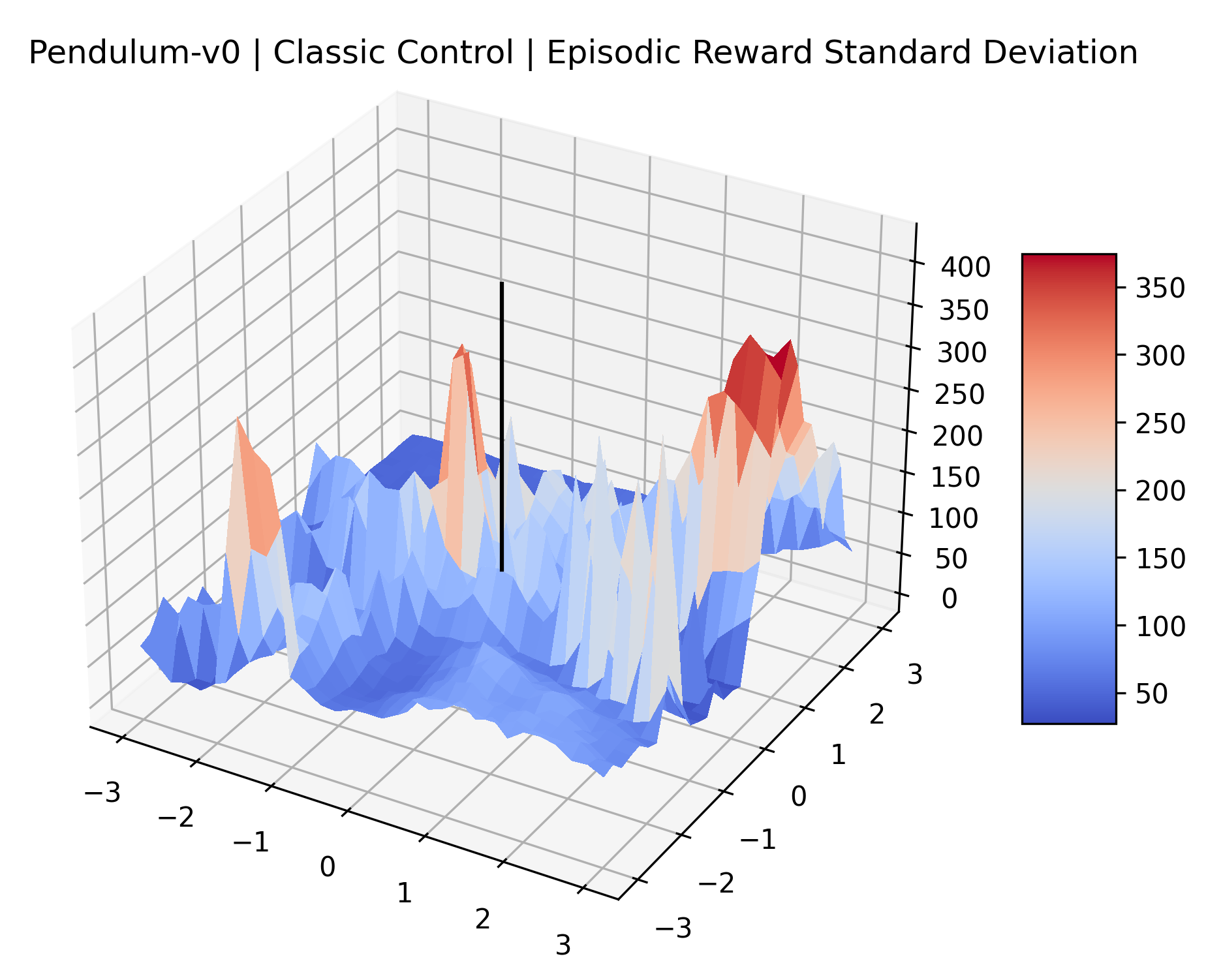} \\
\end{tabular}
\caption{Standard deviation surfaces for 5 Classic Control environments.}
\label{fig:classiccontrol_standarddeviation_table}
\end{figure*}
\pagebreak

\subsection{MuJoCo}
\begin{figure*}[!ht]
\centering
\begin{tabular}{ccc}
 \includegraphics[width=\surfacescale]{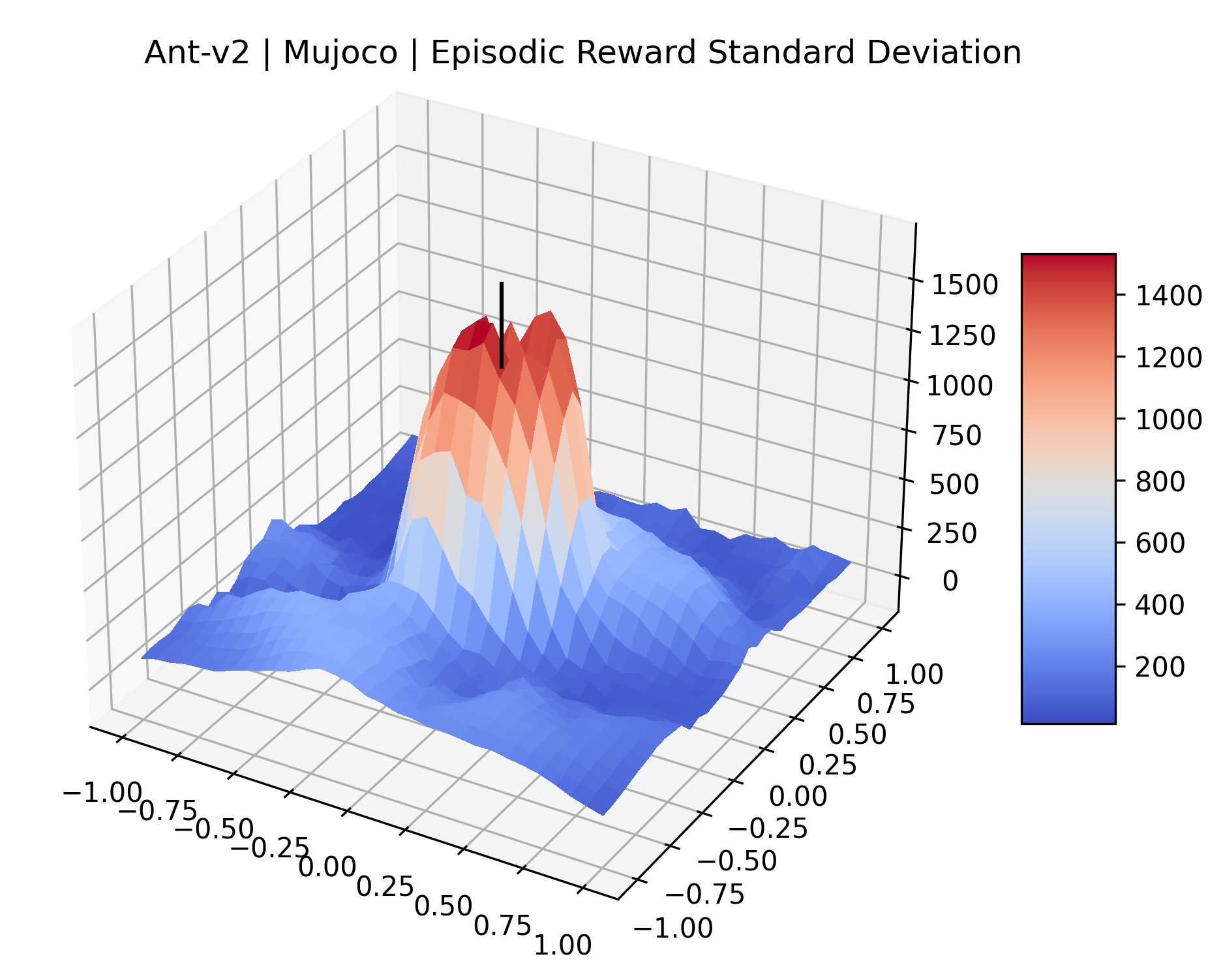} &
 \includegraphics[width=\surfacescale]{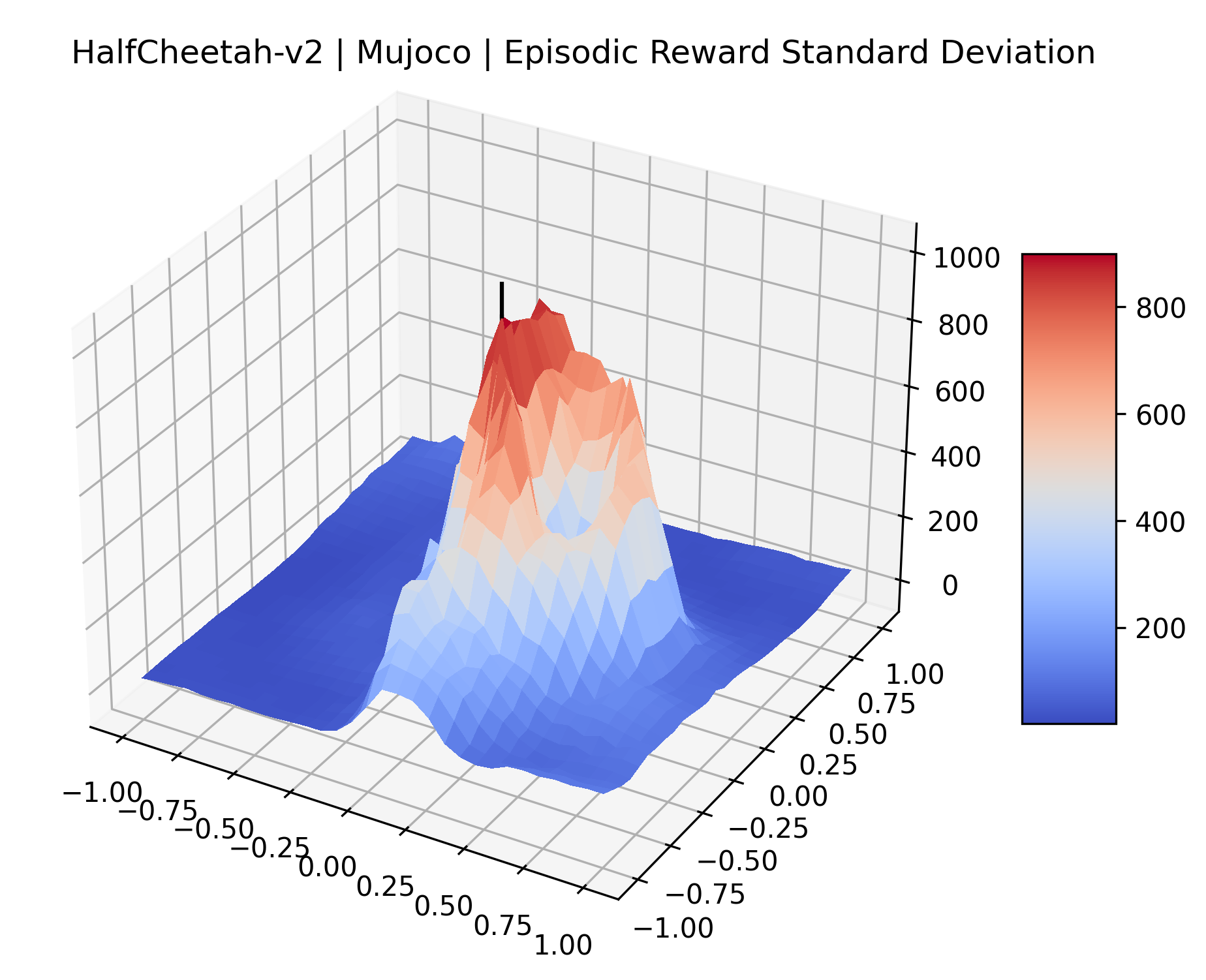} &
 \includegraphics[width=\surfacescale]{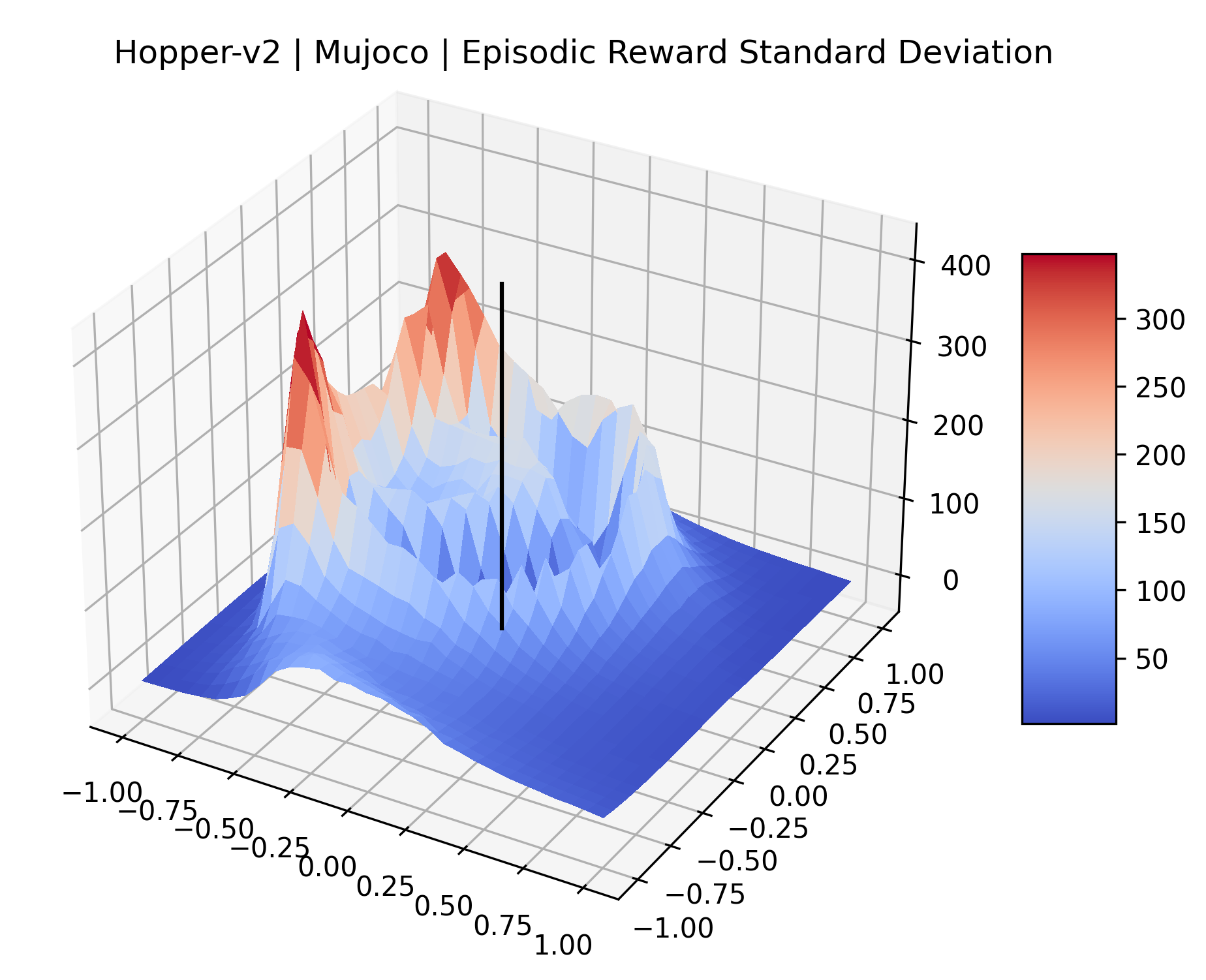} \\
 \includegraphics[width=\surfacescale]{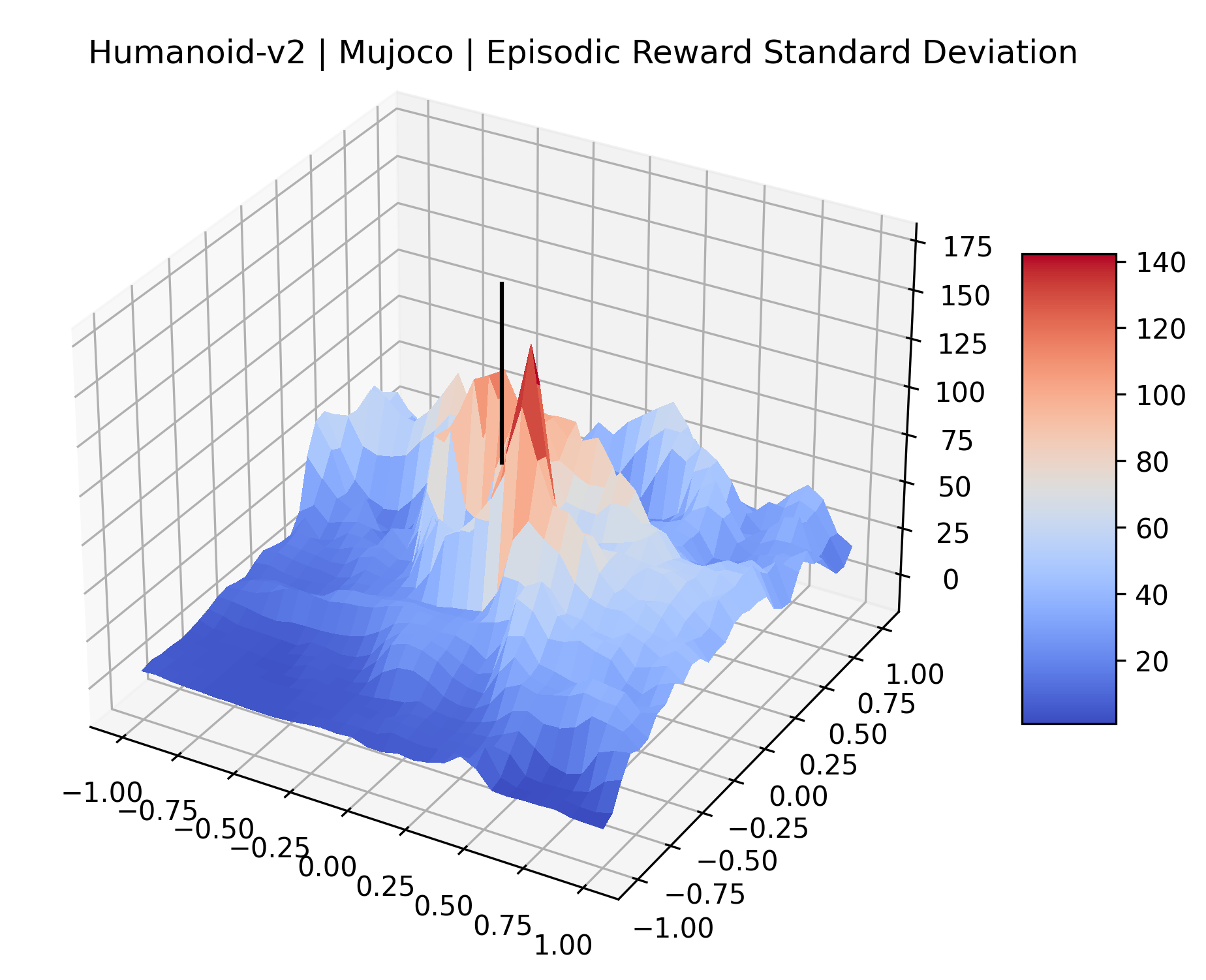} &
 \includegraphics[width=\surfacescale]{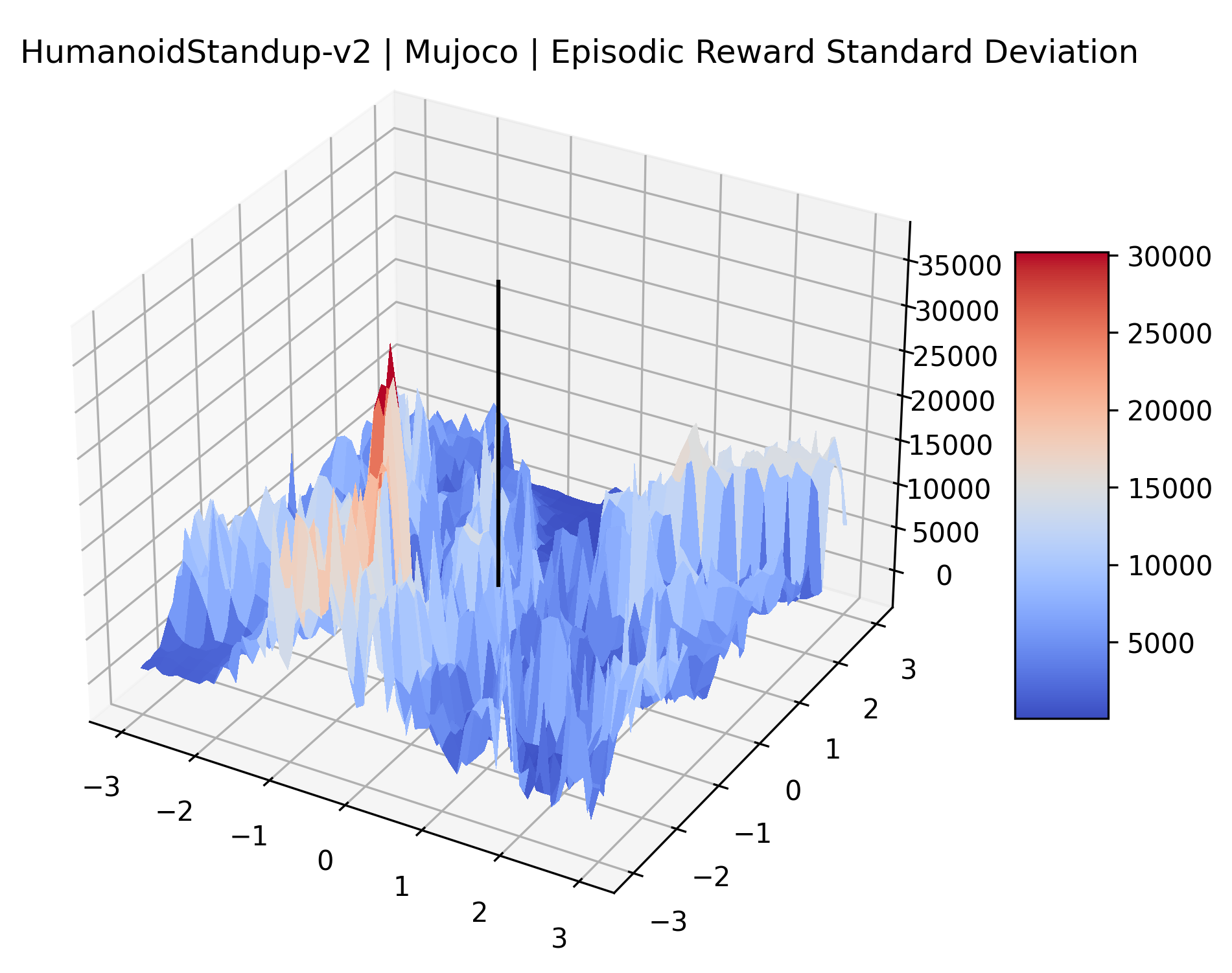} &
 \includegraphics[width=\surfacescale]{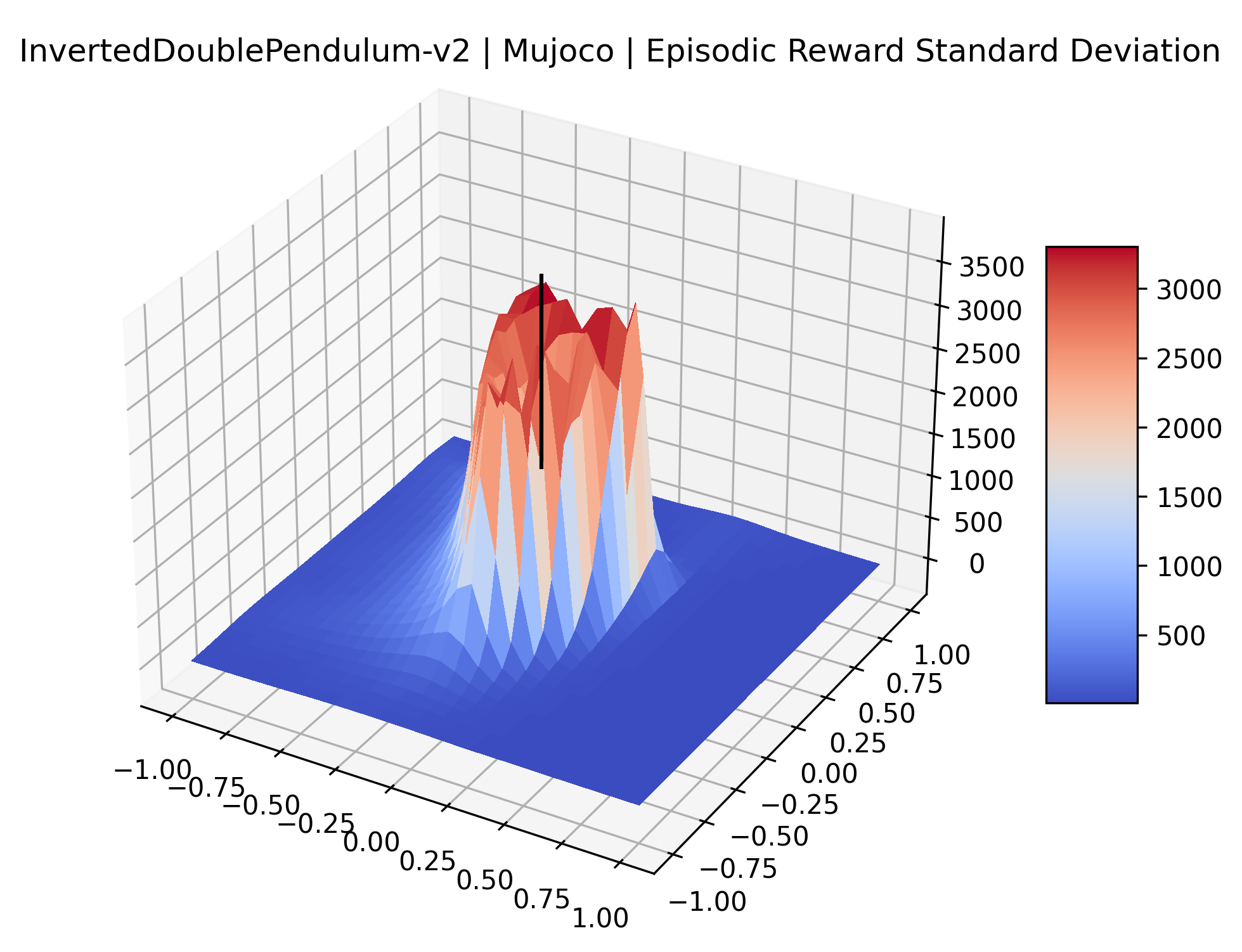} \\
 \includegraphics[width=\surfacescale]{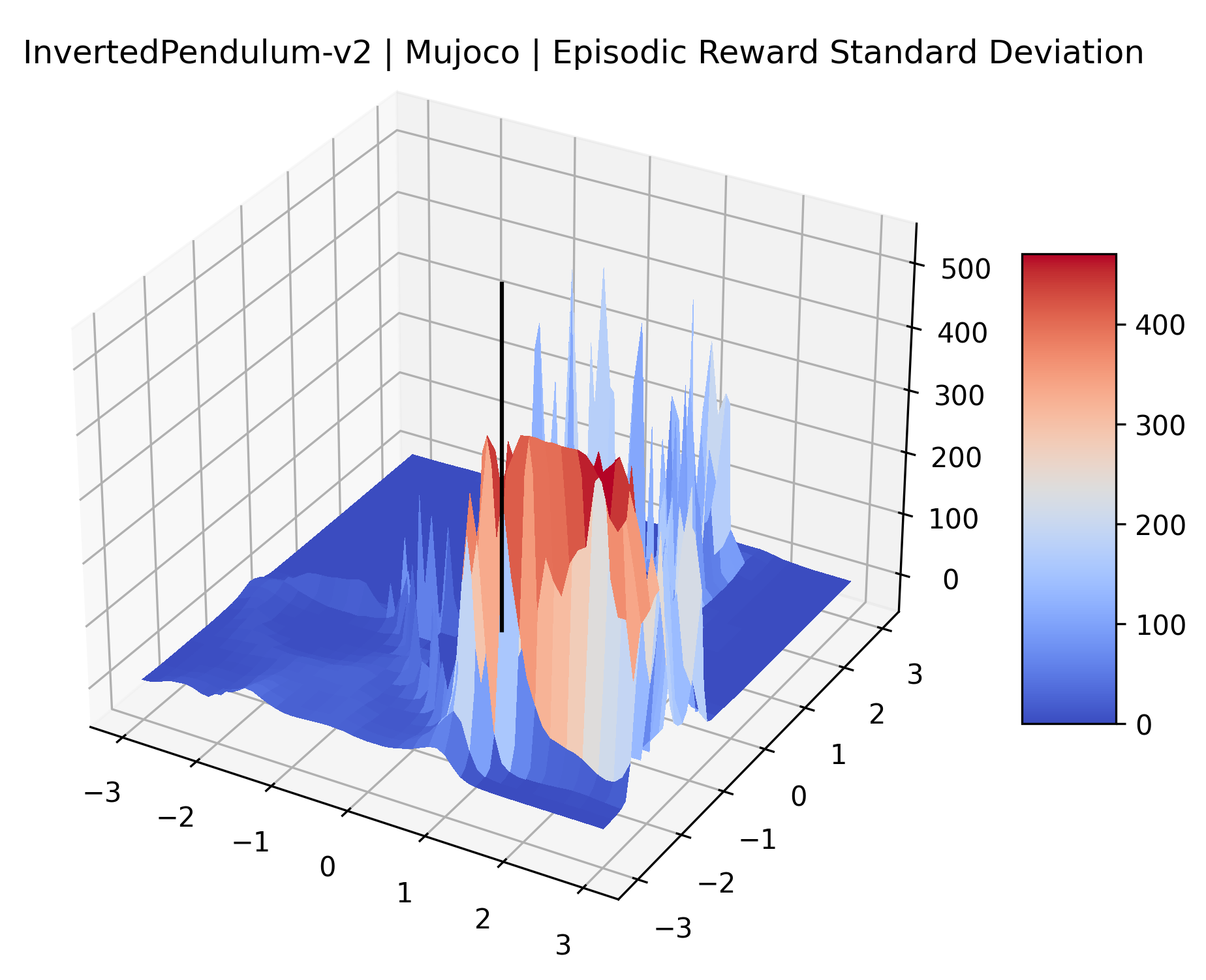} &
 \includegraphics[width=\surfacescale]{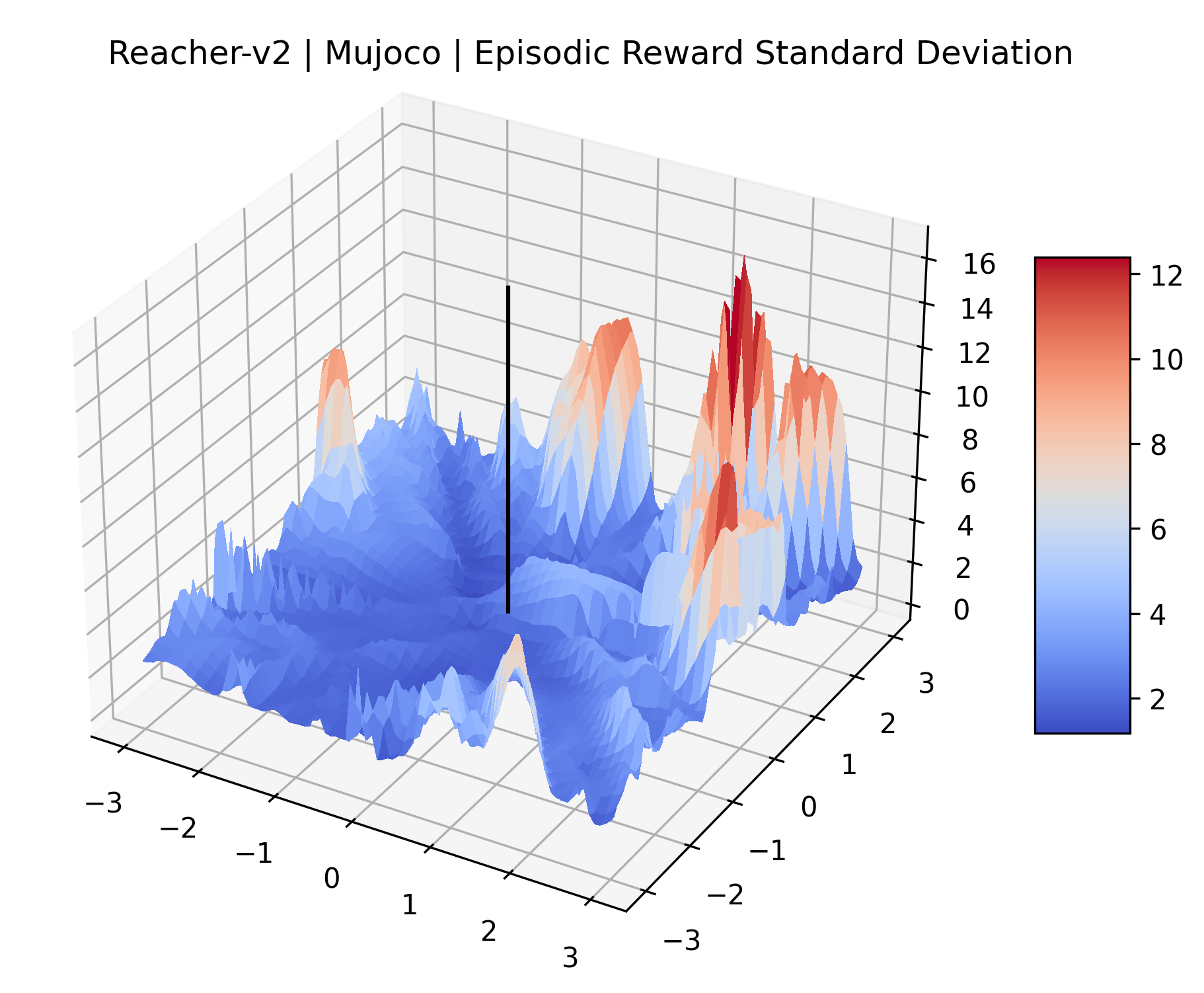} &
 \includegraphics[width=\surfacescale]{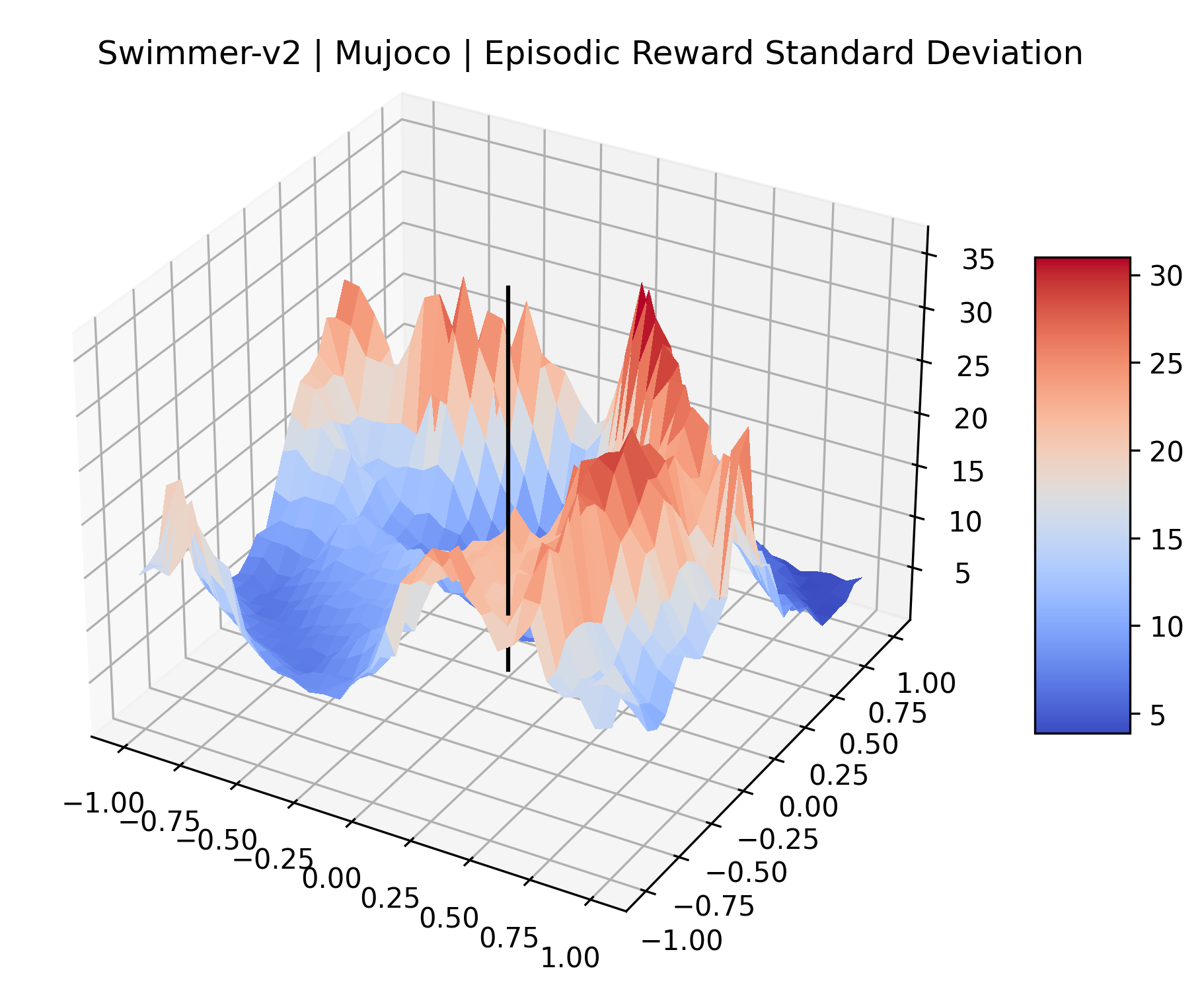} \\ 
 & \includegraphics[width=\surfacescale]{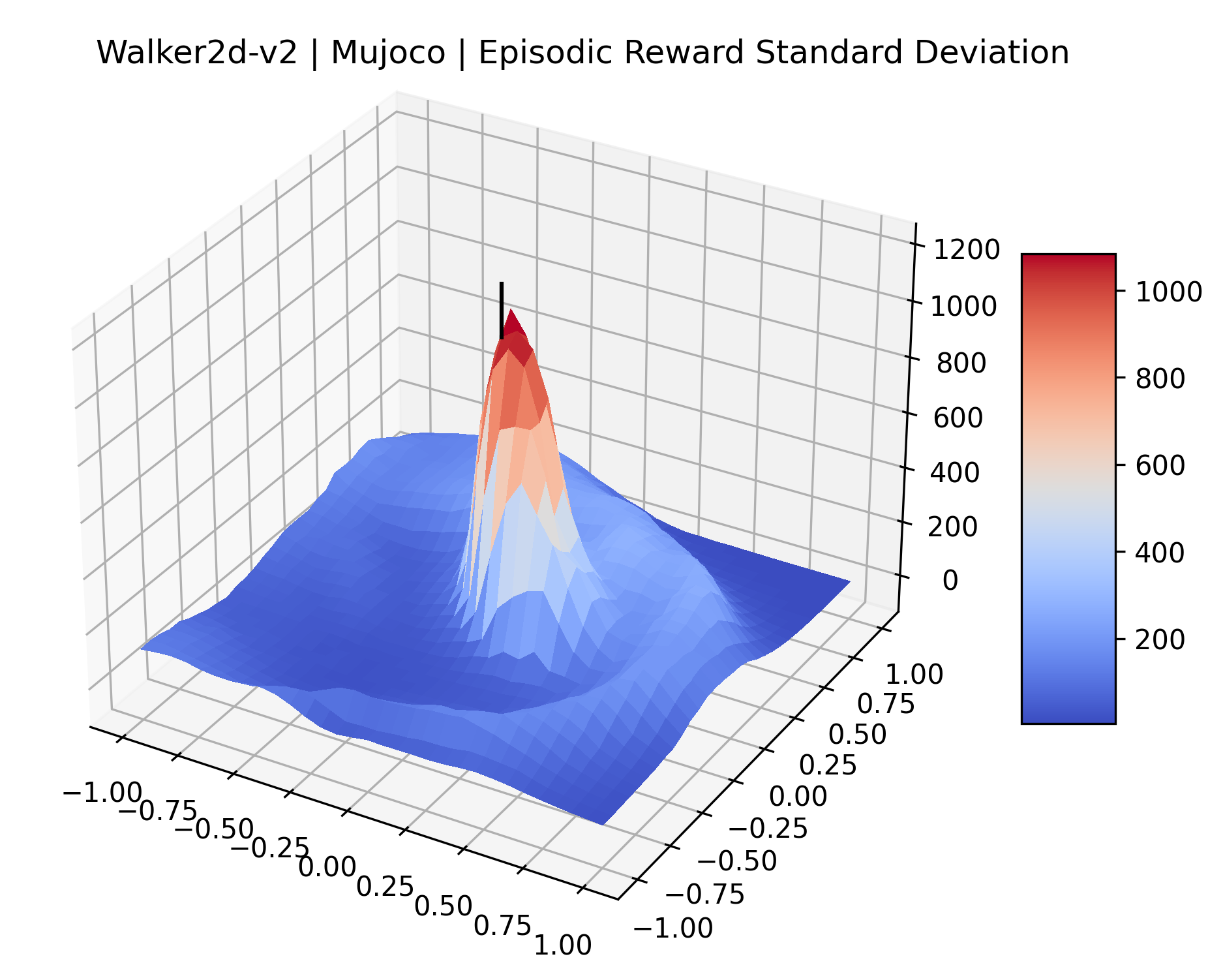} & \\
\end{tabular}
\caption{Standard deviation surfaces for 10 MuJoCo environments.}
\label{fig:mujoco_standarddeviation_table}
\end{figure*}
\pagebreak

\subsection{Atari}
\begin{figure*}[!ht]
\centering
\begin{tabular}{ccc}
 \includegraphics[width=\surfacescale]{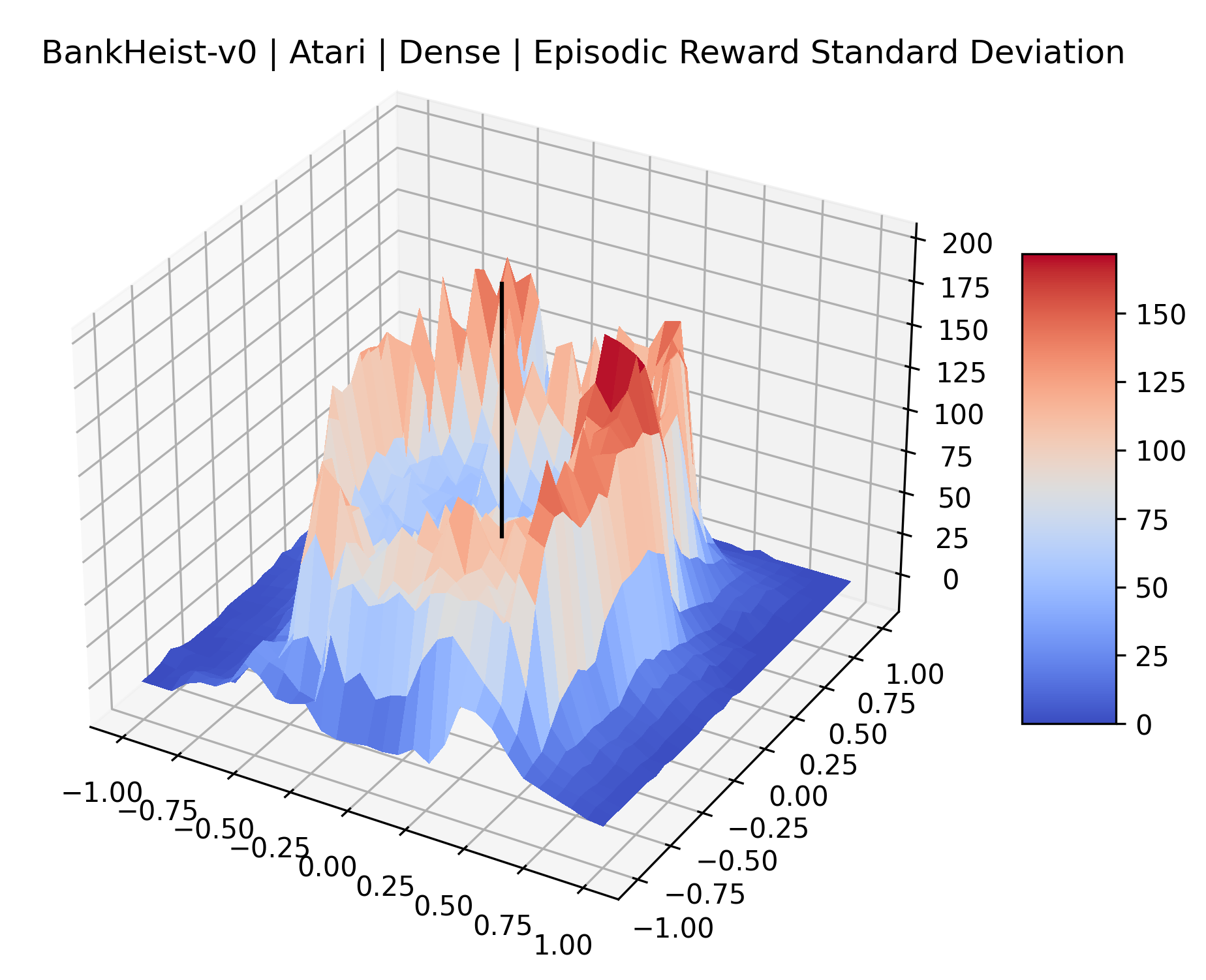} &
 \includegraphics[width=\surfacescale]{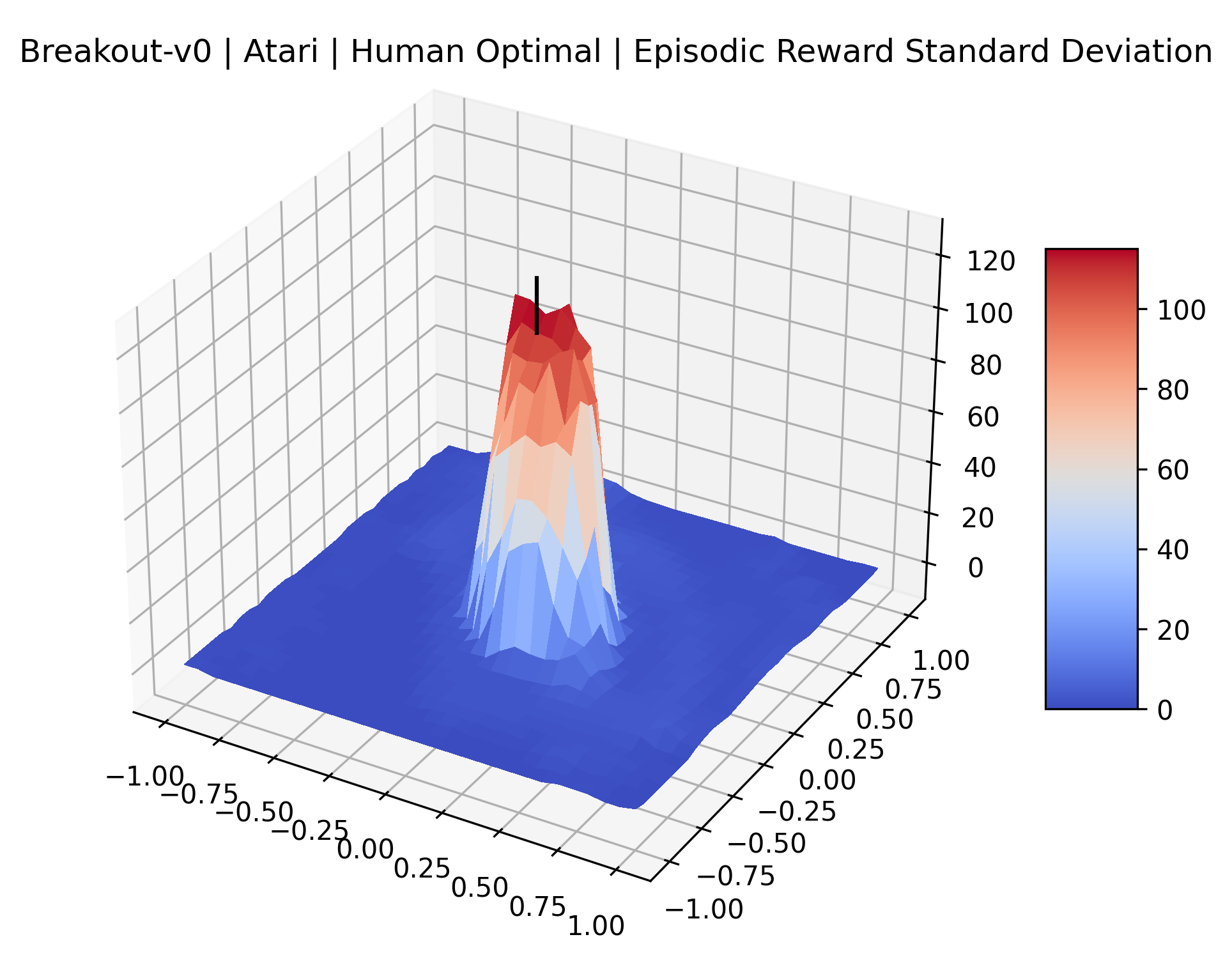} &
 \includegraphics[width=\surfacescale]{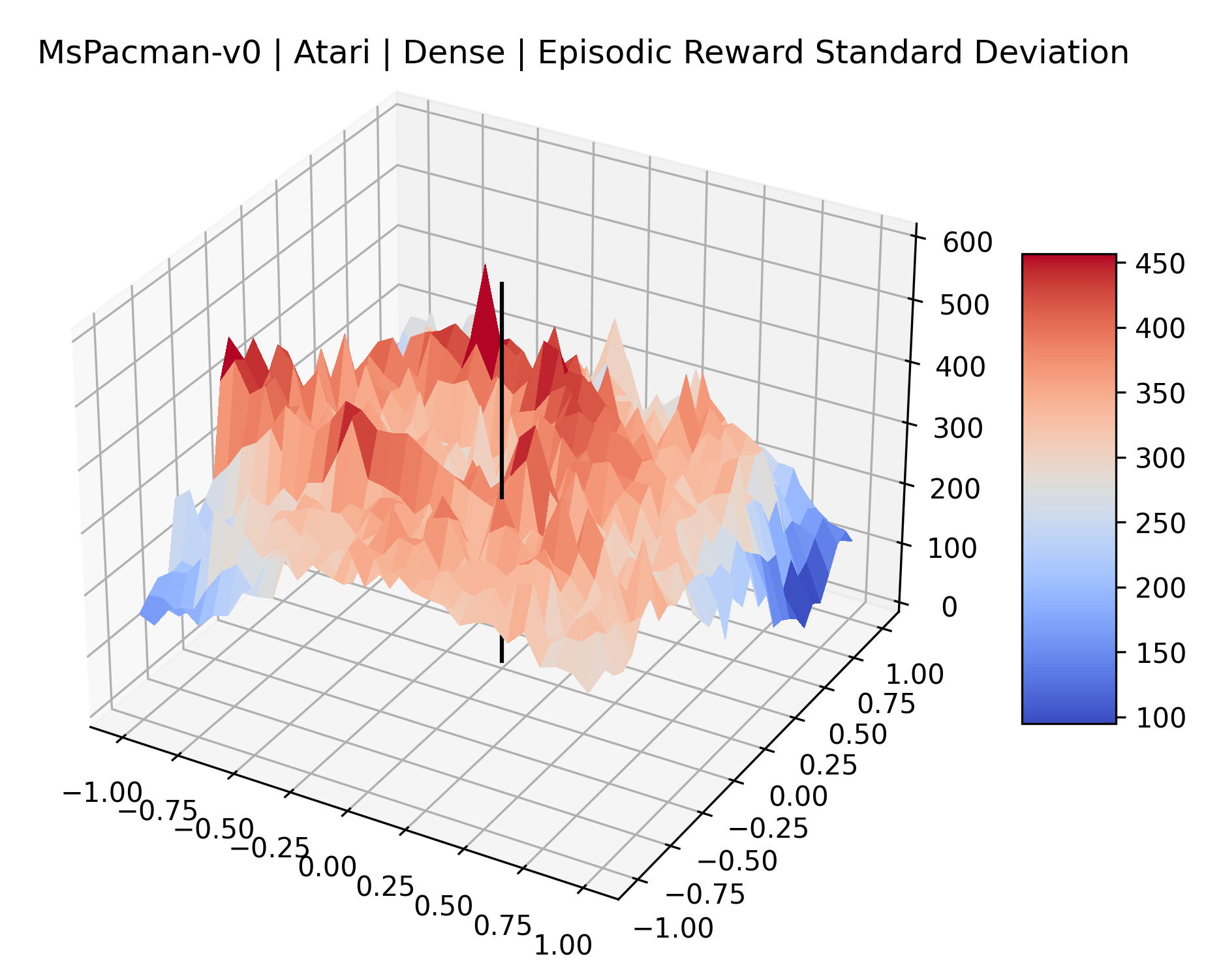} \\
 \includegraphics[width=\surfacescale]{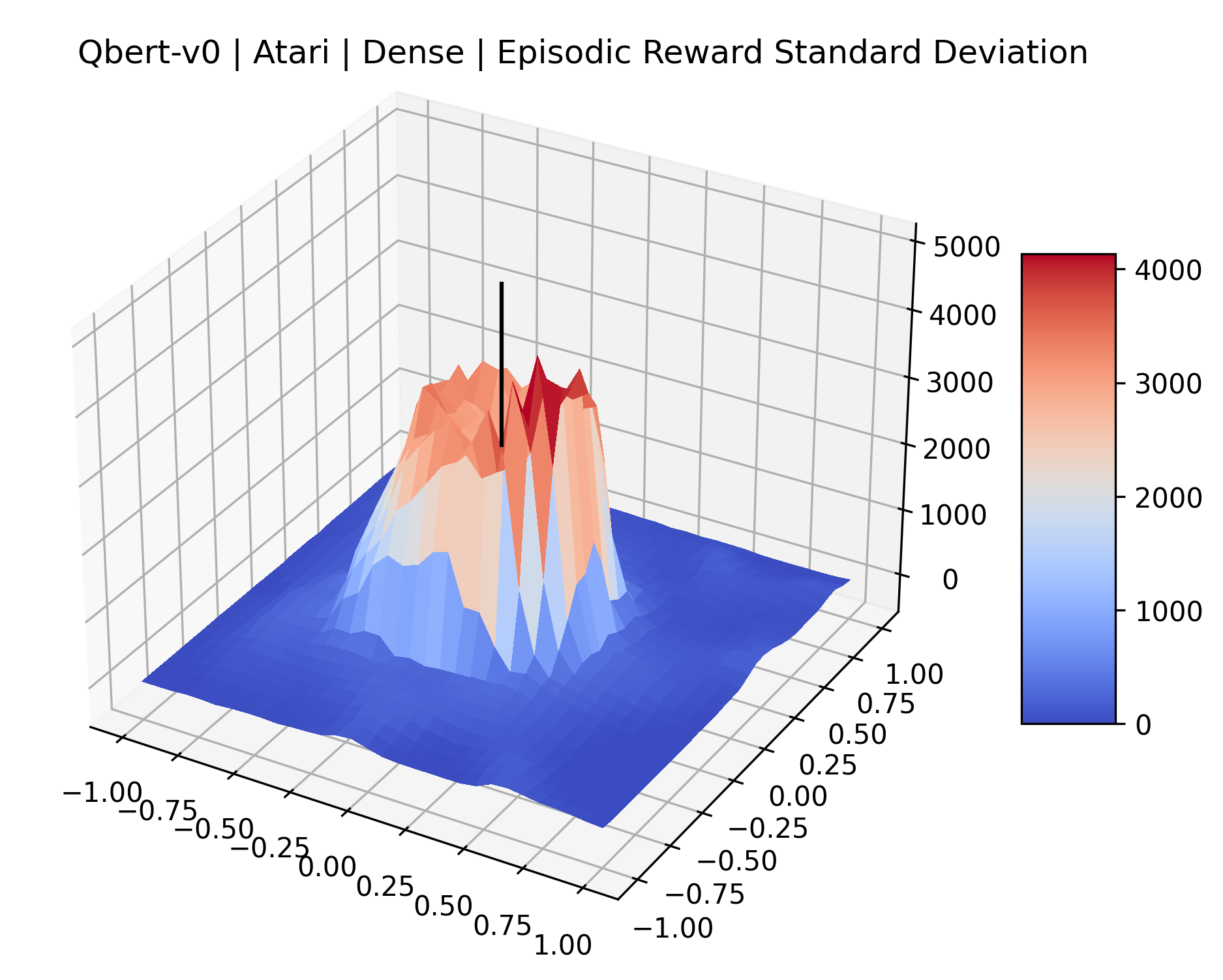} &
 \includegraphics[width=\surfacescale]{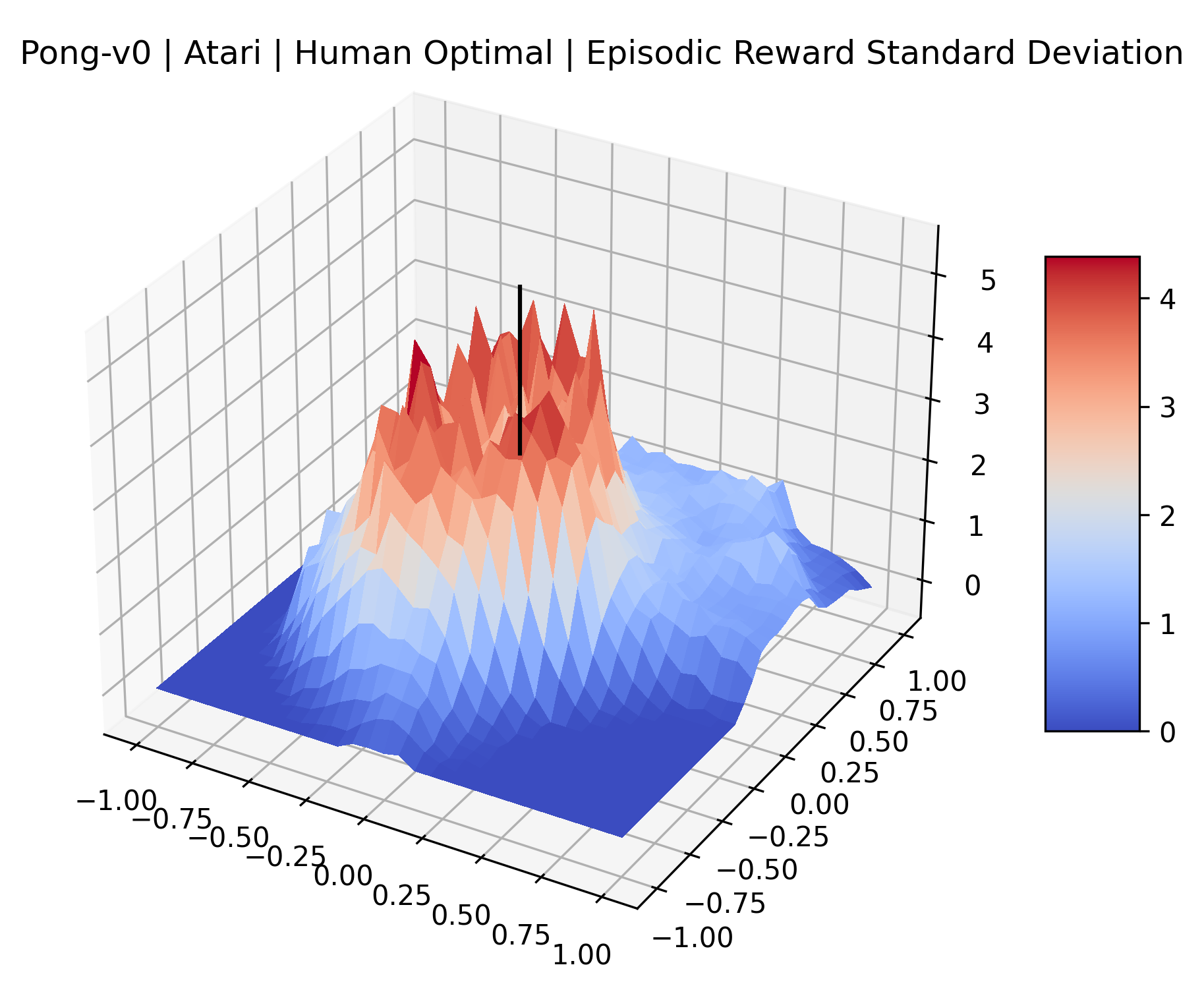} &
 \includegraphics[width=\surfacescale]{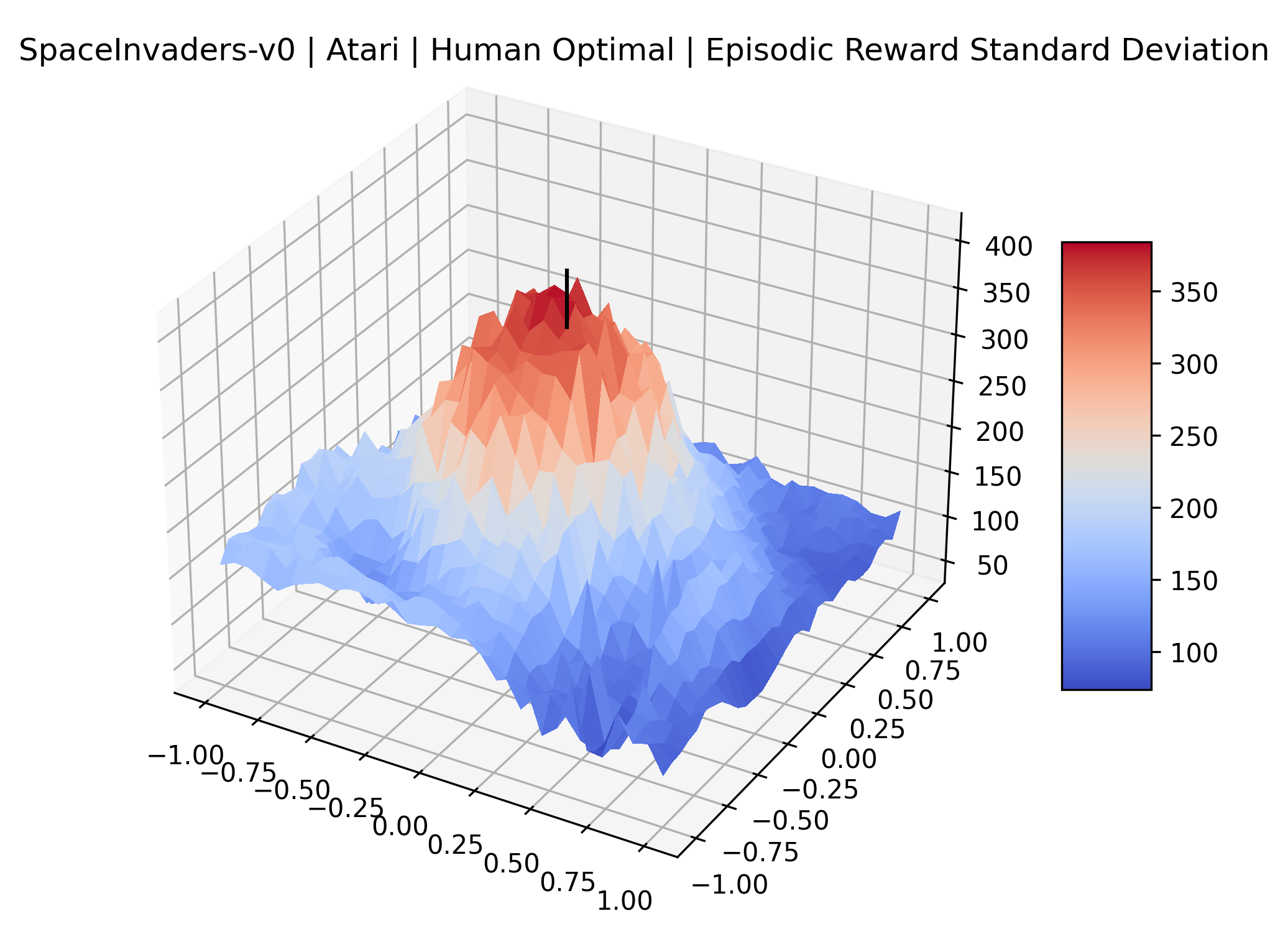} \\
 \includegraphics[width=\surfacescale]{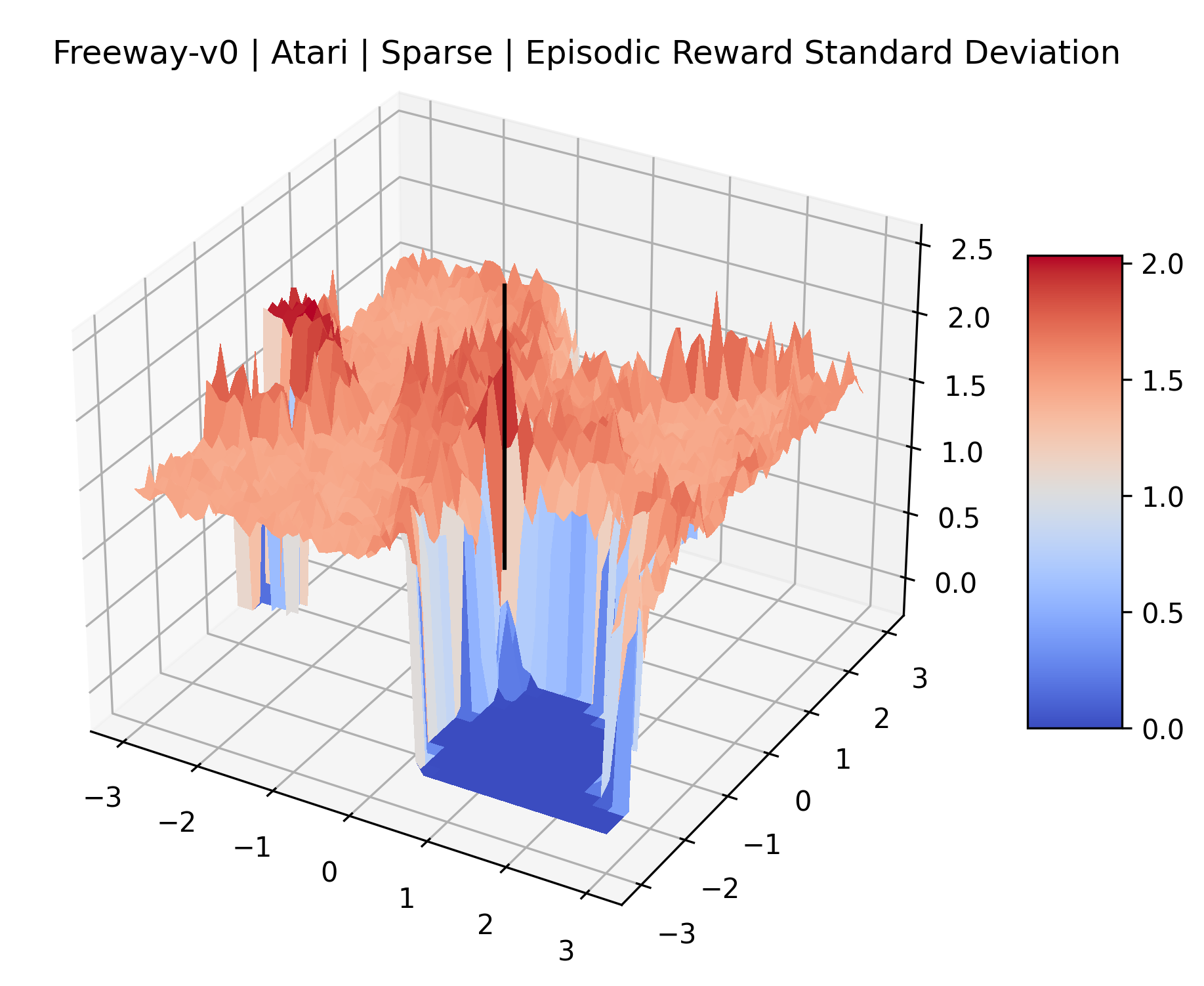} &
 \includegraphics[width=\surfacescale]{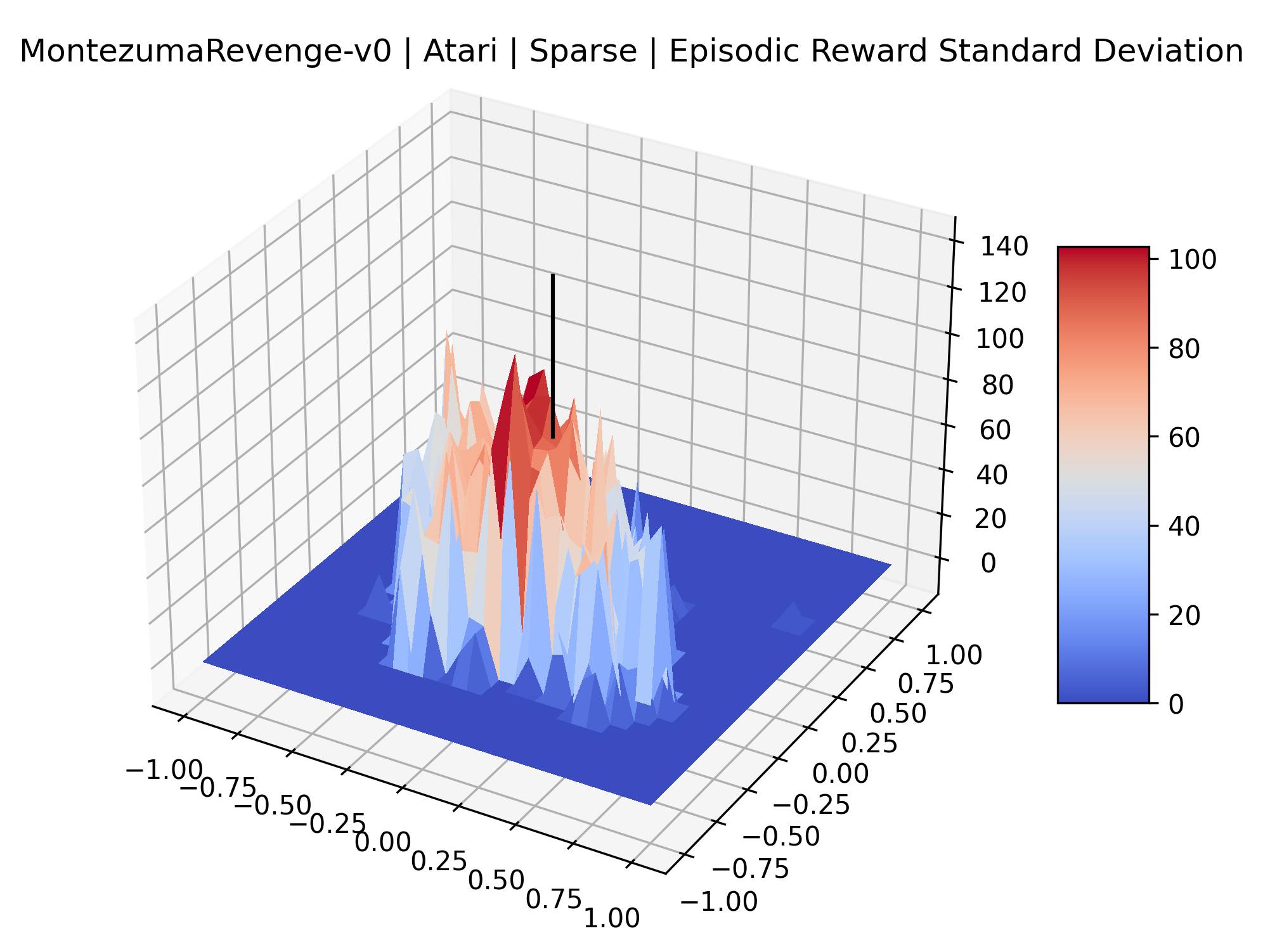} &
 \includegraphics[width=\surfacescale]{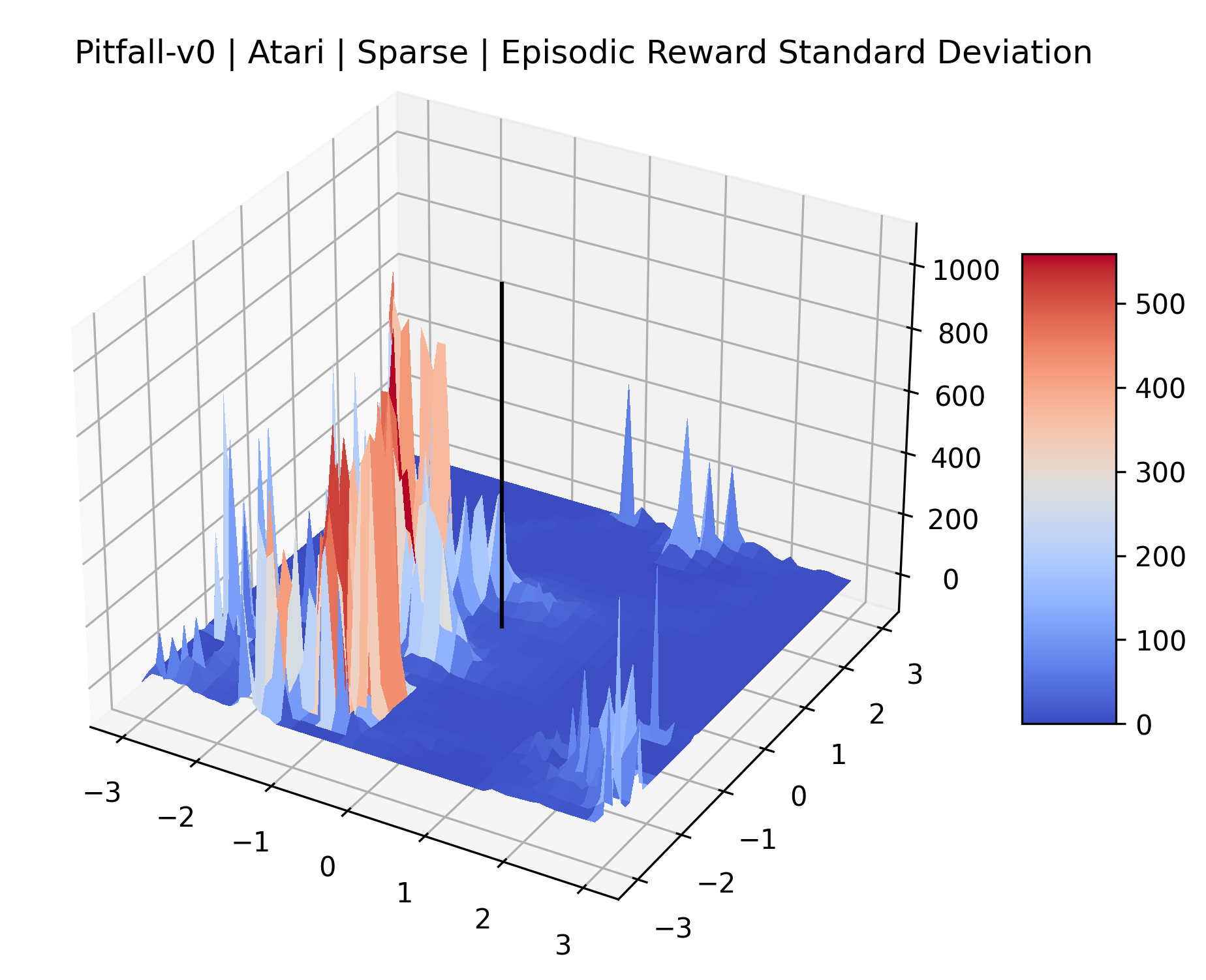} \\ \includegraphics[width=\surfacescale]{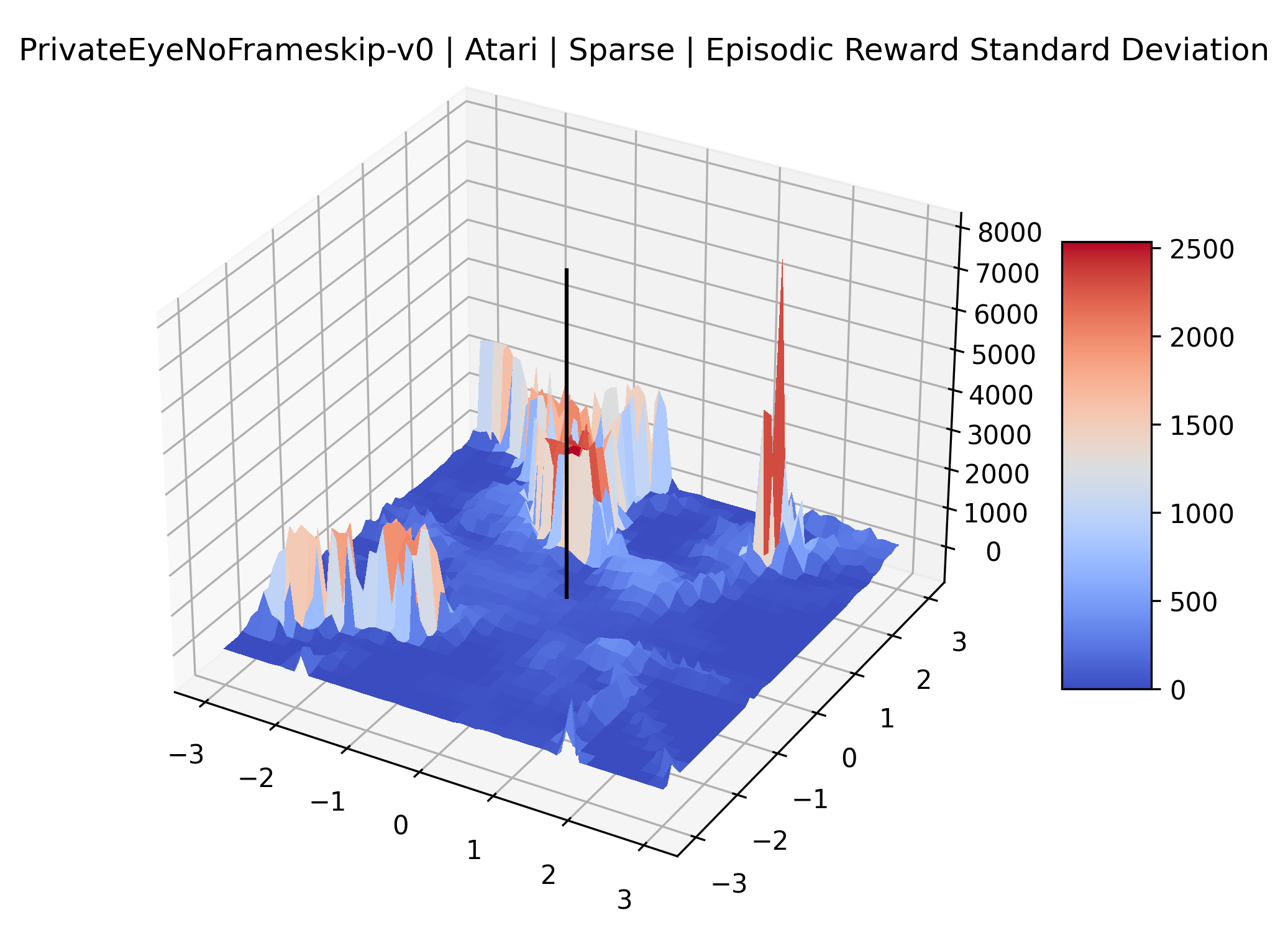} &
 \includegraphics[width=\surfacescale]{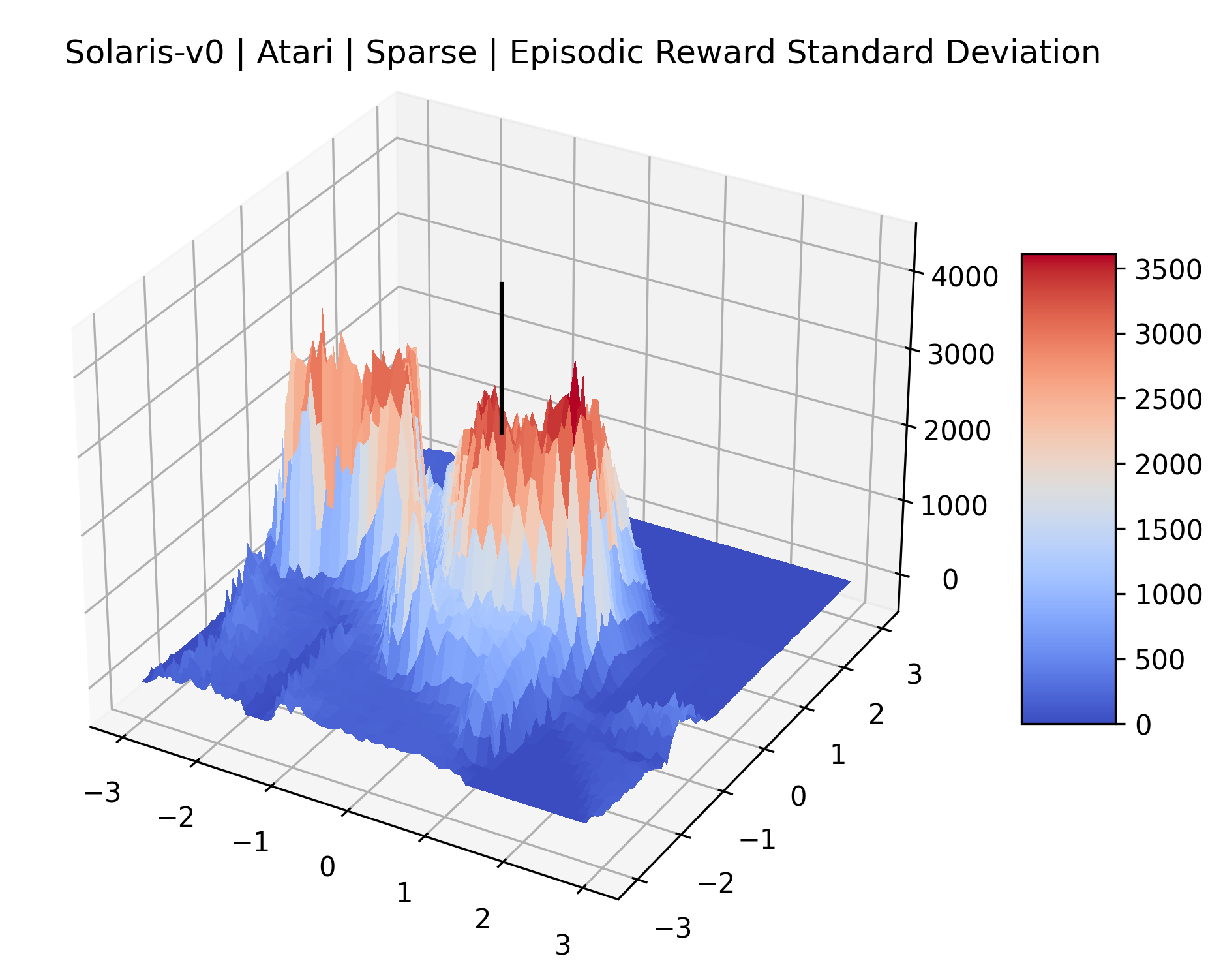} &
 \includegraphics[width=\surfacescale]{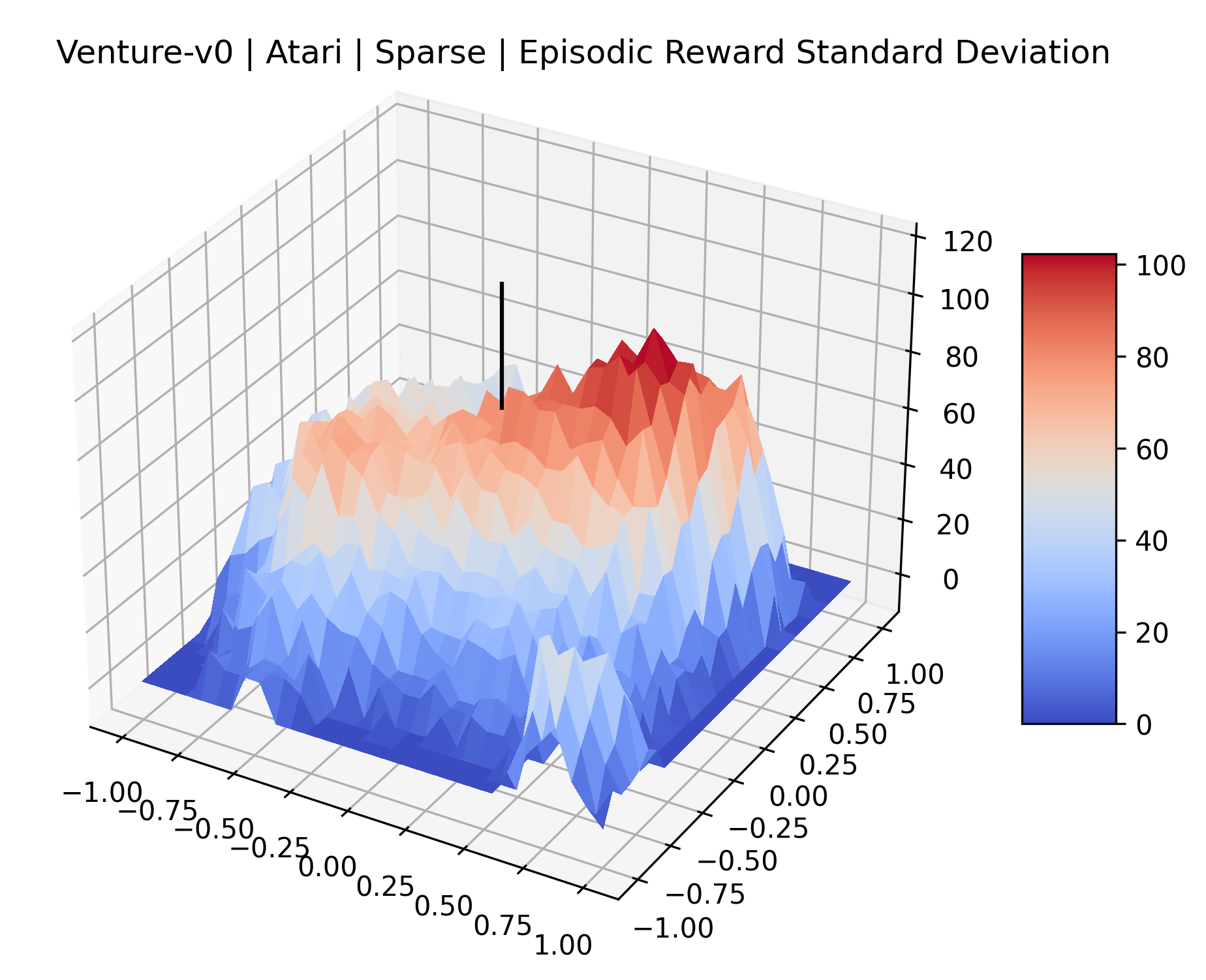} \\
\end{tabular}
\caption{Standard deviation surfaces for 12 Atari environments.}
\label{fig:atari_standarddeviation_table}
\end{figure*}
\pagebreak

\subsection{Standard Deviation and Error Relative to Rewards}
\begin{table*}[ht]
\centering
\begin{tabular}{|c|c|c|}
 \hline
 Environment & Average Standard Deviation / Mean & Average Standard Error / Mean \\
 \hline
 Acrobot & 30.58\% & 0.77\% \\
 CartPole & 37.51\% &  0.67\% \\
 MountainCar & 14.30\% &  0.37\% \\
 MountainCarContinuous & 21.33\% &  0.77\% \\
 Pendulum & 6.82\% &  0.22\% \\
 \hline
 Ant & 39.45\% &  2.53\% \\
 HalfCheetah & 108.68\% &  7.76\% \\
 Hopper & 26.81\% &  0.82\% \\
 Humanoid & 21.08\% &  0.26\% \\
 HumanoidStandup & 11.65\% &  0.52\% \\
 InvertedDoublePendulum & 42.23\% &  0.74\% \\
 InvertedPendulum & 31.88\% &  0.45\% \\
 Reacher & 13.23\% &  0.21\% \\
 Swimmer & 139.79\% &  9.88\% \\
 Walker2d & 100.01\% &  2.11\% \\
 \hline
 Breakout & 119.43\% &  21.76\% \\
 Pong & 15.80\% &  2.09\% \\
 SpaceInvaders & 50.02\% &  2.99\% \\
 BankHeist & 98.40\% &  13.65\% \\
 MsPacman & 52.12\% &  2.83\% \\
 Q*bert & 184.20\% &  8.82\% \\
 Freeway & 18.00\% &  0.81\% \\
 Montezuma's Revenge & 559.64\% &  65.11\% \\
 Pitfall! & 180.45\% &  8.56\% \\
 Private Eye & 448.63\% &  23.26\% \\
 Solaris & 166.94\% &  21.49\% \\
 Venture & 742.24\% &  31.78\% \\
 \hline

\end{tabular}
\caption{Standard deviation and Standard error as a percentage of the rewards for each point estimate in the reward surface for each environment. We see that aside from some of the sparse Atari environments, the standard error of our reward surfaces is fairly low on average.}
\label{fig:stddev_table}
\end{table*}
\pagebreak

\section{Reproducibility}
\label{appendix:reproducibility}
\newcommand\variancescale{0.24\linewidth}

\begin{figure*}[!htb]
\centering
\begin{tabular}{ccc}
 \includegraphics[width=\variancescale]{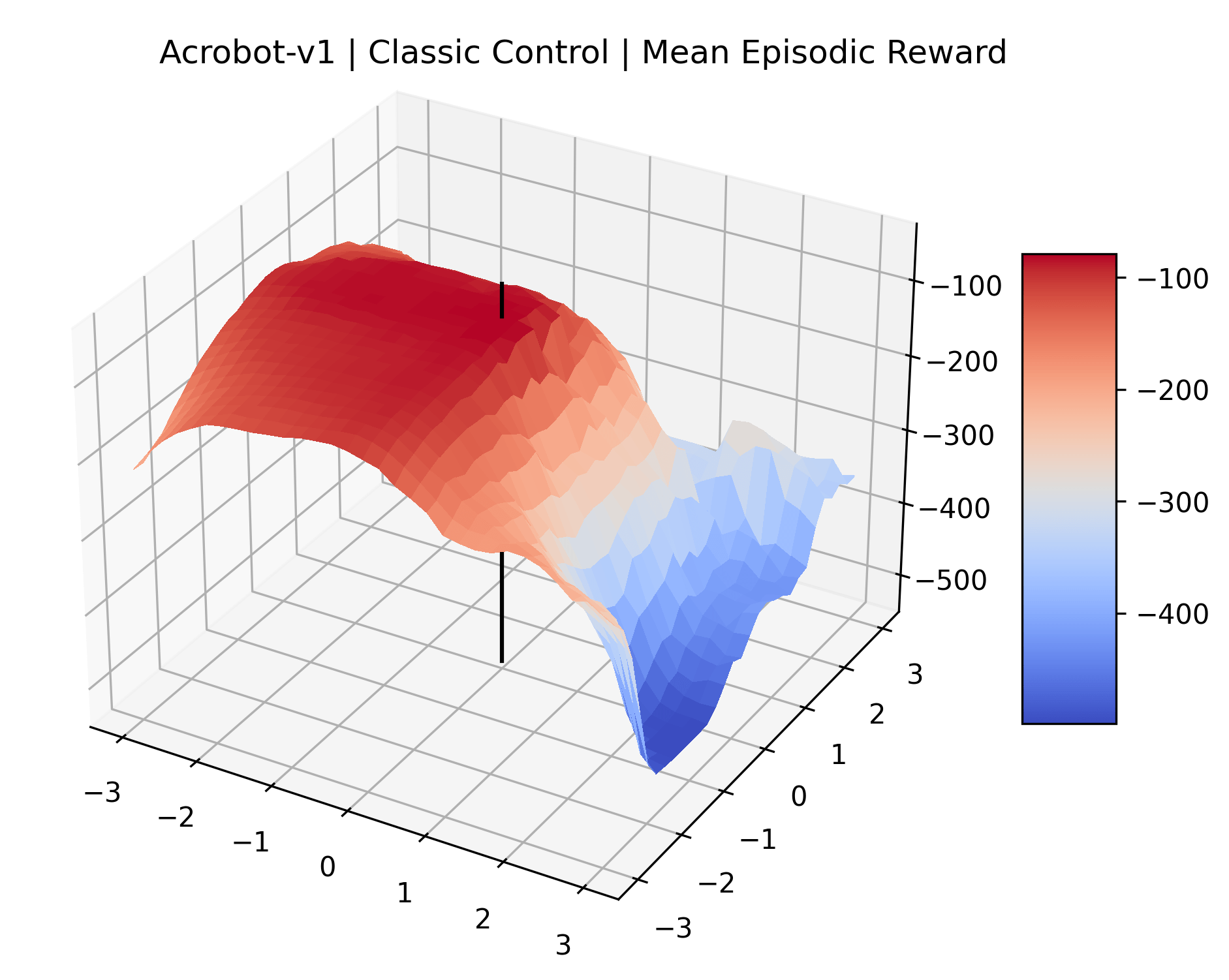} &
 \includegraphics[width=\variancescale]{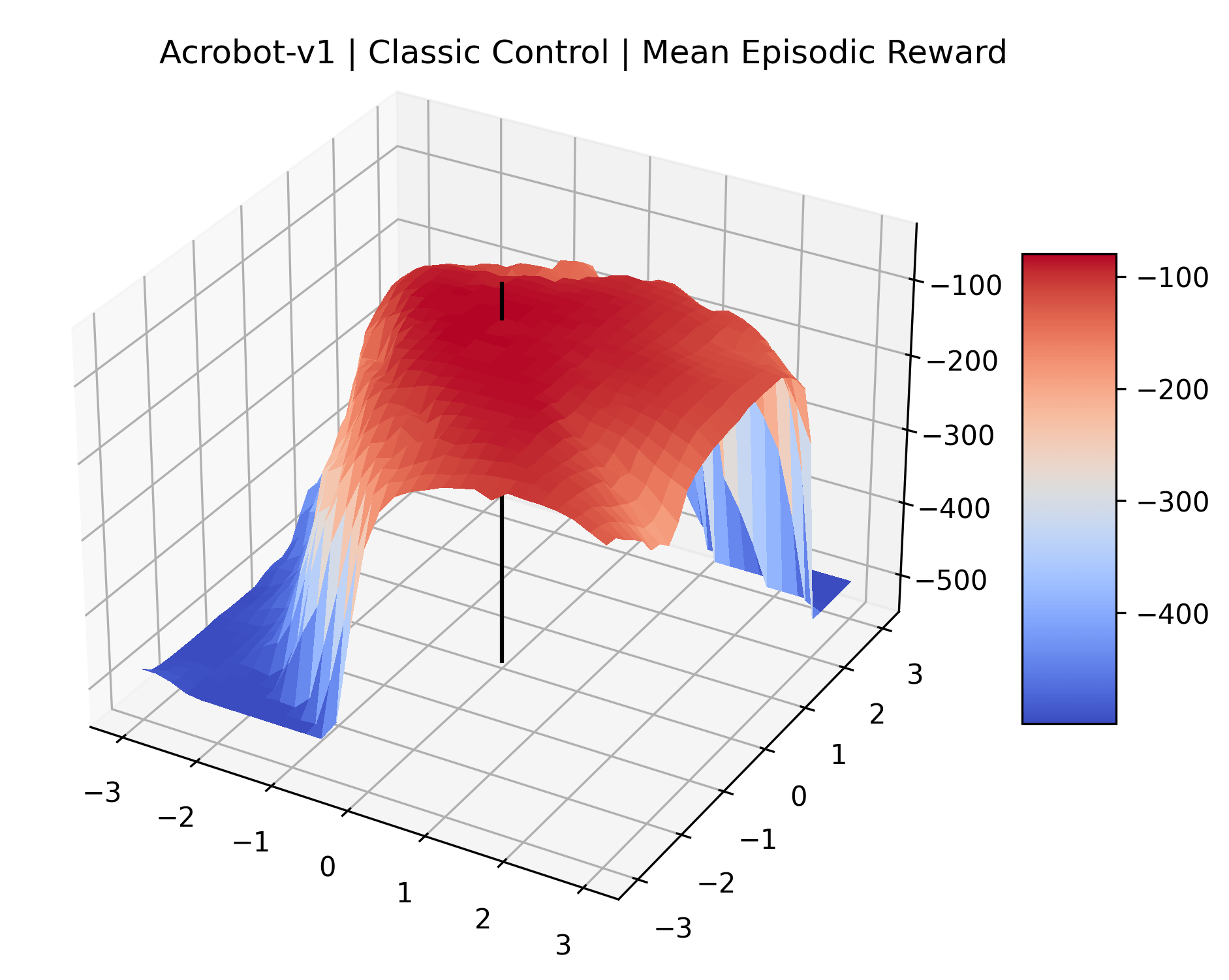} &
 \includegraphics[width=\variancescale]{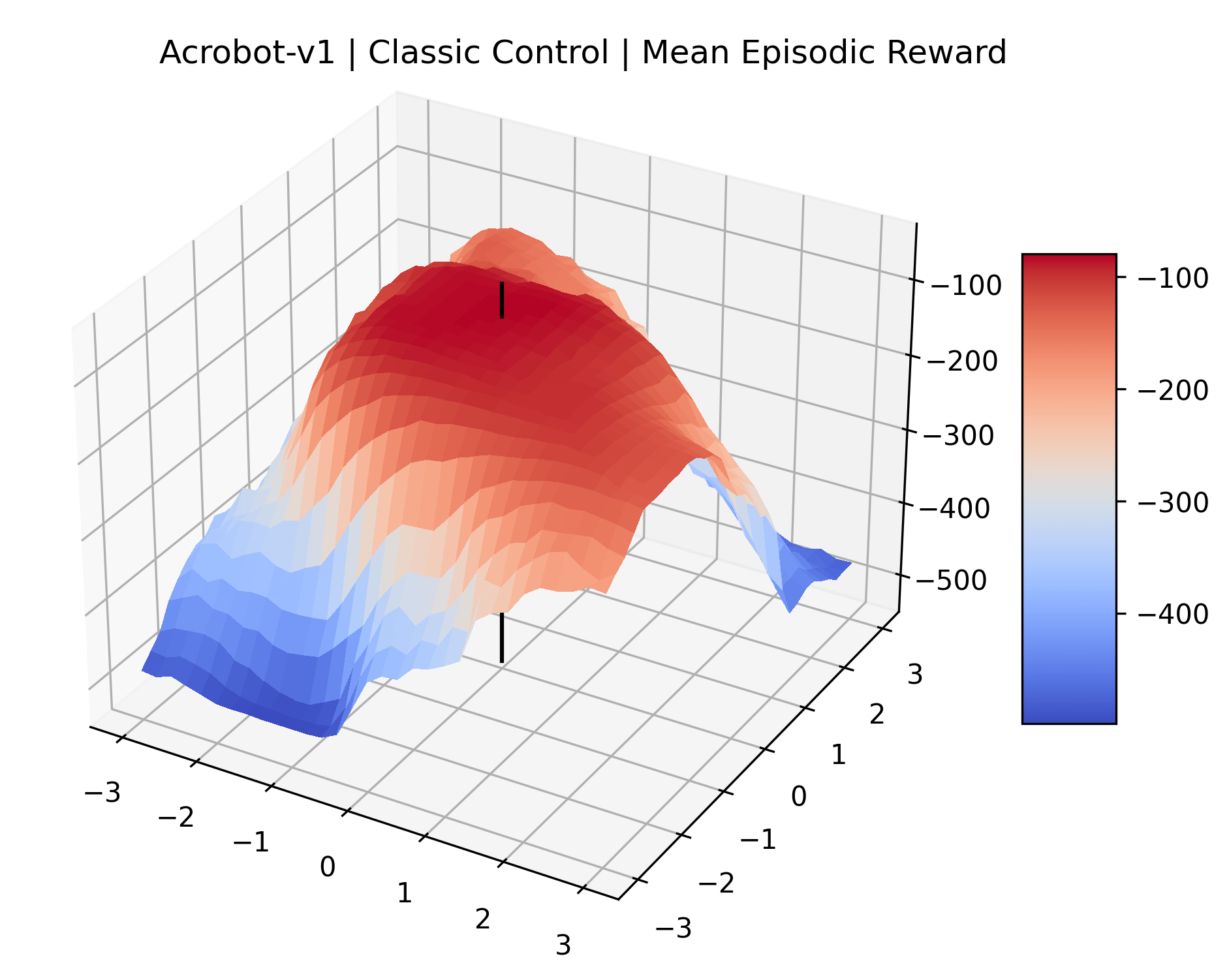} \\
\end{tabular}
\begin{tabular}{ccc}
 \includegraphics[width=\variancescale]{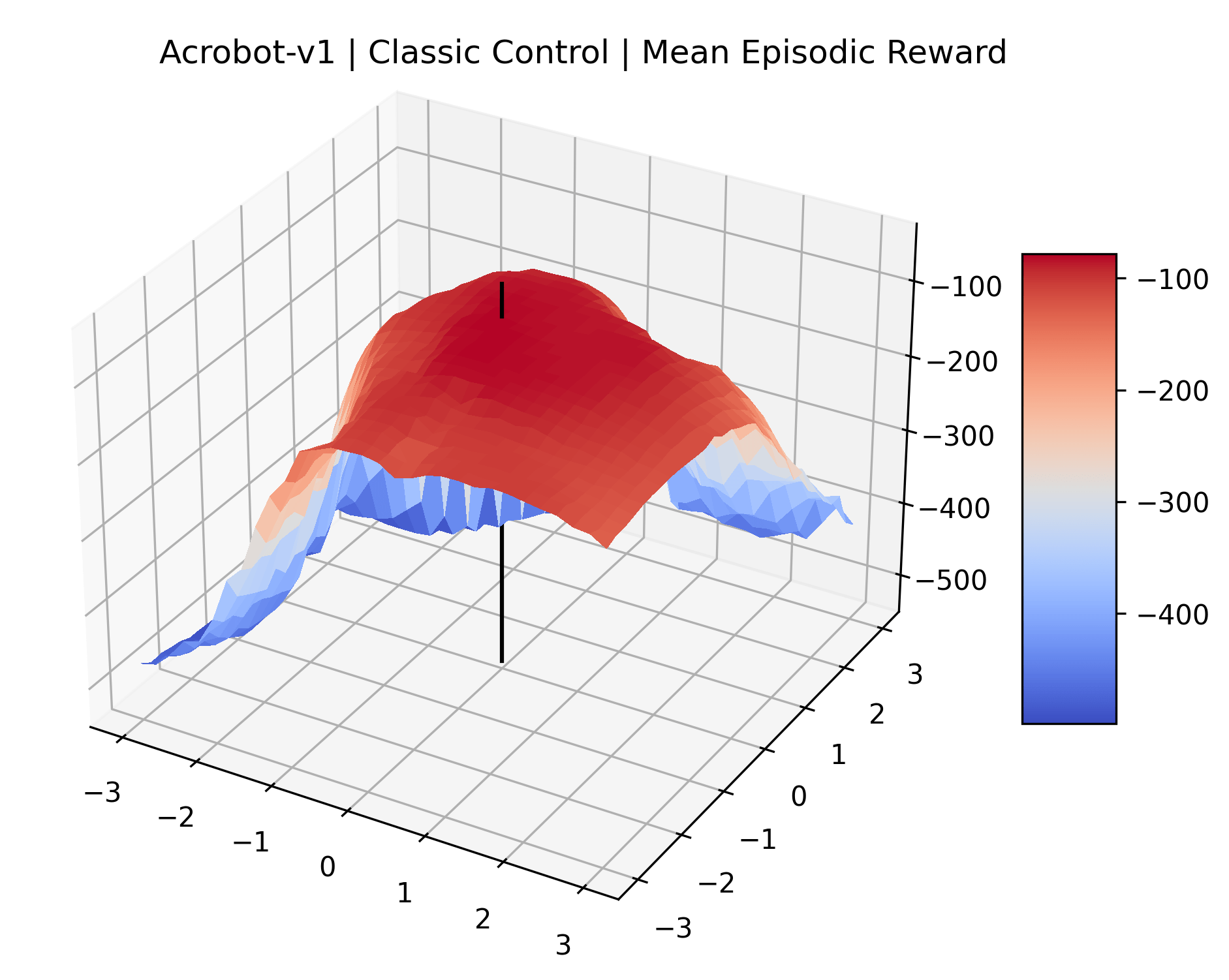} &
 \includegraphics[width=\variancescale]{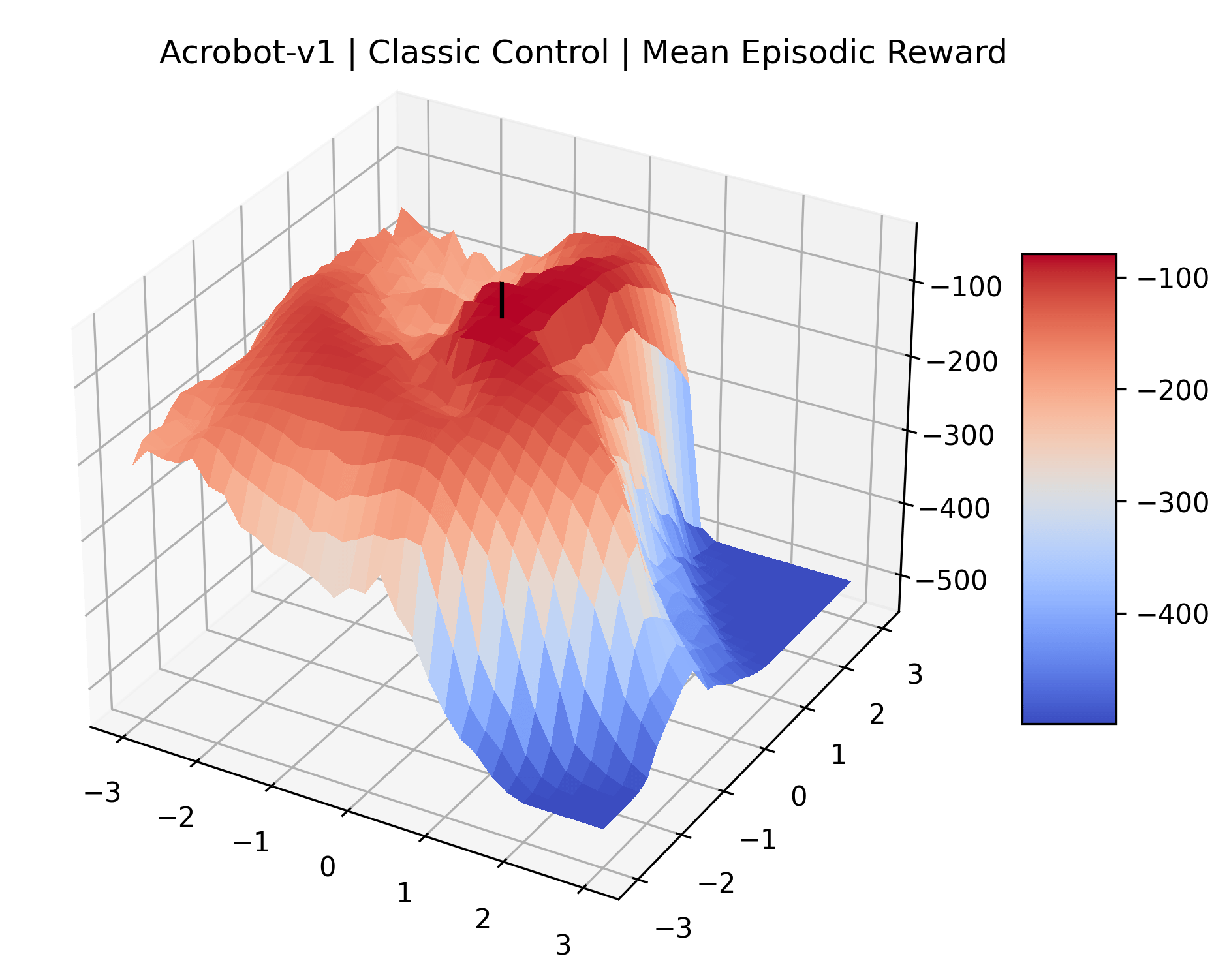} &
 \includegraphics[width=\variancescale]{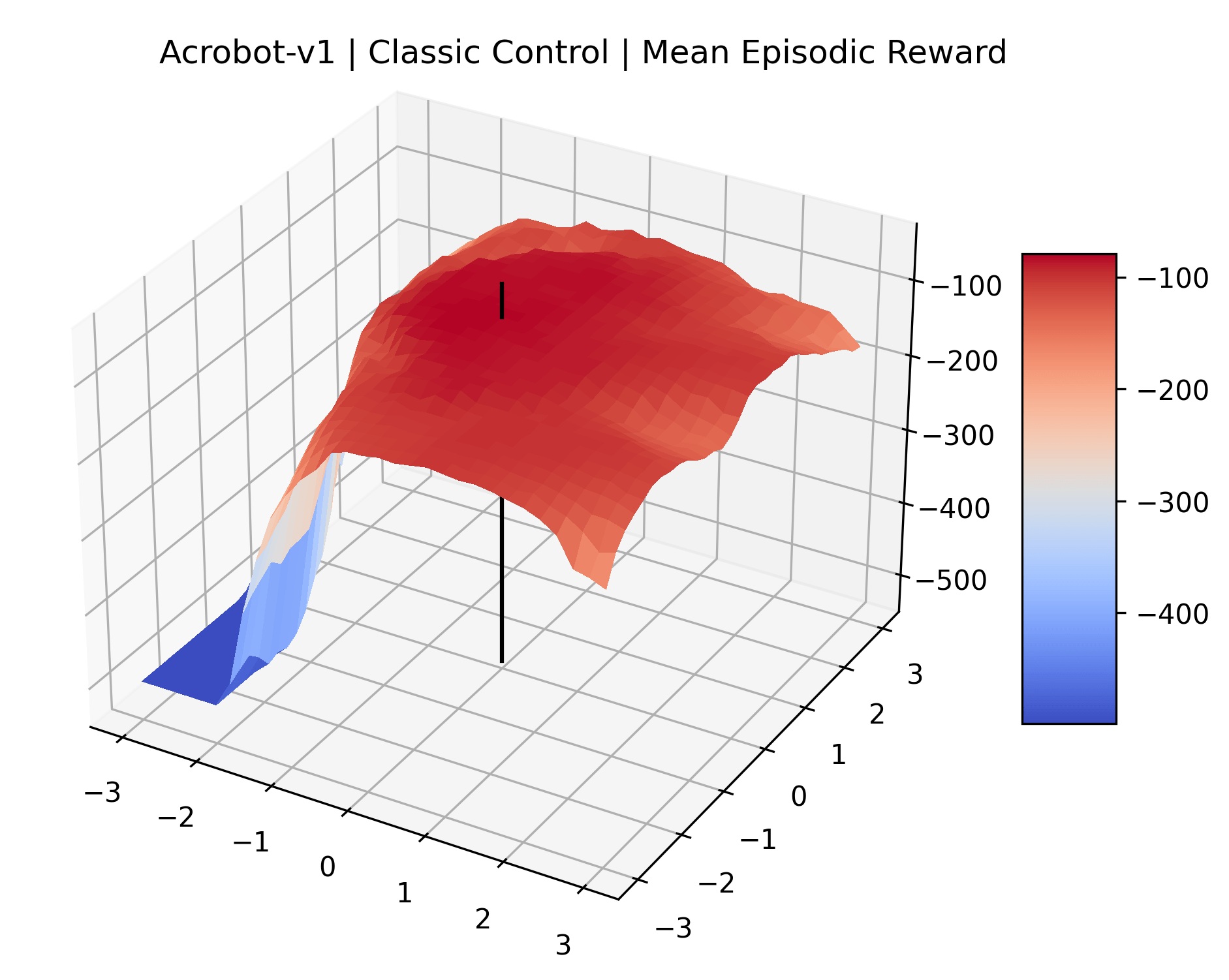} \\
\end{tabular}
\begin{tabular}{ccc}
 \includegraphics[width=\variancescale]{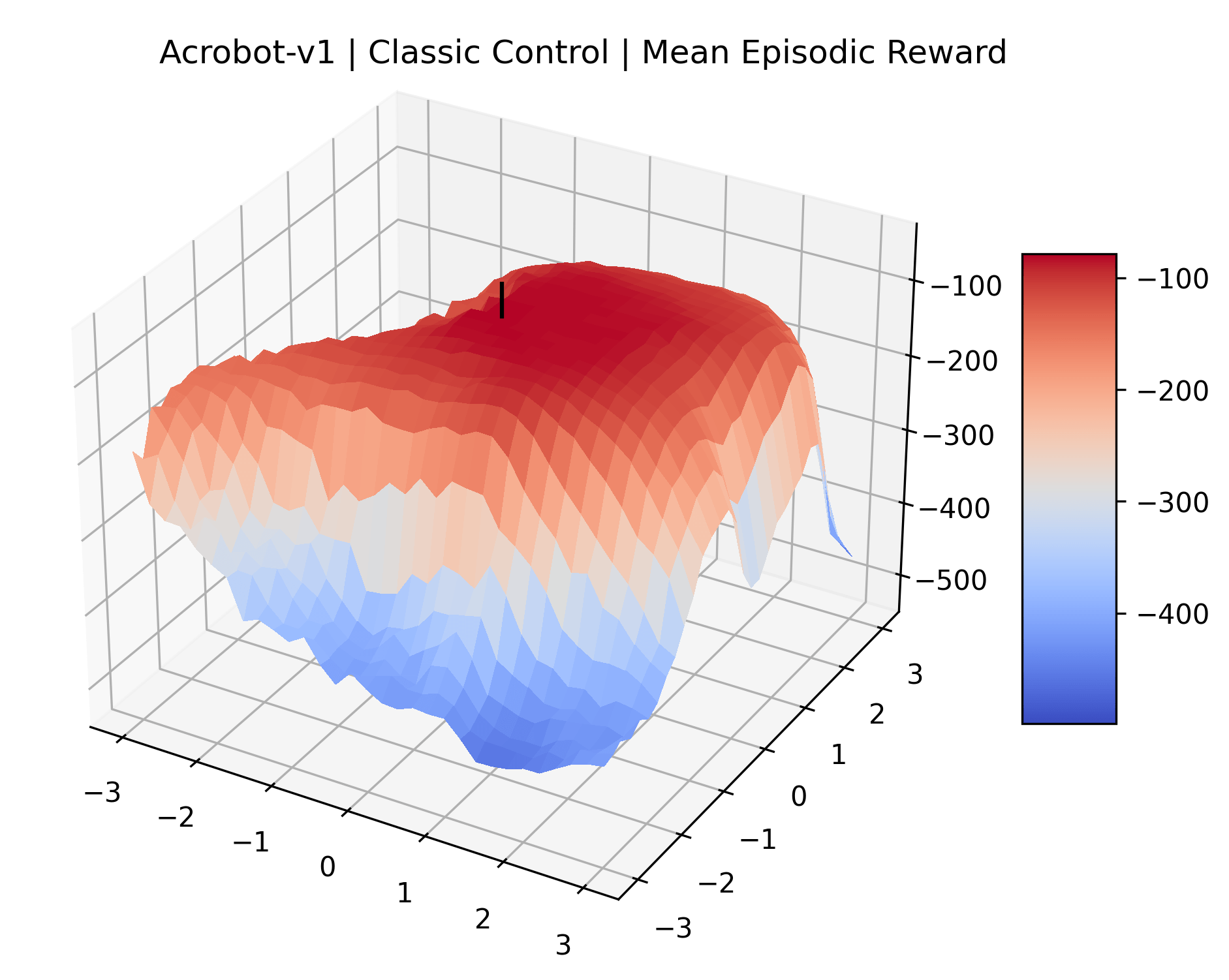} &
 \includegraphics[width=\variancescale]{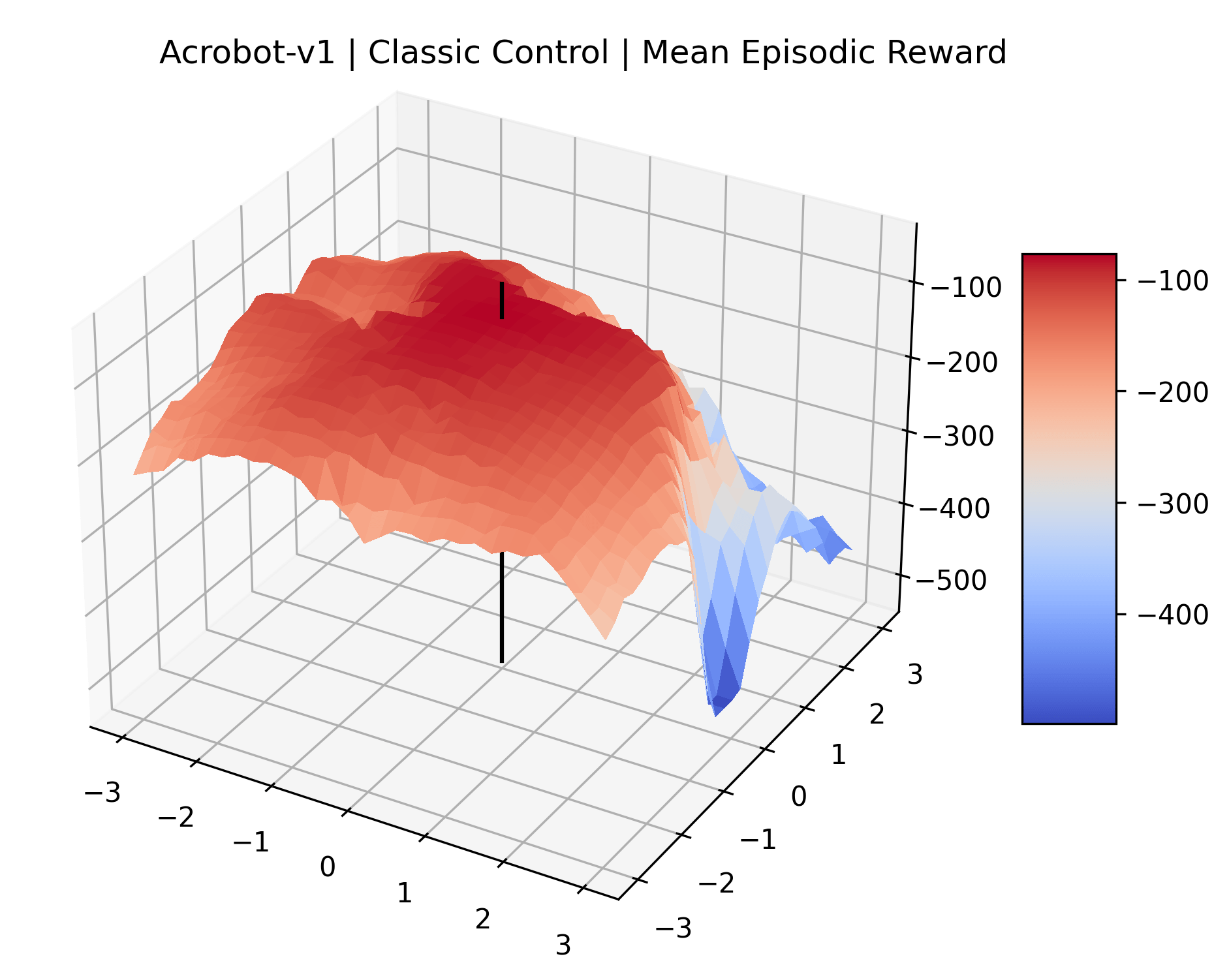} &
 \includegraphics[width=\variancescale]{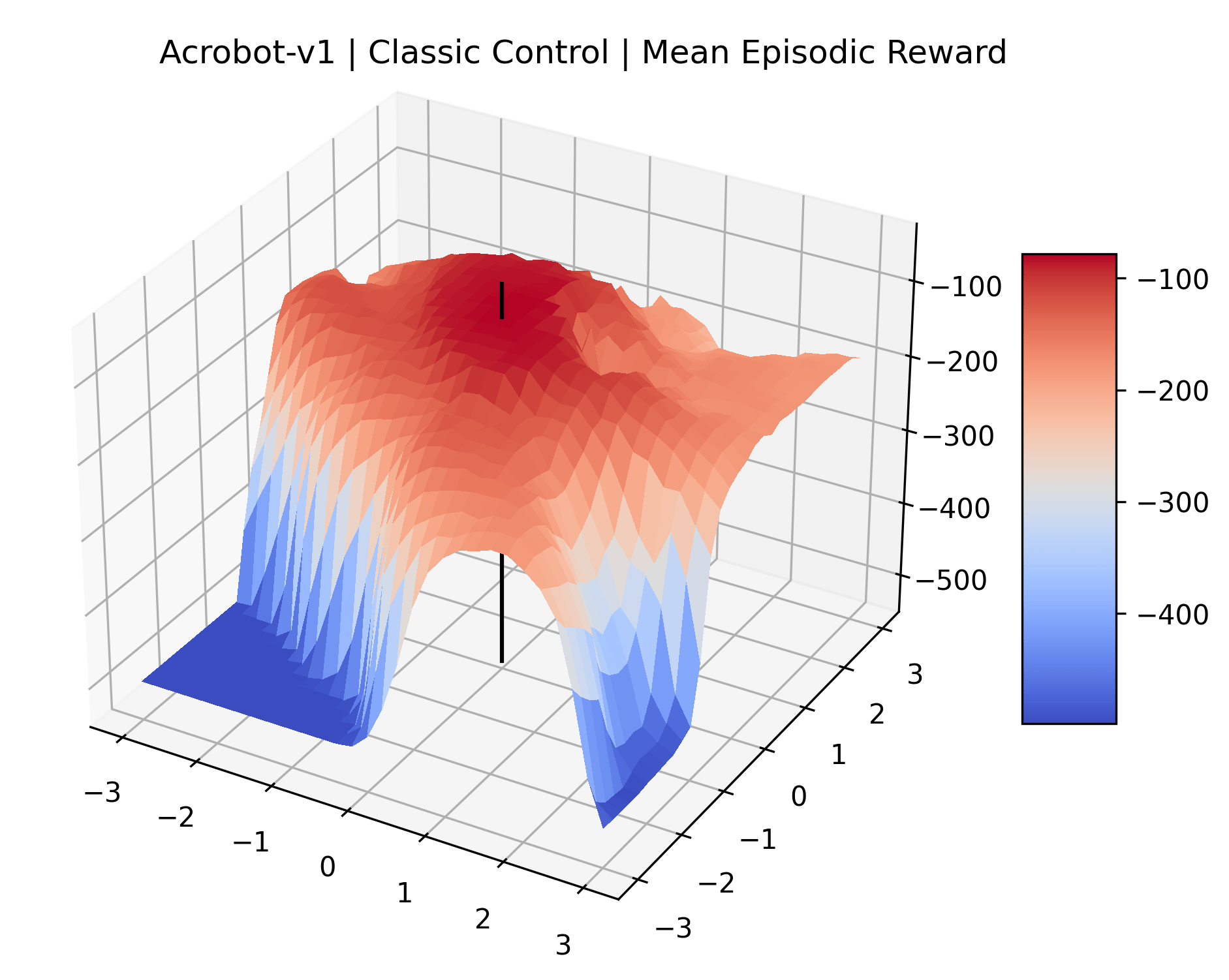} \\
\end{tabular}
\begin{tabular}{ccc}
 \includegraphics[width=\variancescale]{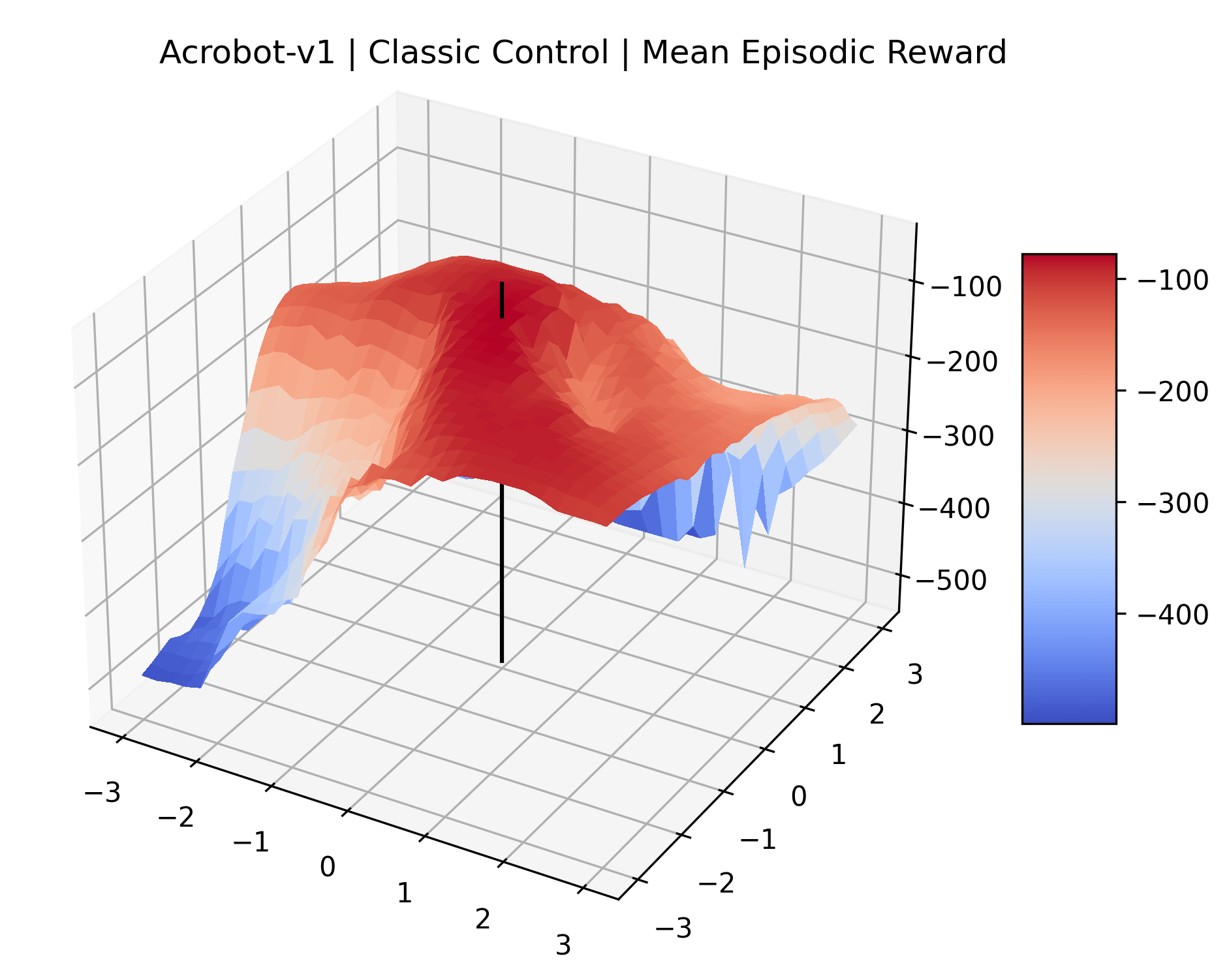} &
 \includegraphics[width=\variancescale]{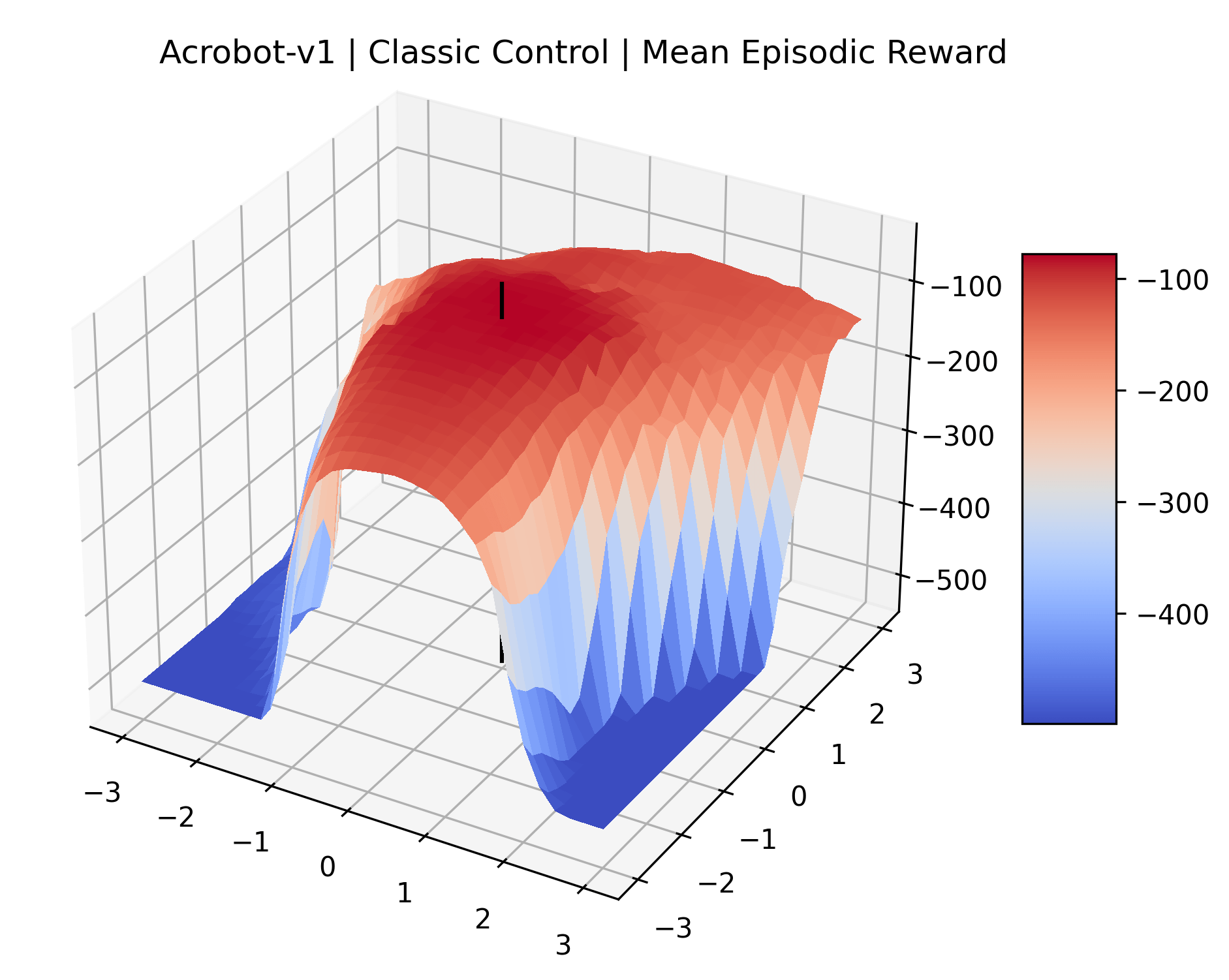} &
 \includegraphics[width=\variancescale]{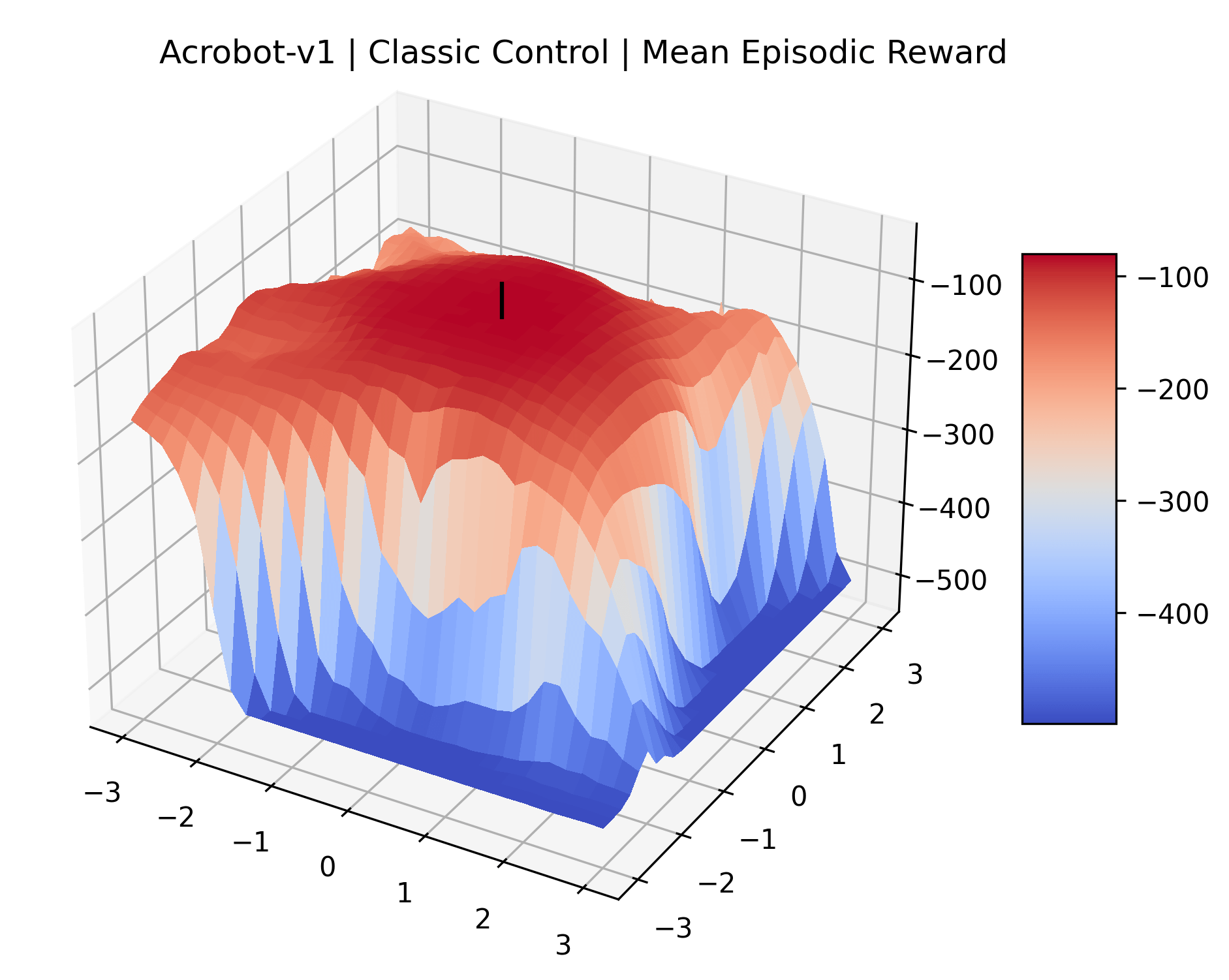} \\
\end{tabular}
\begin{tabular}{ccc}
 \includegraphics[width=\variancescale]{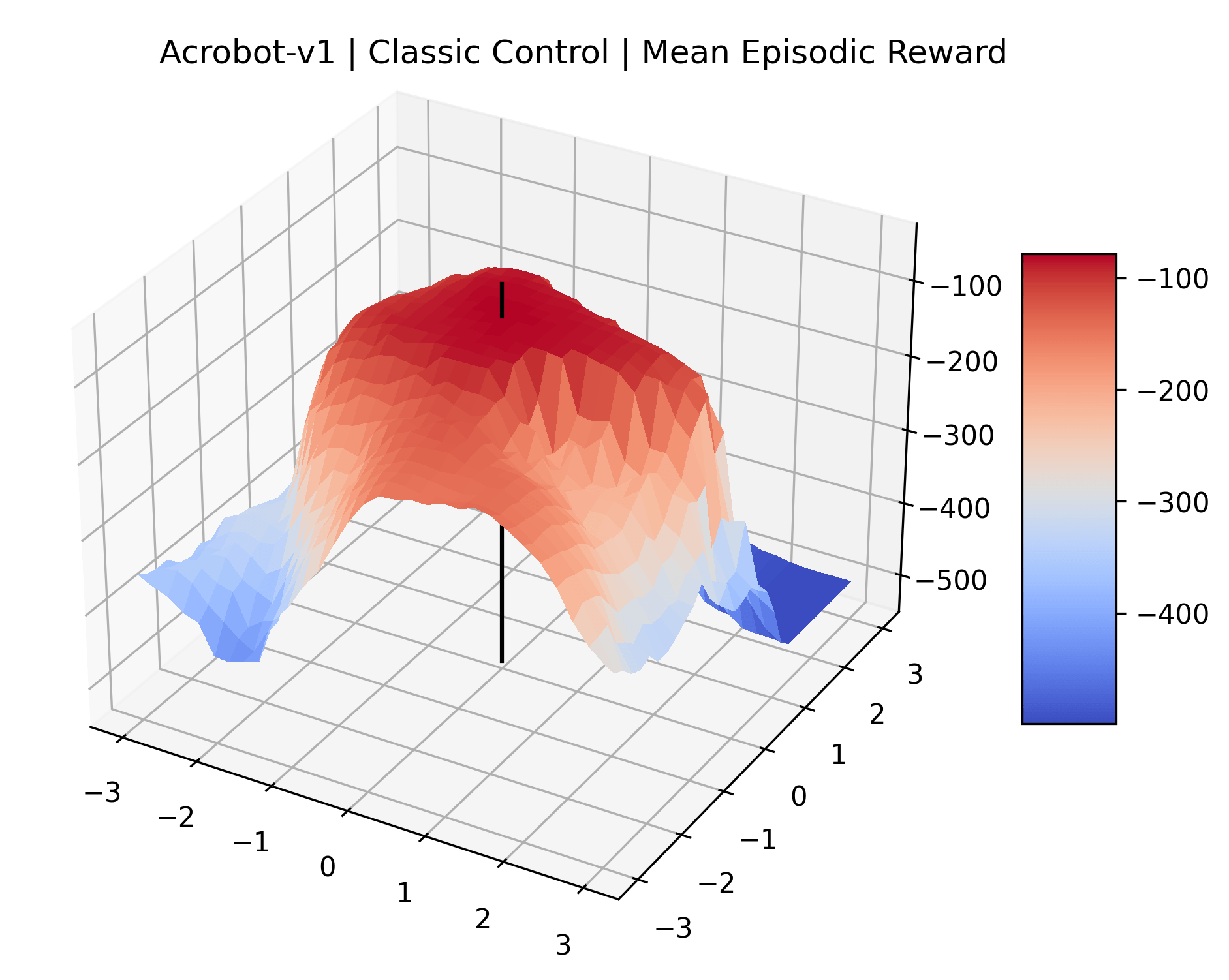} &
 \includegraphics[width=\variancescale]{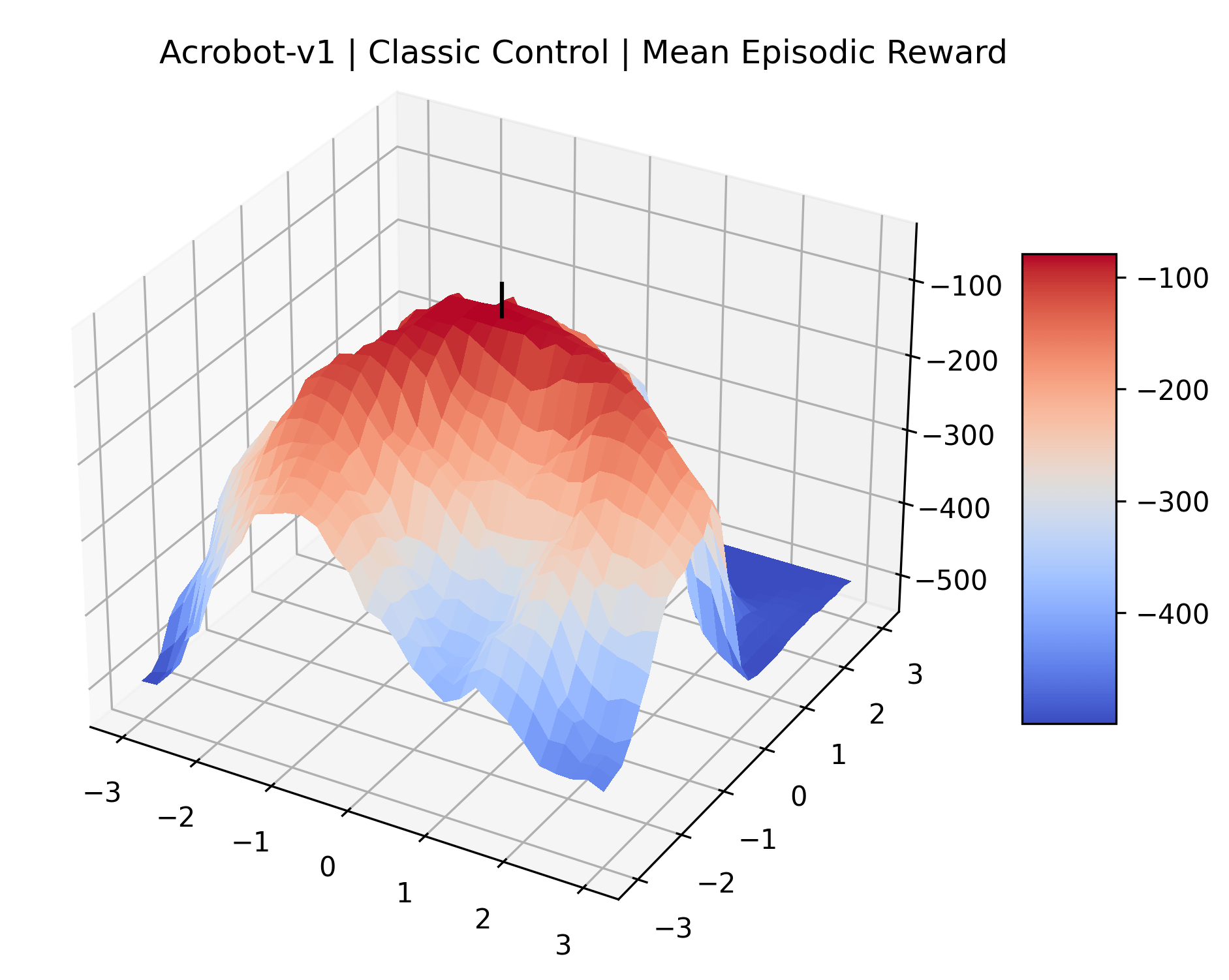} &
 \includegraphics[width=\variancescale]{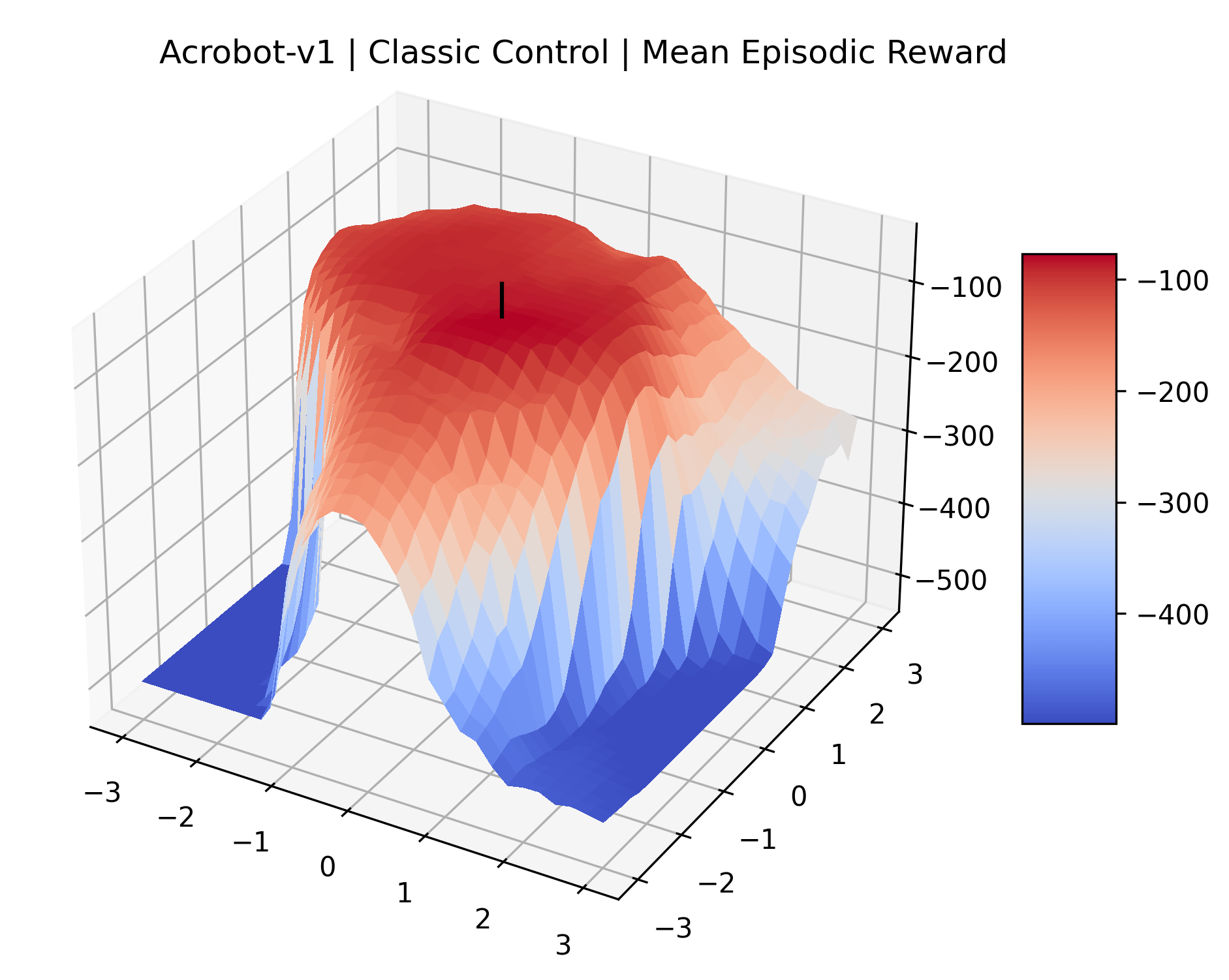} \\
\end{tabular}
\begin{tabular}{ccc}
 \includegraphics[width=\variancescale]{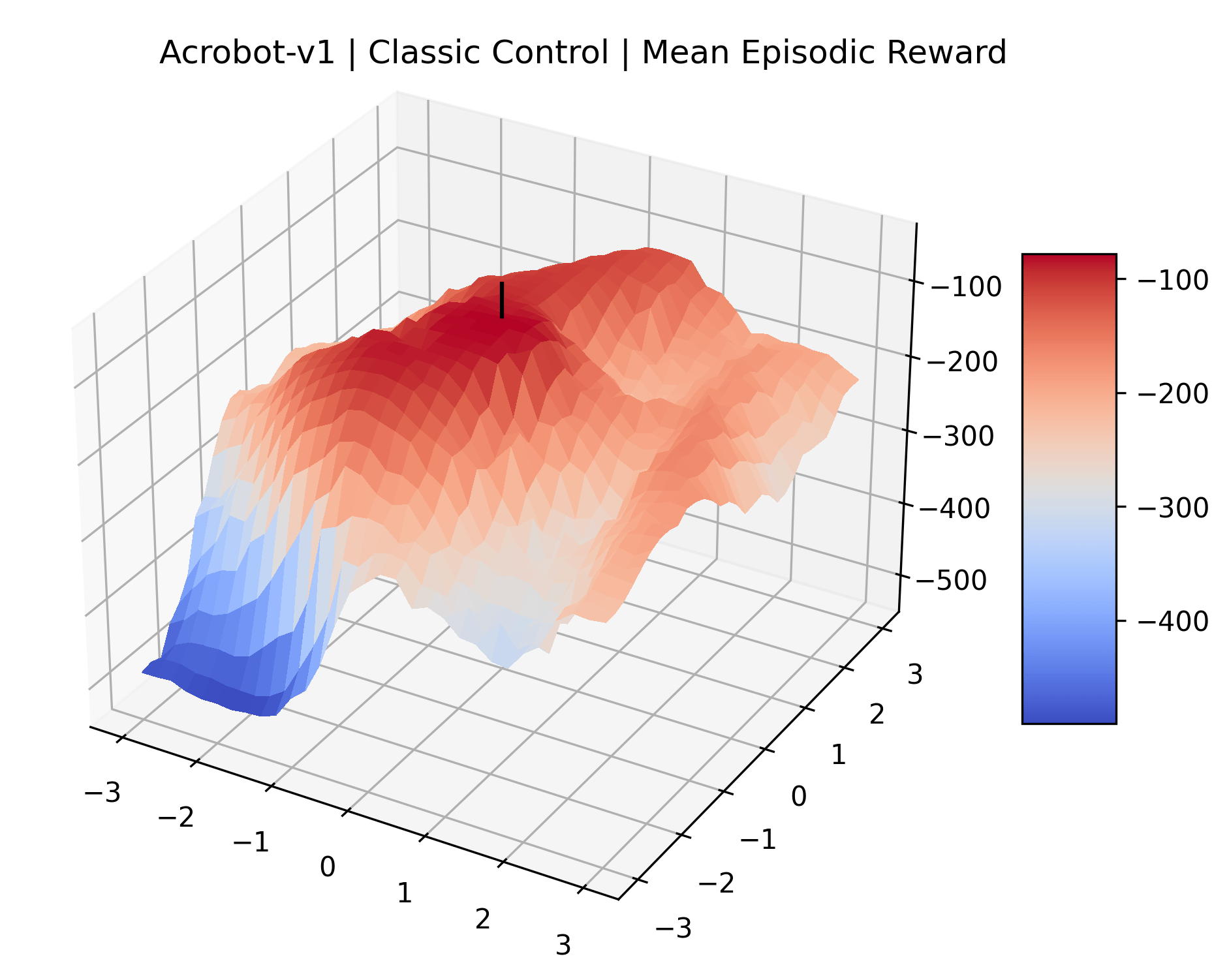} &
 \includegraphics[width=\variancescale]{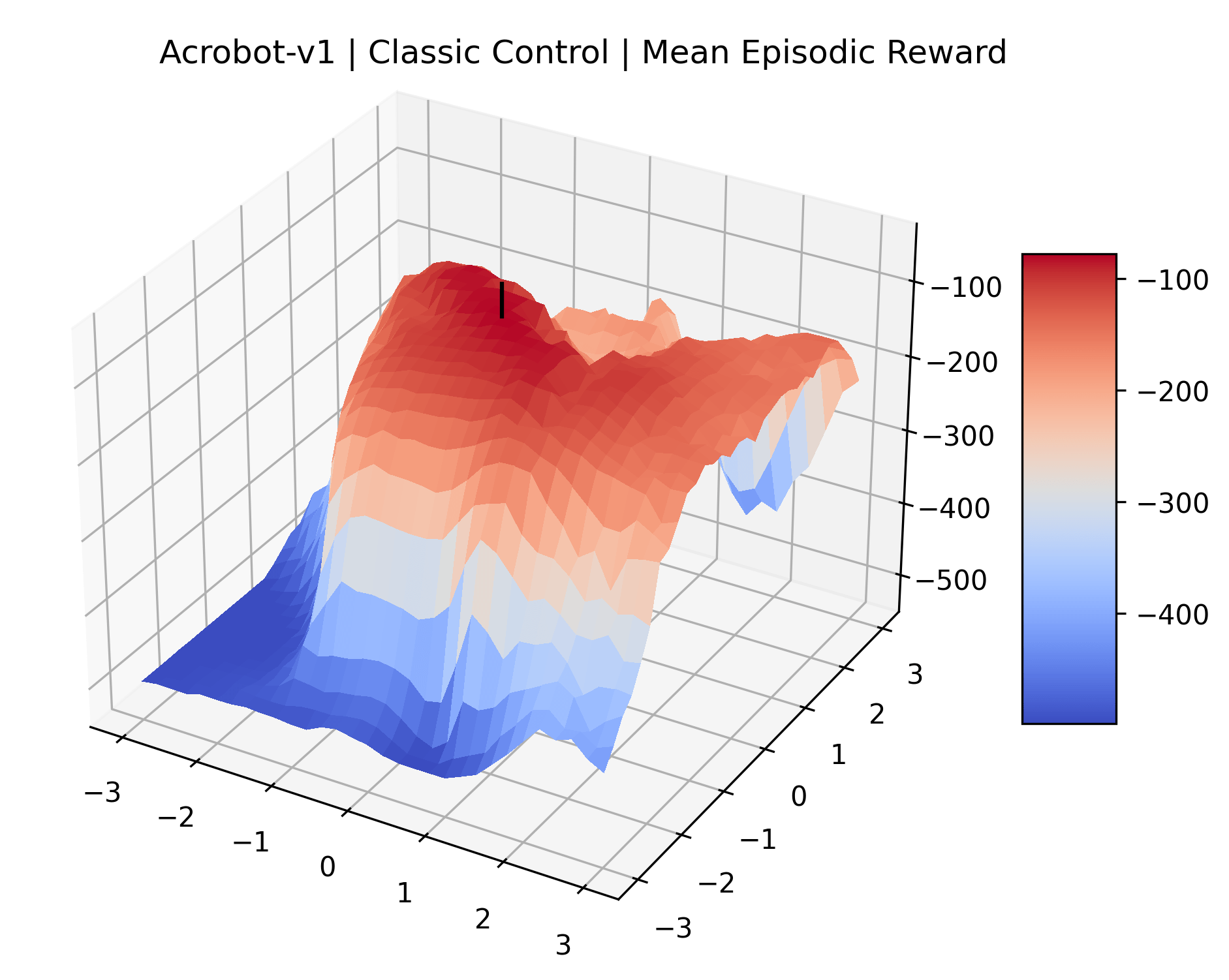} &
 \includegraphics[width=\variancescale]{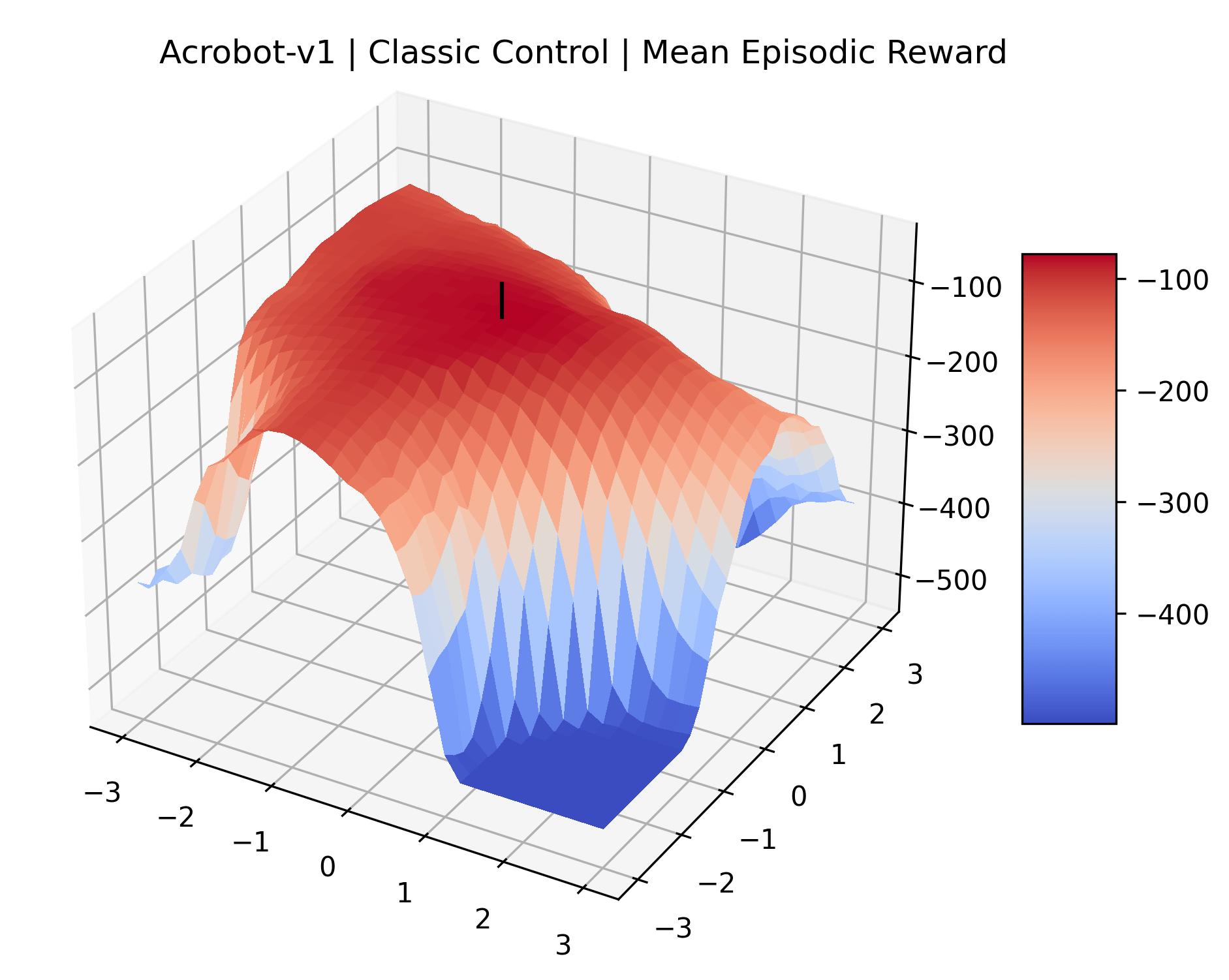} \\
\end{tabular}
\caption{18 training and plotting runs for the Classic Control Acrobot-v1 environment.}
\label{fig:classiccontrol_variance_table}
\end{figure*}

\pagebreak

\begin{figure*}[!htb]
\centering
\begin{tabular}{ccc}
 \includegraphics[width=\variancescale]{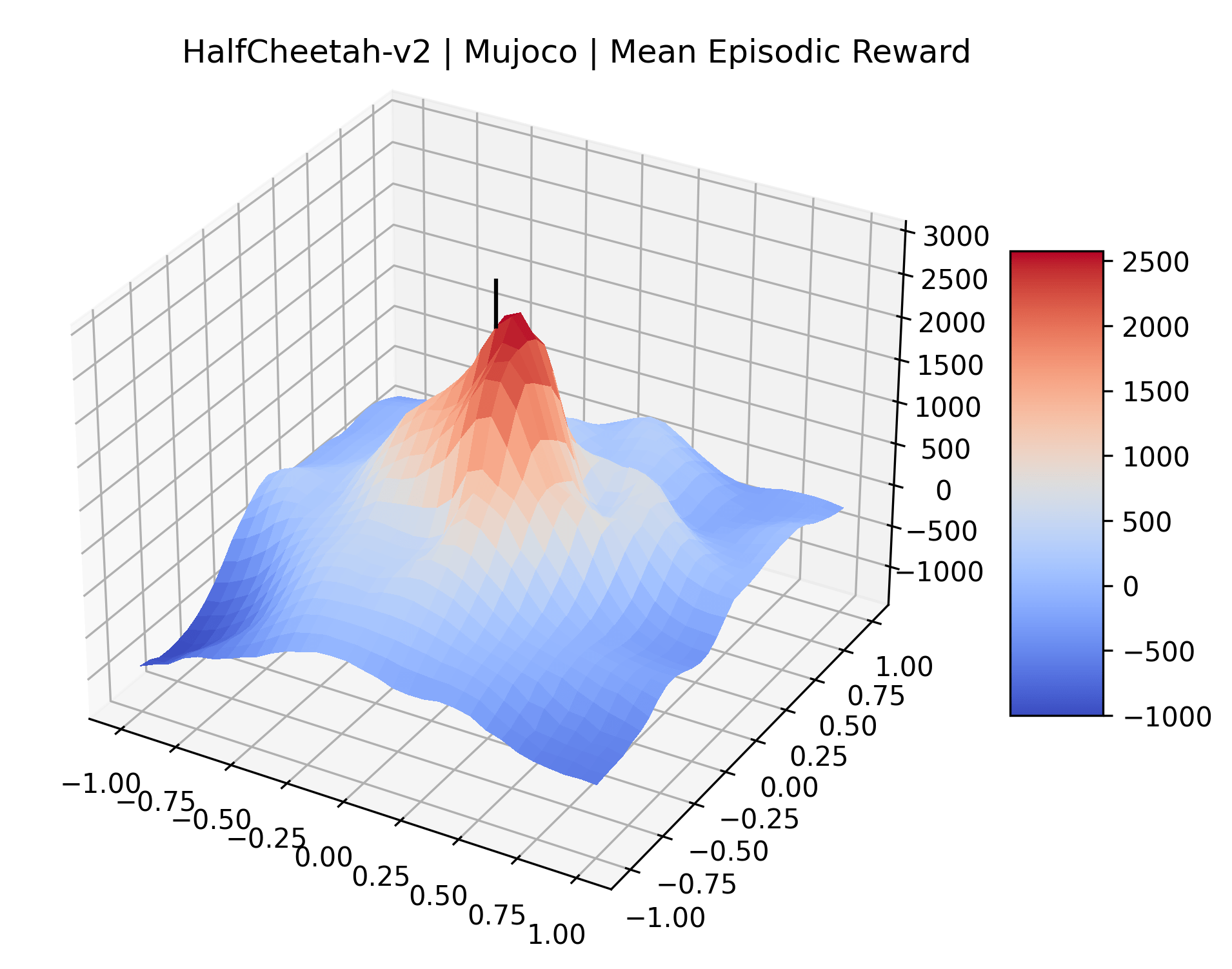} &
 \includegraphics[width=\variancescale]{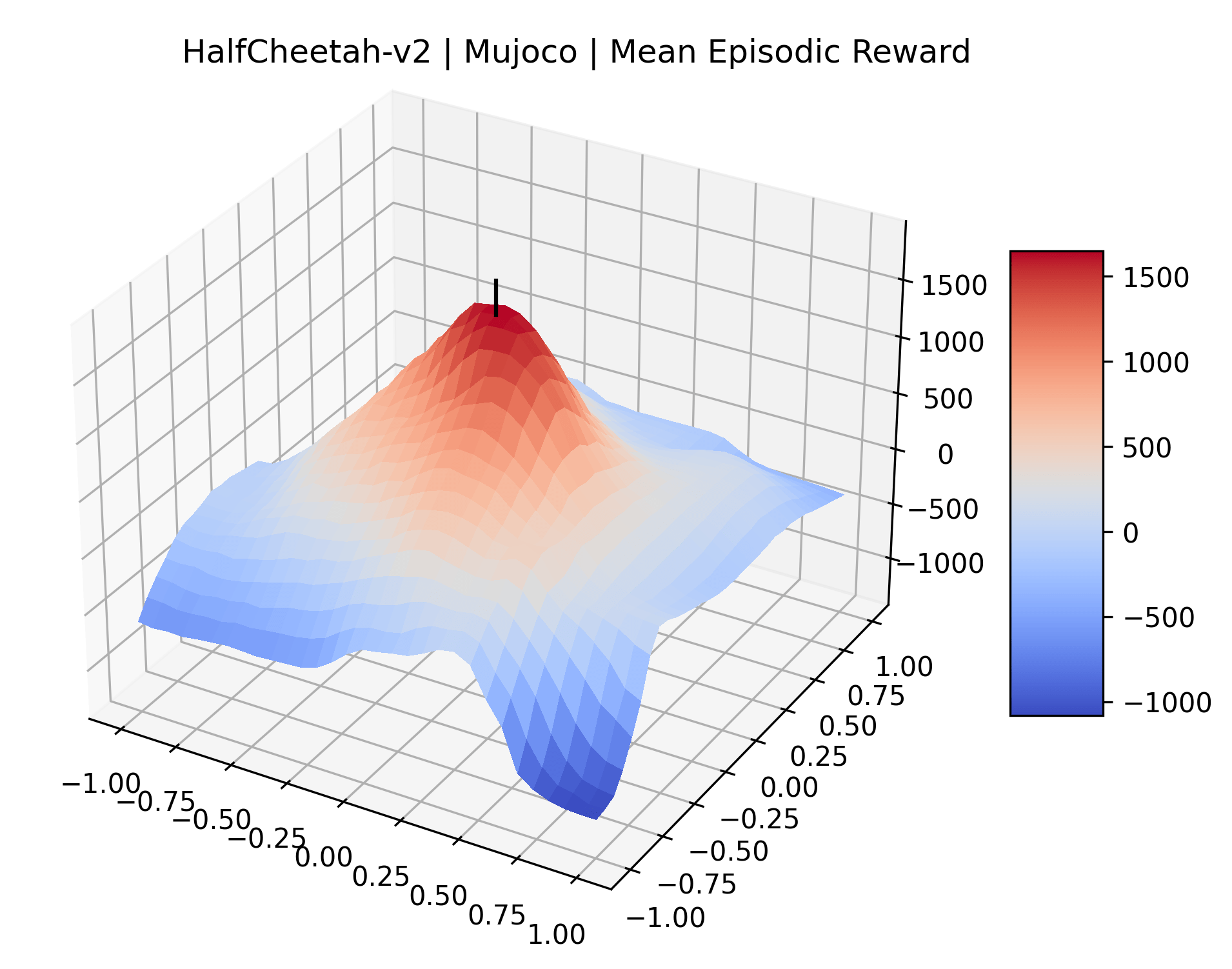} &
 \includegraphics[width=\variancescale]{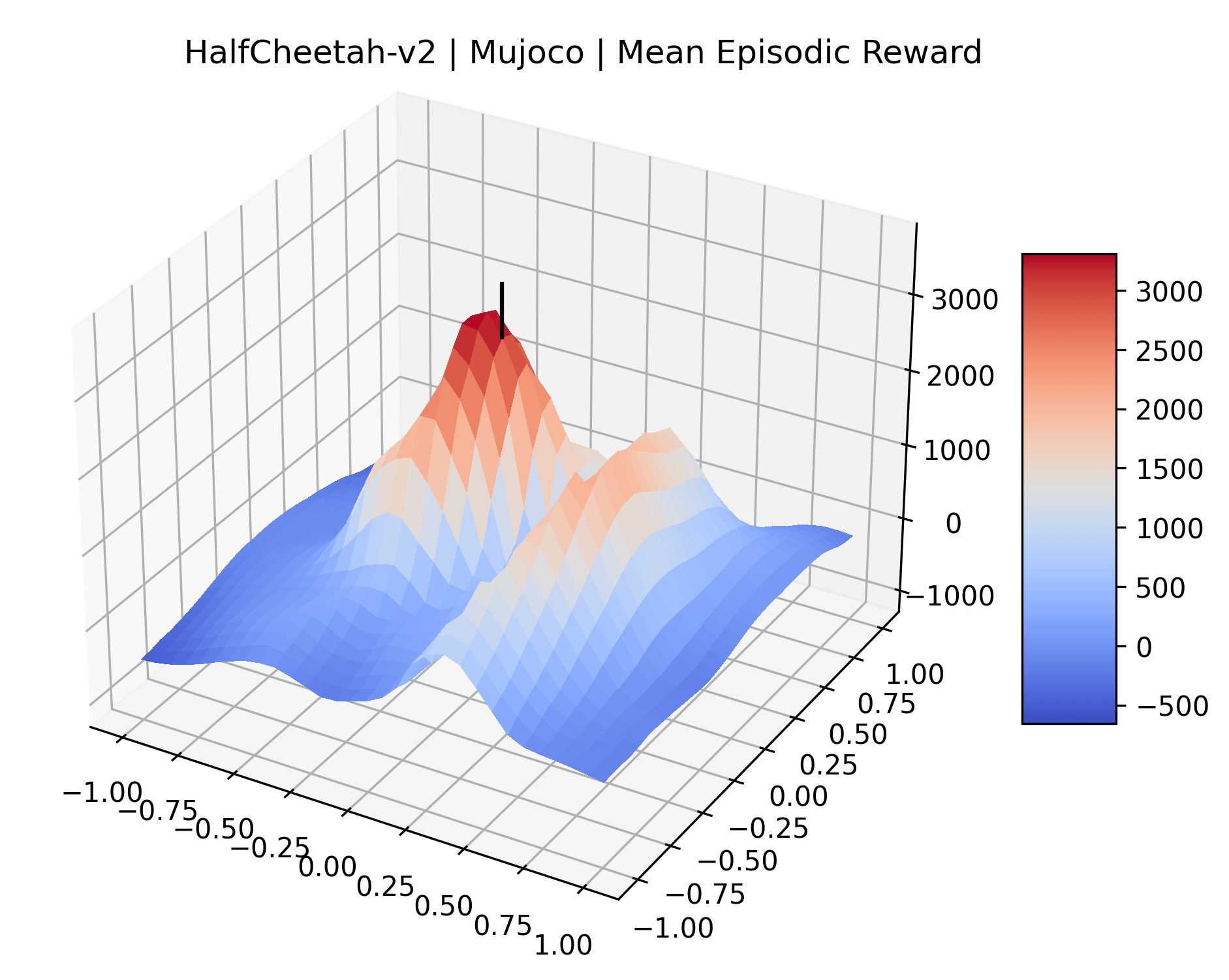} \\
\end{tabular}
\begin{tabular}{ccc}
 \includegraphics[width=\variancescale]{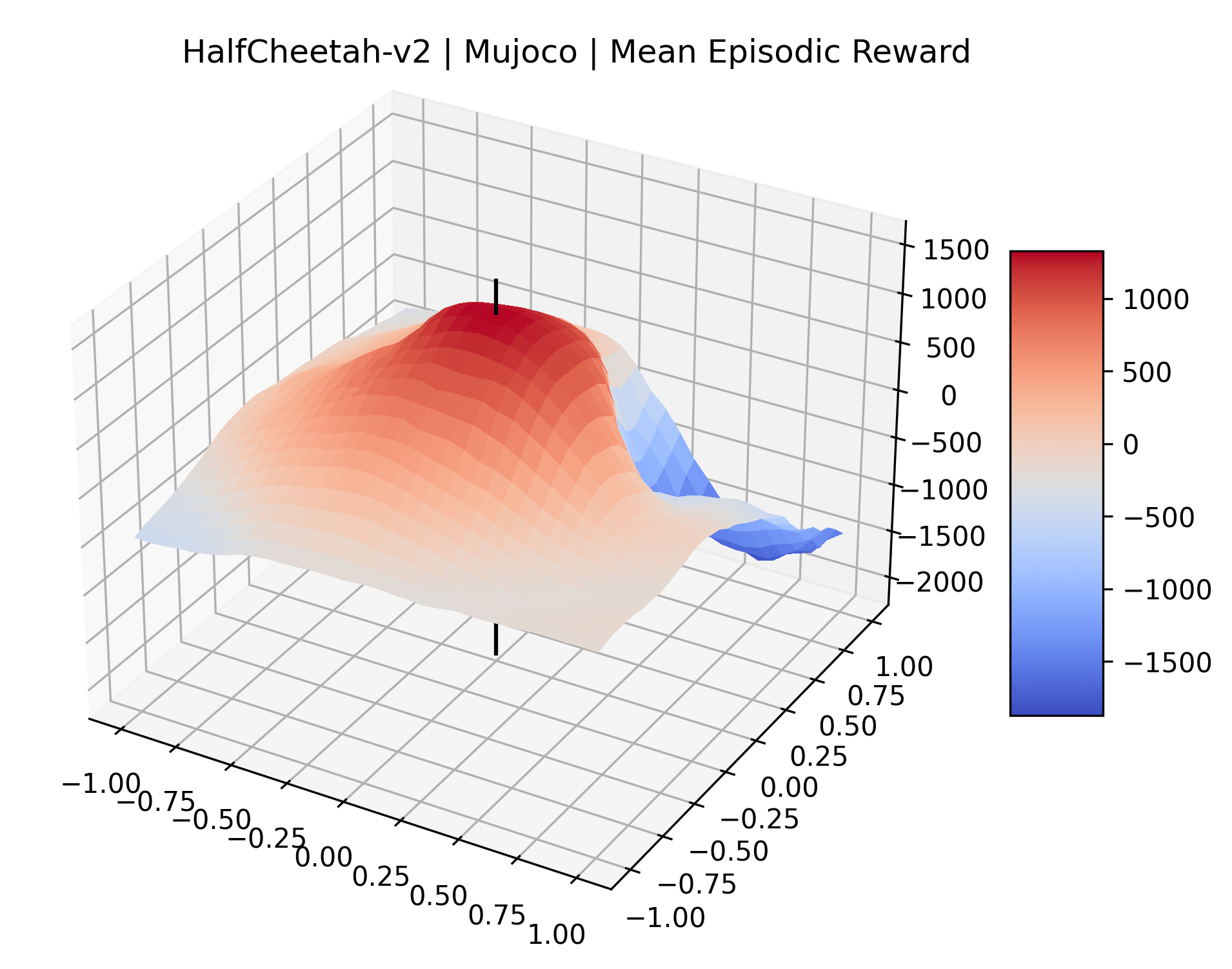} &
 \includegraphics[width=\variancescale]{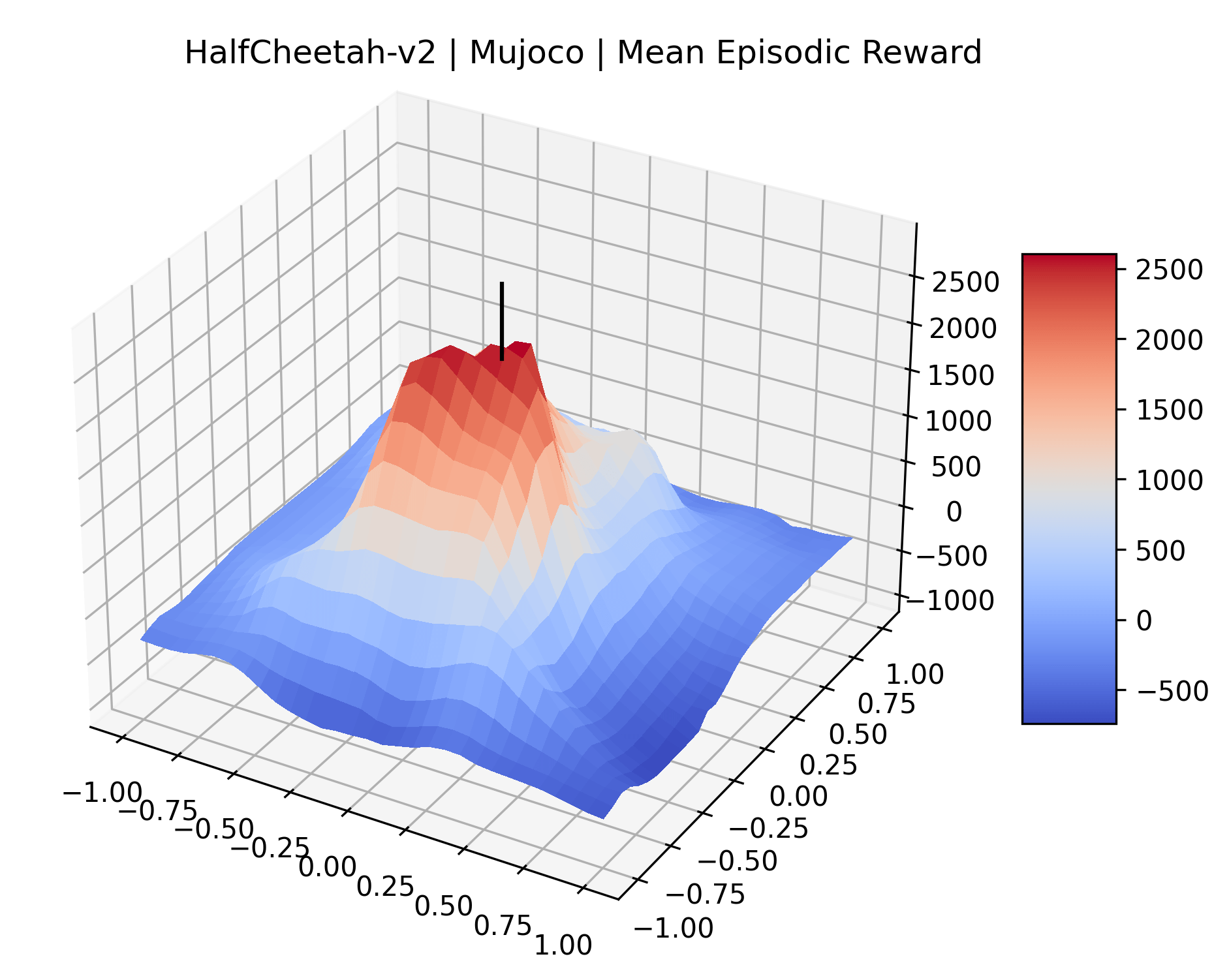} &
 \includegraphics[width=\variancescale]{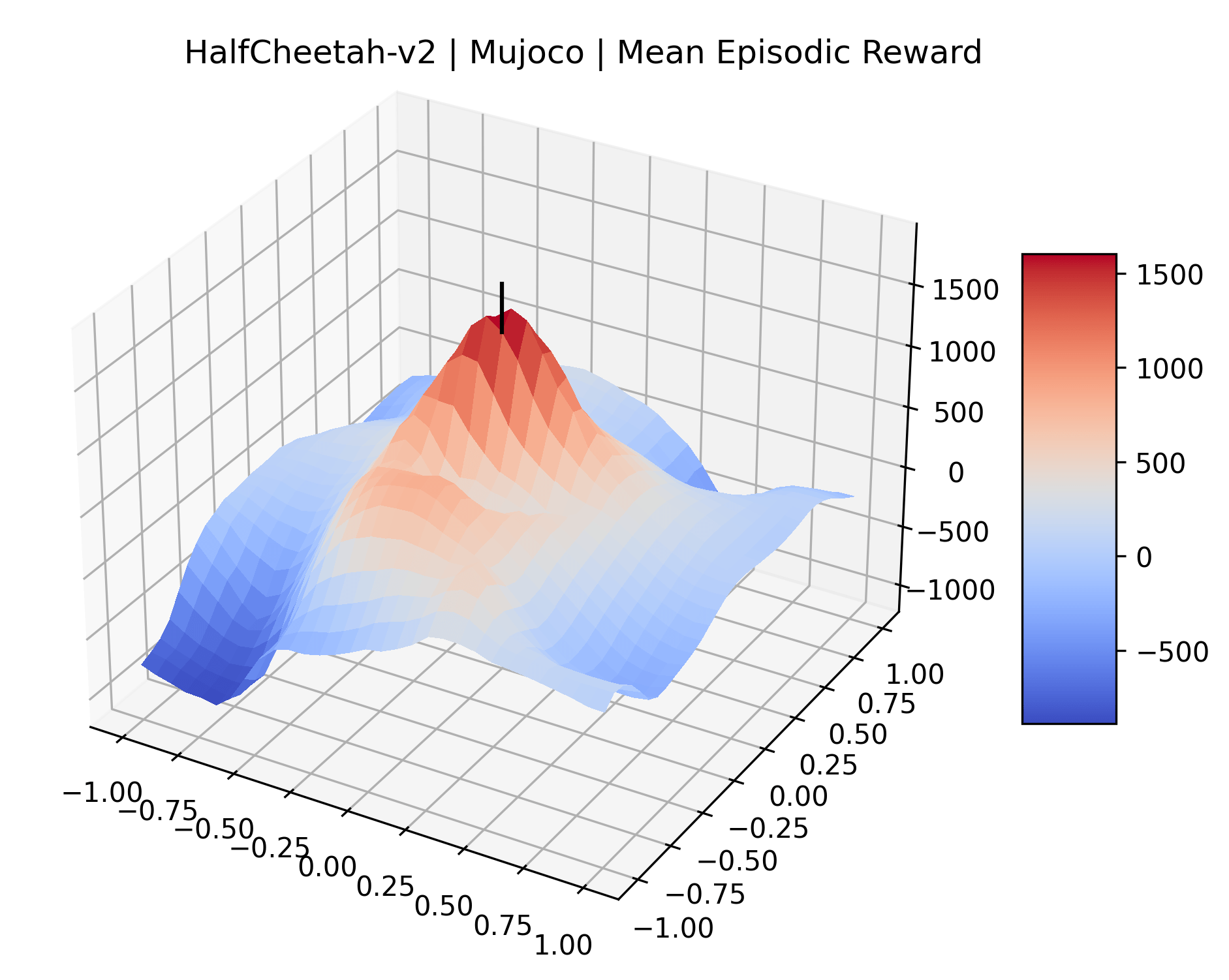} \\
\end{tabular}
\begin{tabular}{ccc}
 \includegraphics[width=\variancescale]{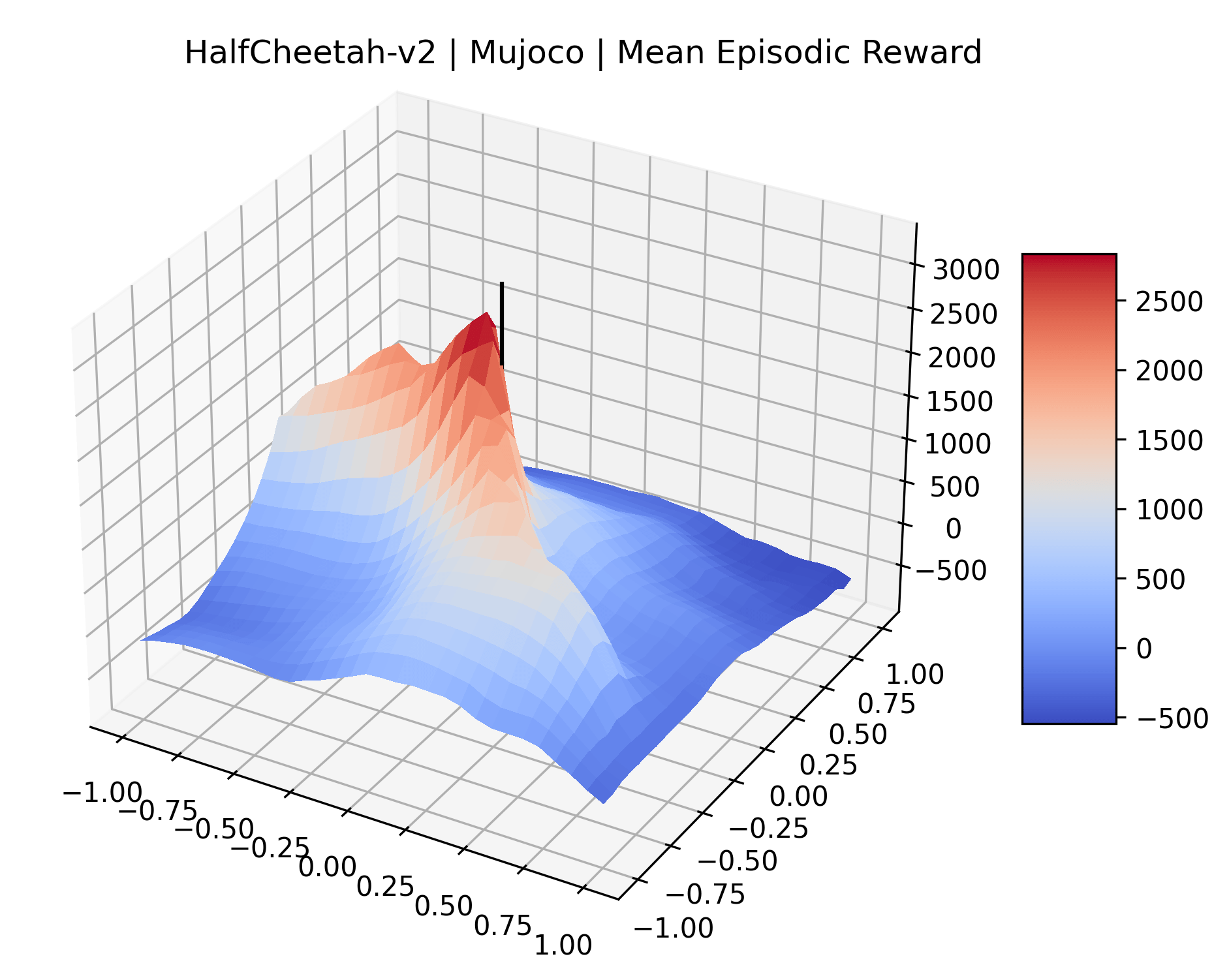} &
 \includegraphics[width=\variancescale]{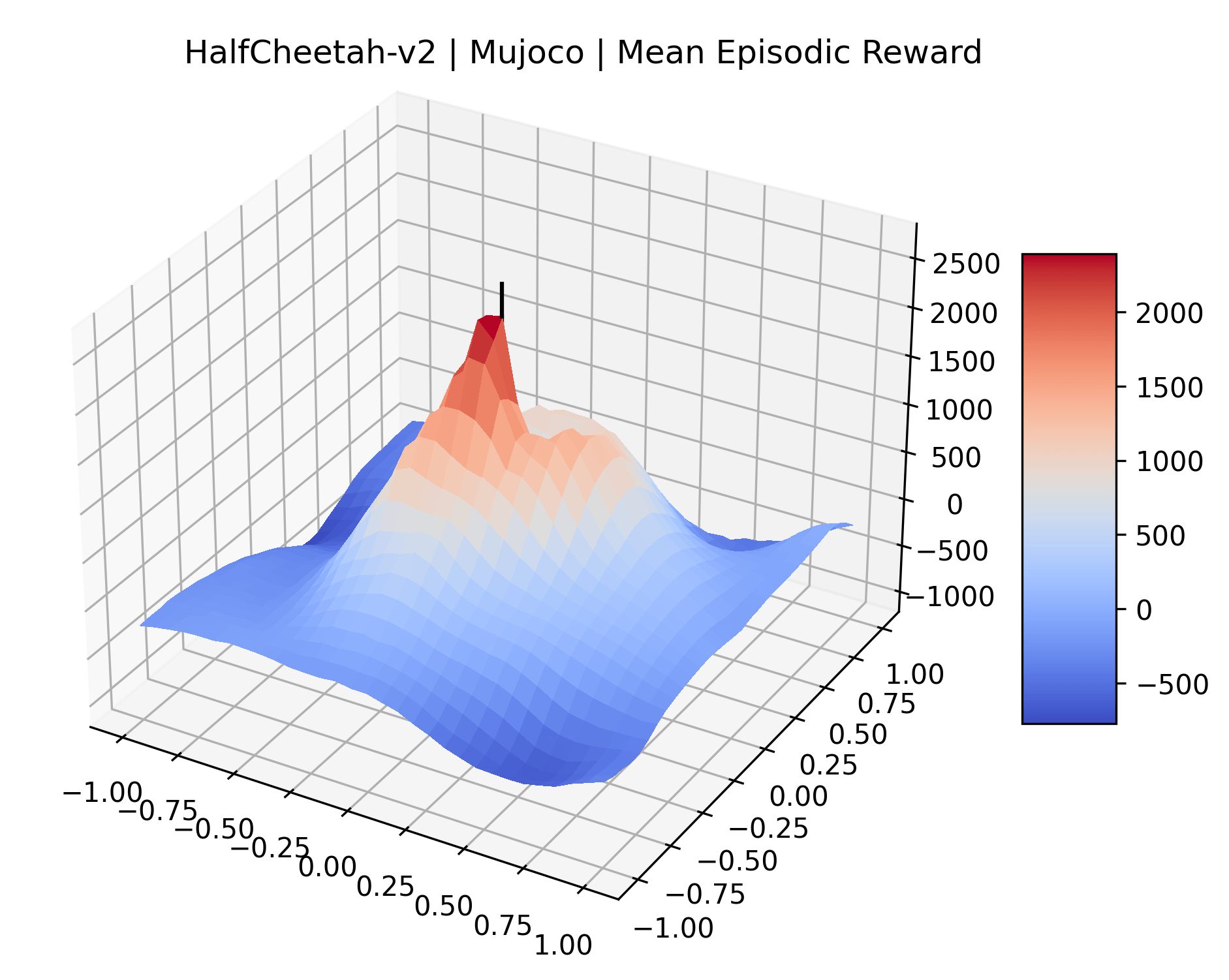} &
 \includegraphics[width=\variancescale]{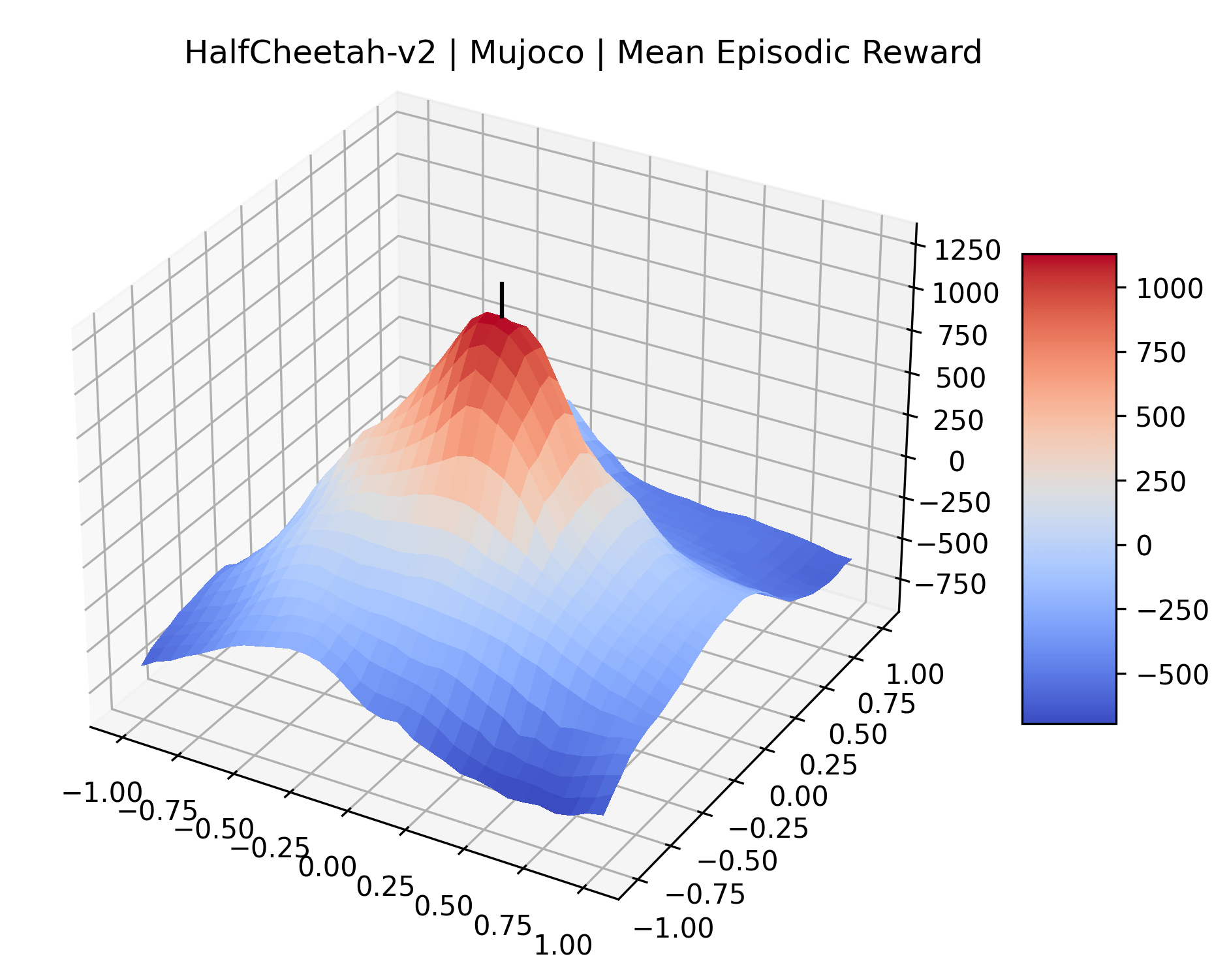} \\
\end{tabular}
\begin{tabular}{ccc}
 \includegraphics[width=\variancescale]{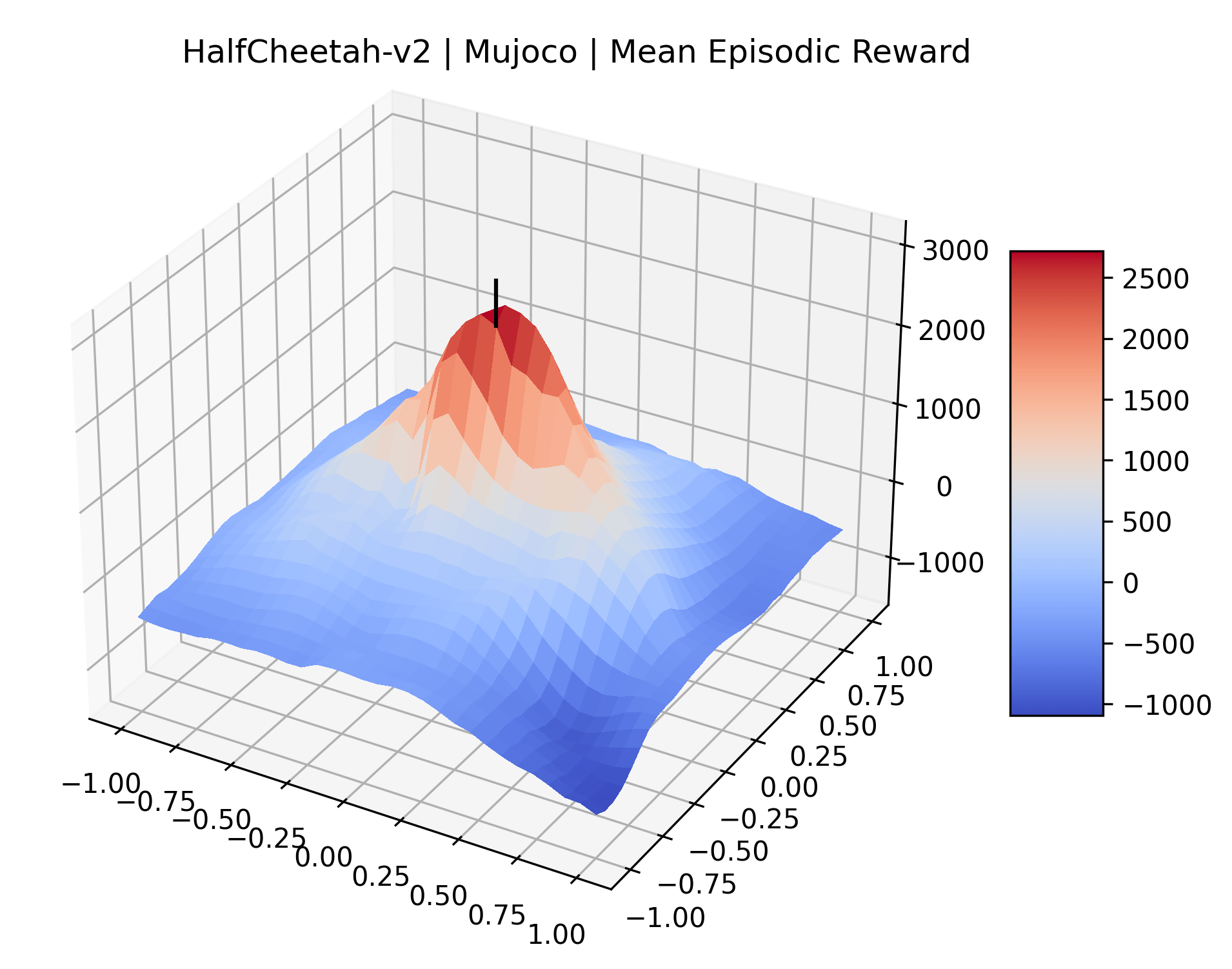} &
 \includegraphics[width=\variancescale]{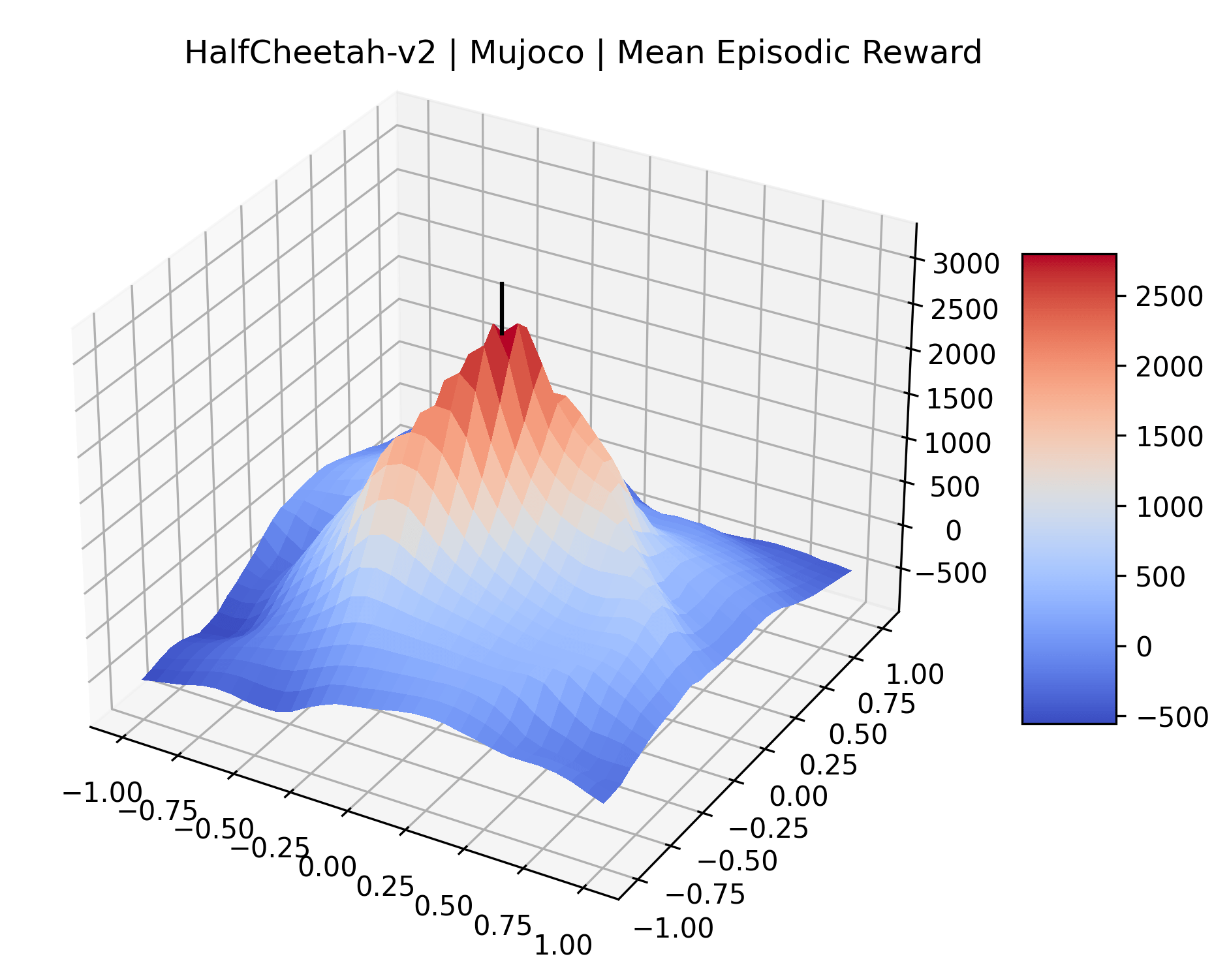} &
 \includegraphics[width=\variancescale]{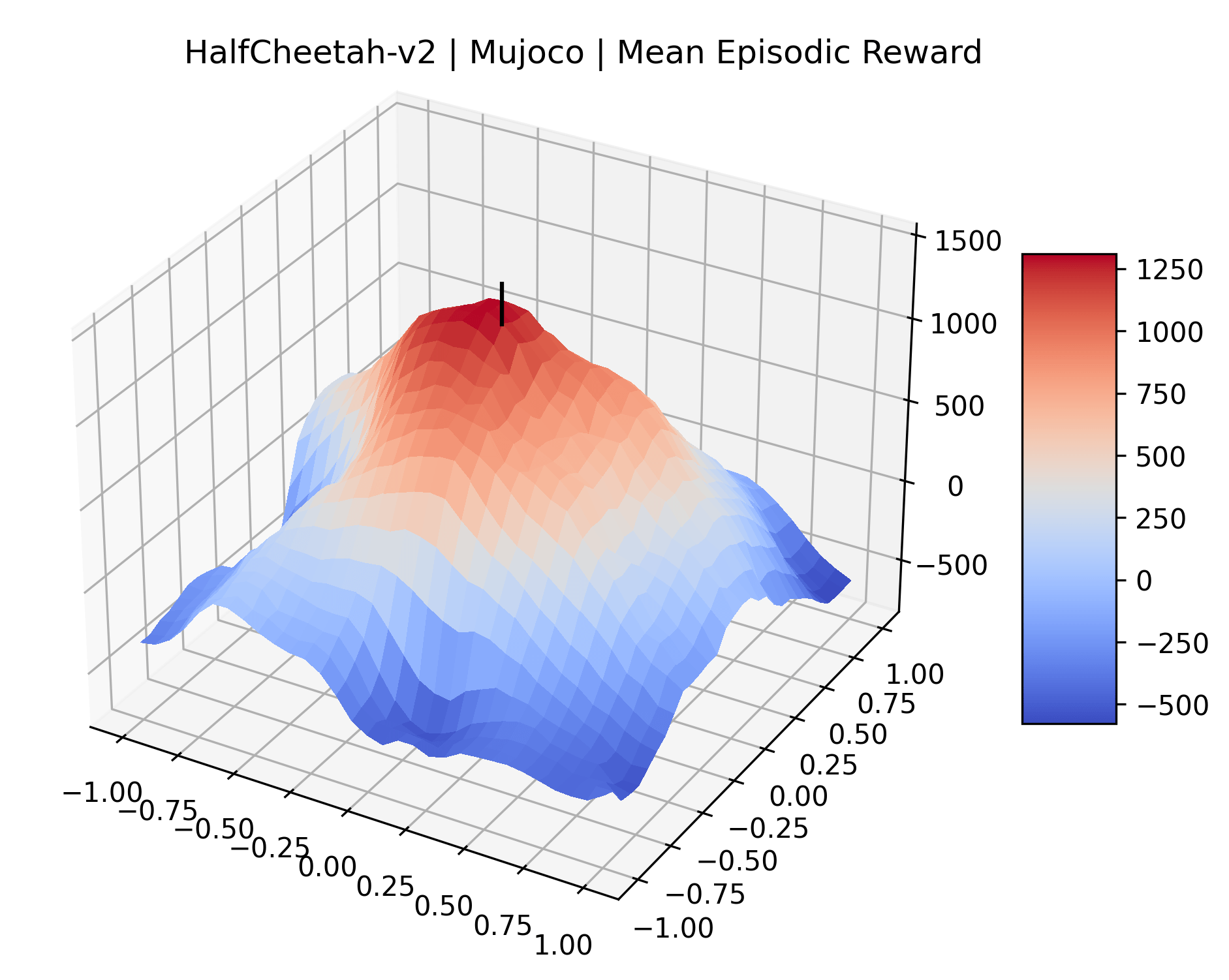} \\
\end{tabular}
\begin{tabular}{ccc}
 \includegraphics[width=\variancescale]{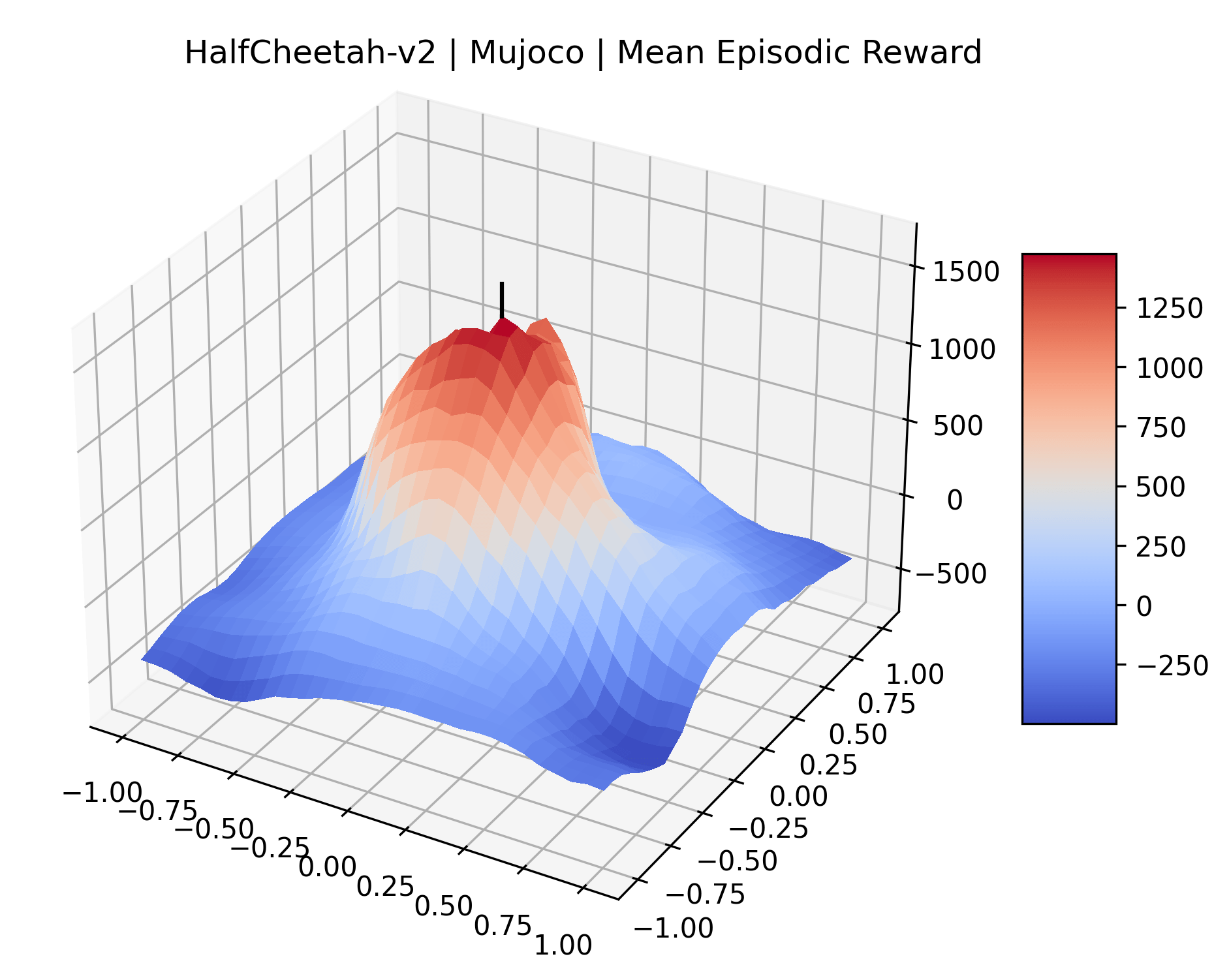} &
 \includegraphics[width=\variancescale]{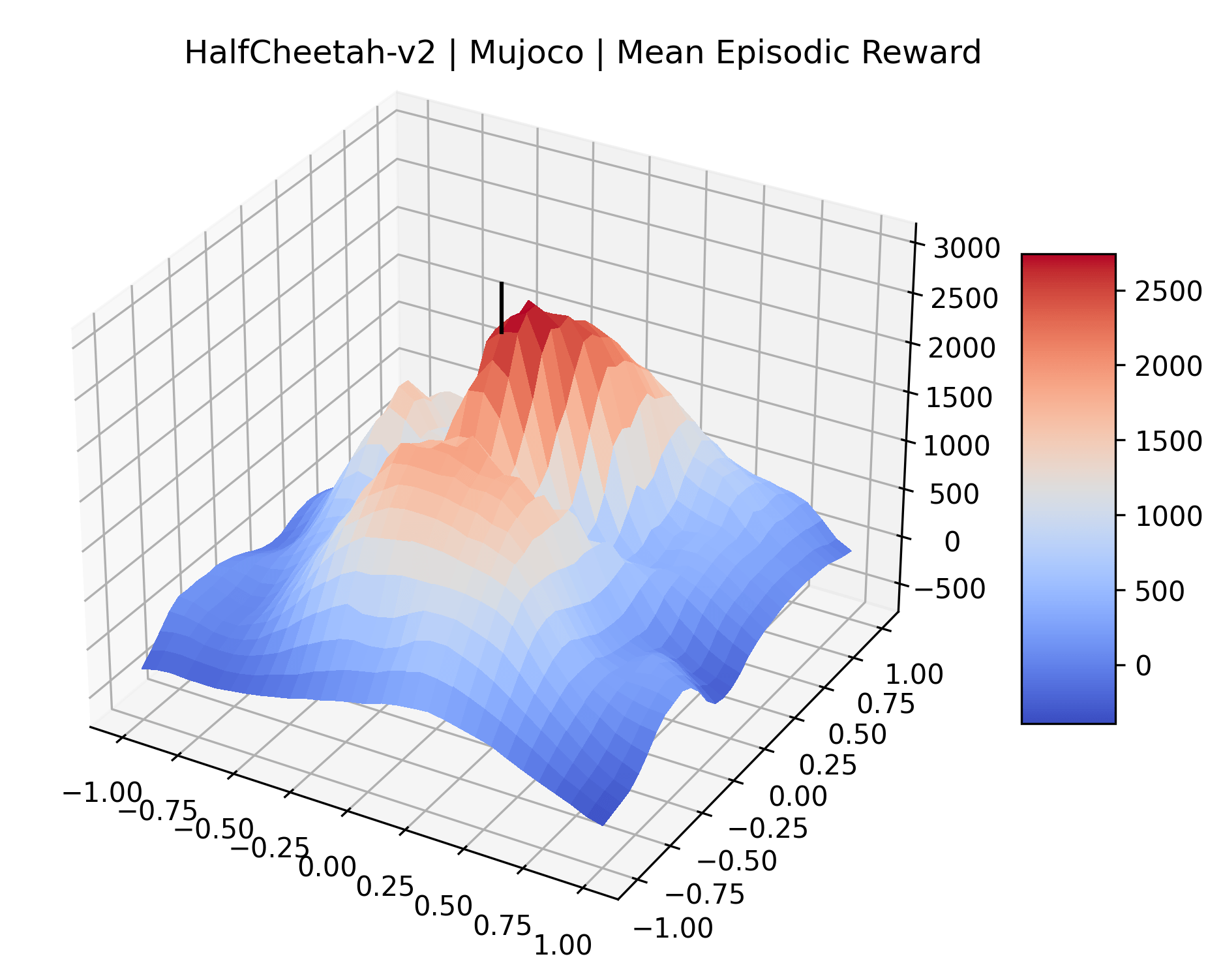} &
 \includegraphics[width=\variancescale]{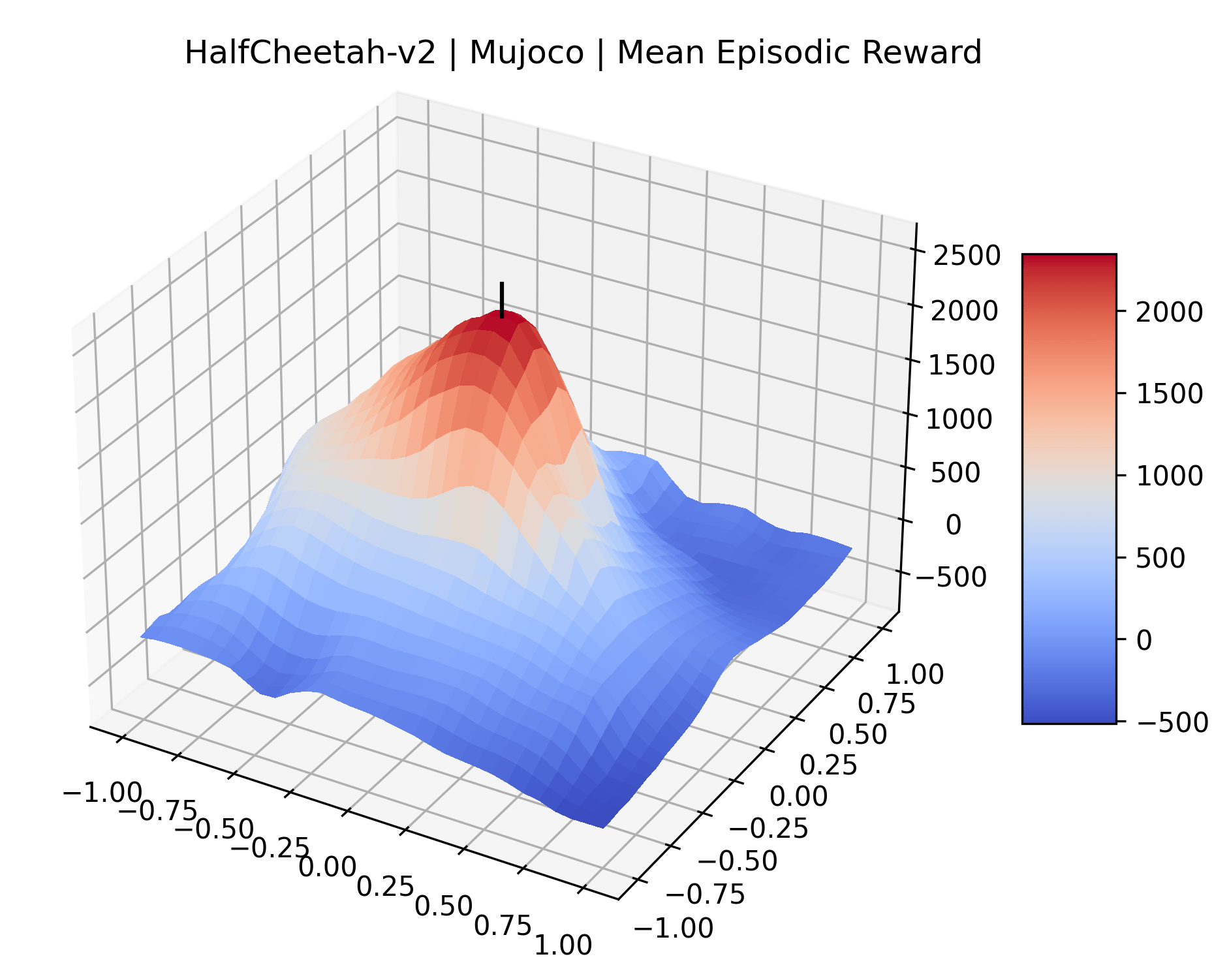} \\
\end{tabular}
\begin{tabular}{ccc}
 \includegraphics[width=\variancescale]{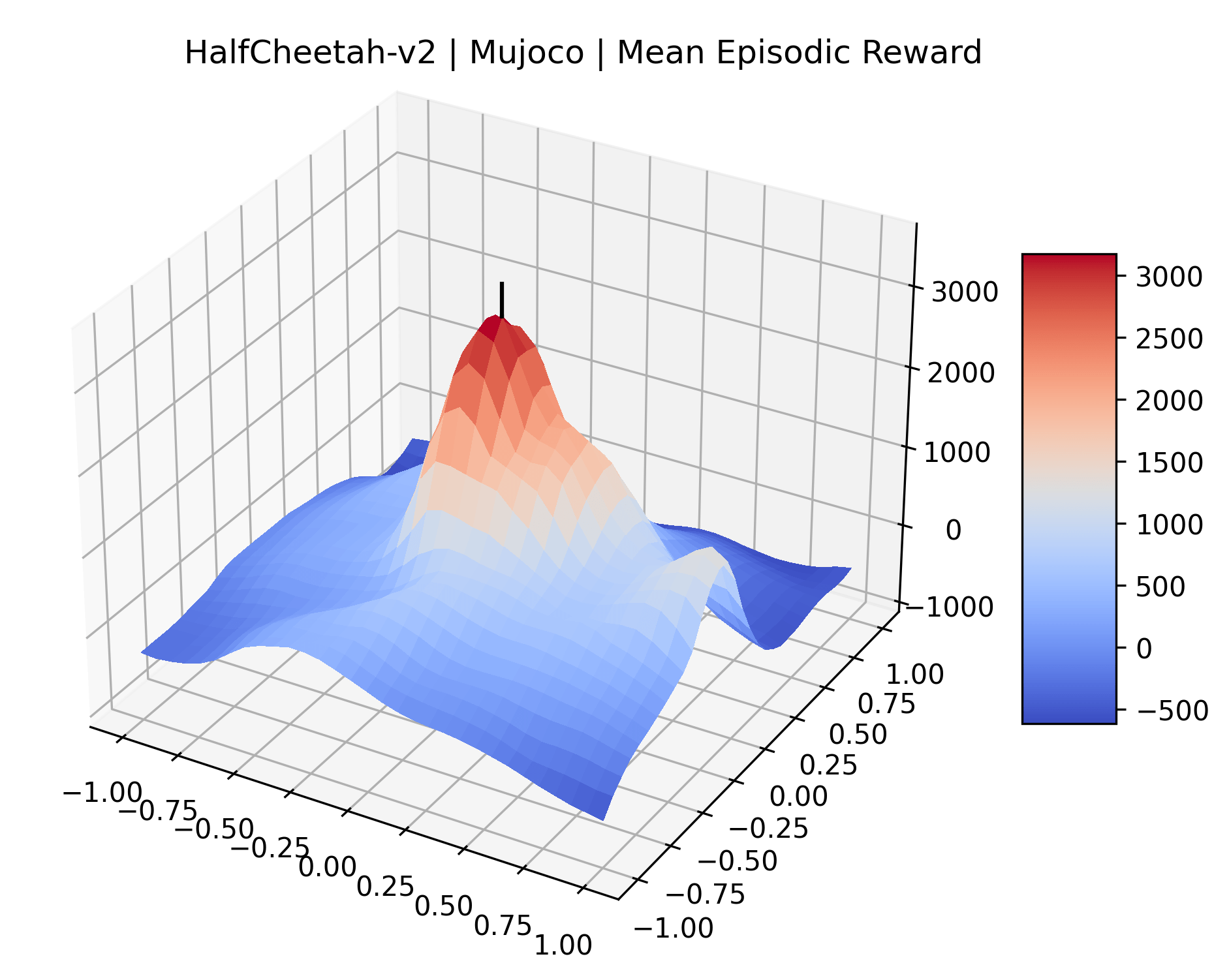} &
 \includegraphics[width=\variancescale]{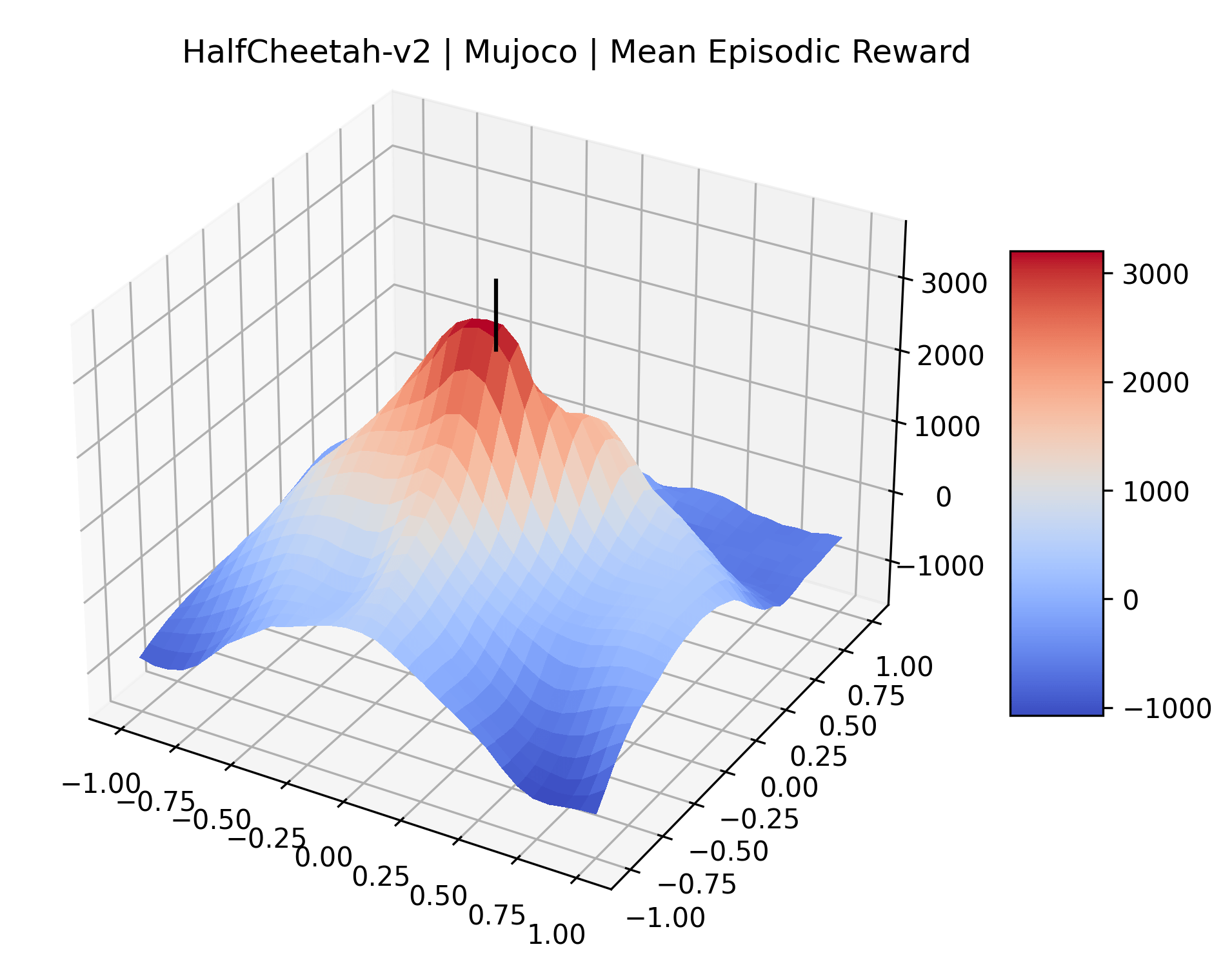} &
 \includegraphics[width=\variancescale]{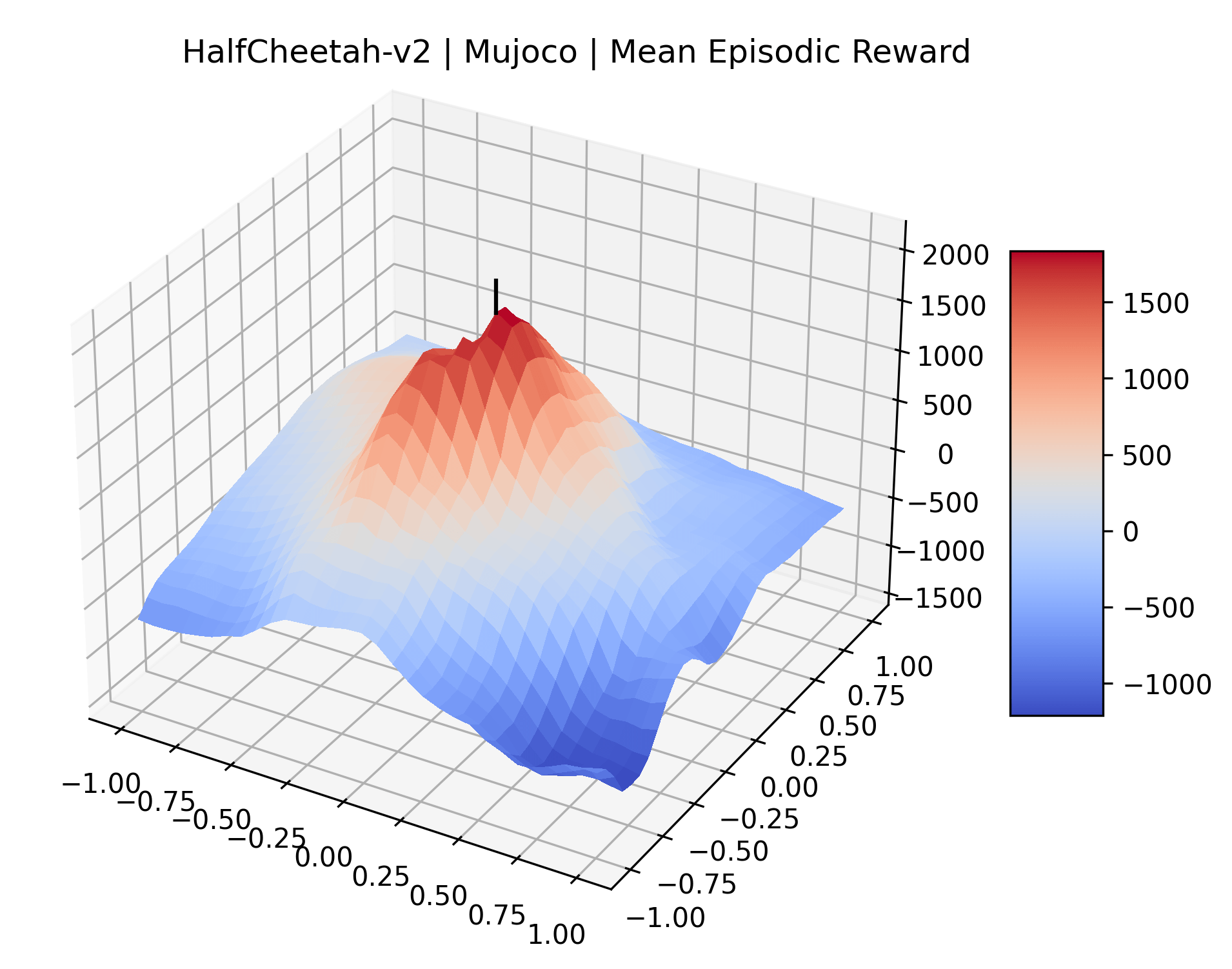} \\
\end{tabular}
\caption{18 training and plotting runs for the MuJoCo HalfCheetah-v2 environment.}
\label{fig:mujococ_variance_table}
\end{figure*}

\pagebreak

\begin{figure*}[!htb]
\centering
\begin{tabular}{ccc}
 \includegraphics[width=\variancescale]{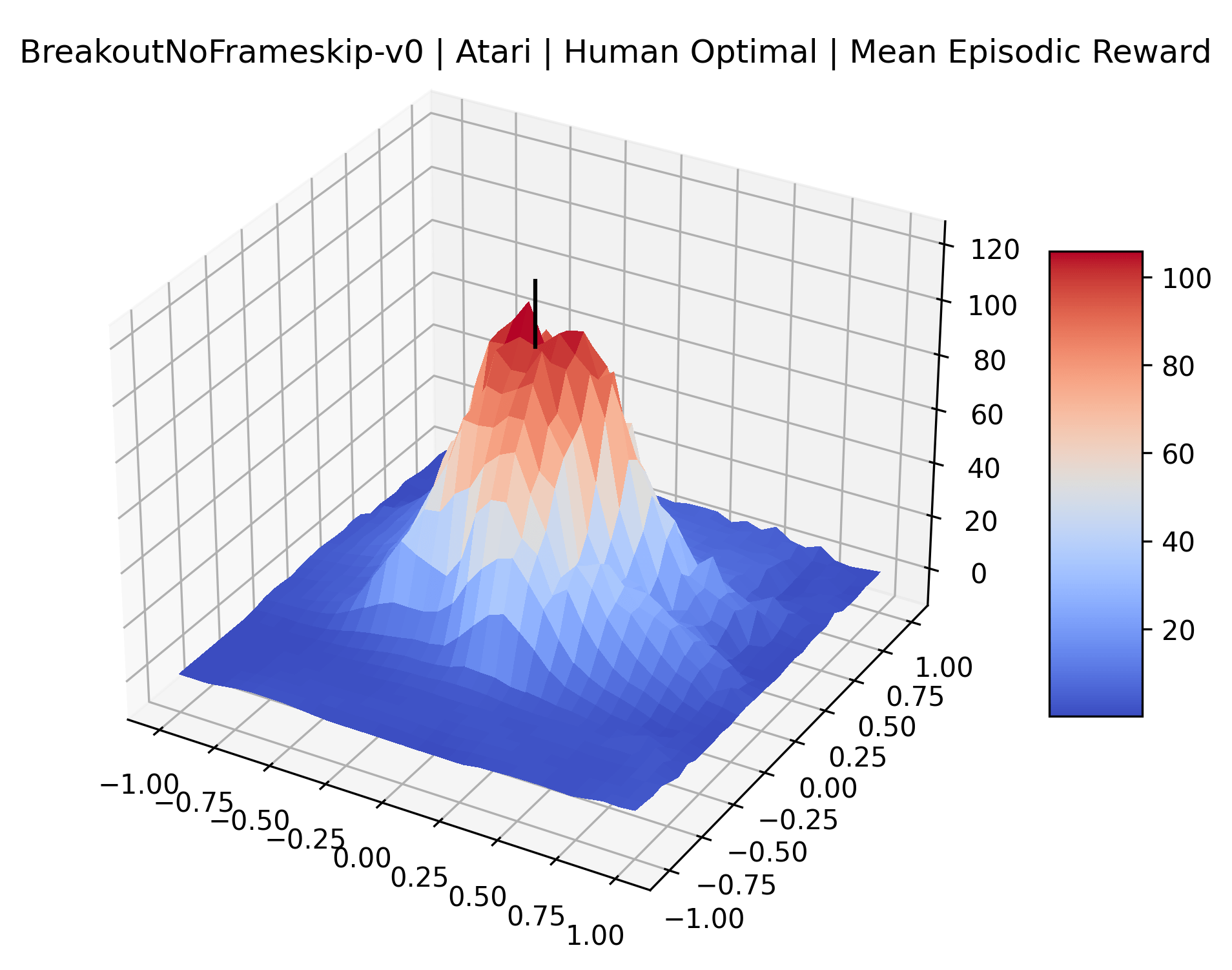} &
 \includegraphics[width=\variancescale]{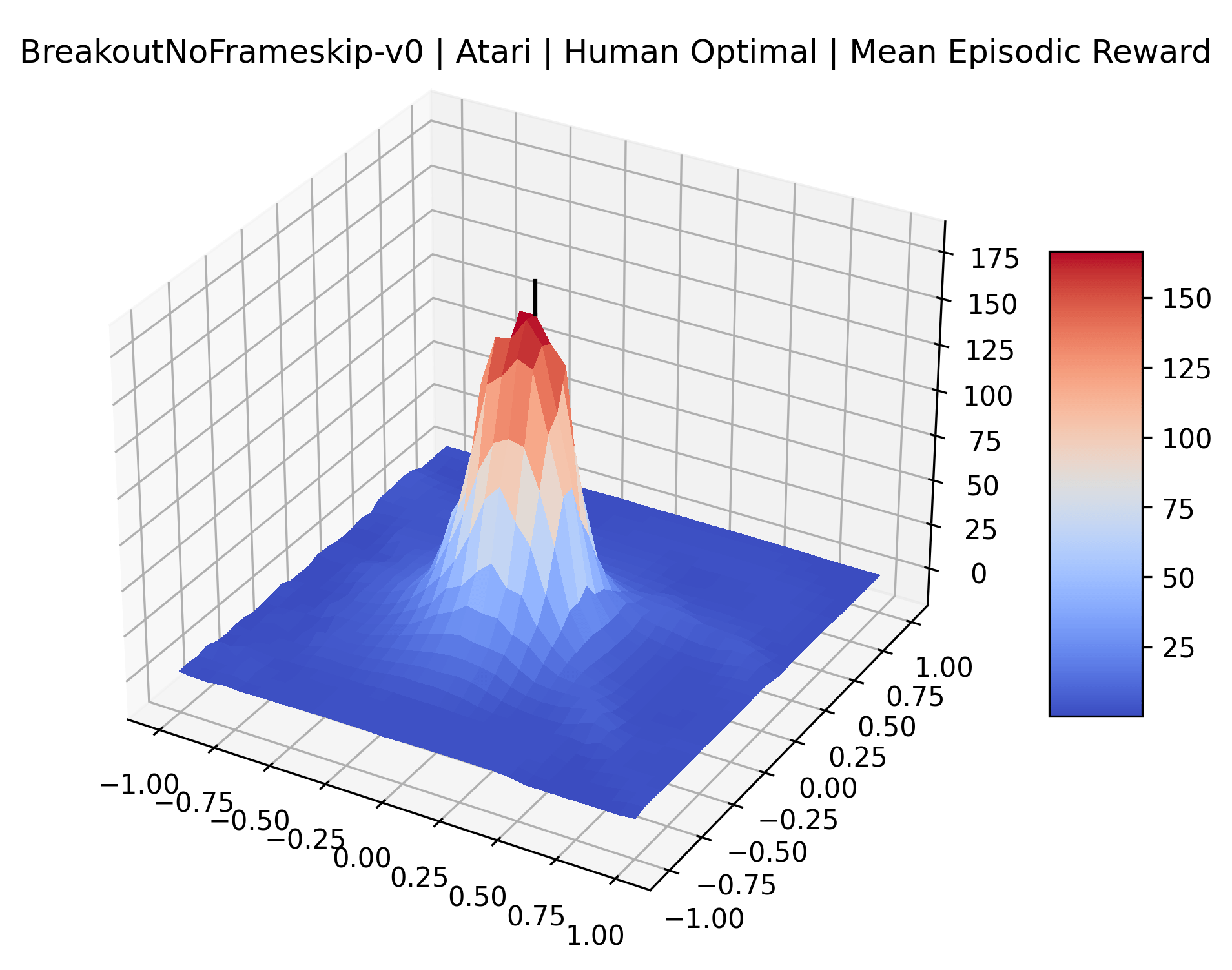} &
 \includegraphics[width=\variancescale]{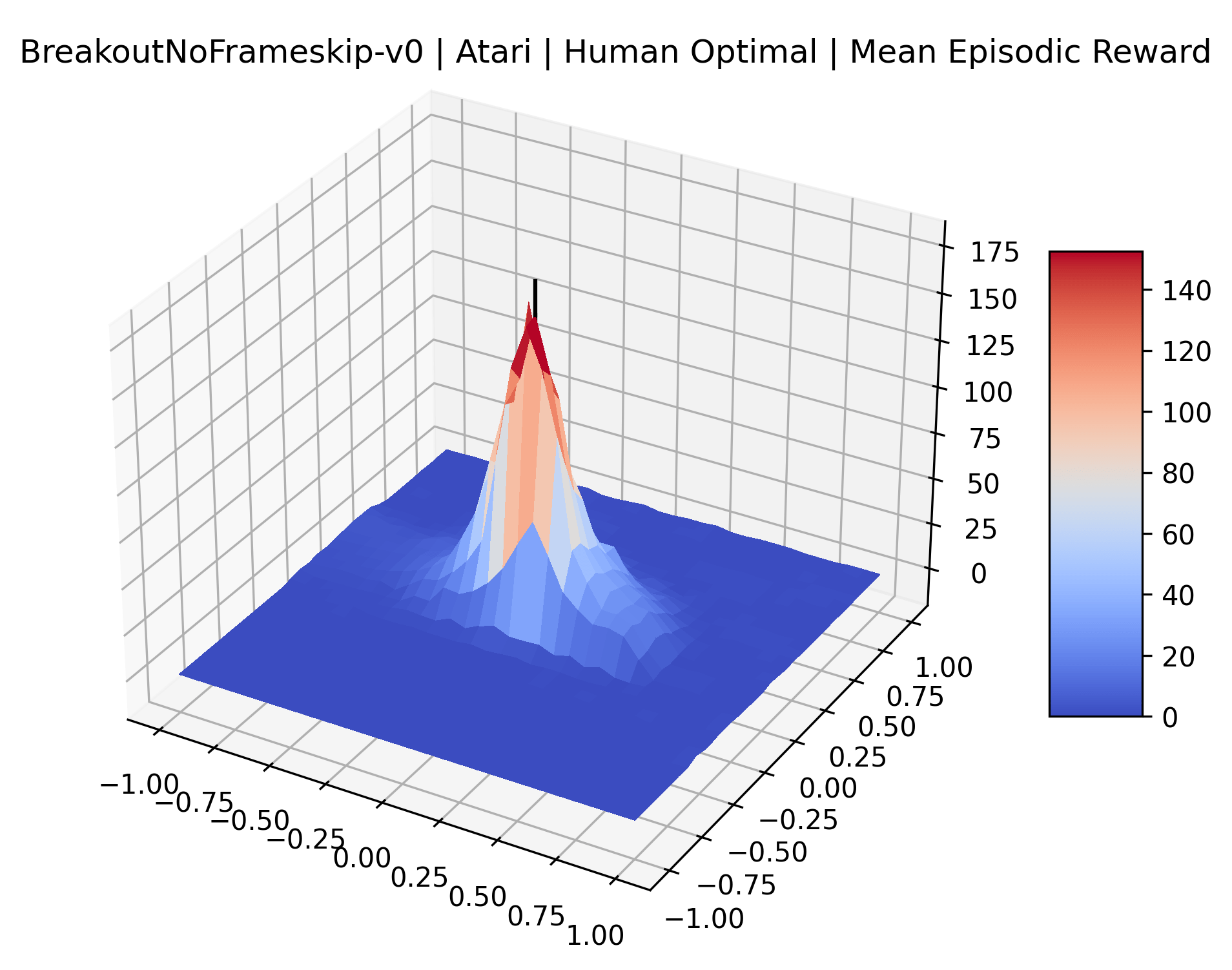} \\
\end{tabular}
\begin{tabular}{ccc}
 \includegraphics[width=\variancescale]{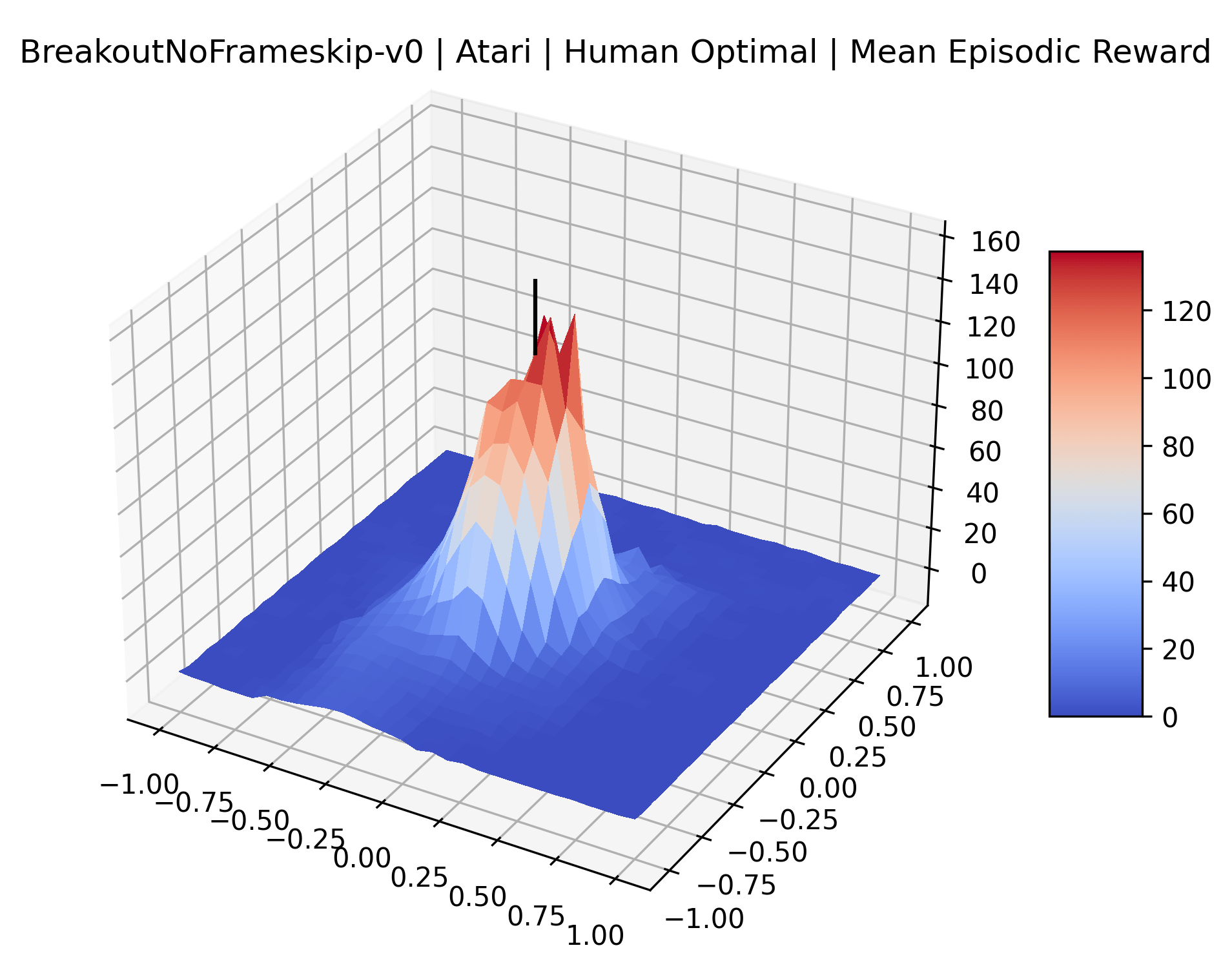} &
 \includegraphics[width=\variancescale]{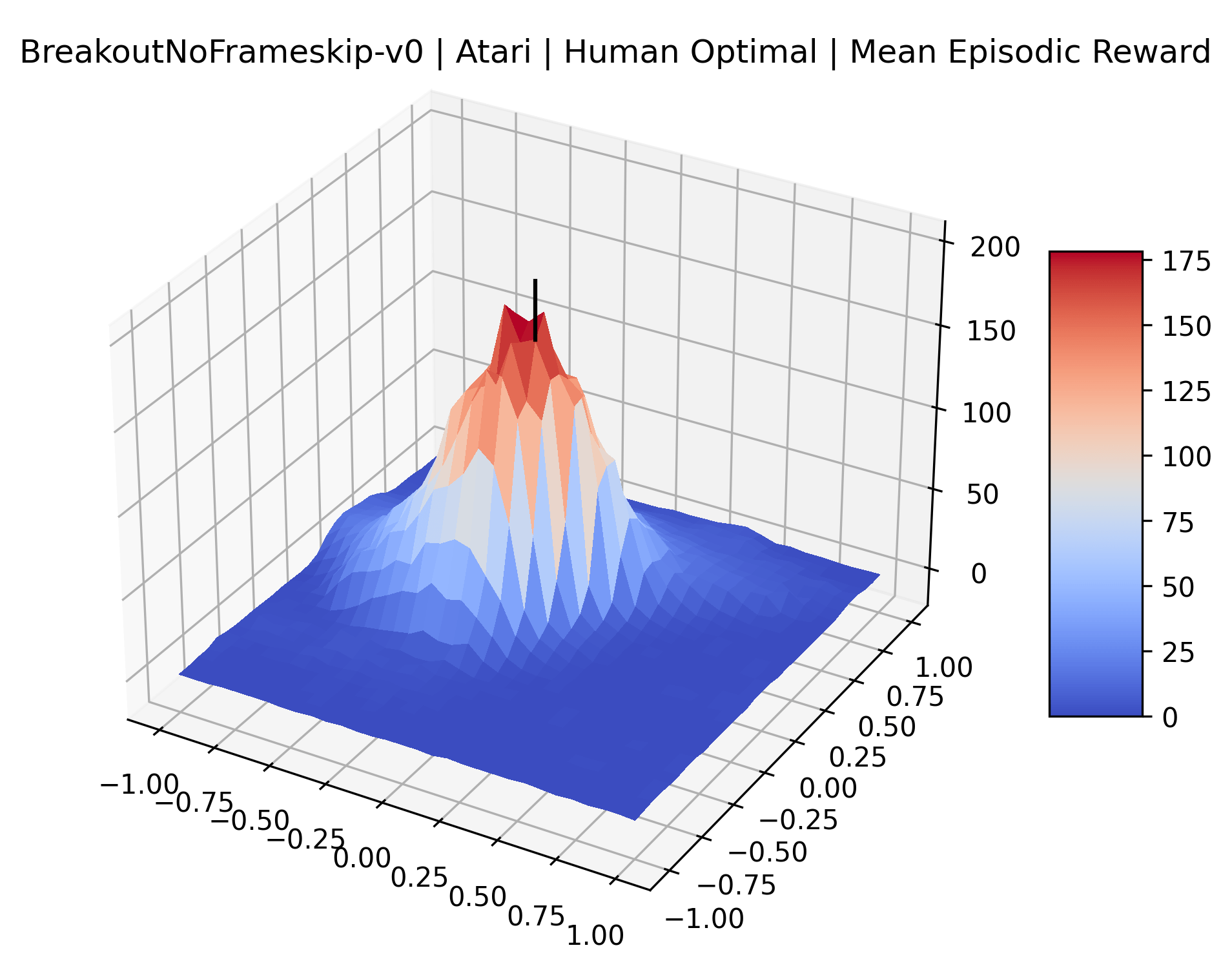} &
 \includegraphics[width=\variancescale]{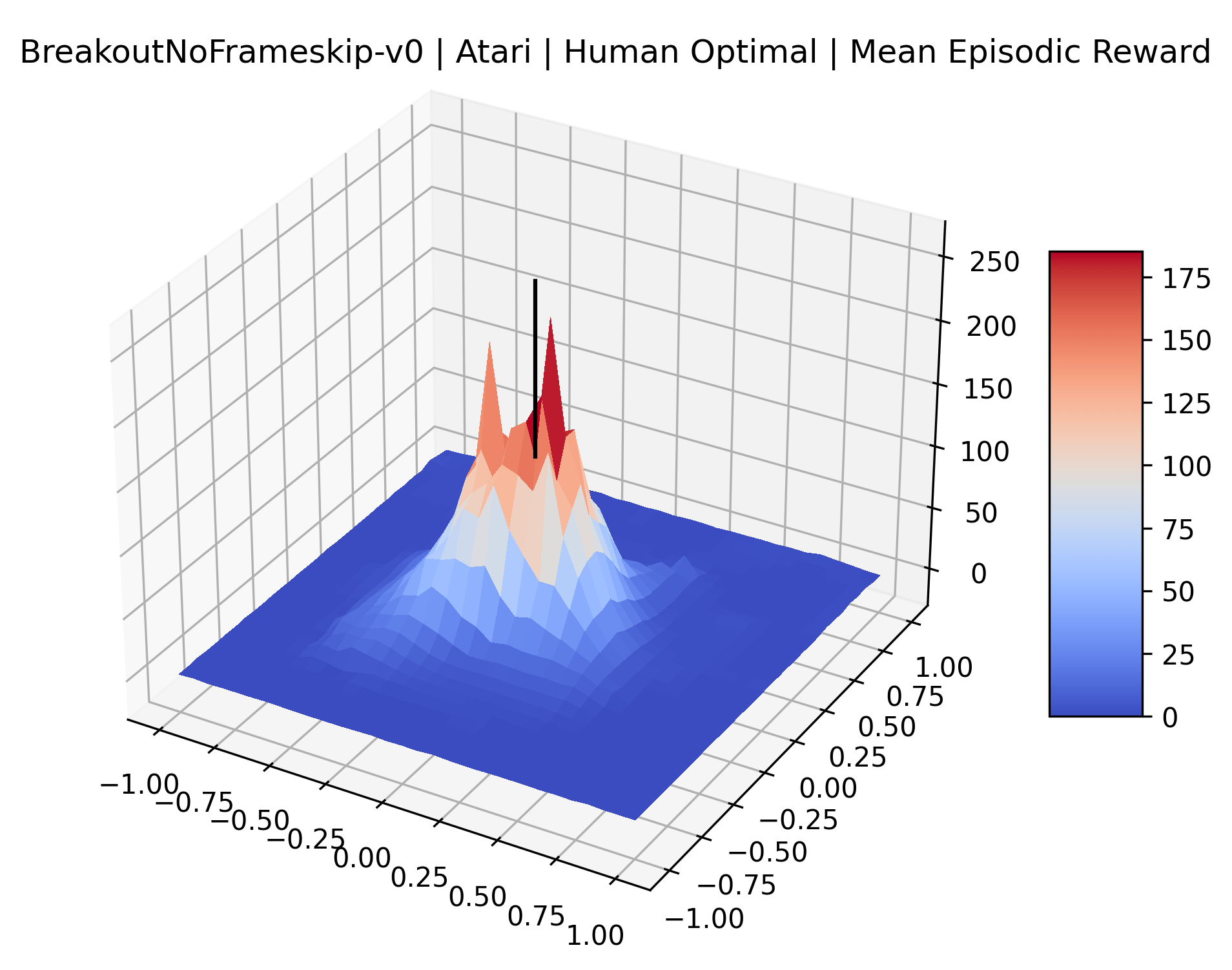} \\
\end{tabular}
\begin{tabular}{ccc}
 \includegraphics[width=\variancescale]{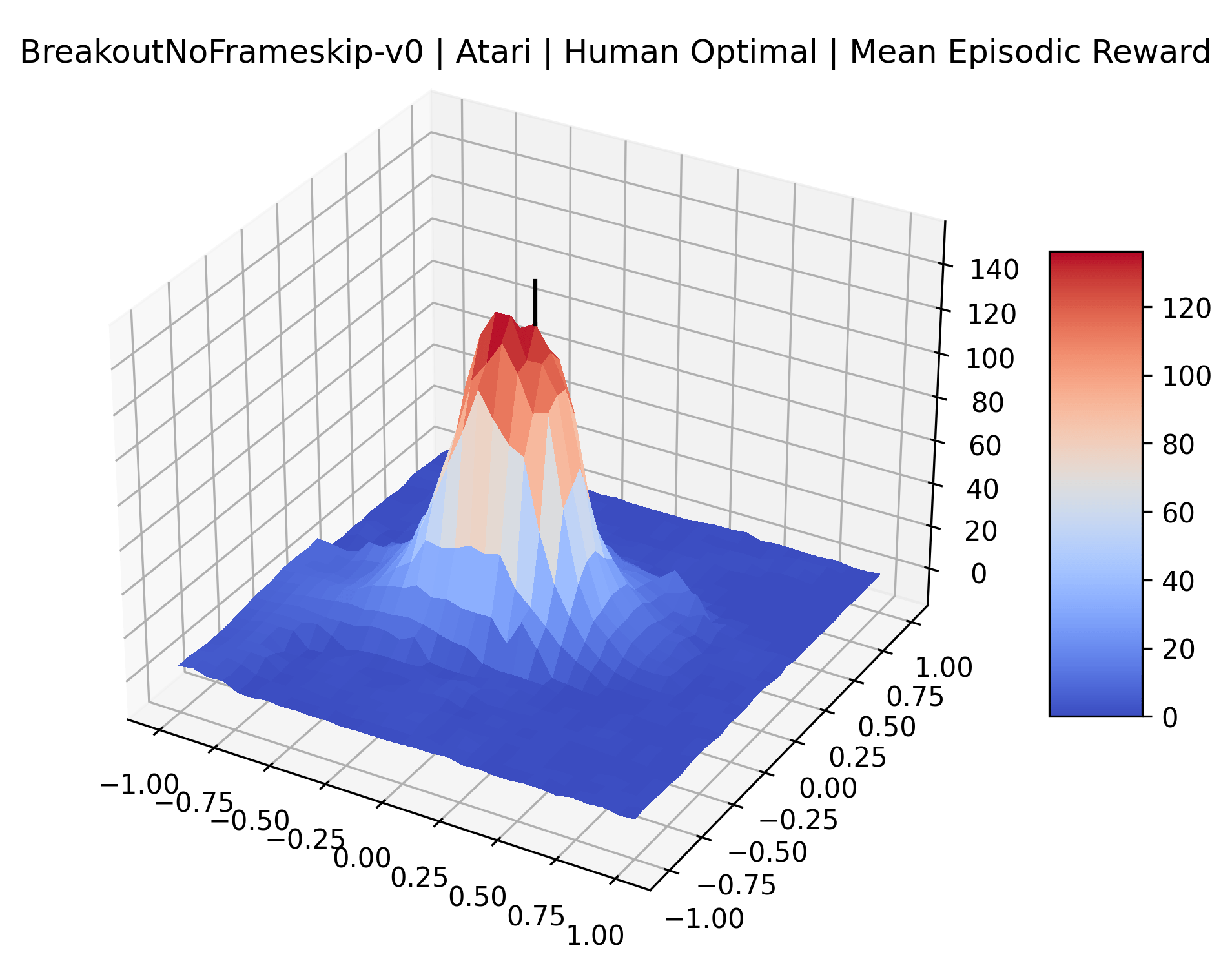} &
 \includegraphics[width=\variancescale]{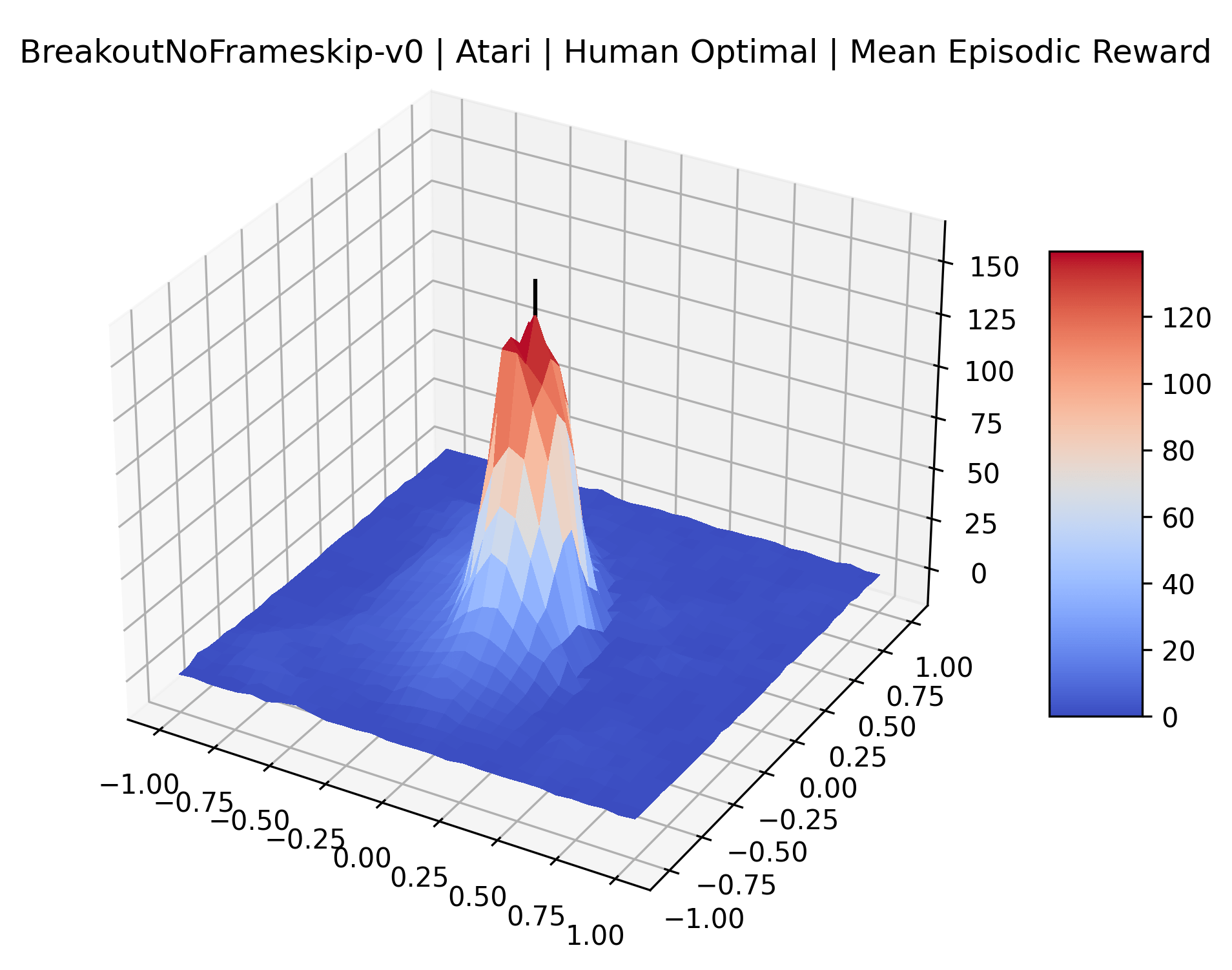} &
 \includegraphics[width=\variancescale]{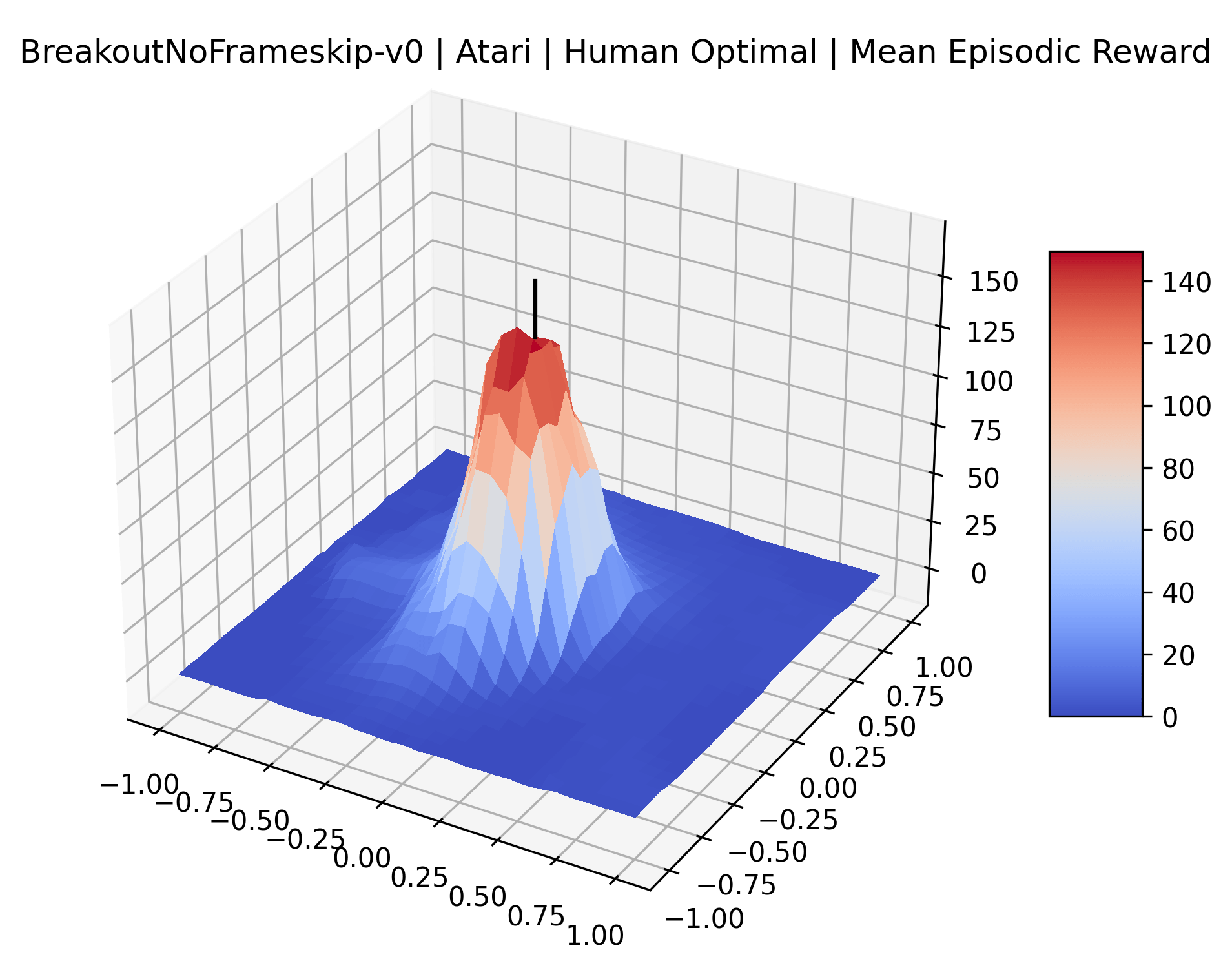} \\
\end{tabular}
\begin{tabular}{ccc}
 \includegraphics[width=\variancescale]{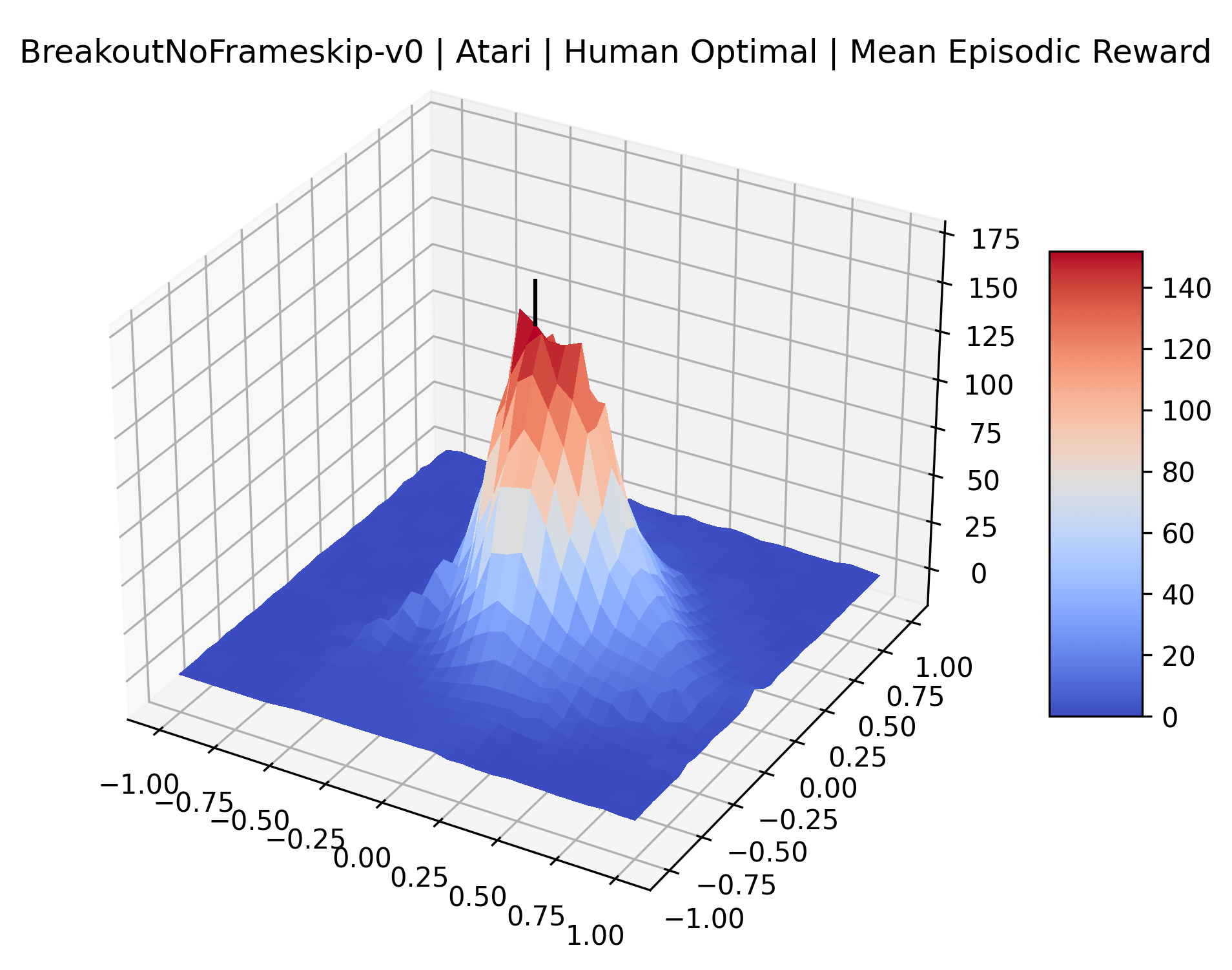} &
 \includegraphics[width=\variancescale]{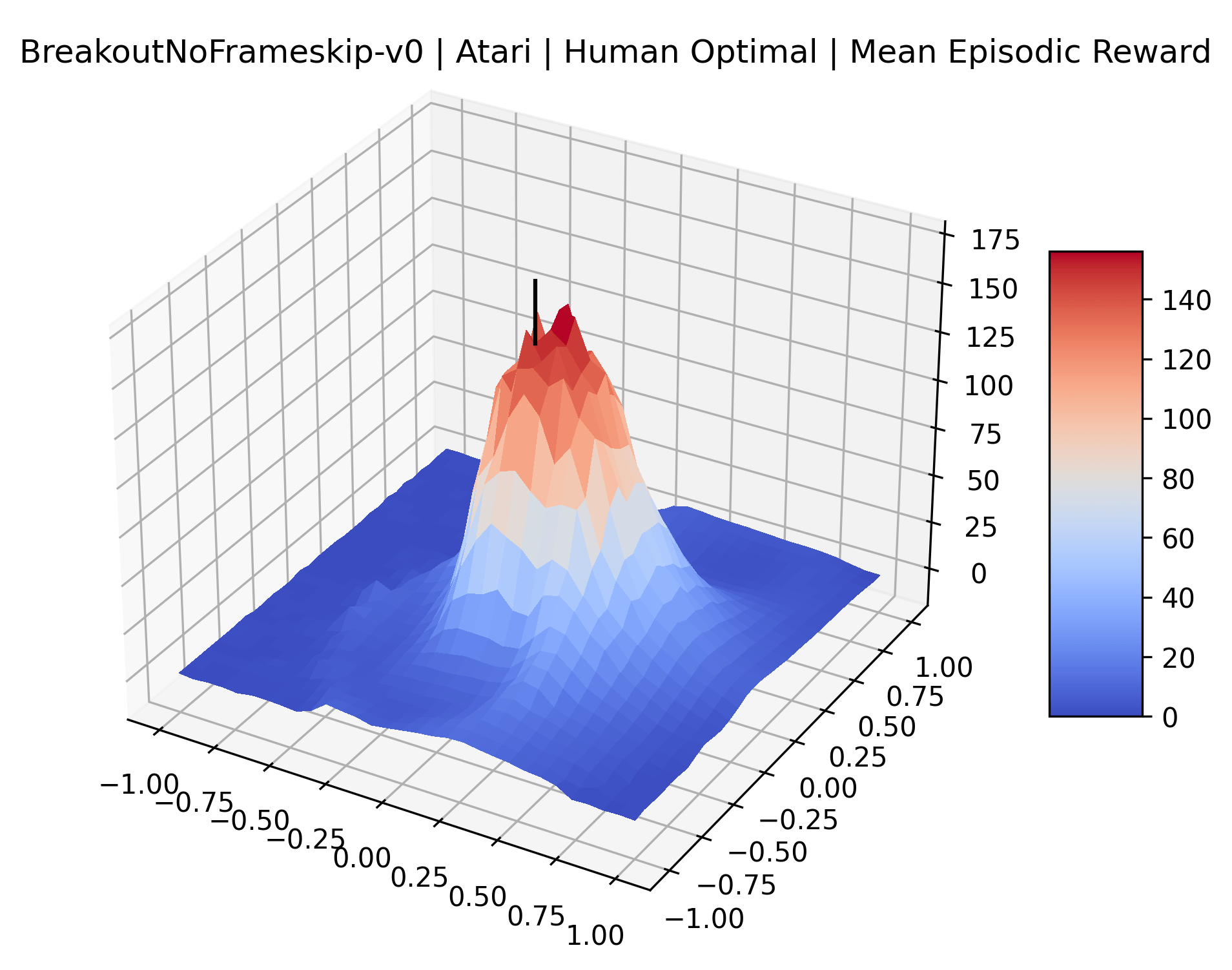} &
 \includegraphics[width=\variancescale]{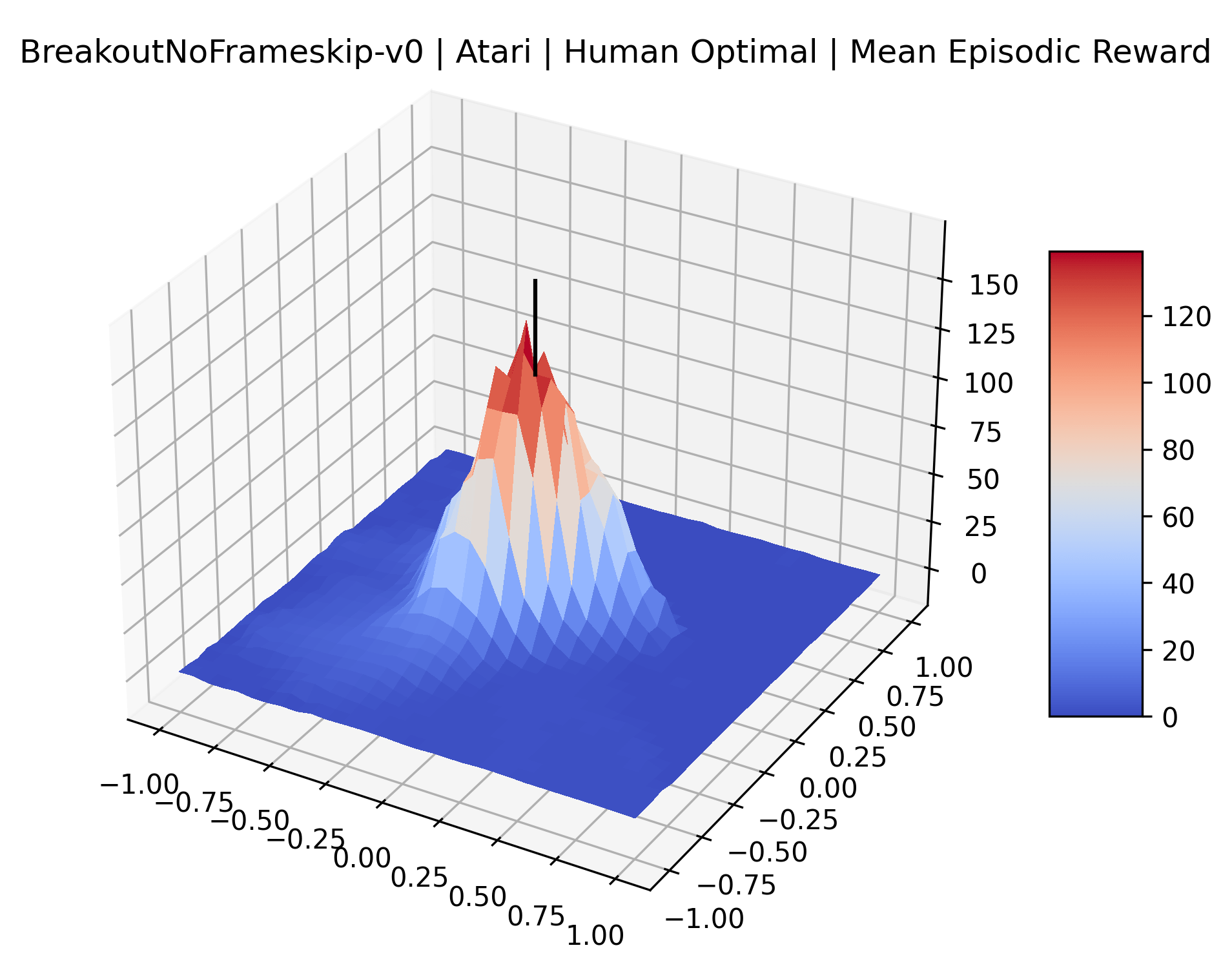} \\
\end{tabular}
\begin{tabular}{ccc}
 \includegraphics[width=\variancescale]{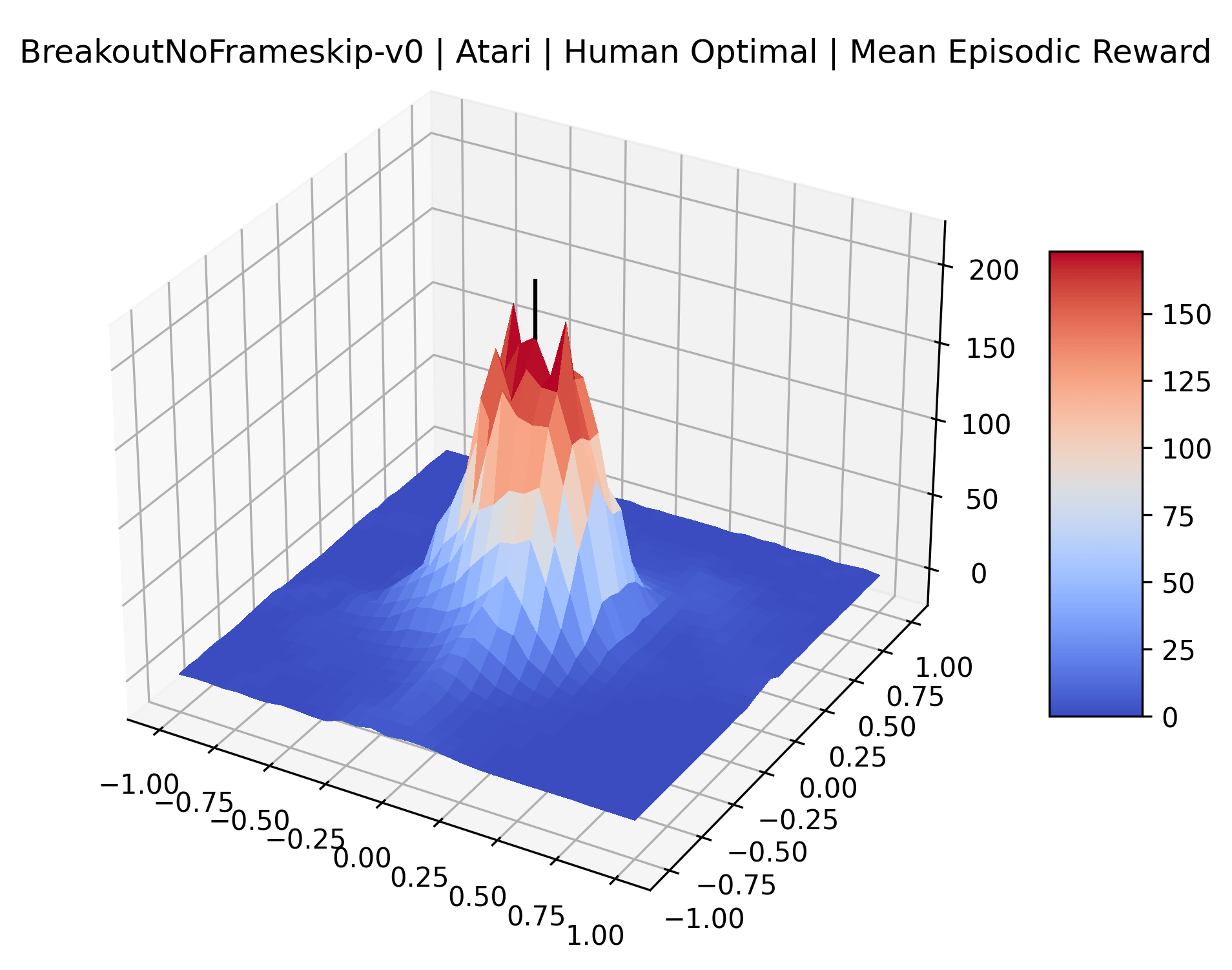} &
 \includegraphics[width=\variancescale]{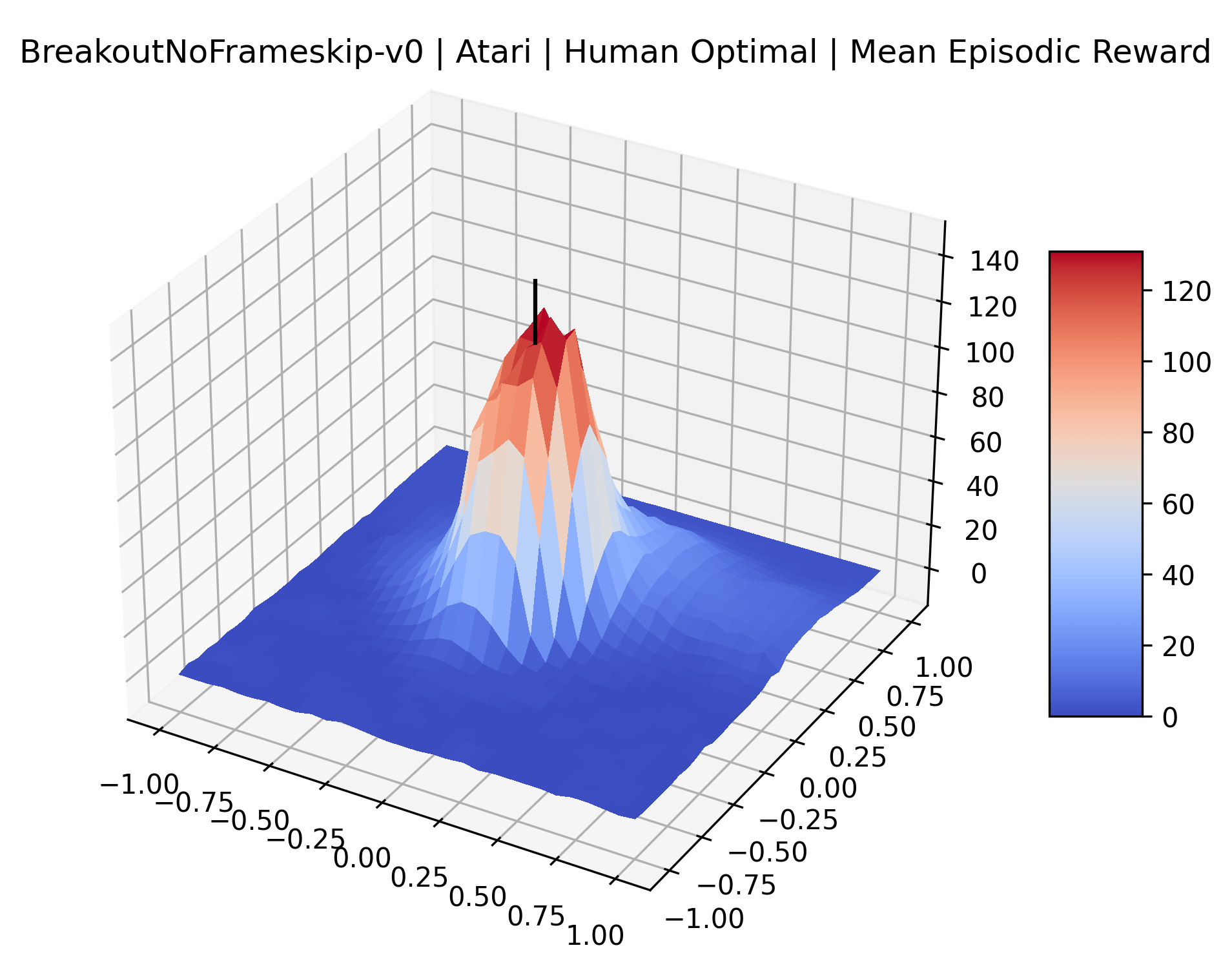} &
 \includegraphics[width=\variancescale]{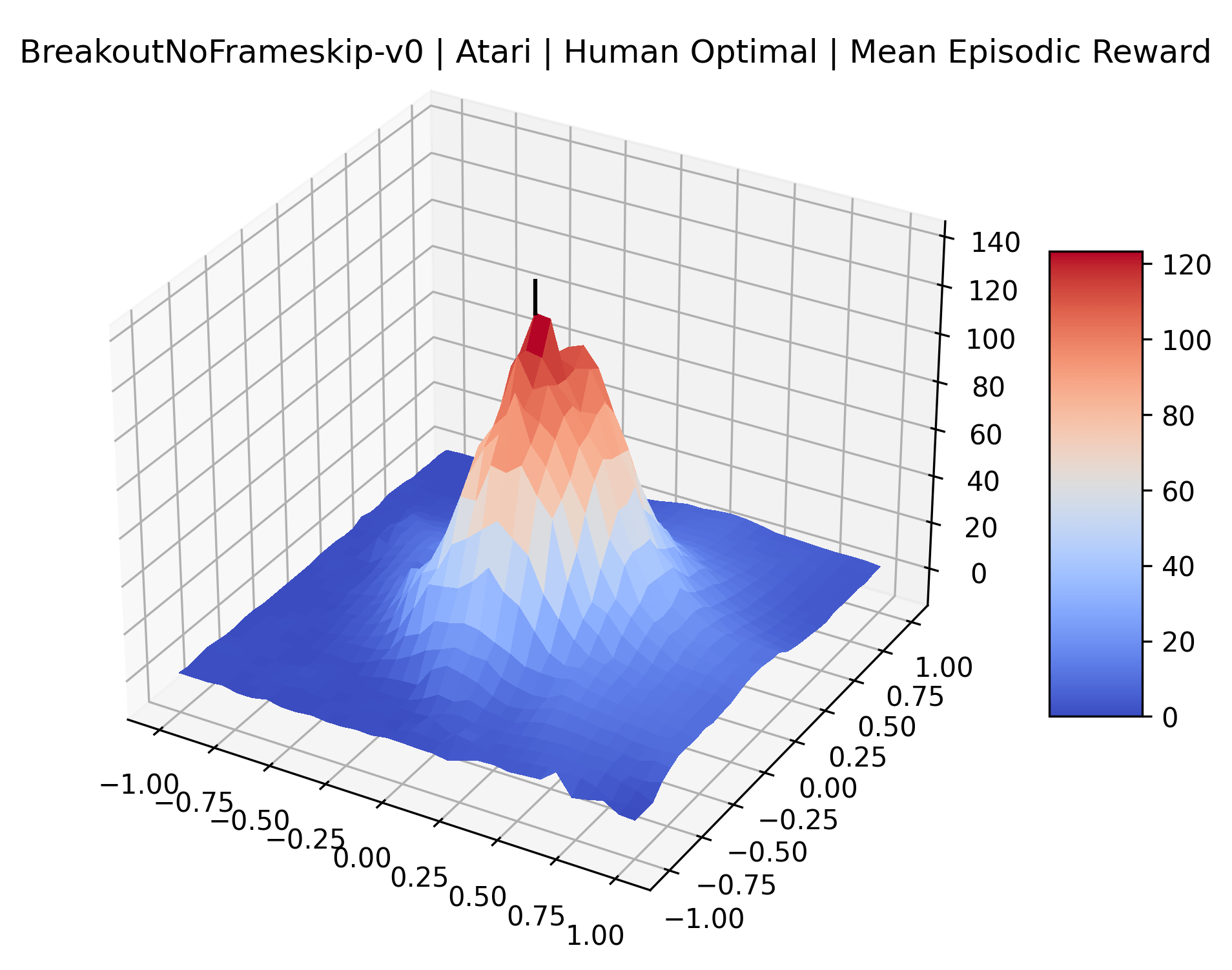} \\
\end{tabular}
\begin{tabular}{ccc}
 \includegraphics[width=\variancescale]{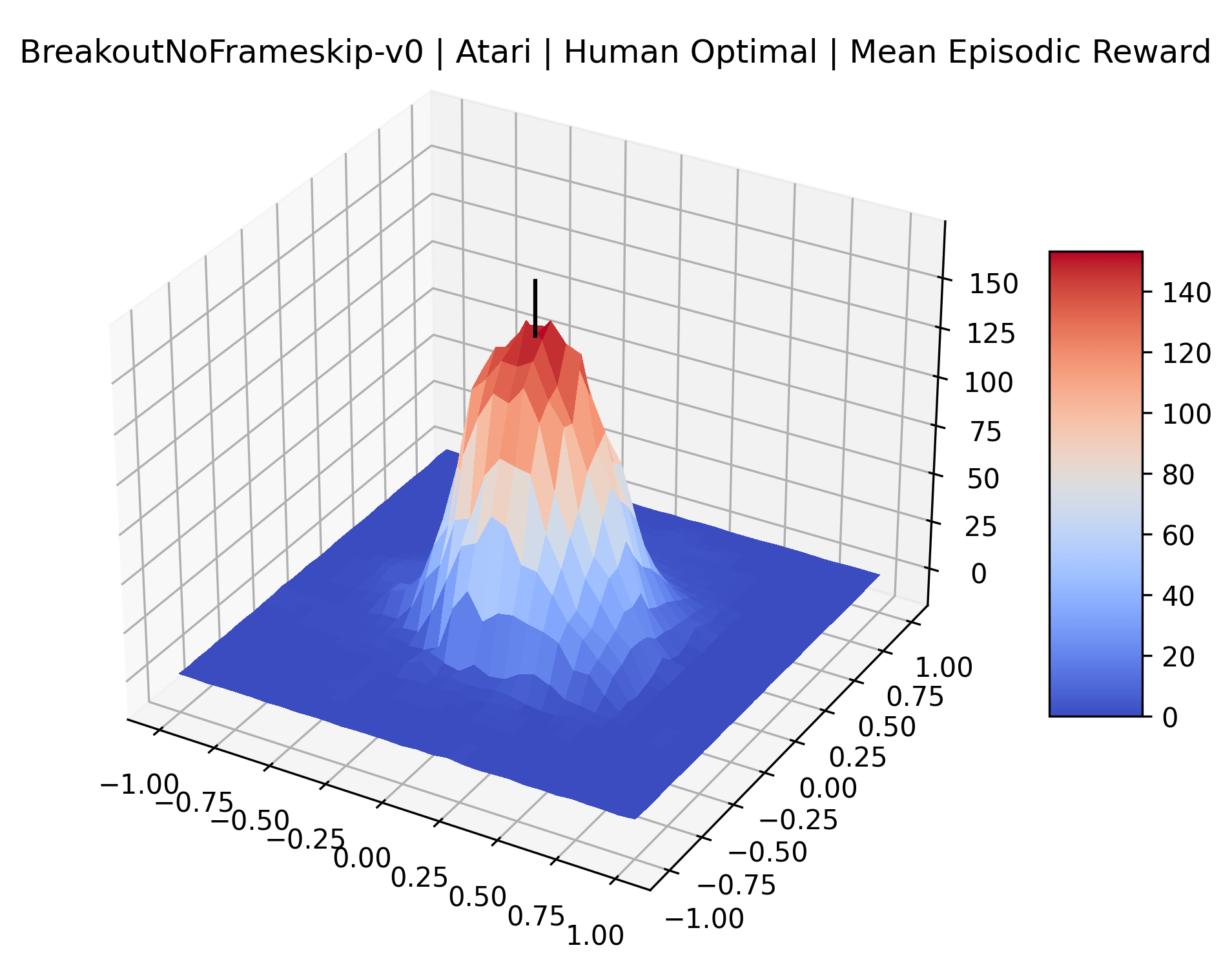} &
 \includegraphics[width=\variancescale]{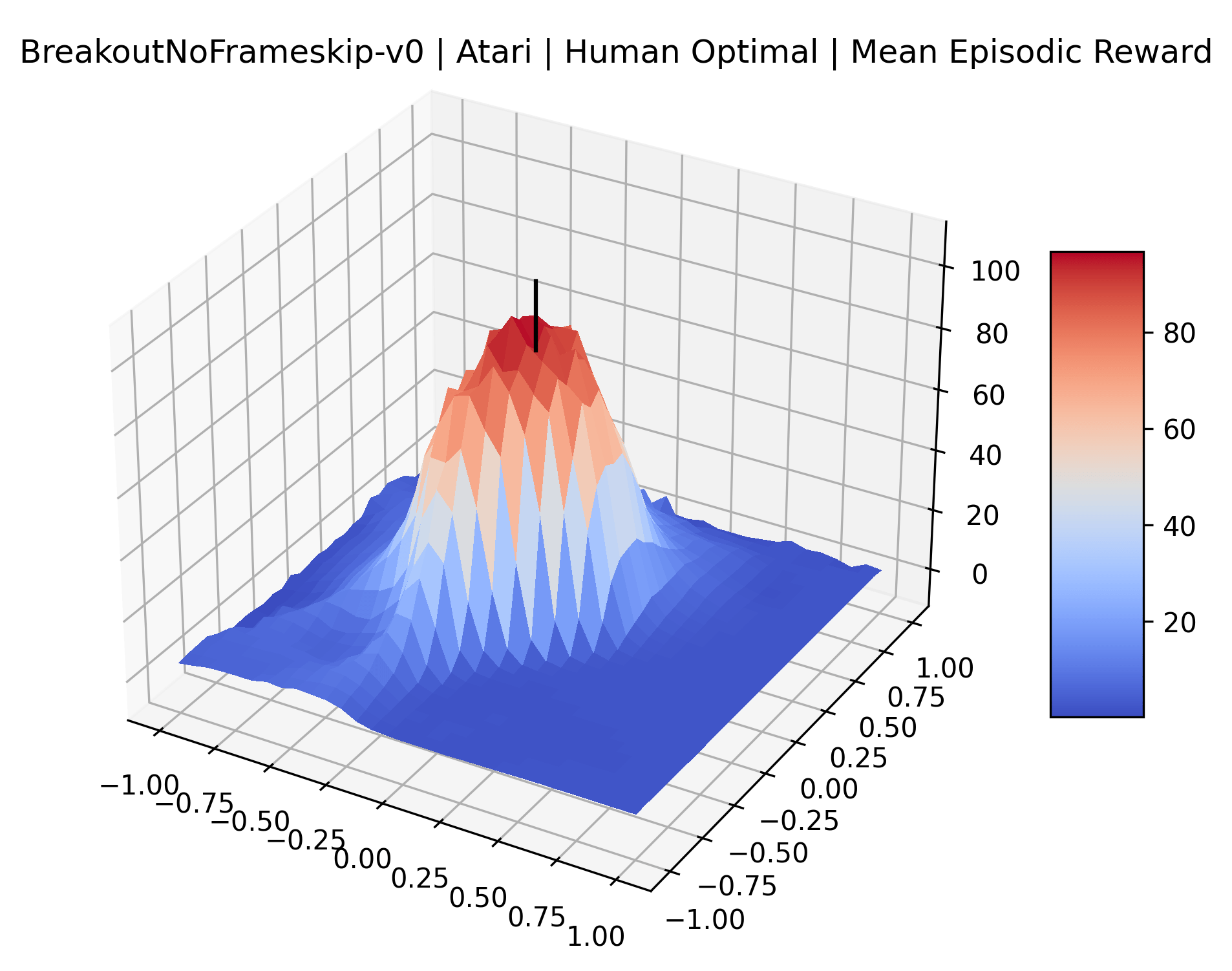} &
 \includegraphics[width=\variancescale]{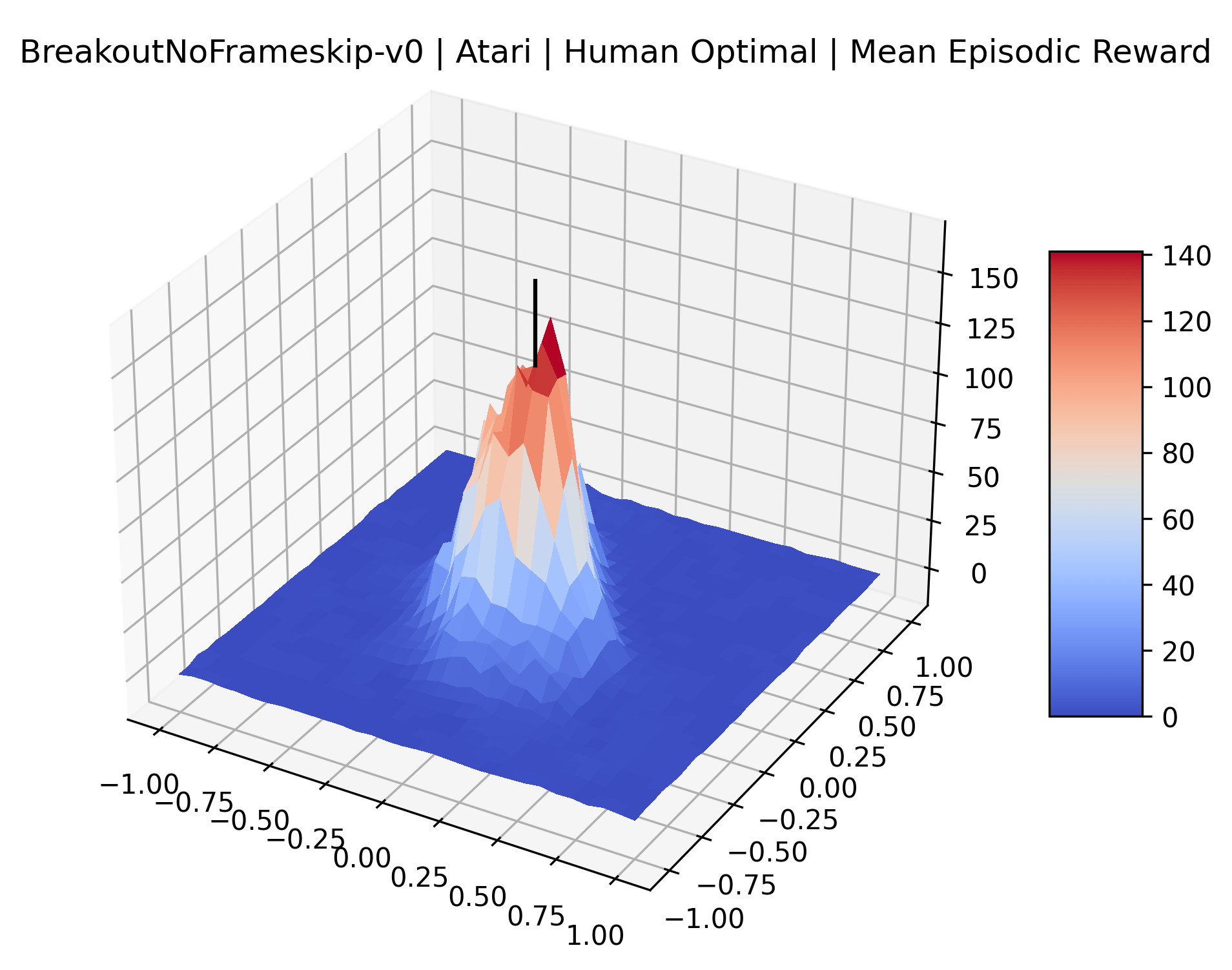} \\
\end{tabular}
\caption{18 training and plotting runs for the Atari Breakout environment.}
\label{fig:dense_atari_variance_table}
\end{figure*}

\pagebreak

\begin{figure*}[!htb]
\centering
\begin{tabular}{ccc}
 \includegraphics[width=\variancescale]{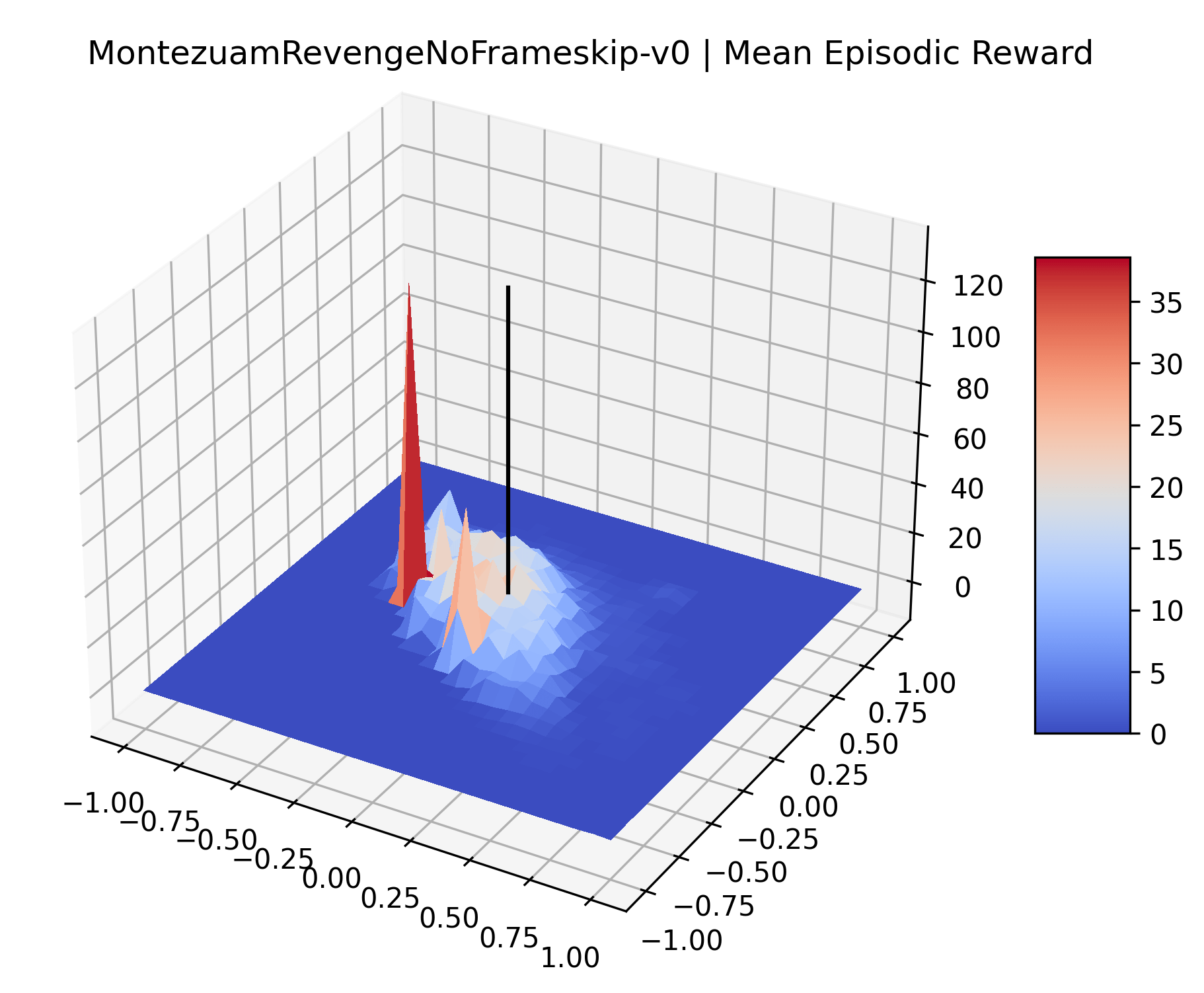} &
 \includegraphics[width=\variancescale]{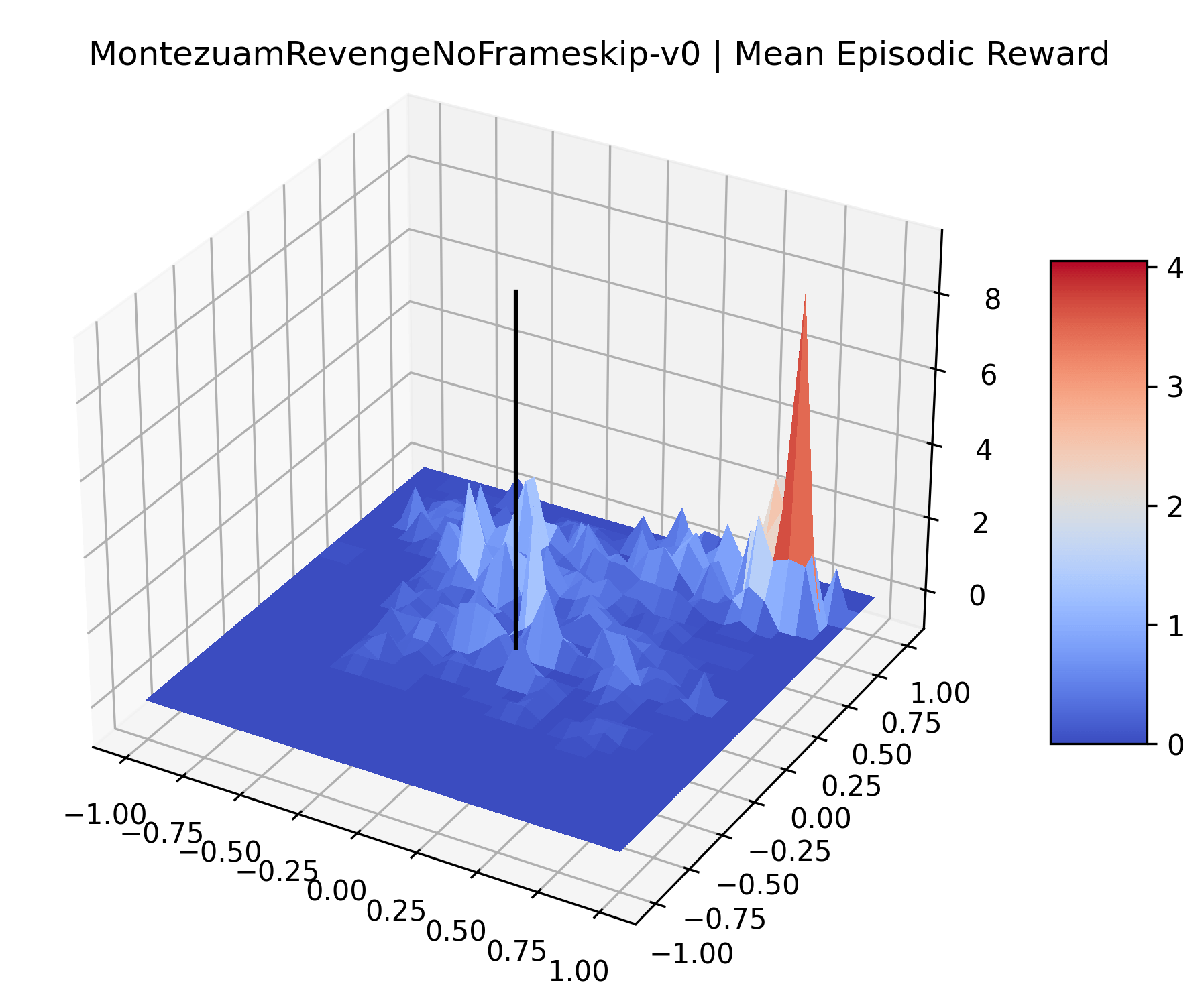} &
 \includegraphics[width=\variancescale]{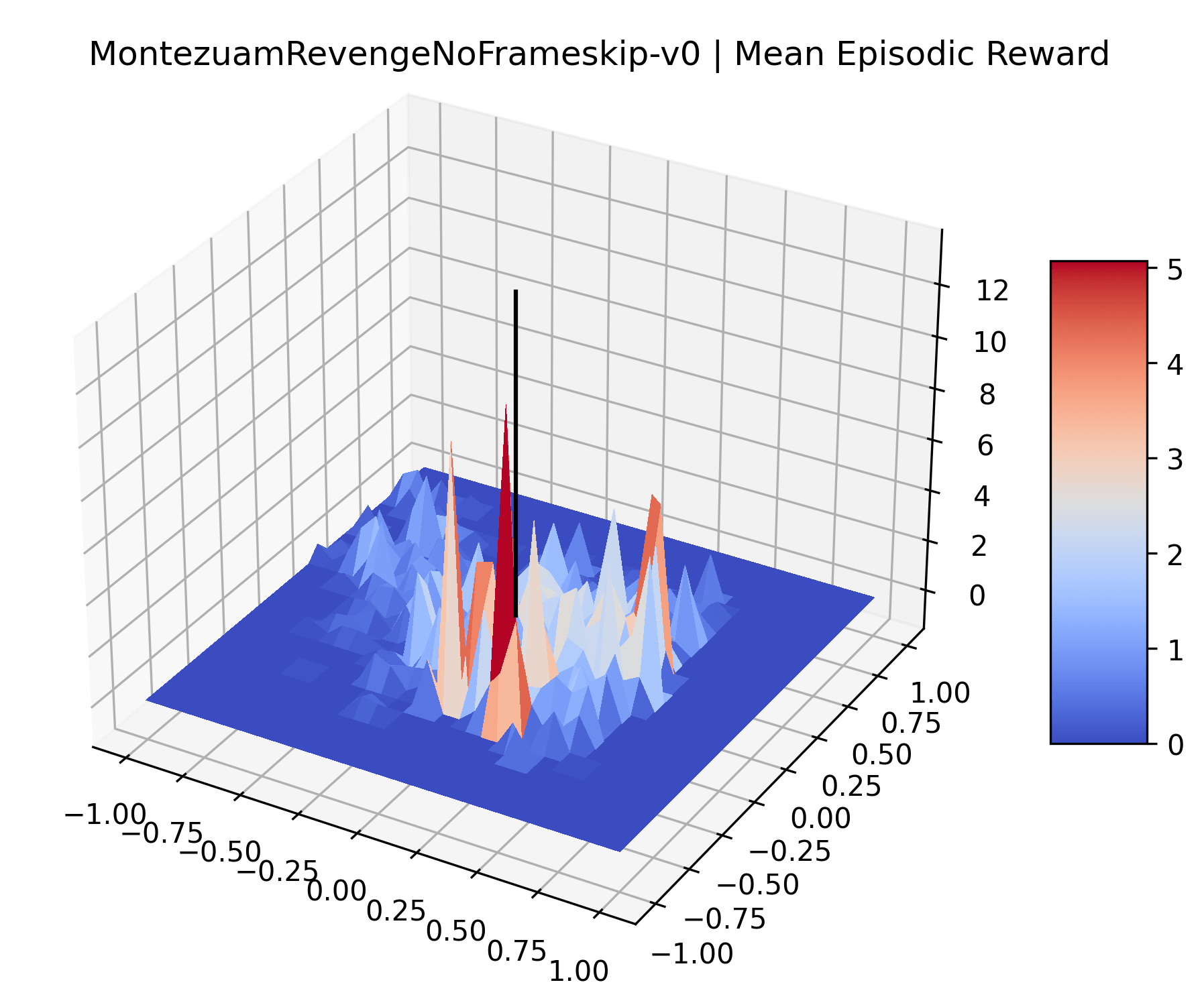} \\
\end{tabular}
\begin{tabular}{ccc}
 \includegraphics[width=\variancescale]{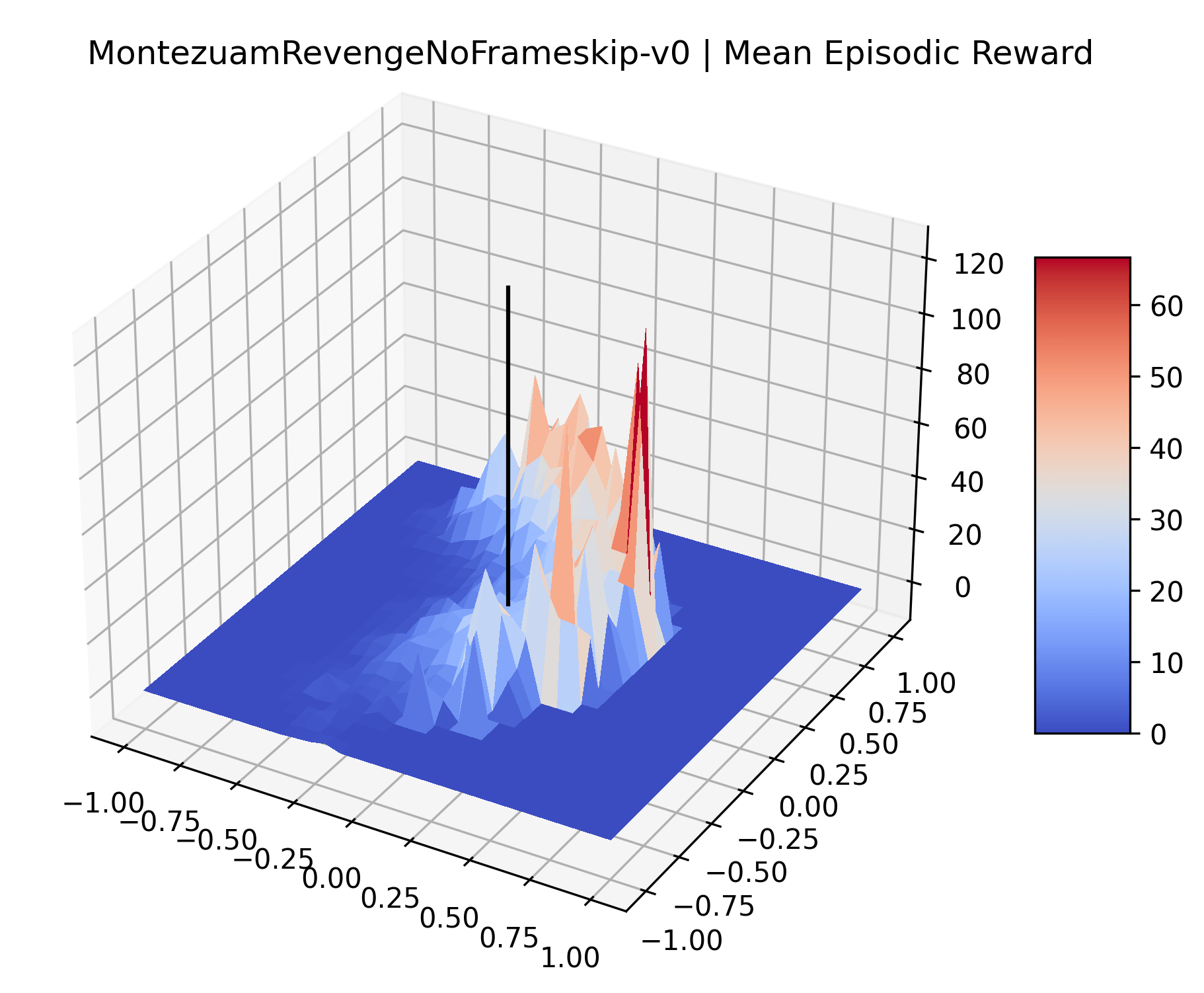} &
 \includegraphics[width=\variancescale]{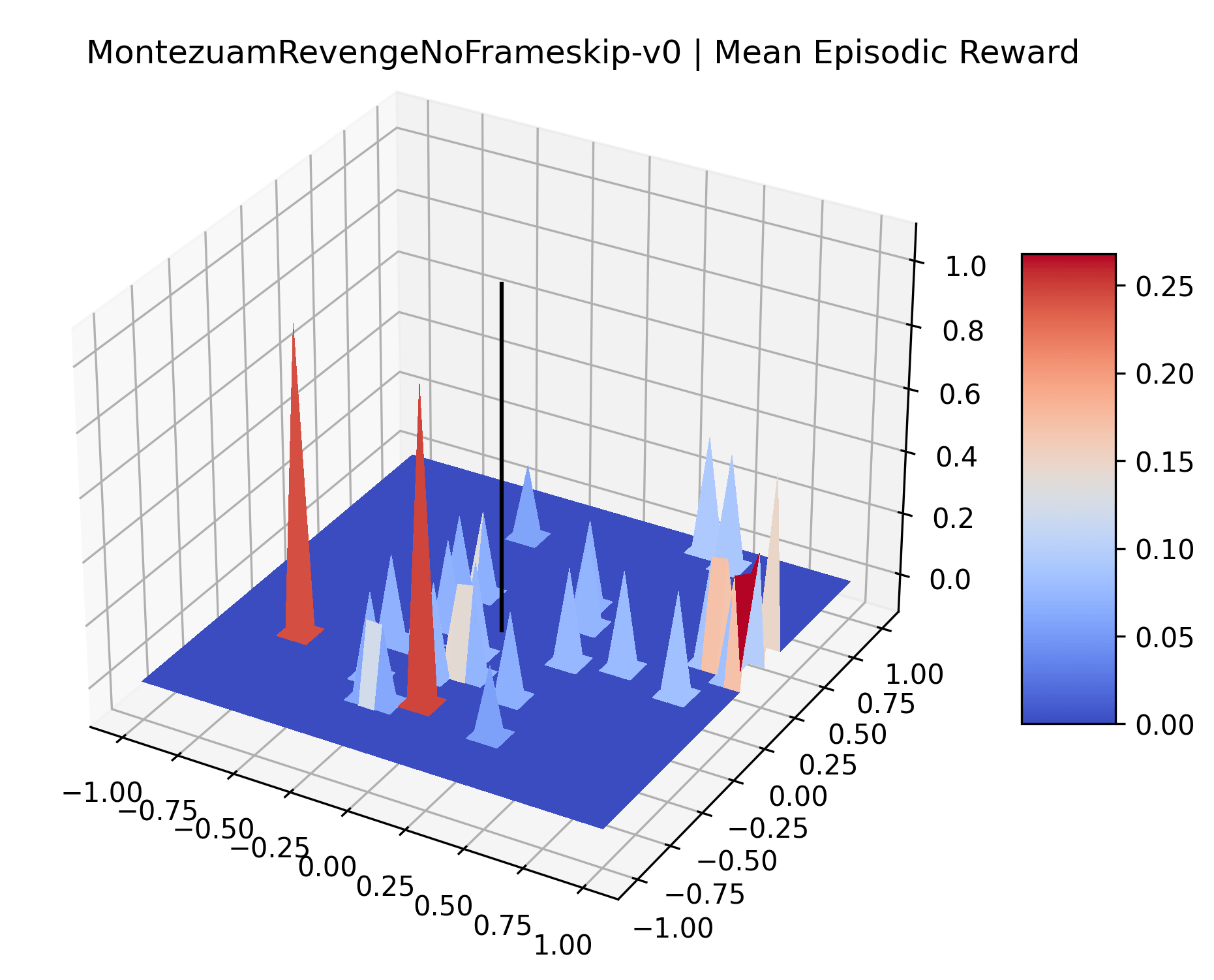} &
 \includegraphics[width=\variancescale]{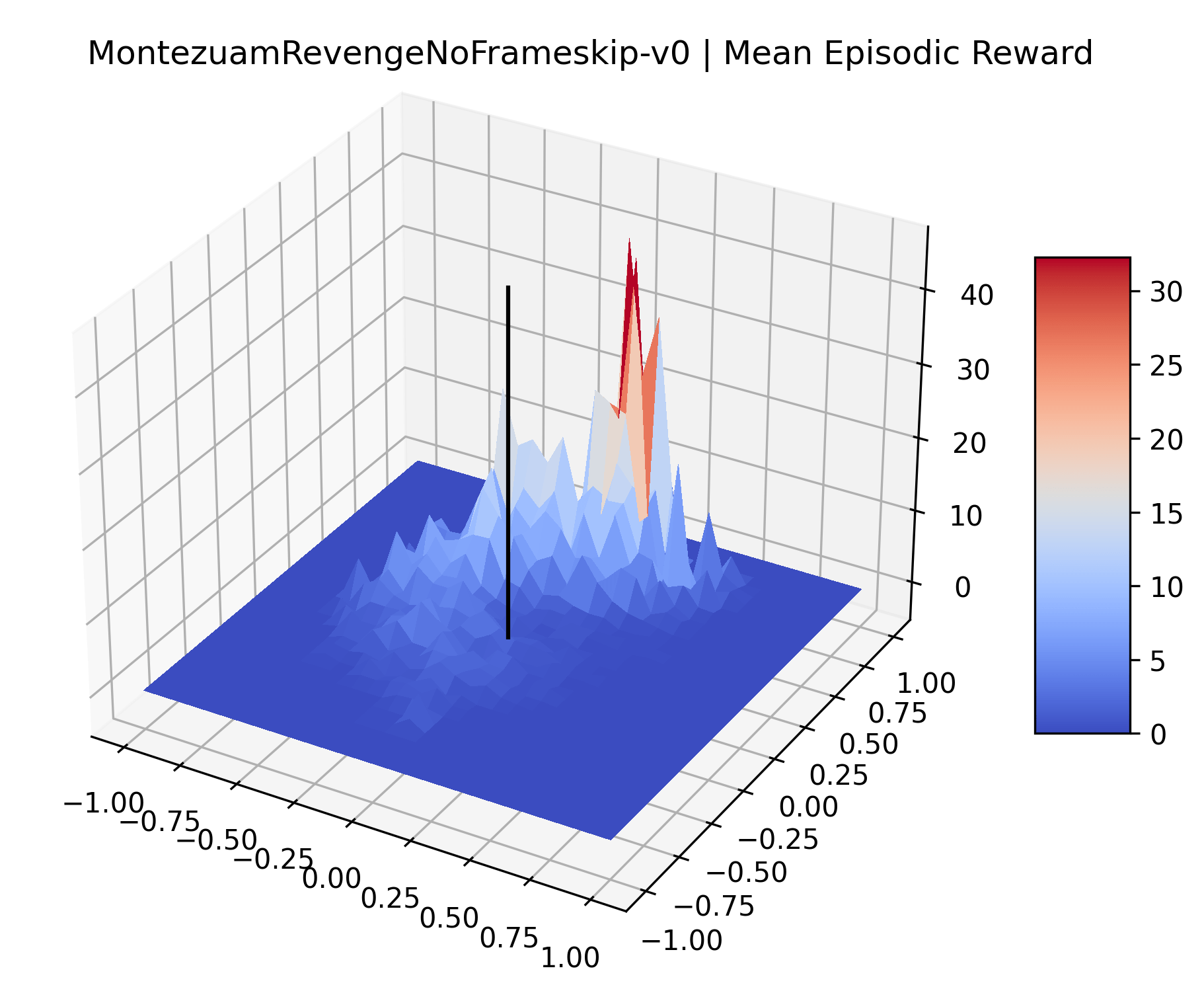} \\
\end{tabular}
\begin{tabular}{ccc}
 \includegraphics[width=\variancescale]{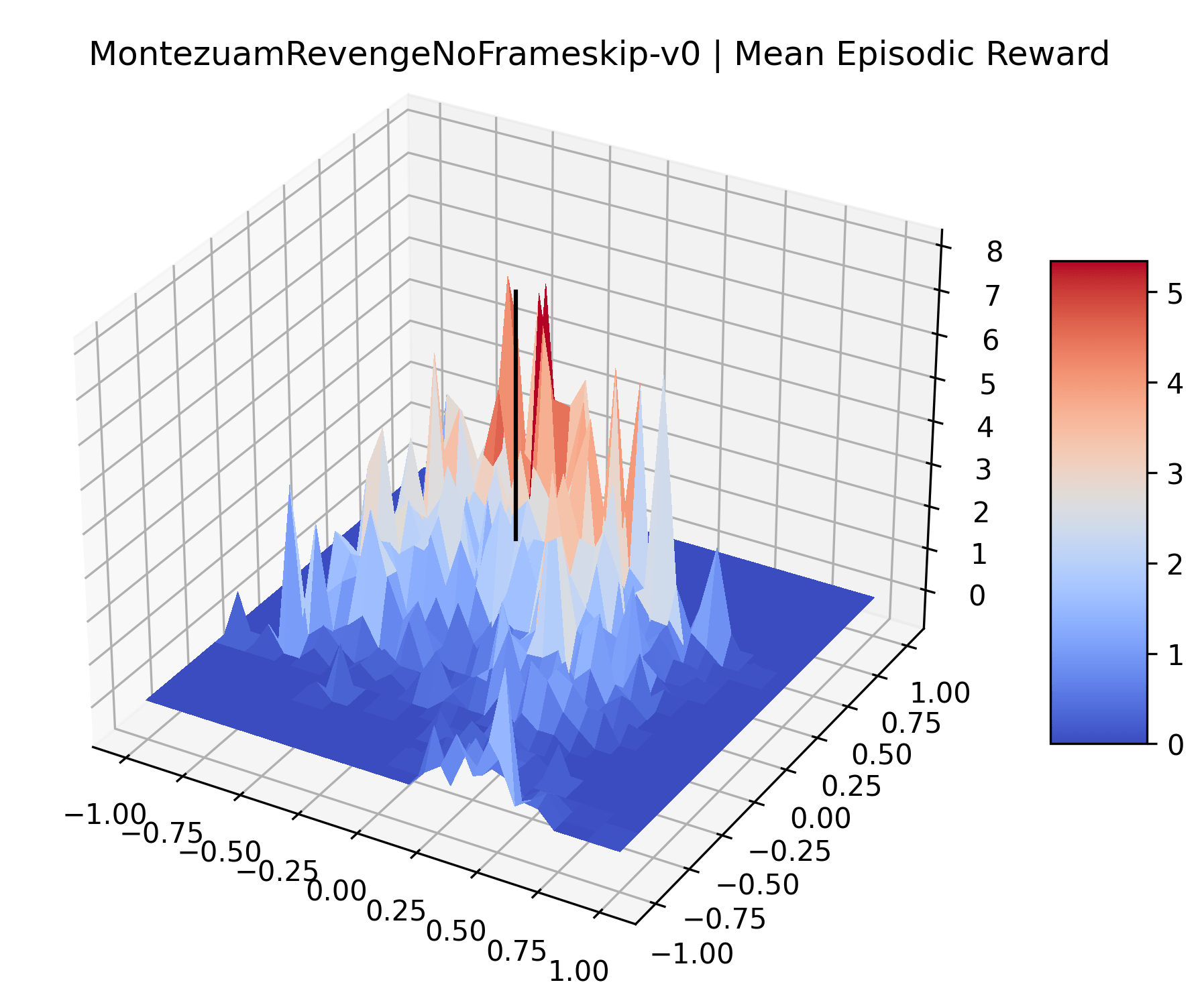} &
 \includegraphics[width=\variancescale]{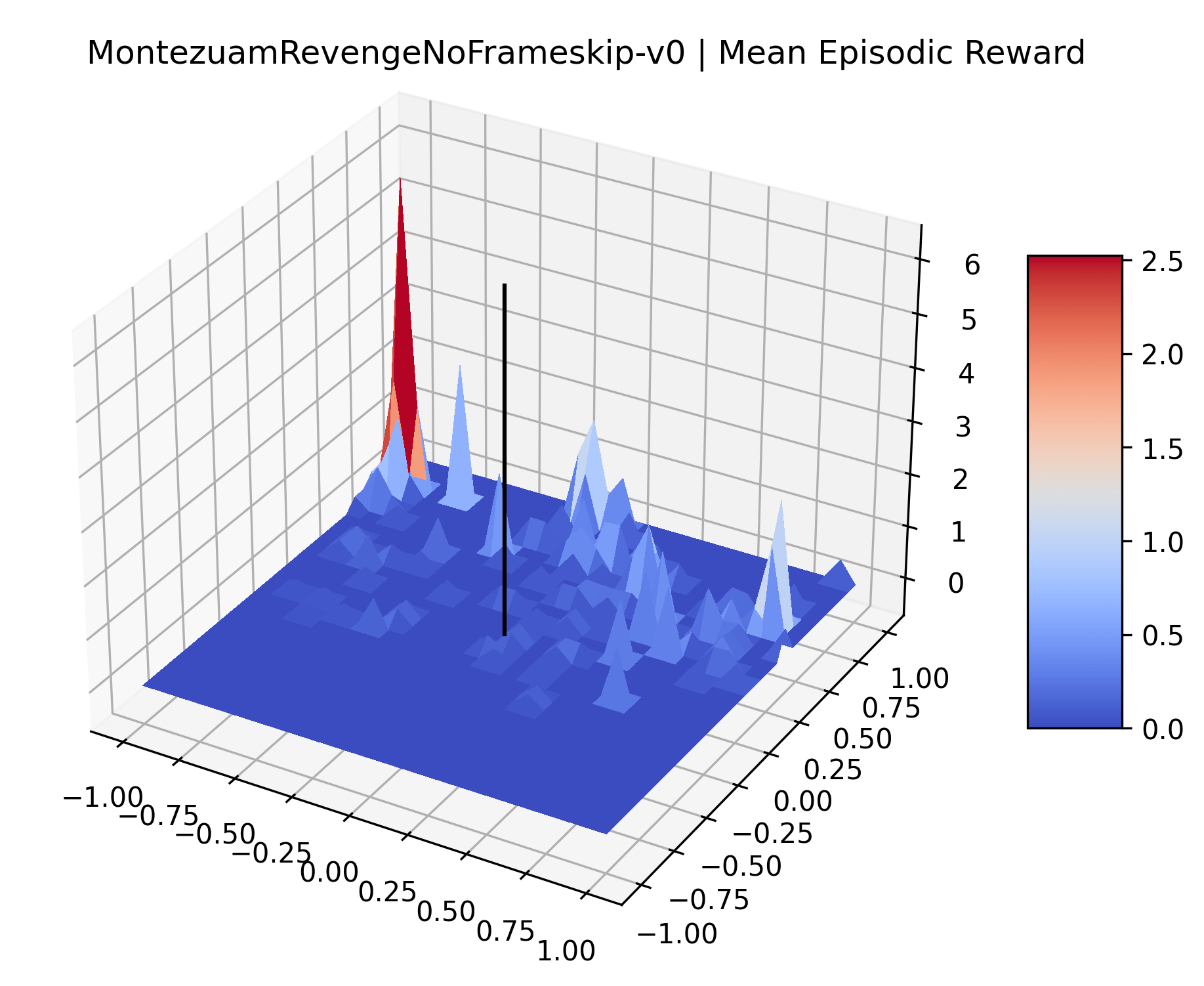} &
 \includegraphics[width=\variancescale]{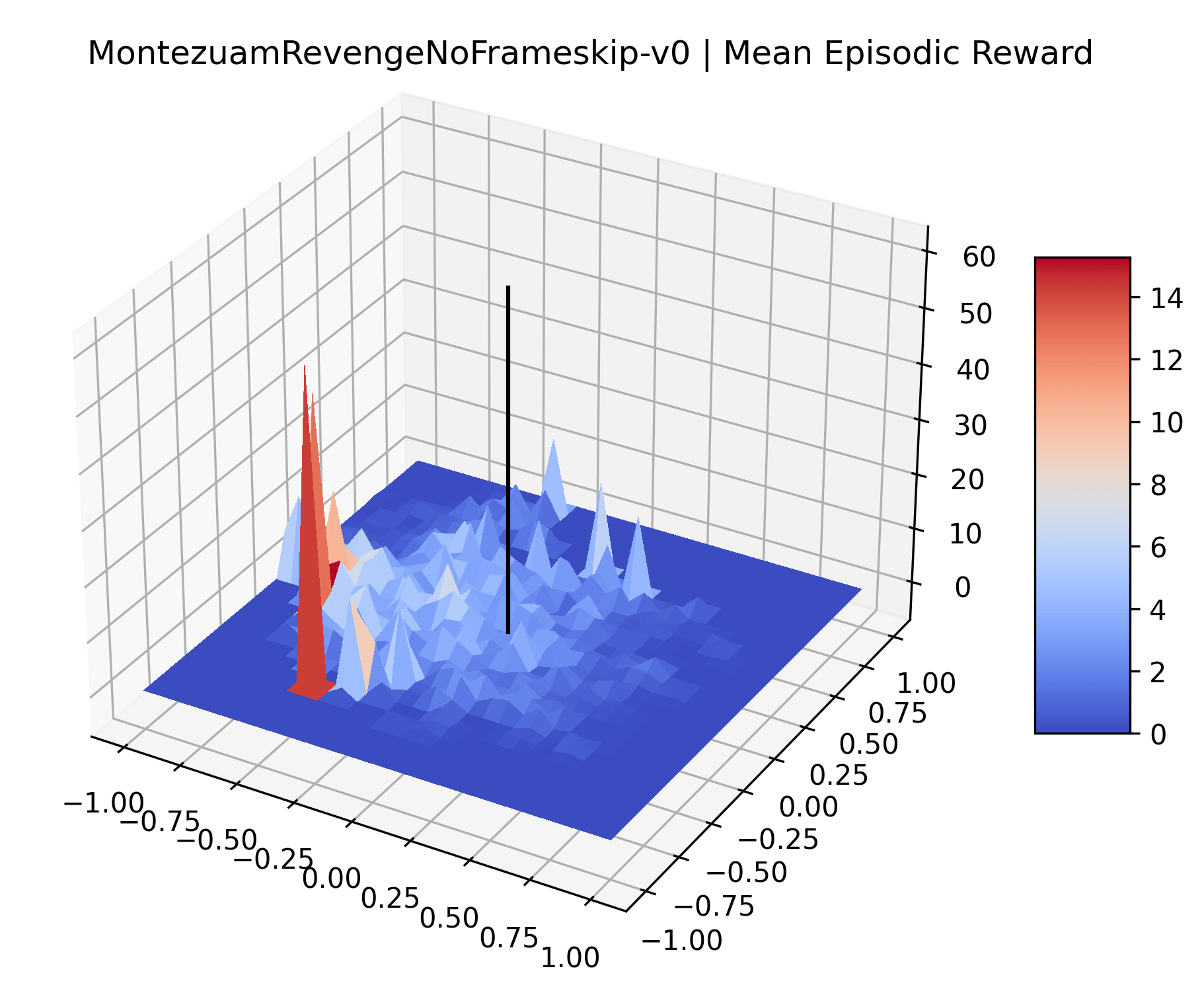} \\
\end{tabular}
\begin{tabular}{ccc}
 \includegraphics[width=\variancescale]{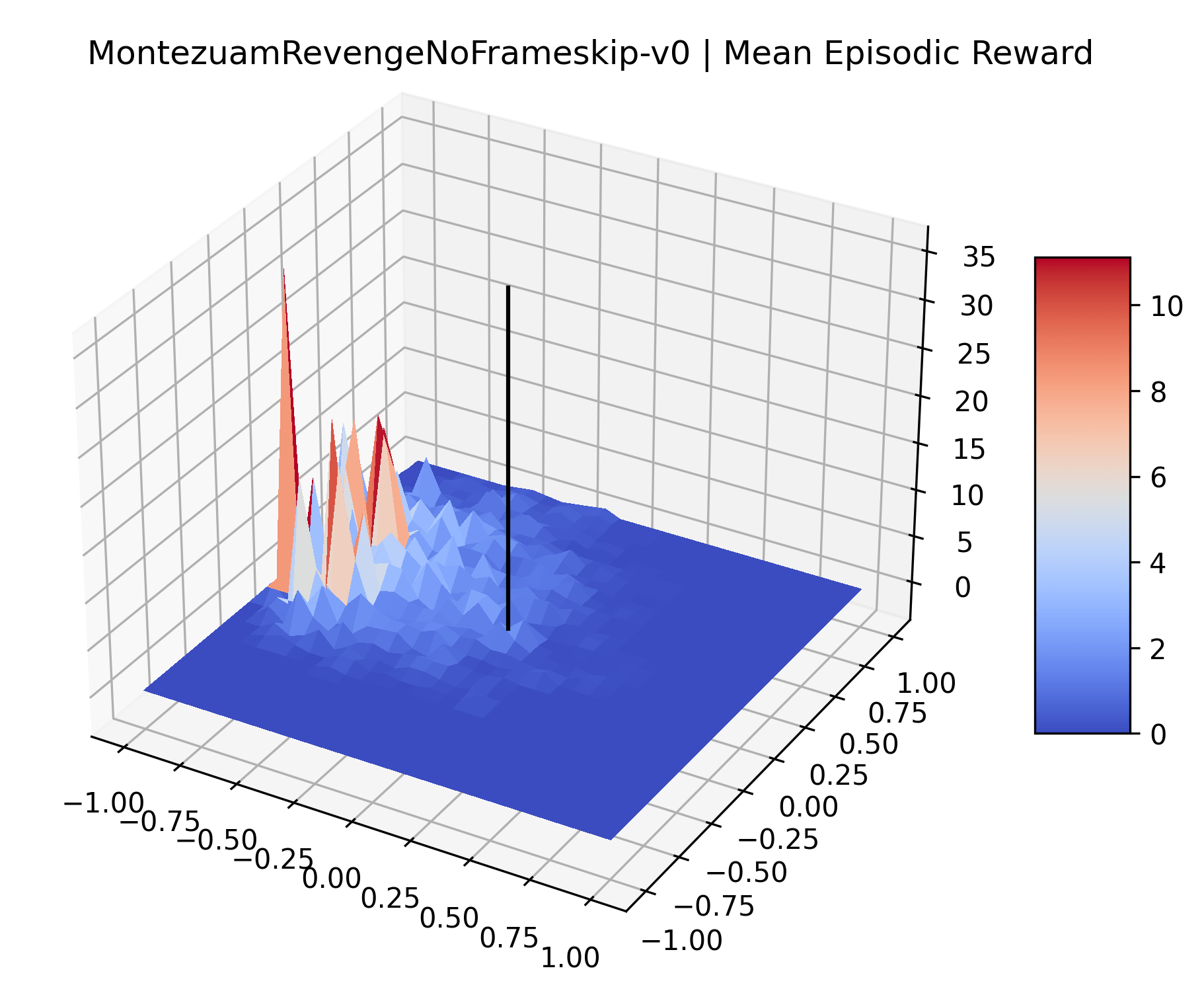} &
 \includegraphics[width=\variancescale]{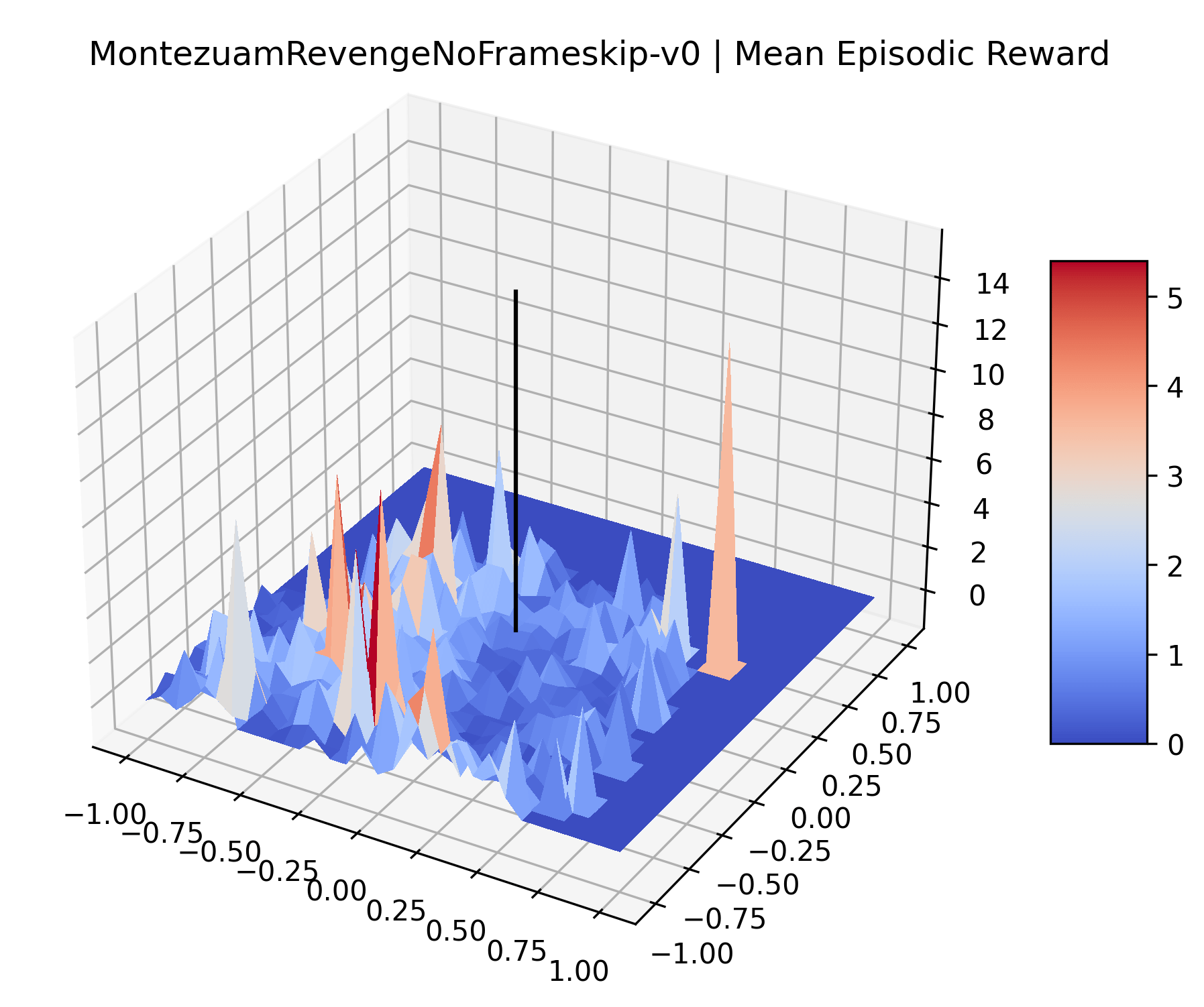} &
 \includegraphics[width=\variancescale]{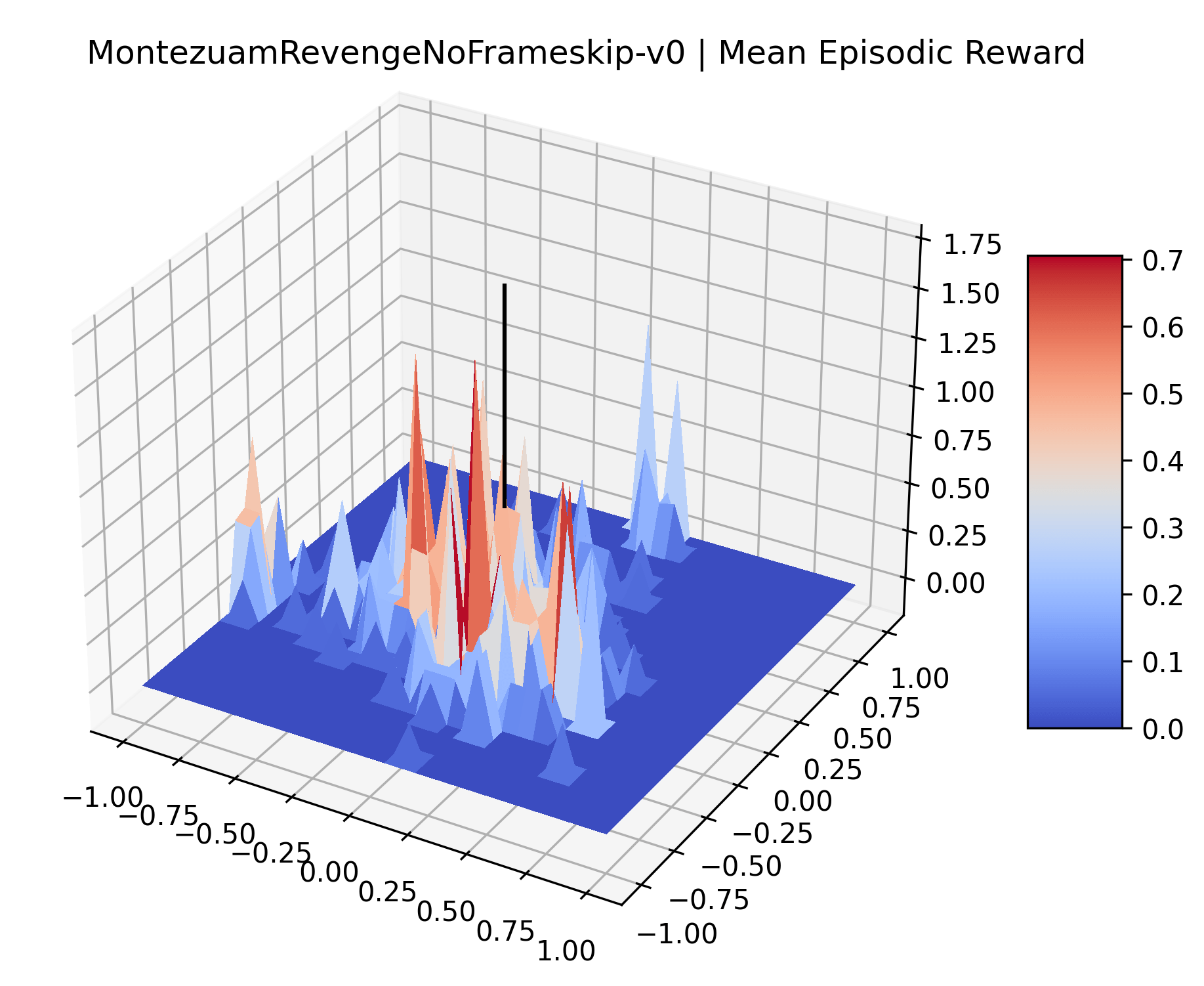} \\
\end{tabular}
\begin{tabular}{ccc}
 \includegraphics[width=\variancescale]{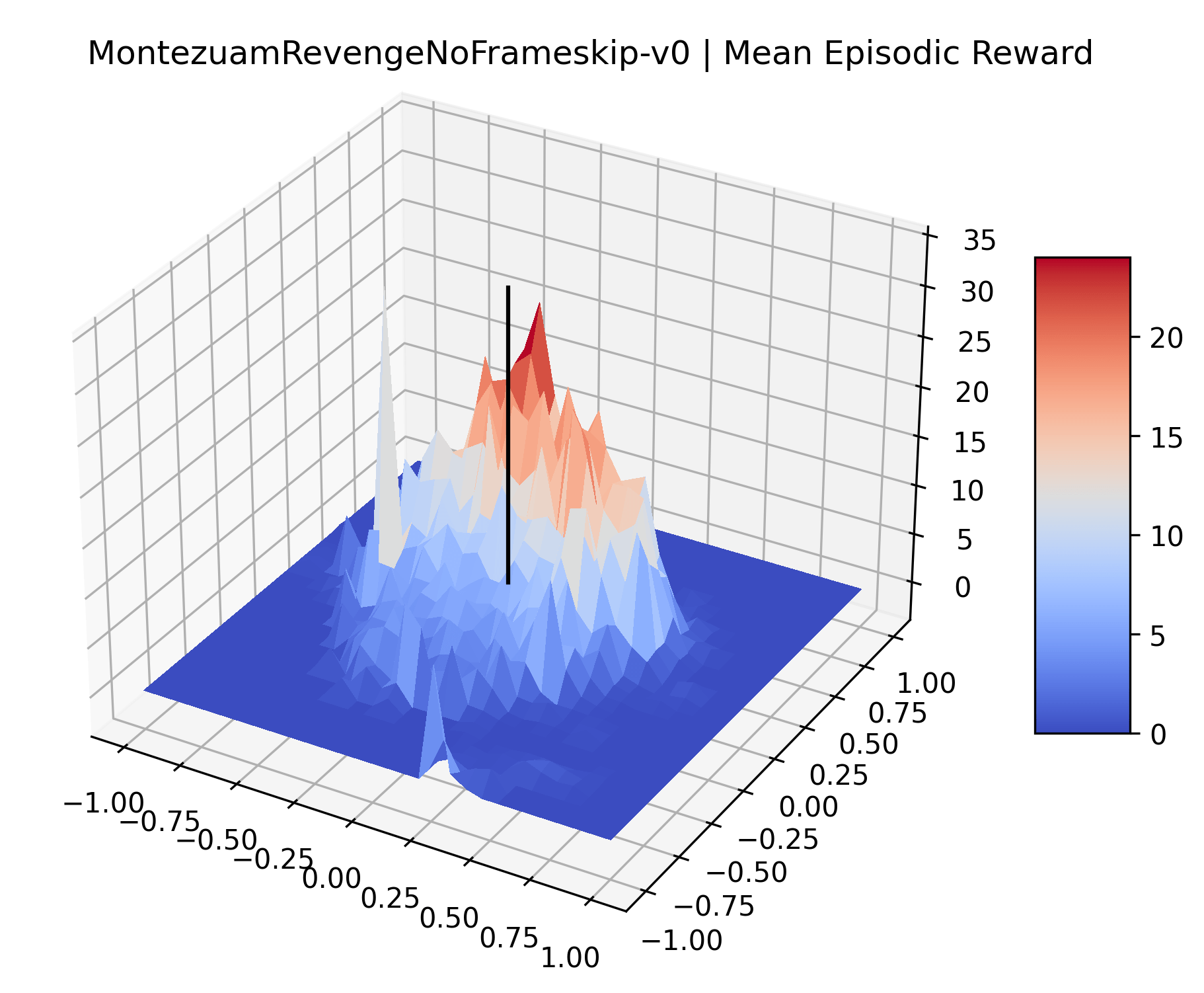} &
 \includegraphics[width=\variancescale]{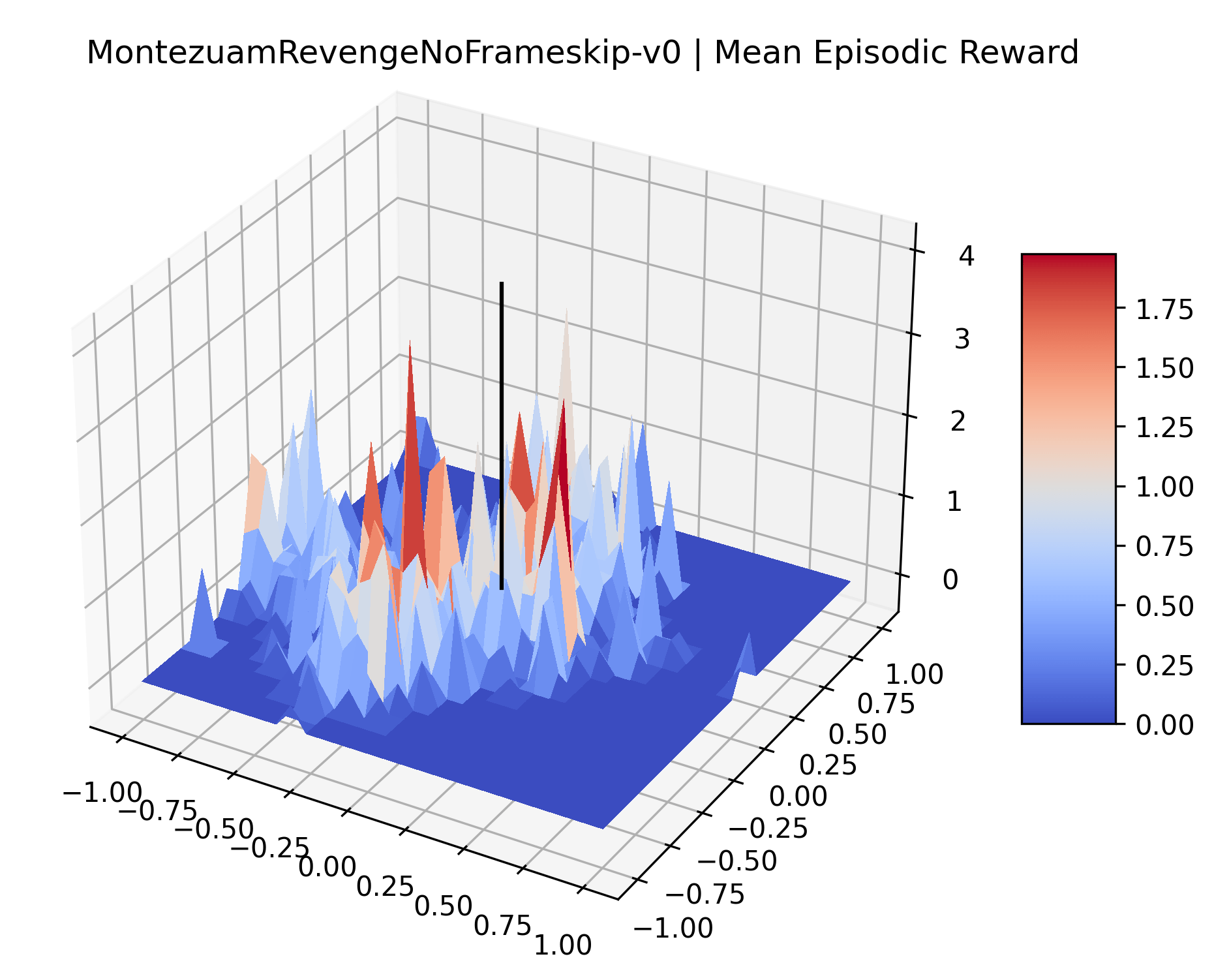} &
 \includegraphics[width=\variancescale]{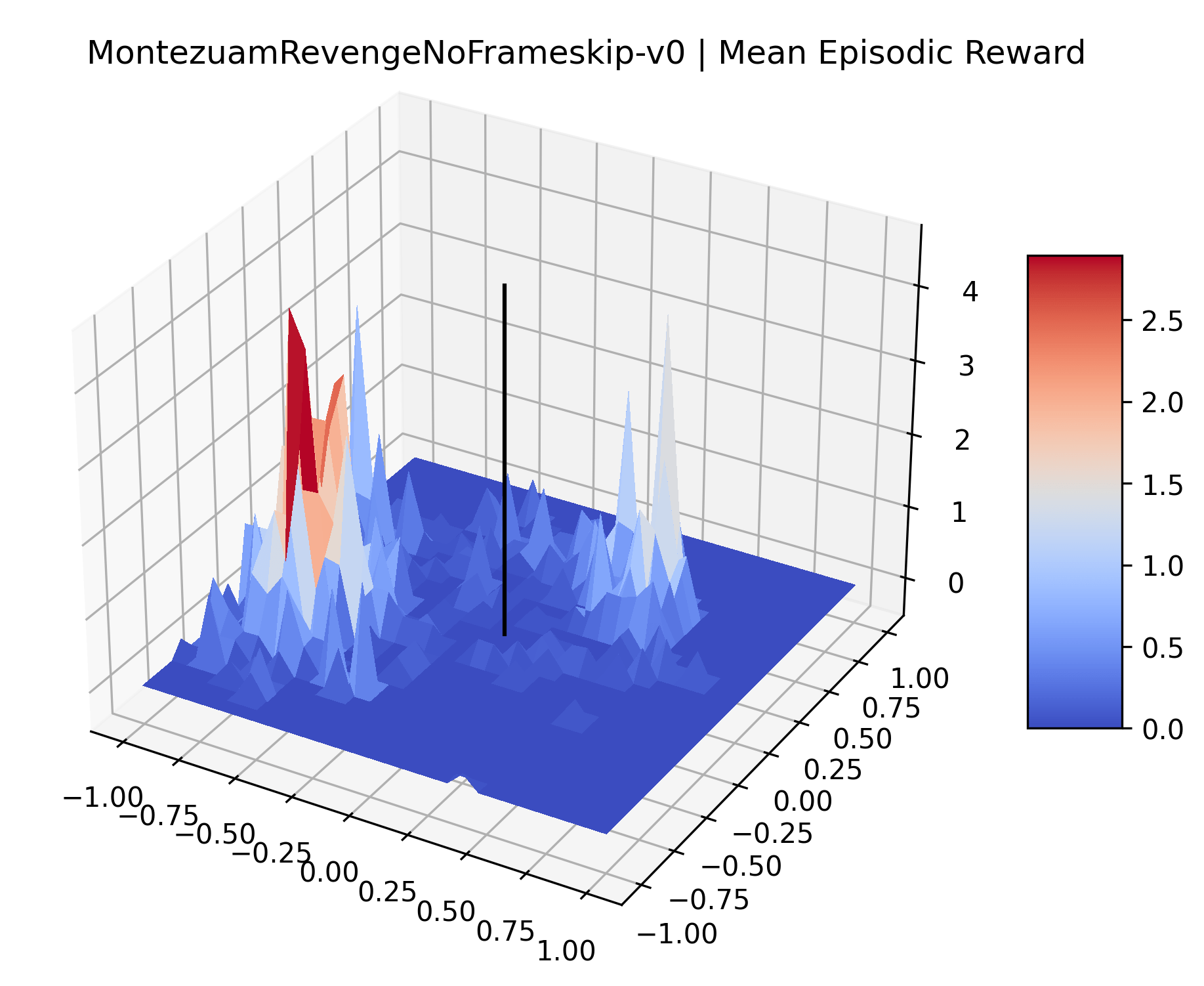} \\
\end{tabular}
\begin{tabular}{ccc}
 \includegraphics[width=\variancescale]{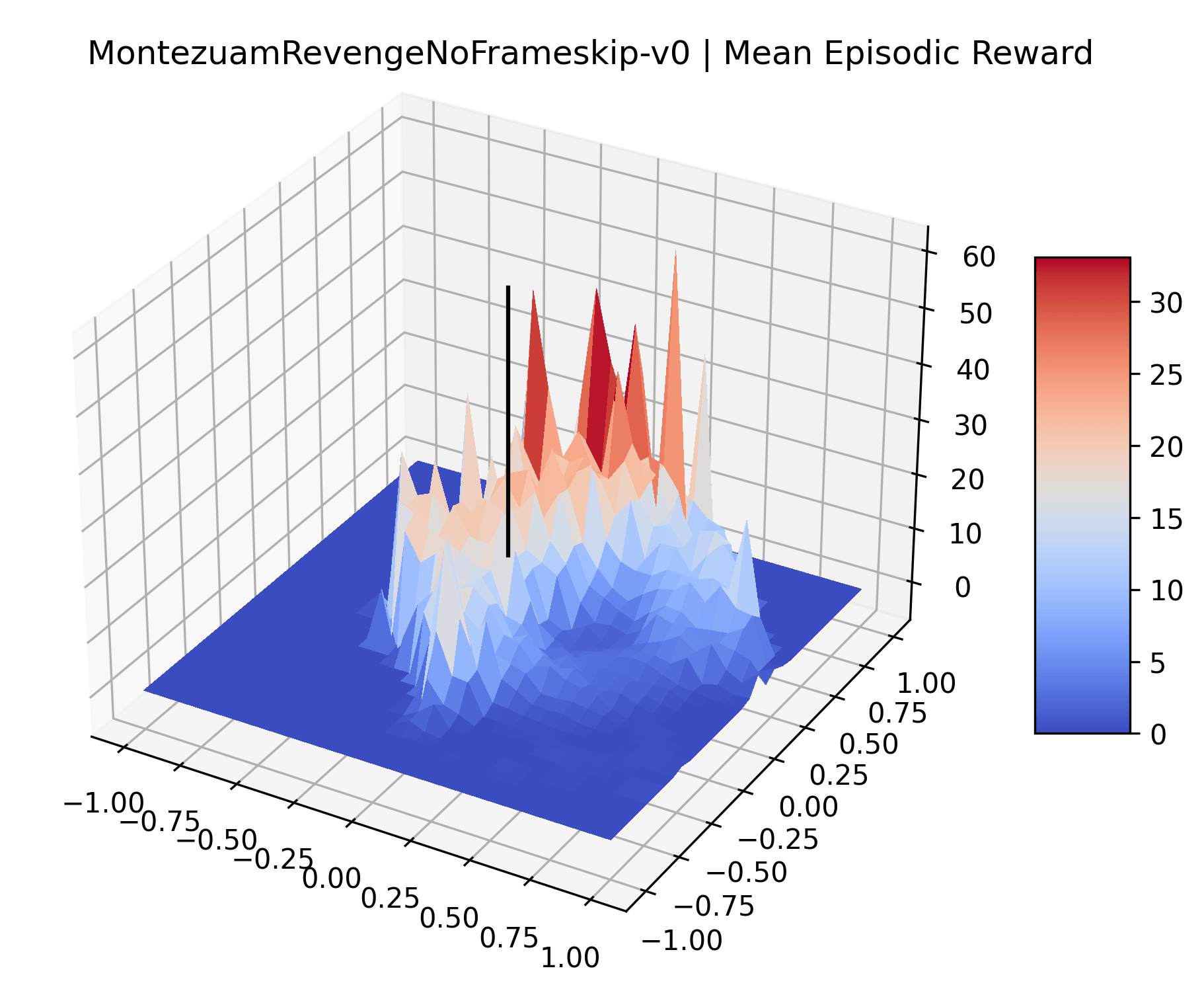} &
 \includegraphics[width=\variancescale]{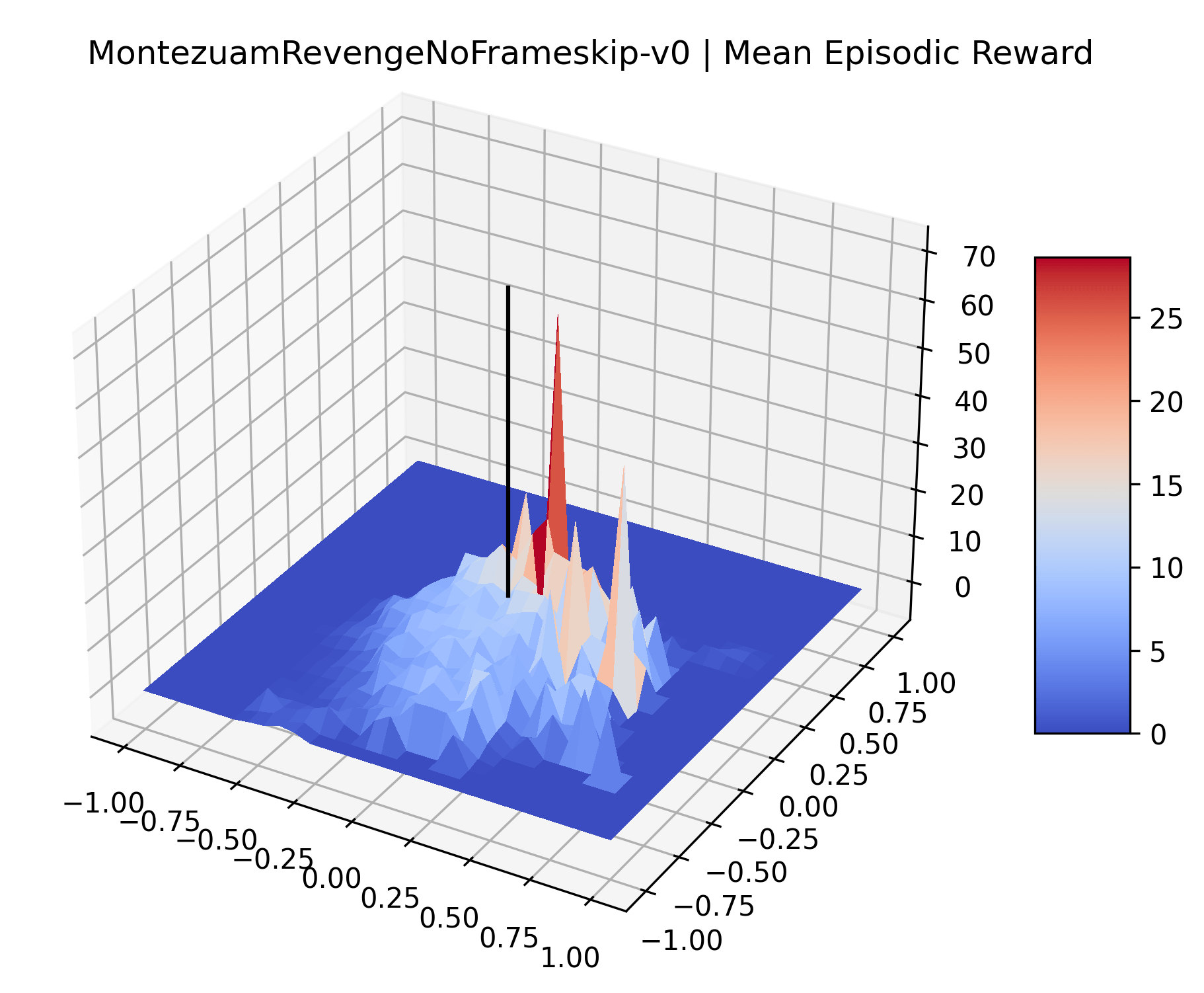} &
 \includegraphics[width=\variancescale]{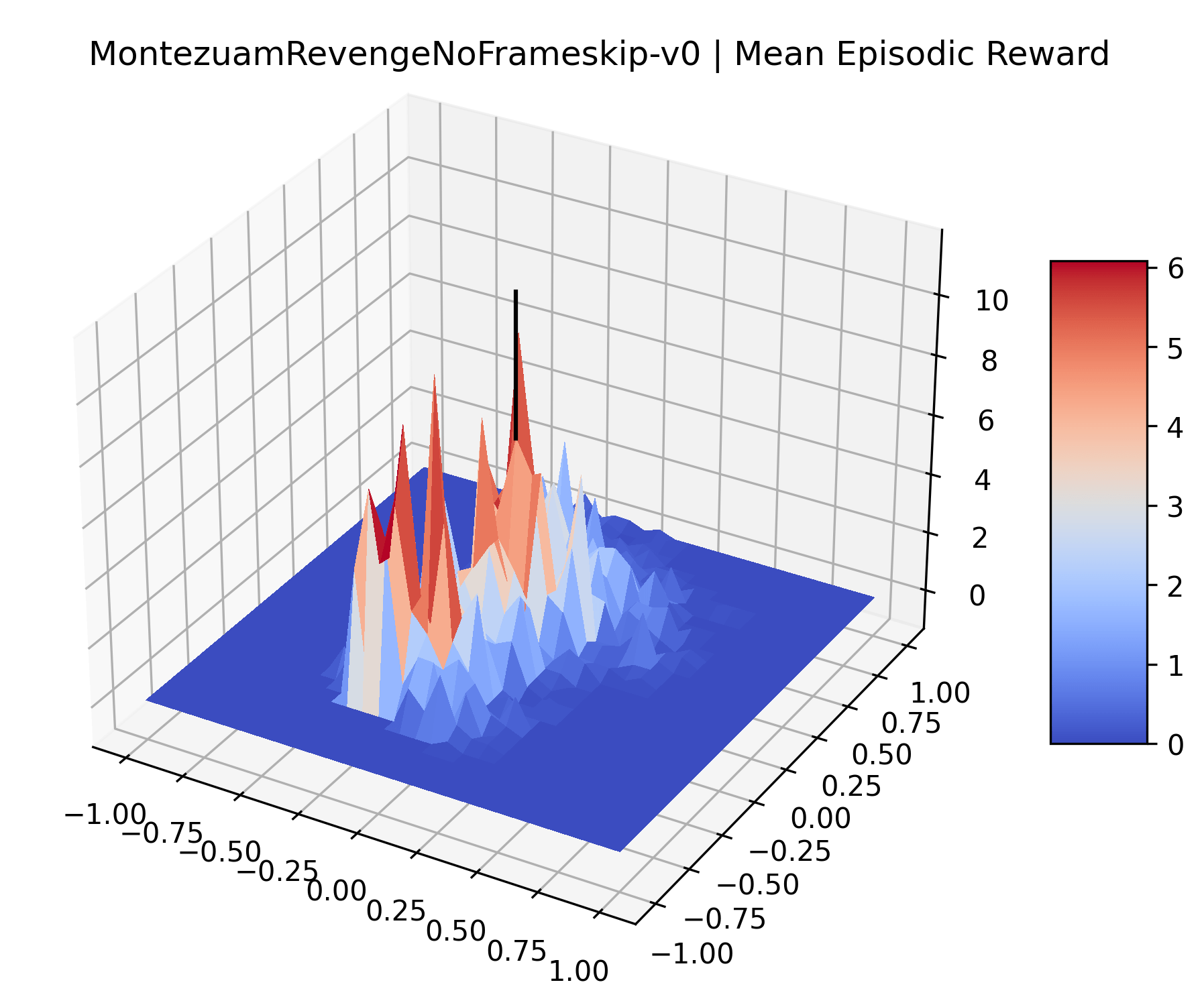} \\
\end{tabular}
\caption{18 training and plotting runs for the Atari Montezuma's Revenge environment.}
\label{fig:sparse_atari_variance_table}
\end{figure*}
\pagebreak

\section{Standard Error Plots}
\label{appendix:standard_error}

\subsection{Classic Control}
\begin{figure*}[!ht]
\centering
\begin{tabular}{ccc}
 \includegraphics[width=\surfacescale]{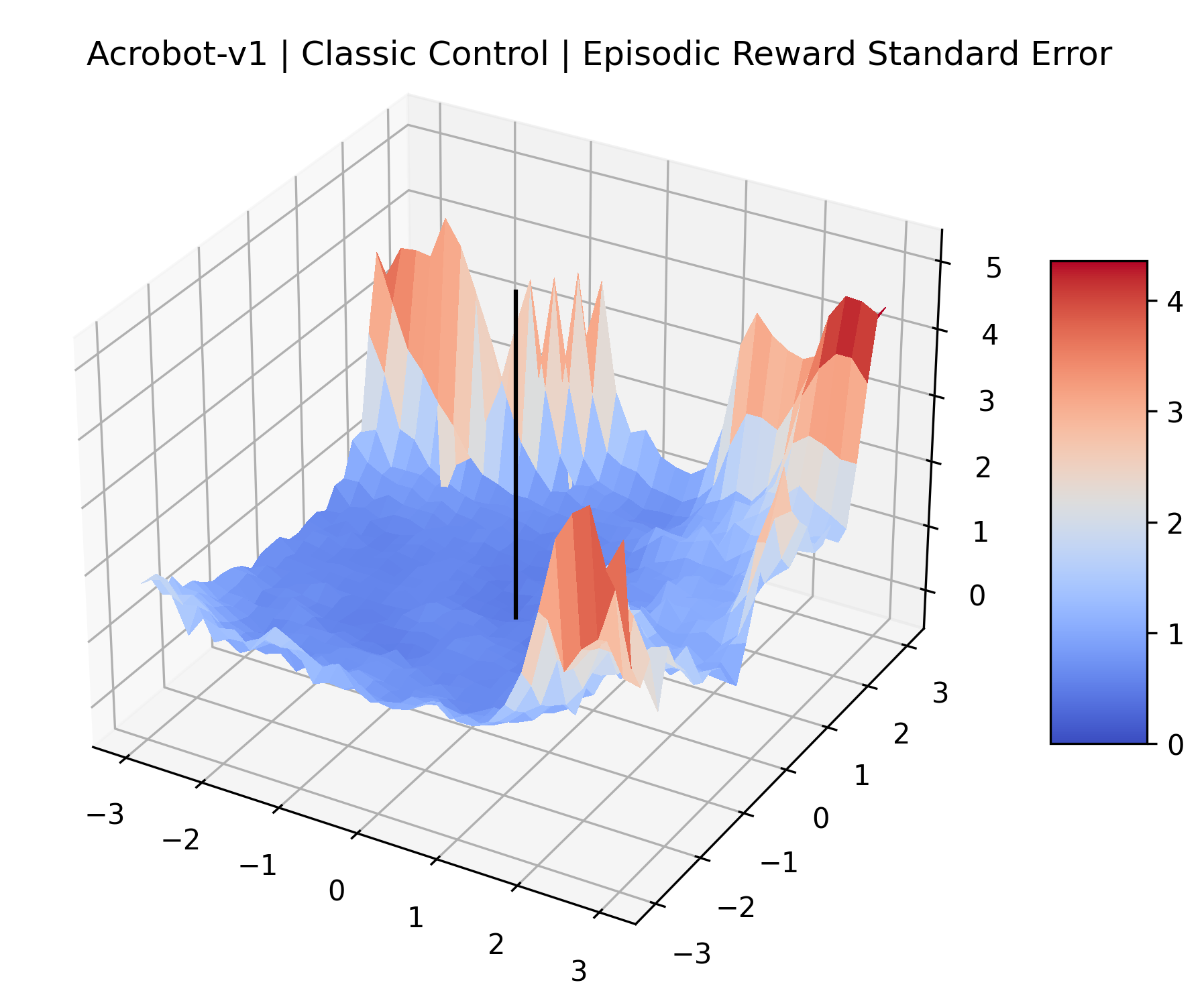} &
 \includegraphics[width=\surfacescale]{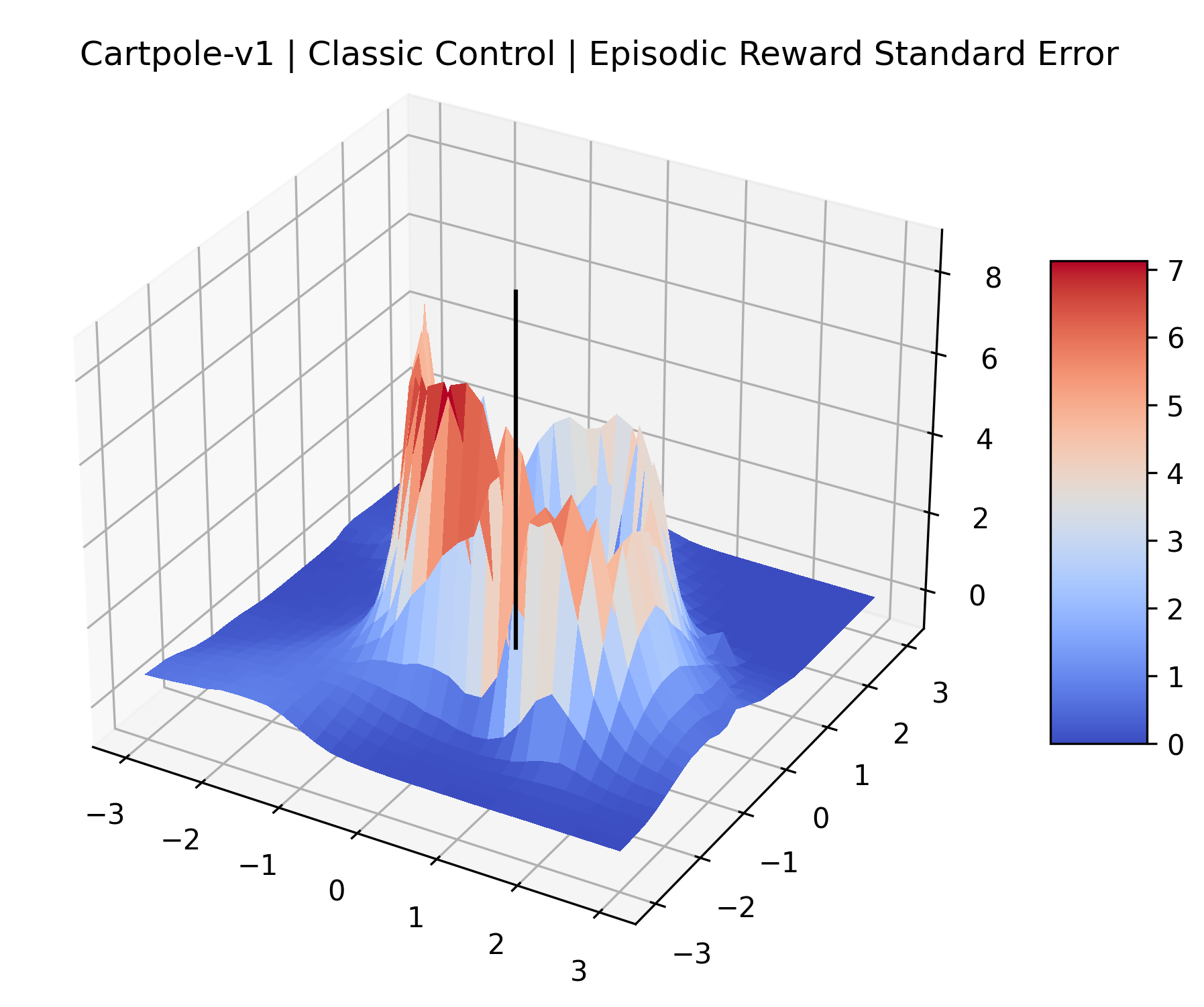} &
 \includegraphics[width=\surfacescale]{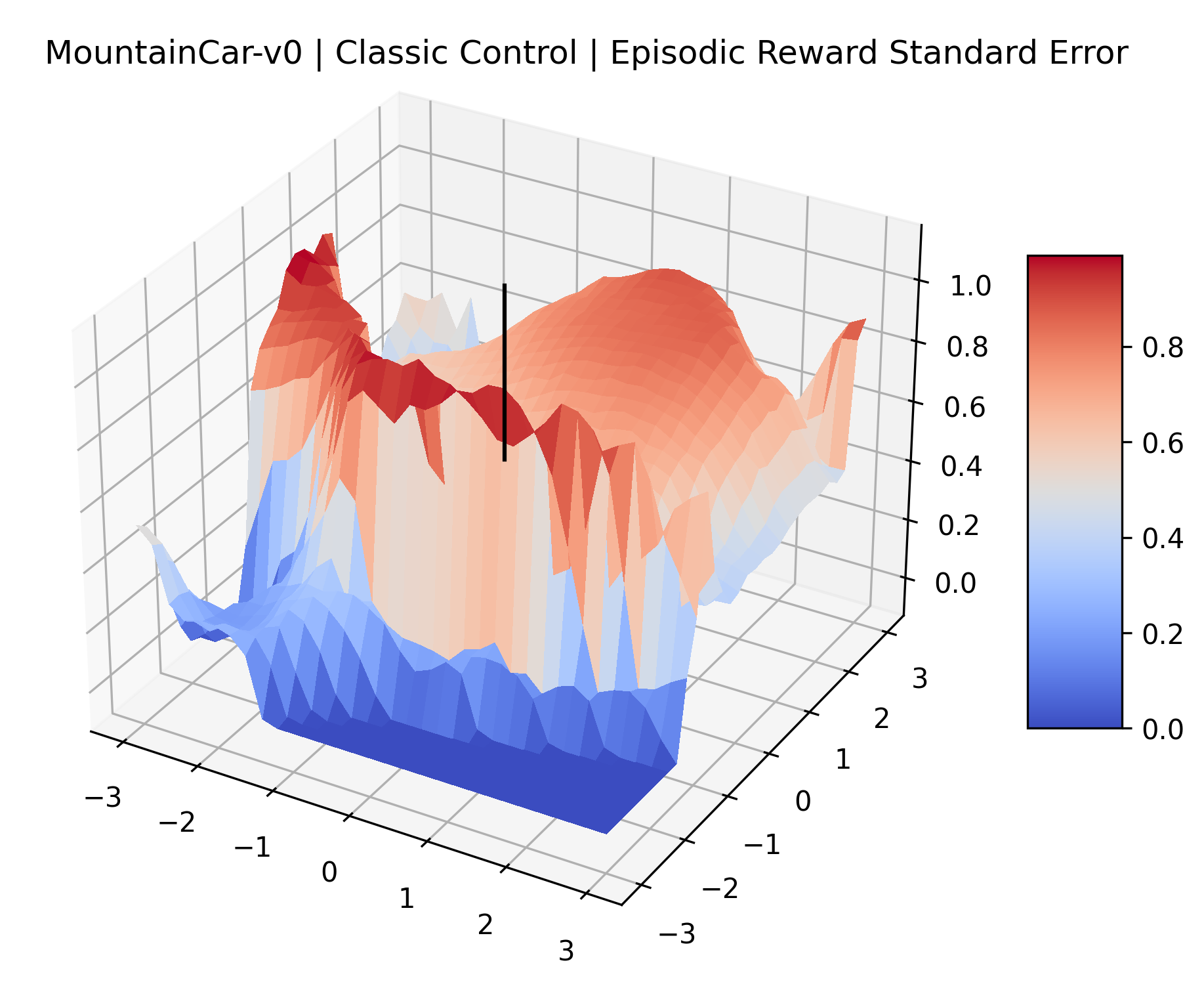} \\
\end{tabular}
\begin{tabular}{cc}
 \includegraphics[width=\surfacescale]{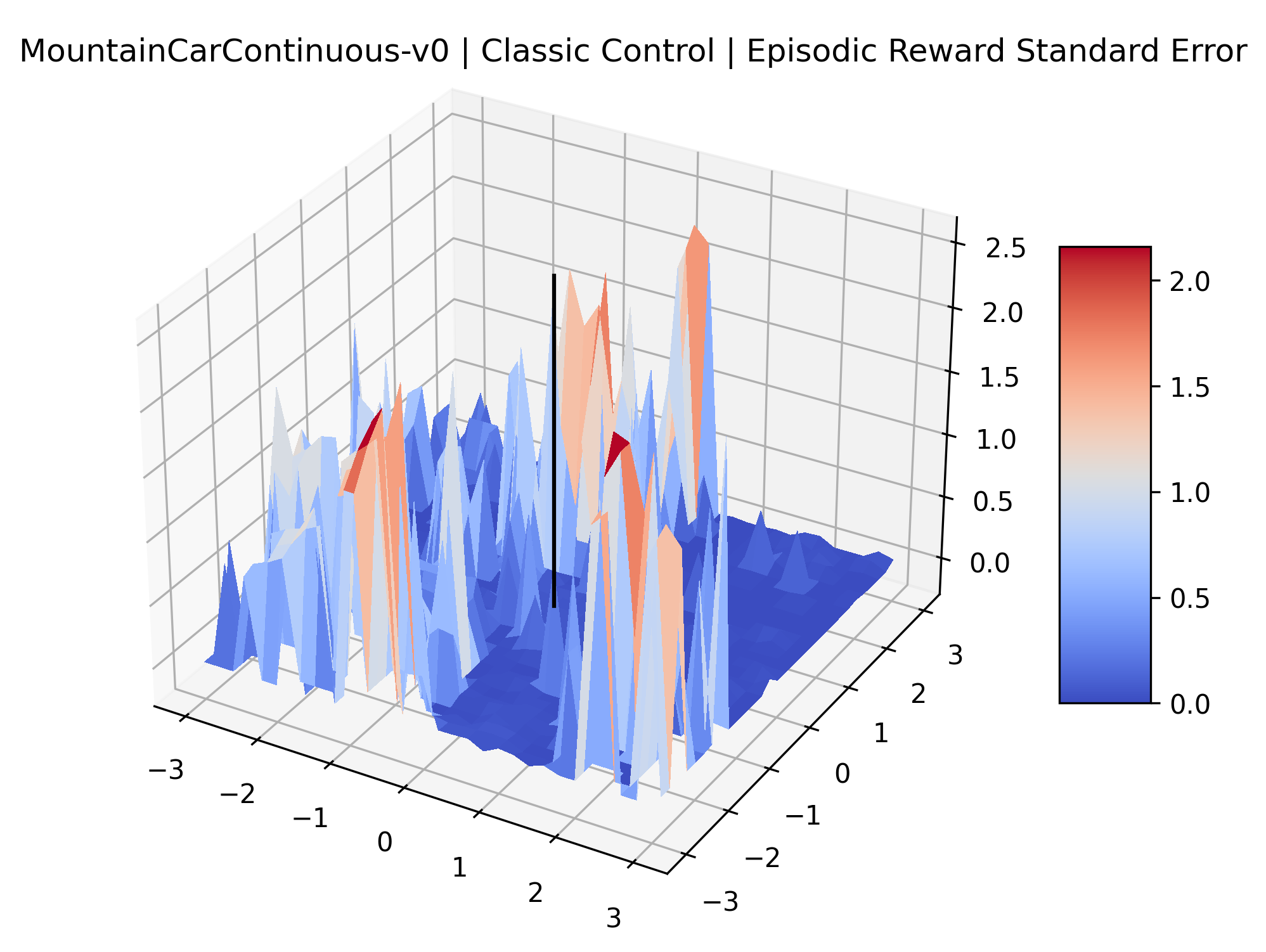} &
 \includegraphics[width=\surfacescale]{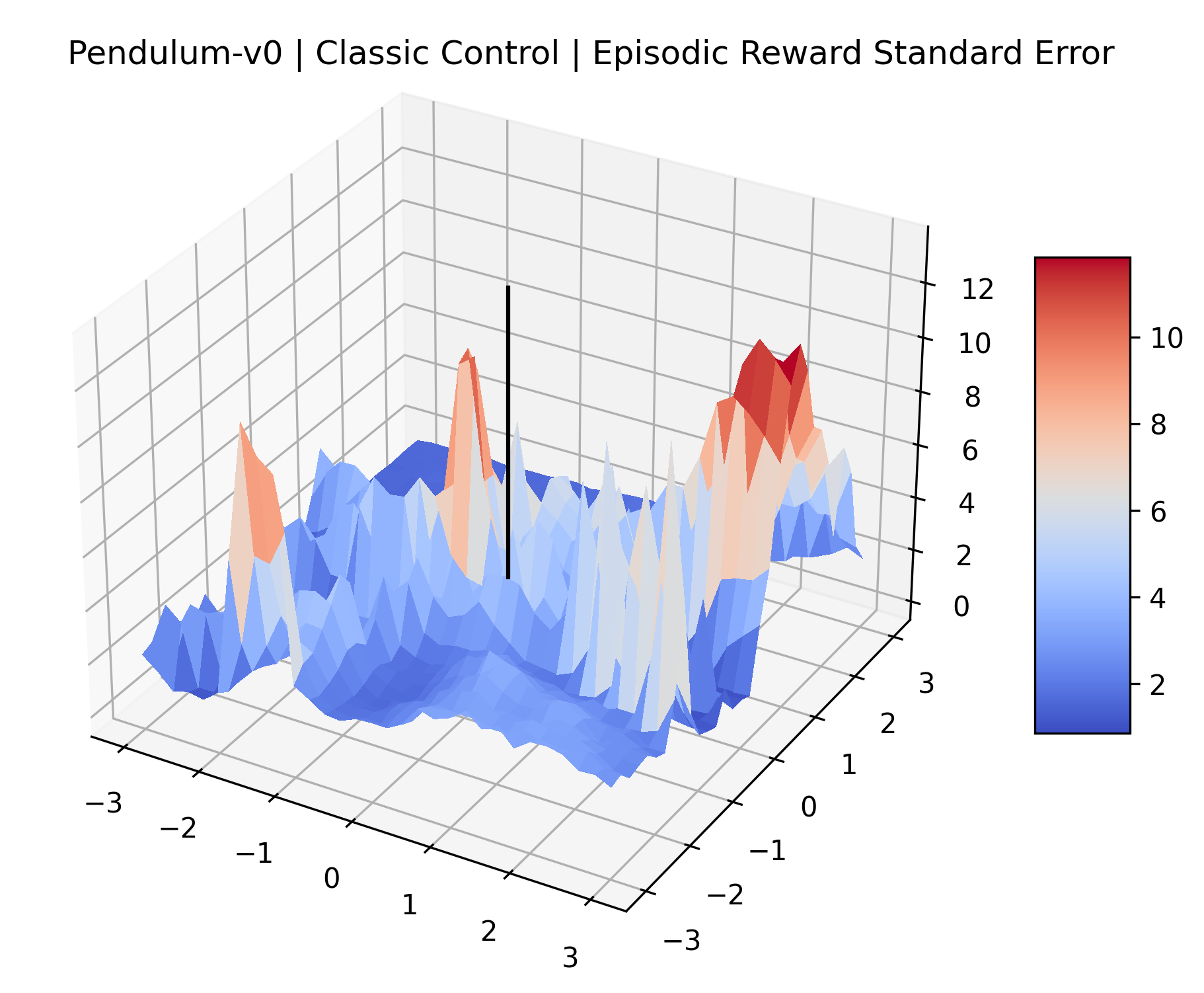} \\
\end{tabular}
\caption{Standard error surfaces for 5 Classic Control environments.}
\label{fig:classiccontrol_standarderror_table}
\end{figure*}
\pagebreak

\subsection{MuJoCo}
\begin{figure*}[!ht]
\centering
\begin{tabular}{ccc}
 \includegraphics[width=\surfacescale]{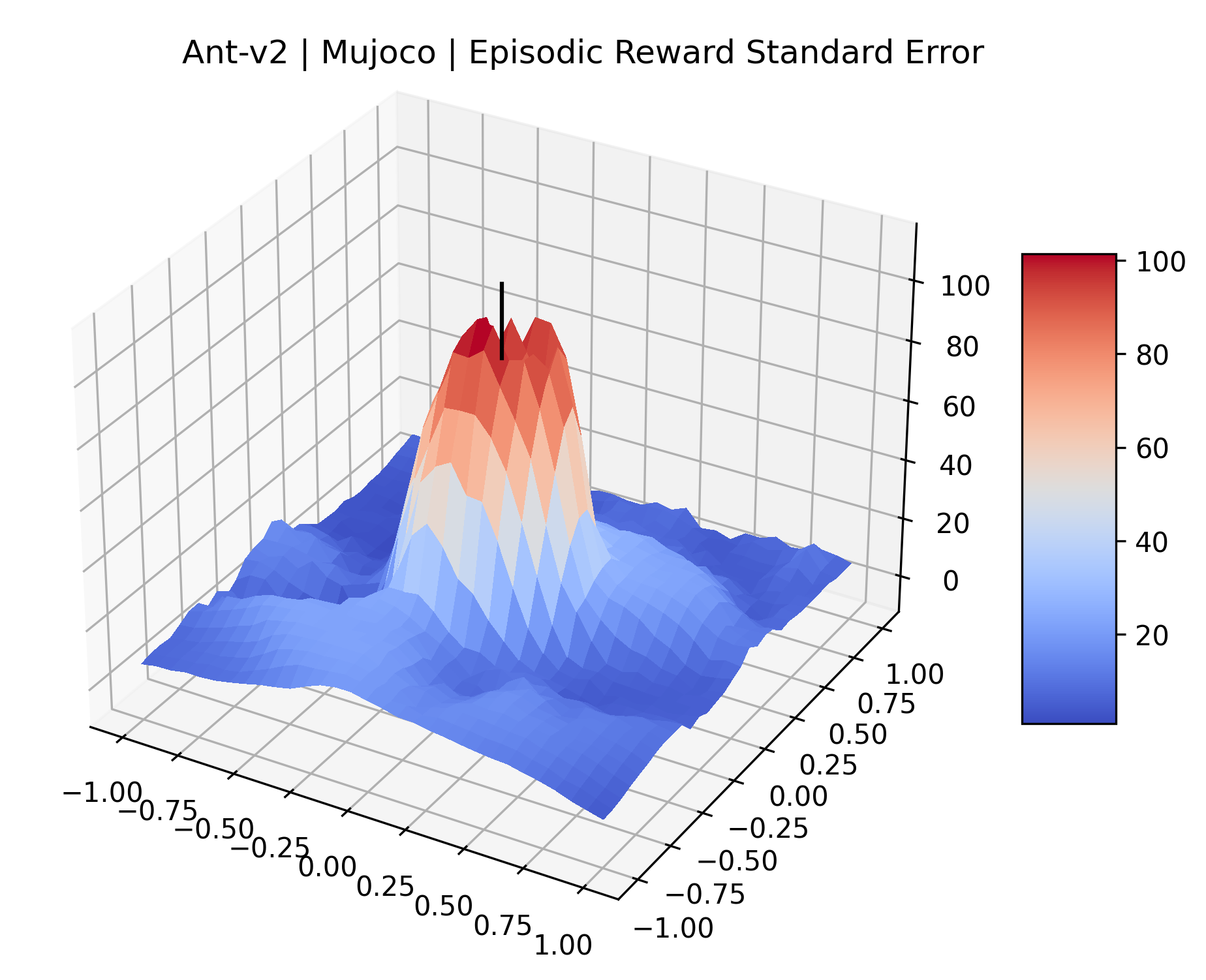} &
 \includegraphics[width=\surfacescale]{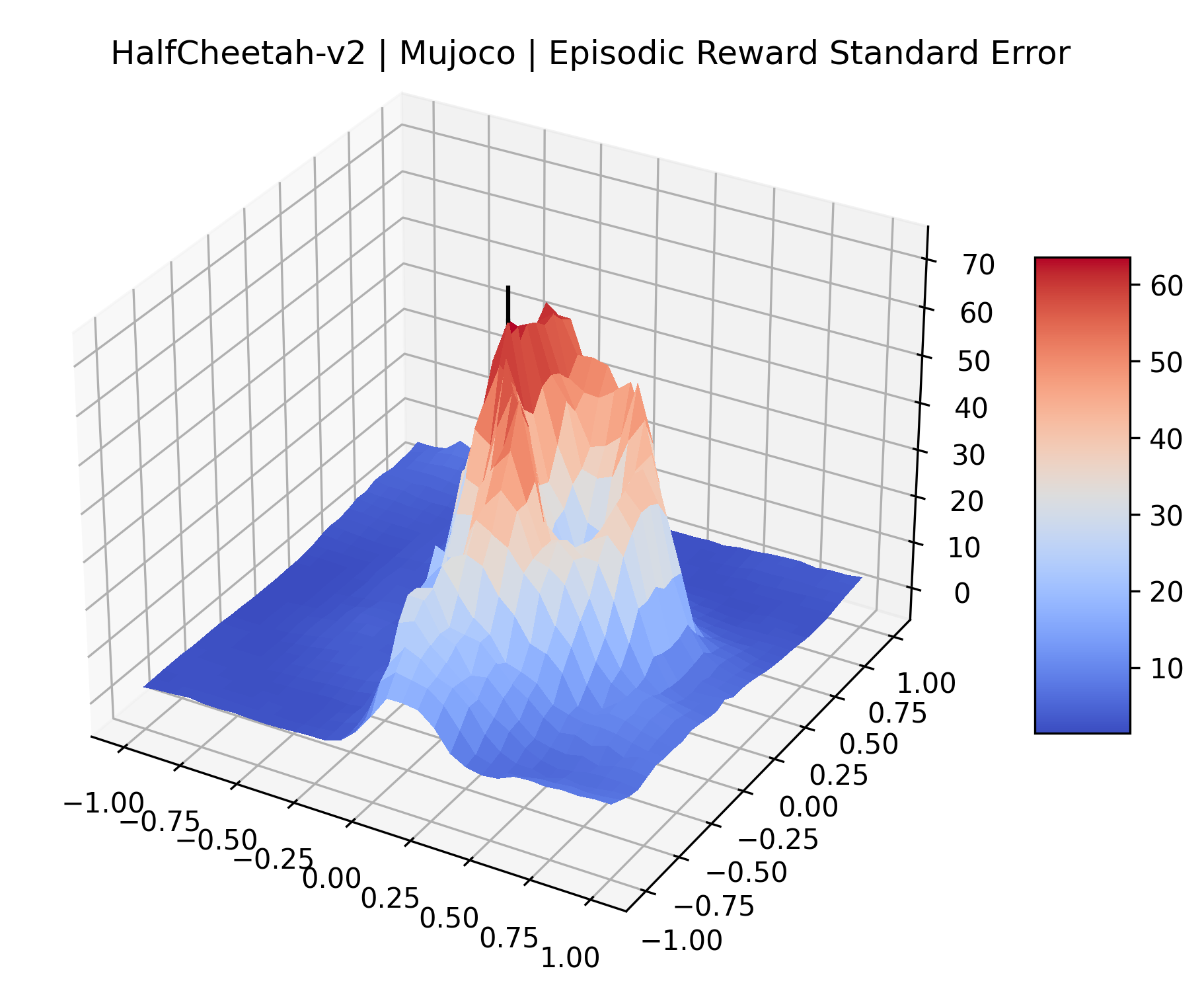} &
 \includegraphics[width=\surfacescale]{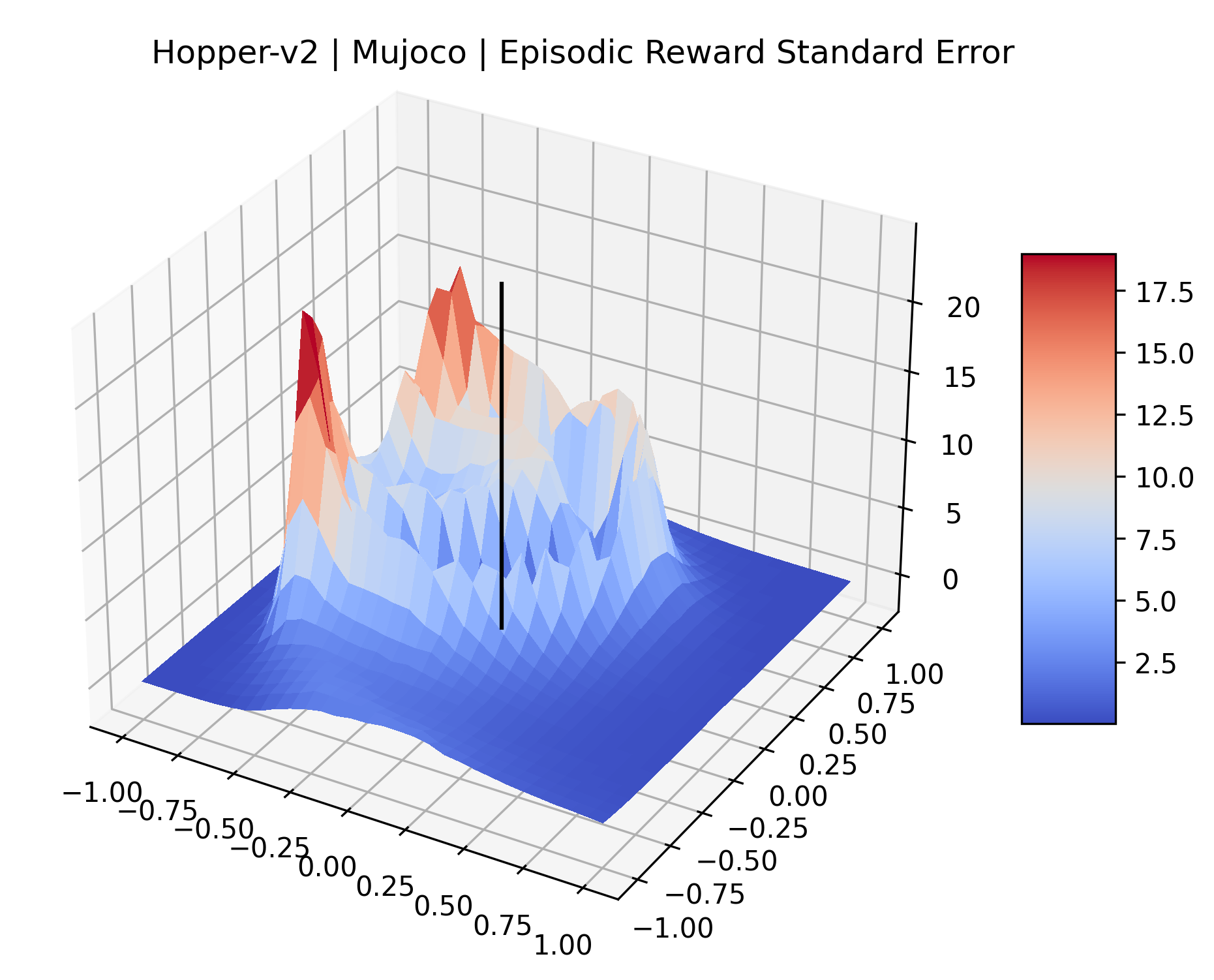} \\
 \includegraphics[width=\surfacescale]{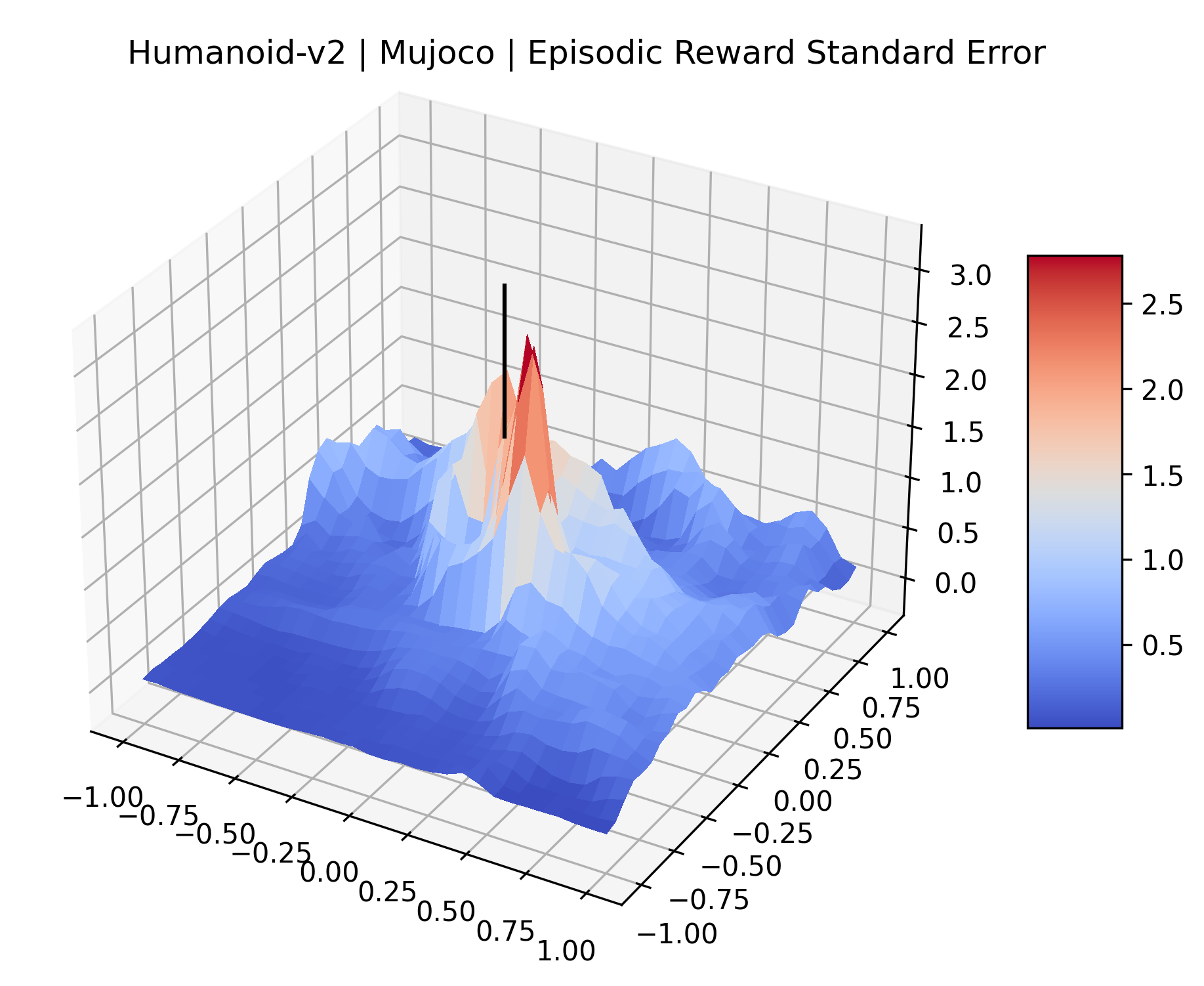} &
 \includegraphics[width=\surfacescale]{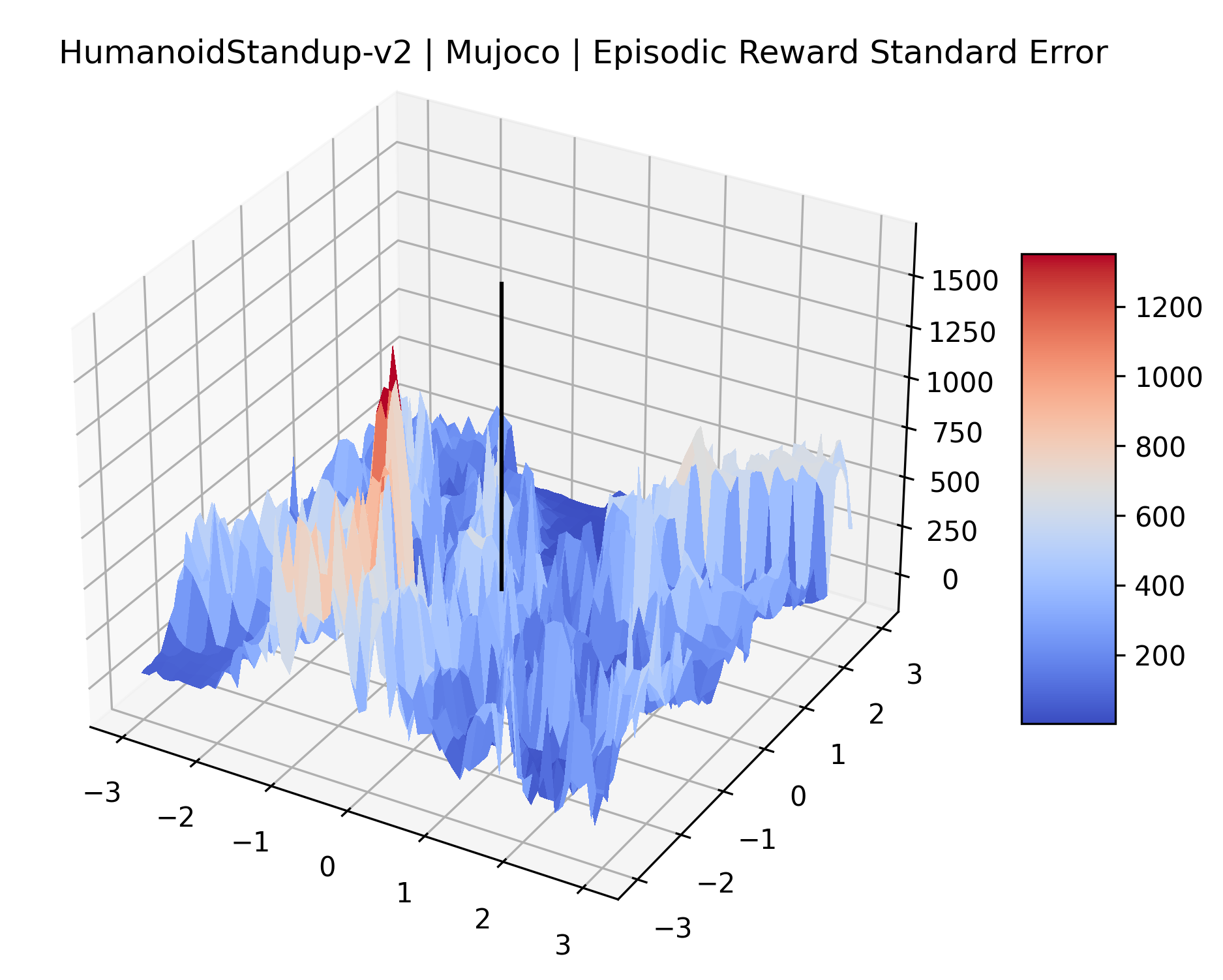} &
 \includegraphics[width=\surfacescale]{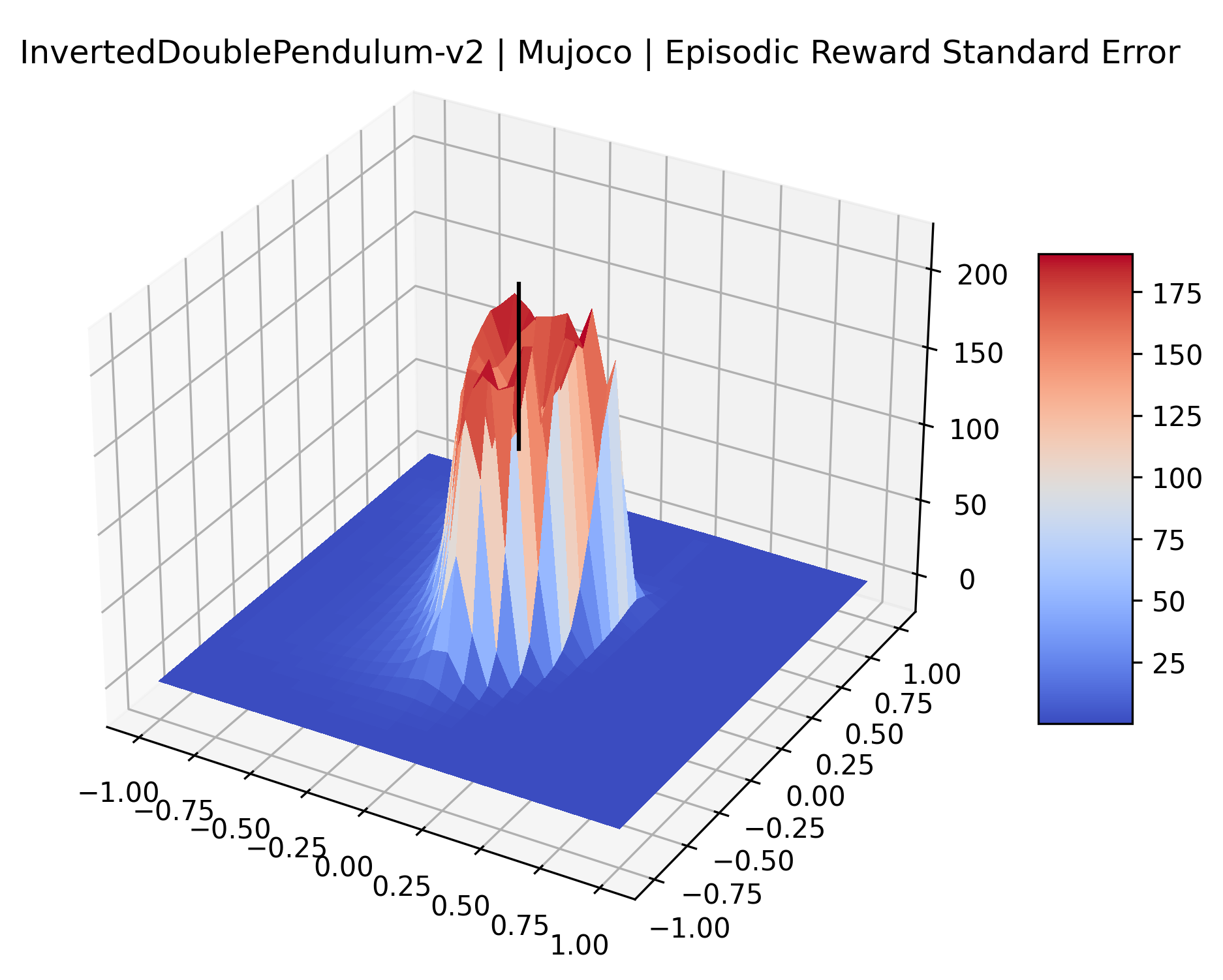} \\
 \includegraphics[width=\surfacescale]{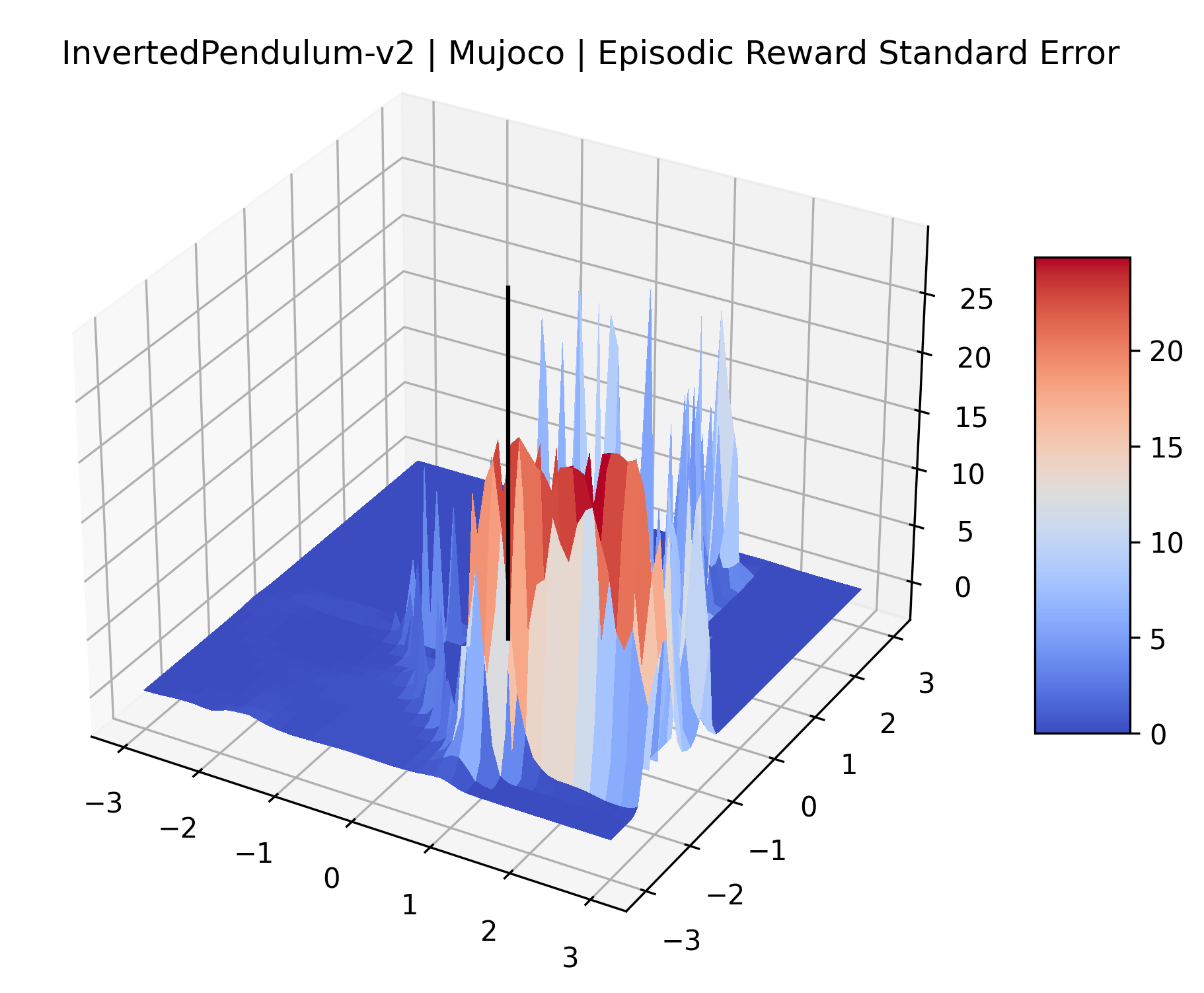} &
 \includegraphics[width=\surfacescale]{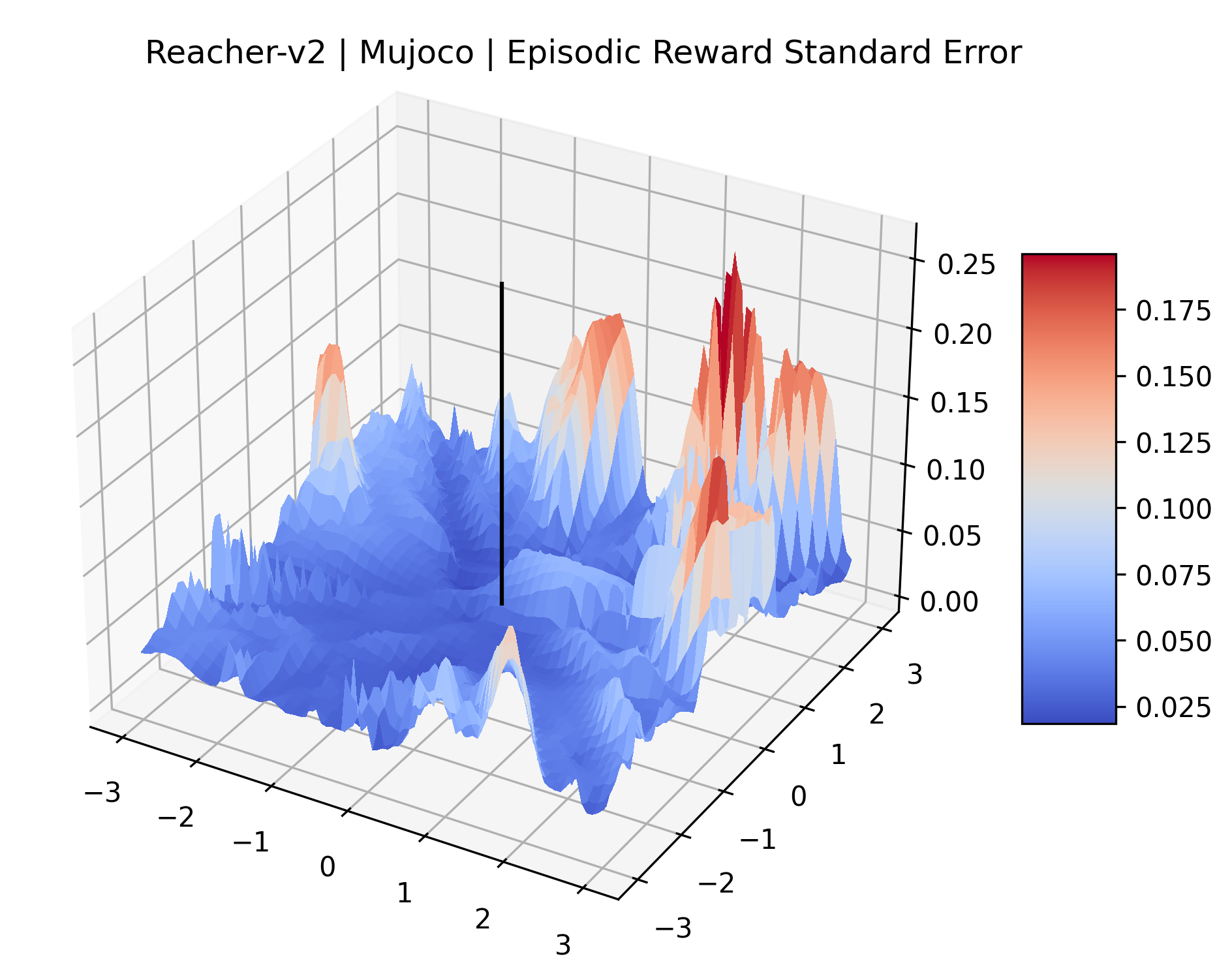} &
 \includegraphics[width=\surfacescale]{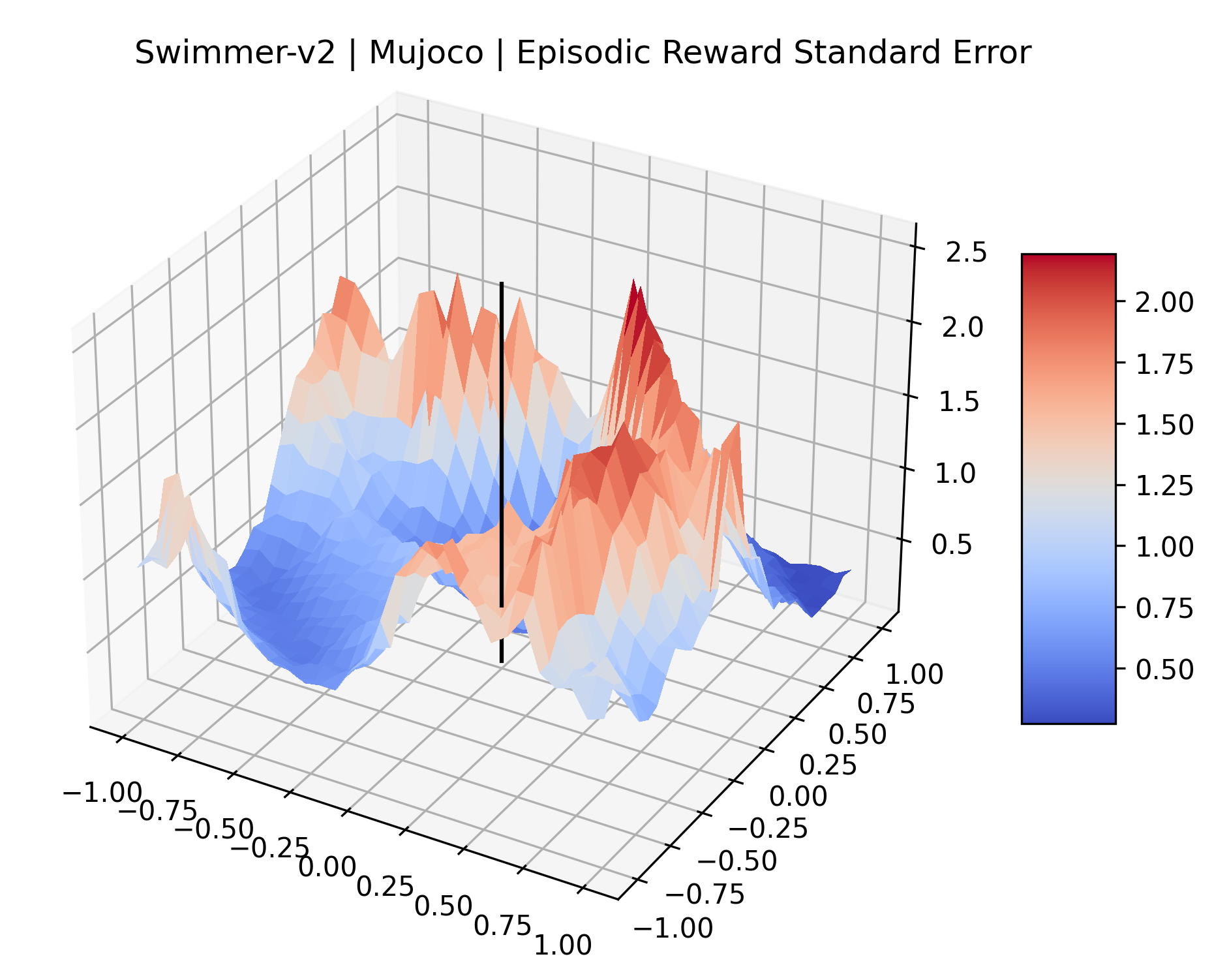} \\ 
 & \includegraphics[width=\surfacescale]{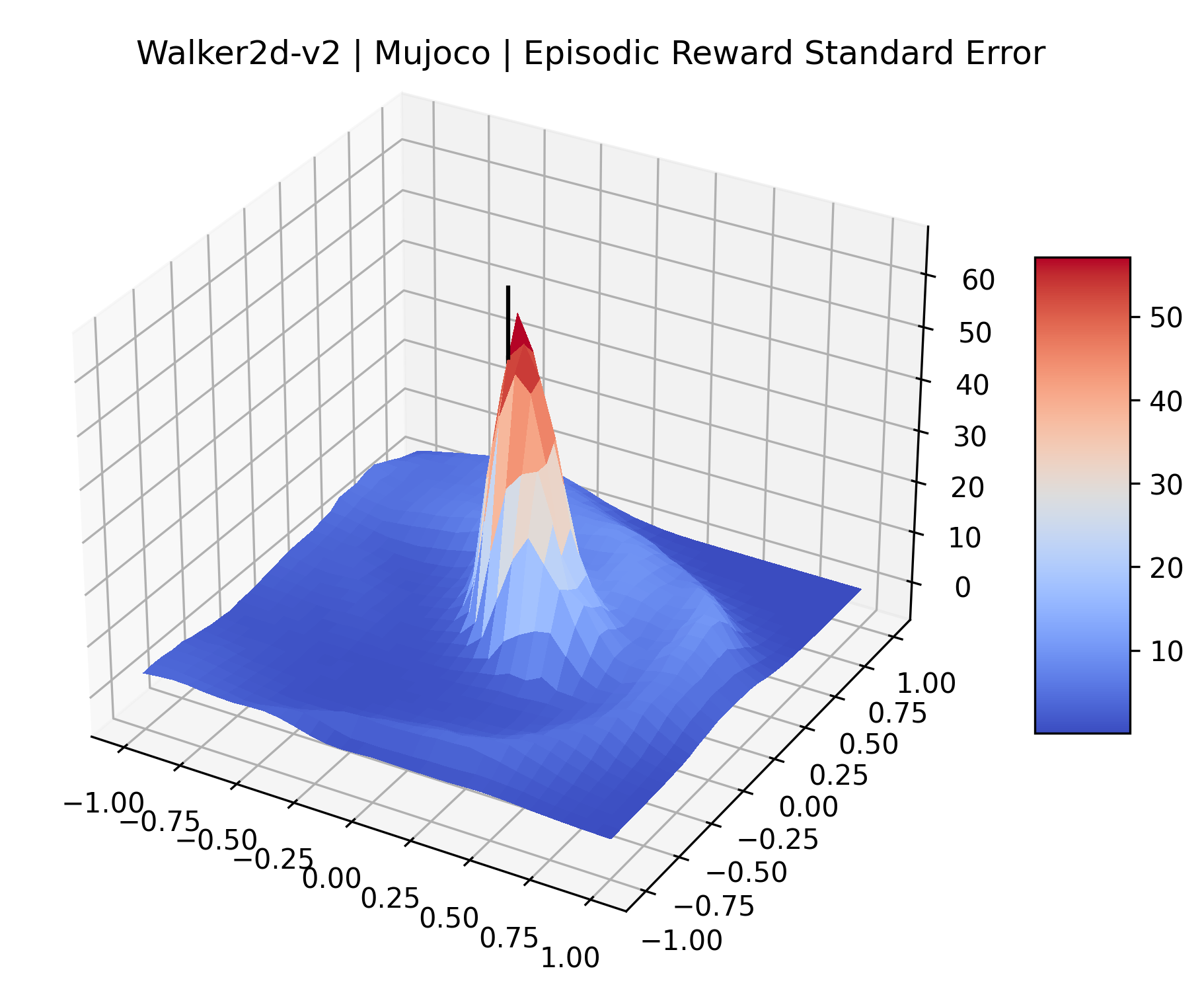} & \\
\end{tabular}
\caption{Standard error surfaces for 10 MuJoCo environments.}
\label{fig:mujococ_standarderror_table}
\end{figure*}
\pagebreak

\subsection{Atari}
\begin{figure*}[!ht]
\centering
\begin{tabular}{ccc}
 \includegraphics[width=\surfacescale]{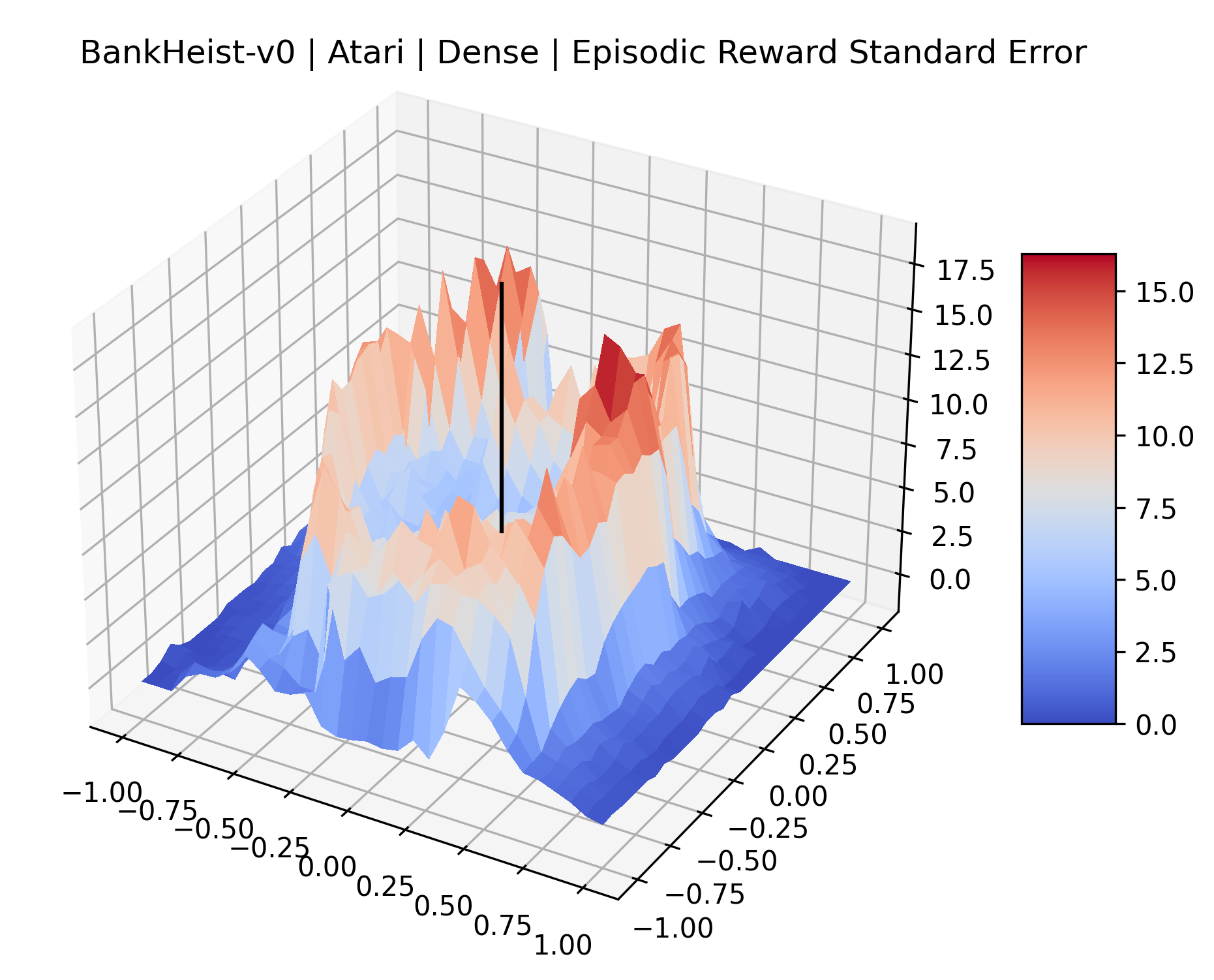} &
 \includegraphics[width=\surfacescale]{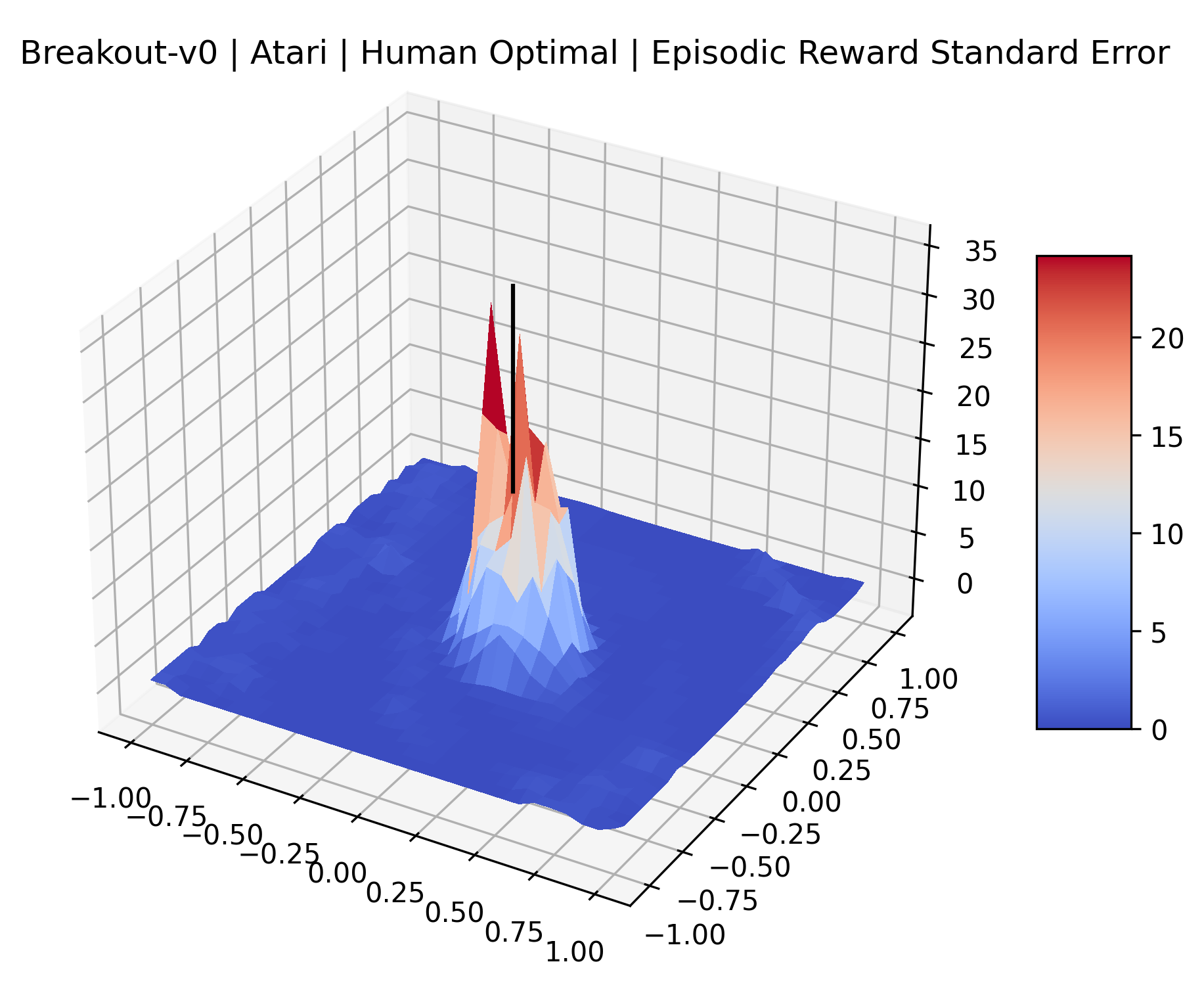} &
 \includegraphics[width=\surfacescale]{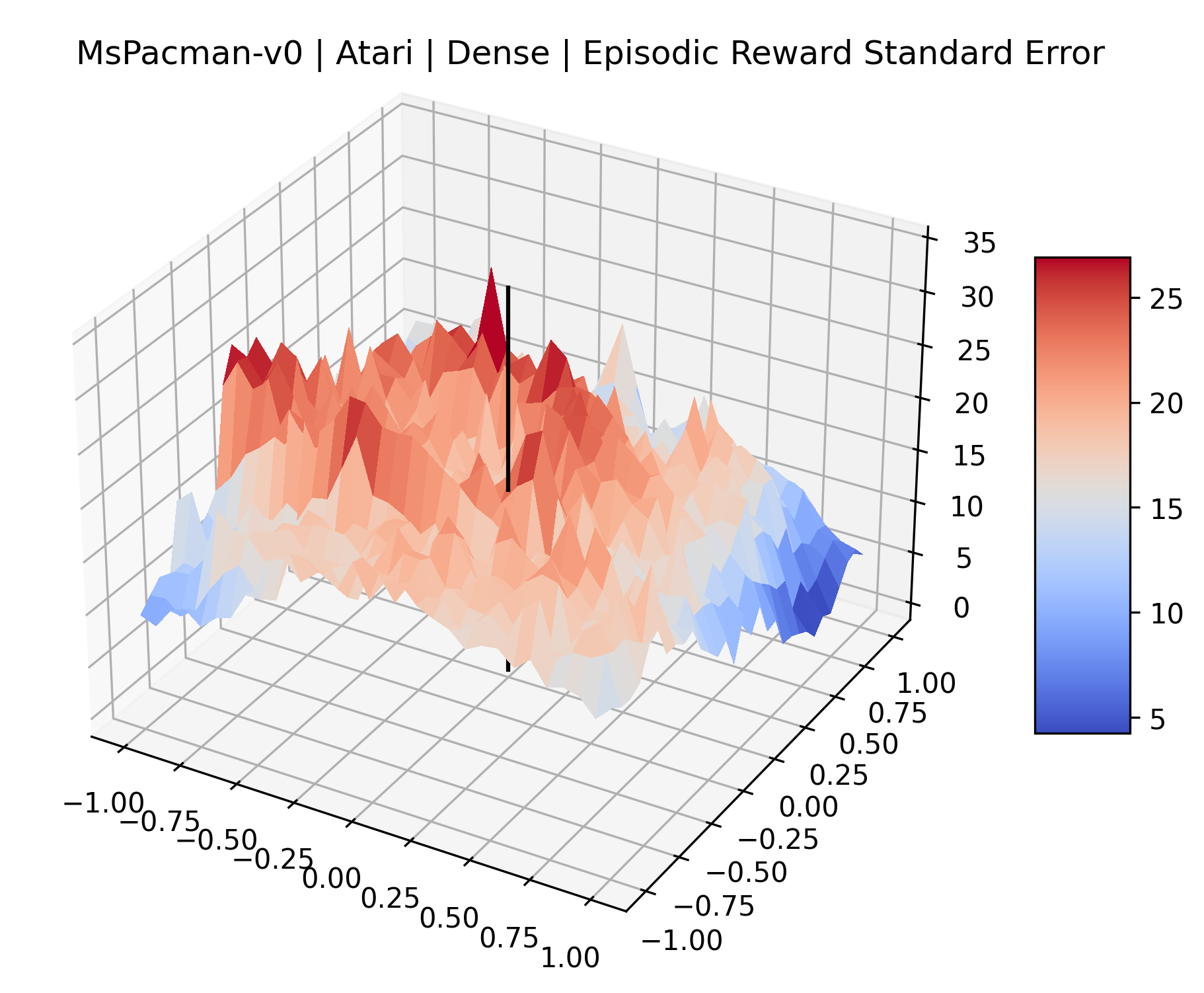} \\
 \includegraphics[width=\surfacescale]{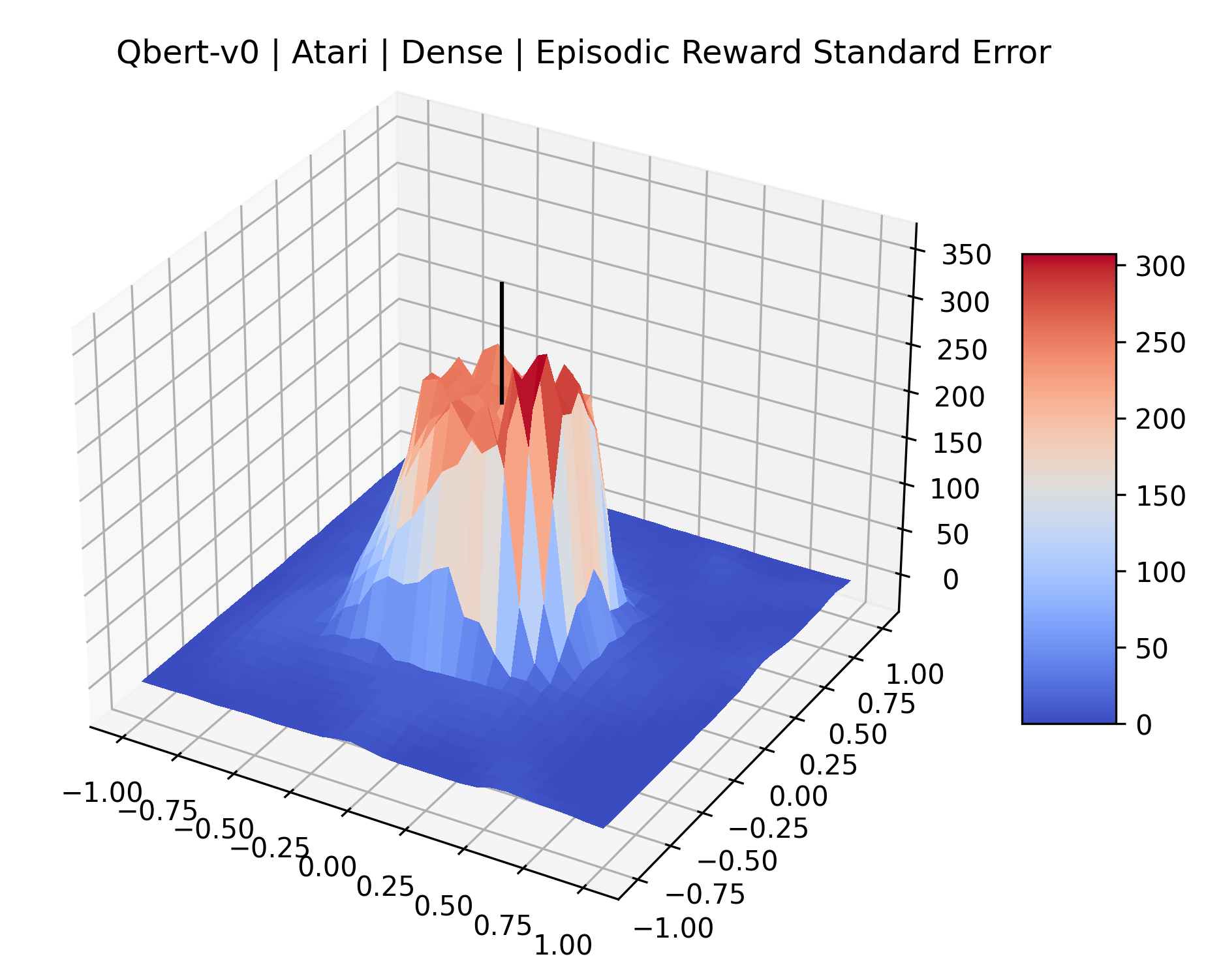} &
 \includegraphics[width=\surfacescale]{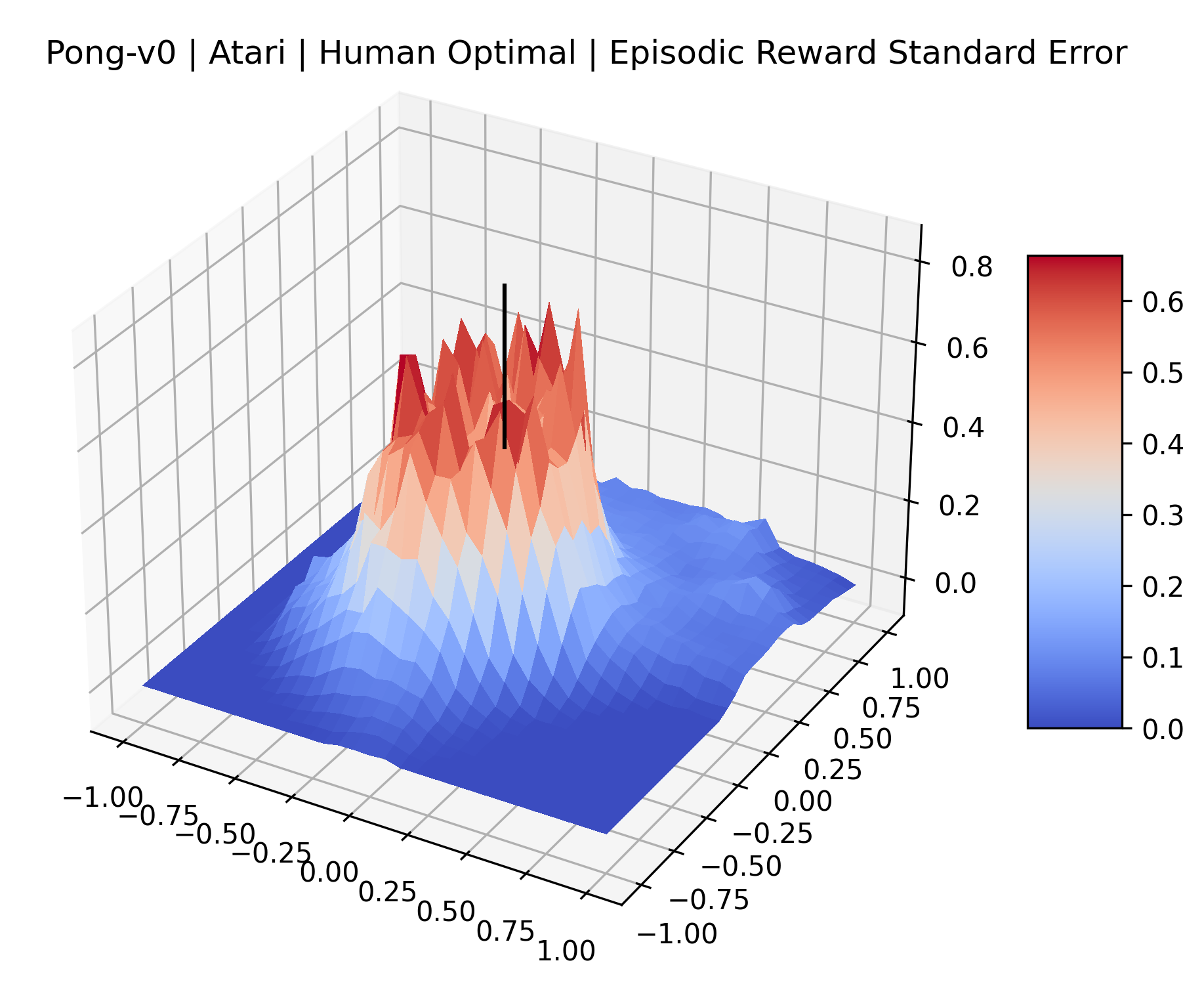} &
 \includegraphics[width=\surfacescale]{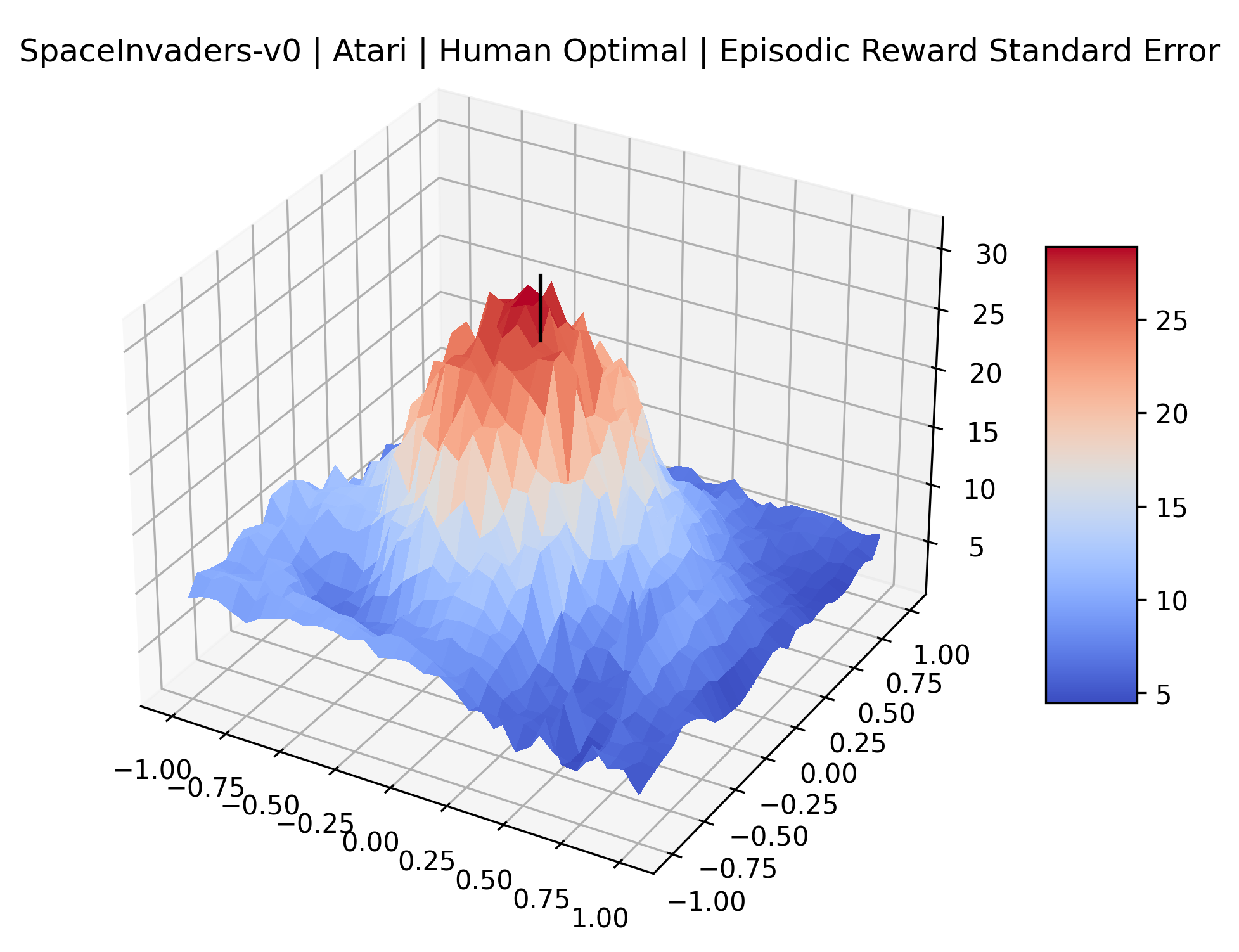} \\
 \includegraphics[width=\surfacescale]{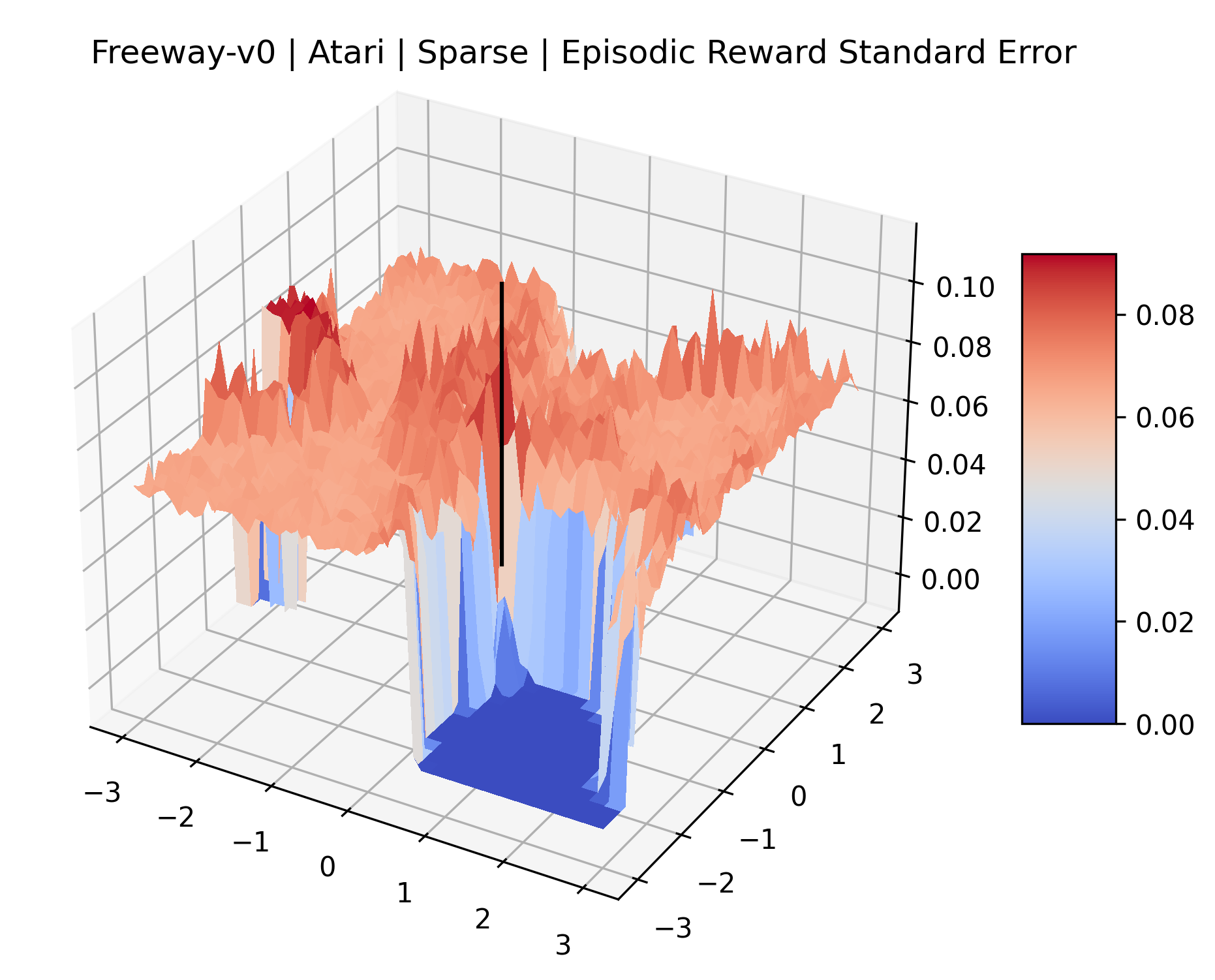} &
 \includegraphics[width=\surfacescale]{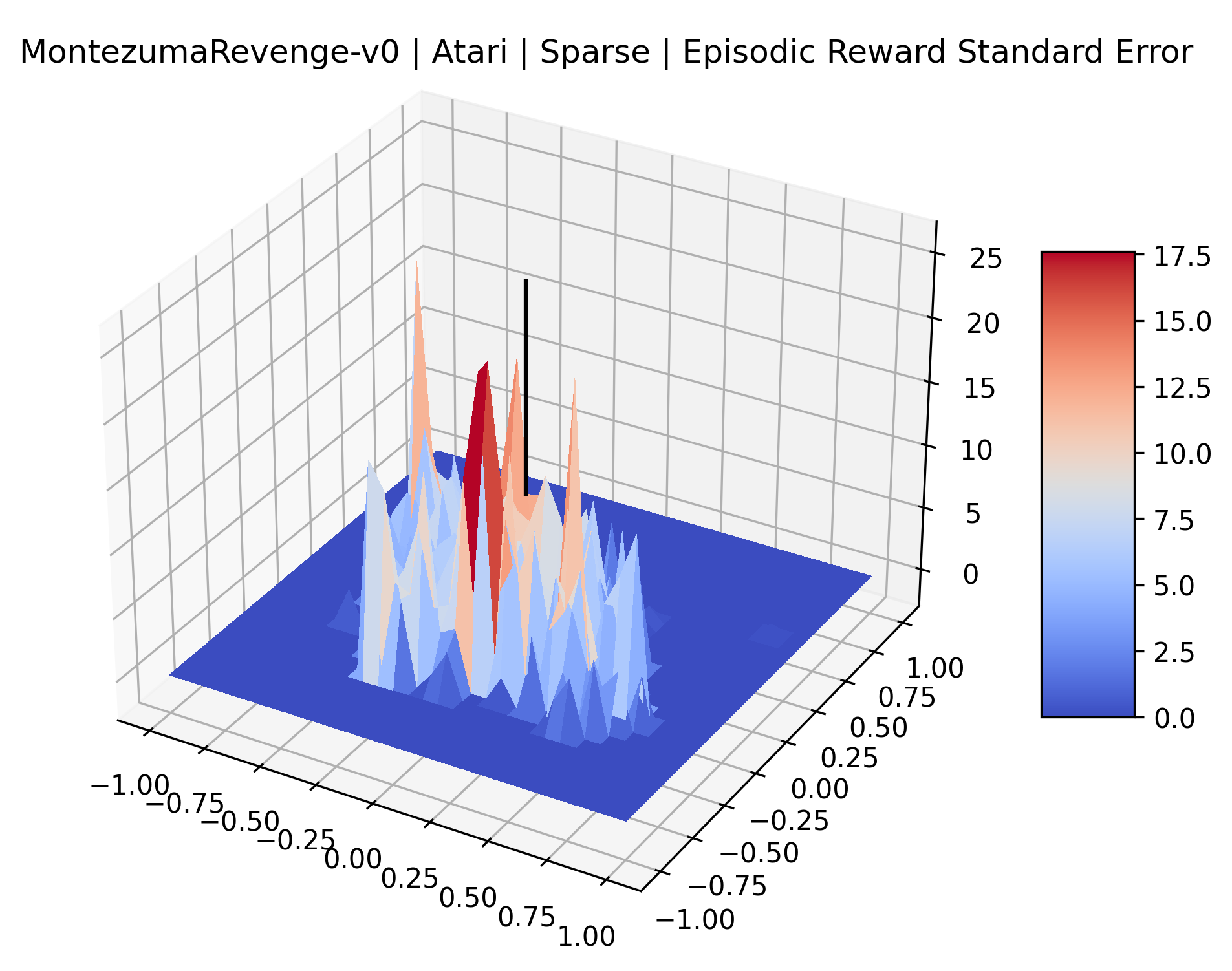} &
 \includegraphics[width=\surfacescale]{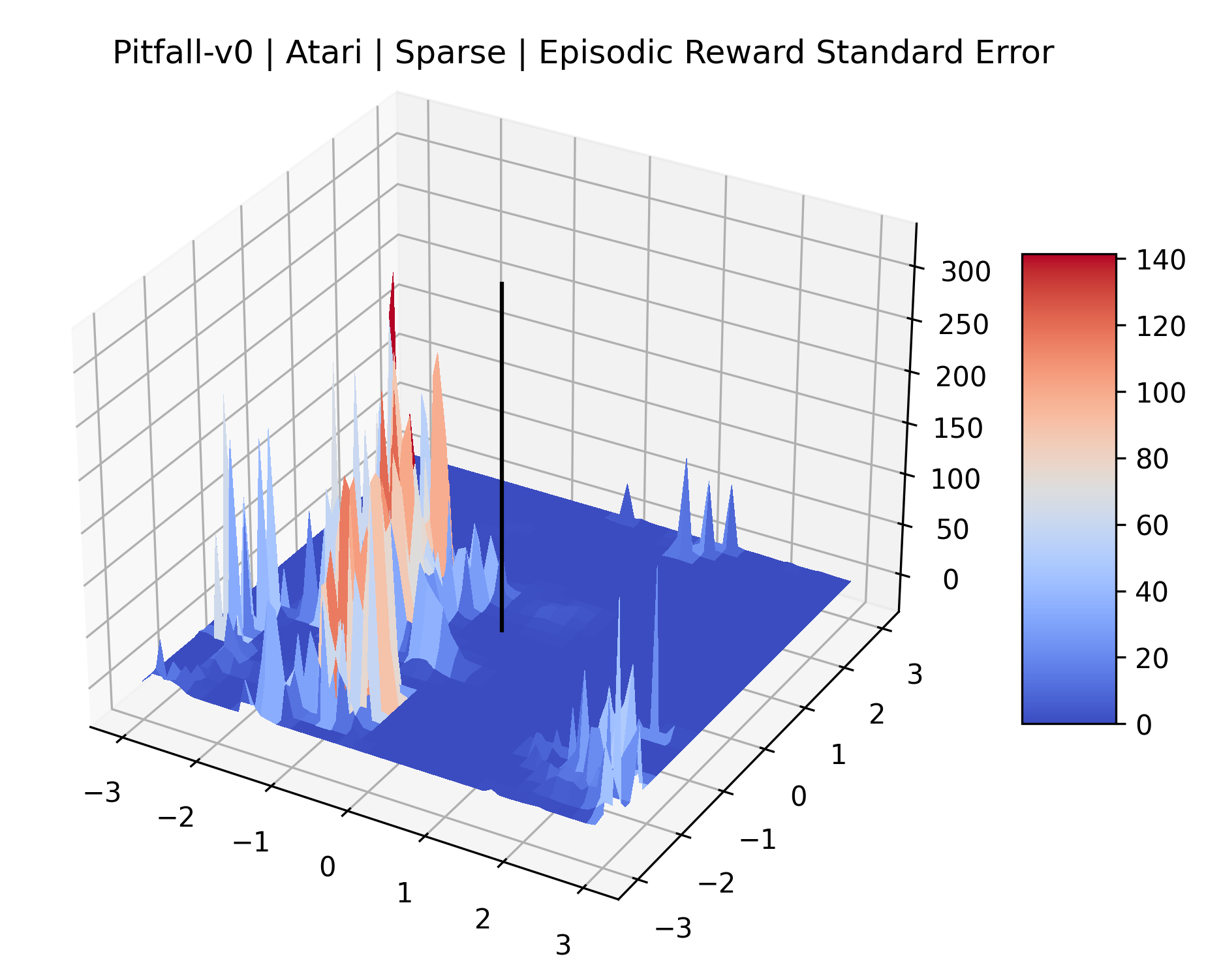} \\ \includegraphics[width=\surfacescale]{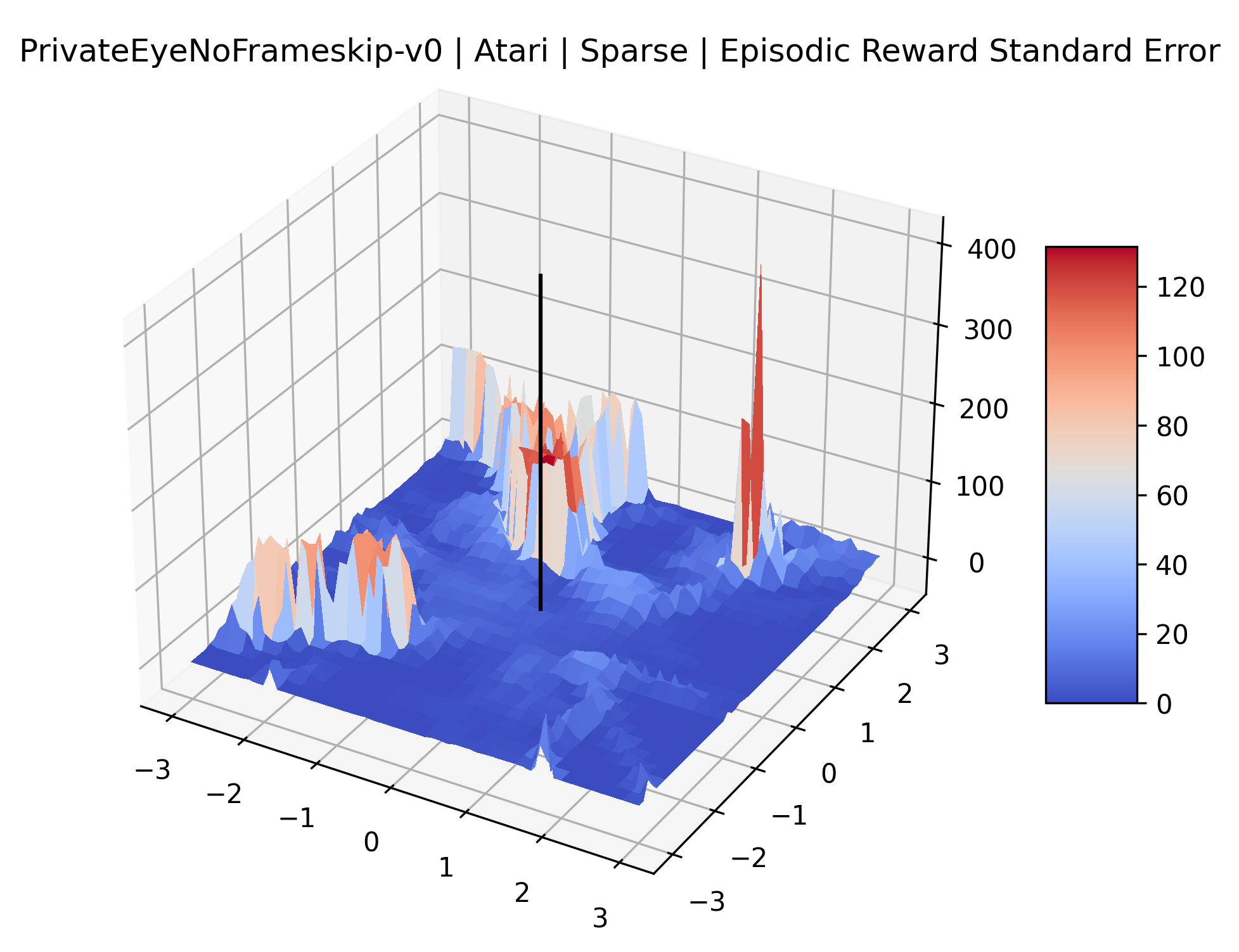} &
 \includegraphics[width=\surfacescale]{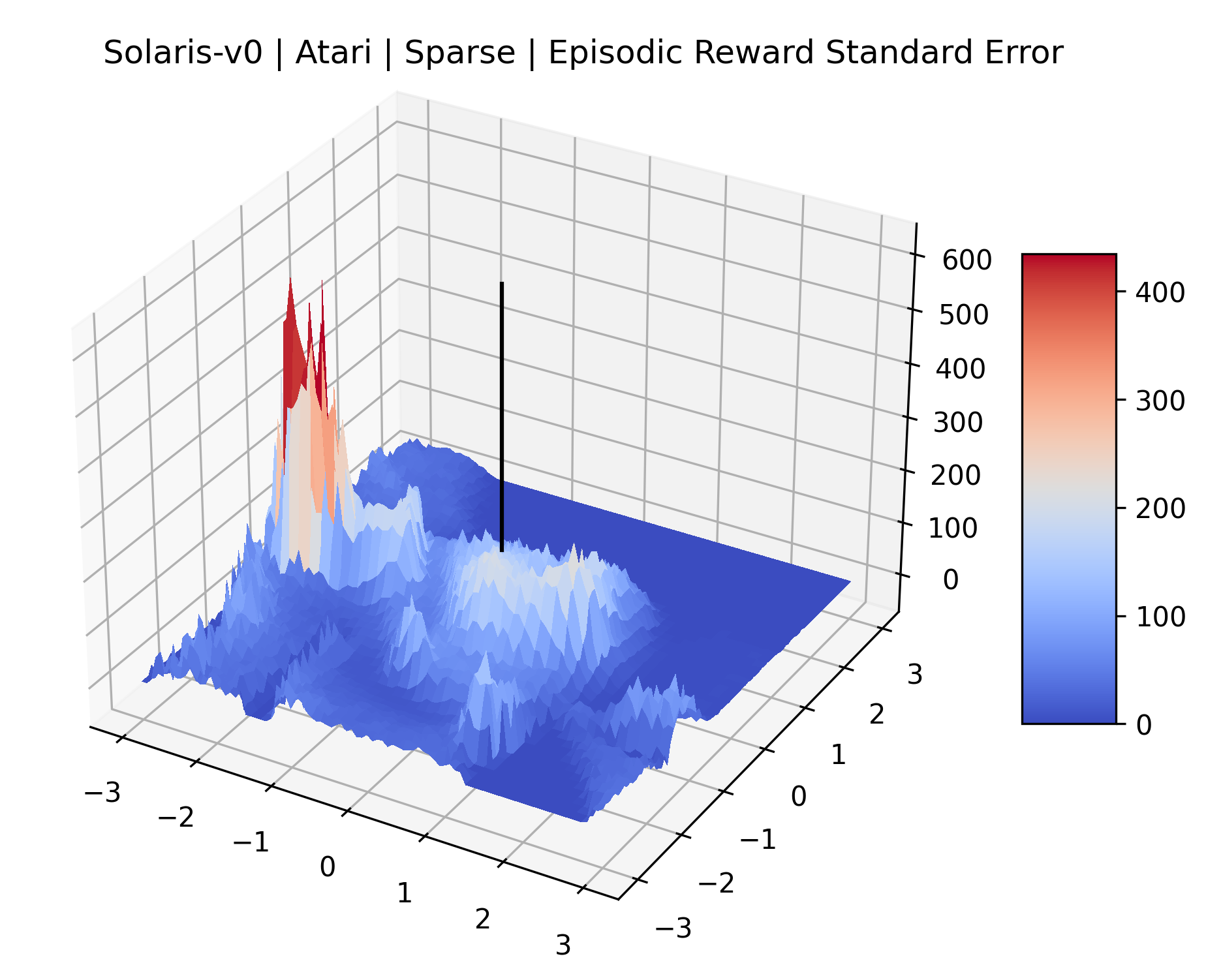} &
 \includegraphics[width=\surfacescale]{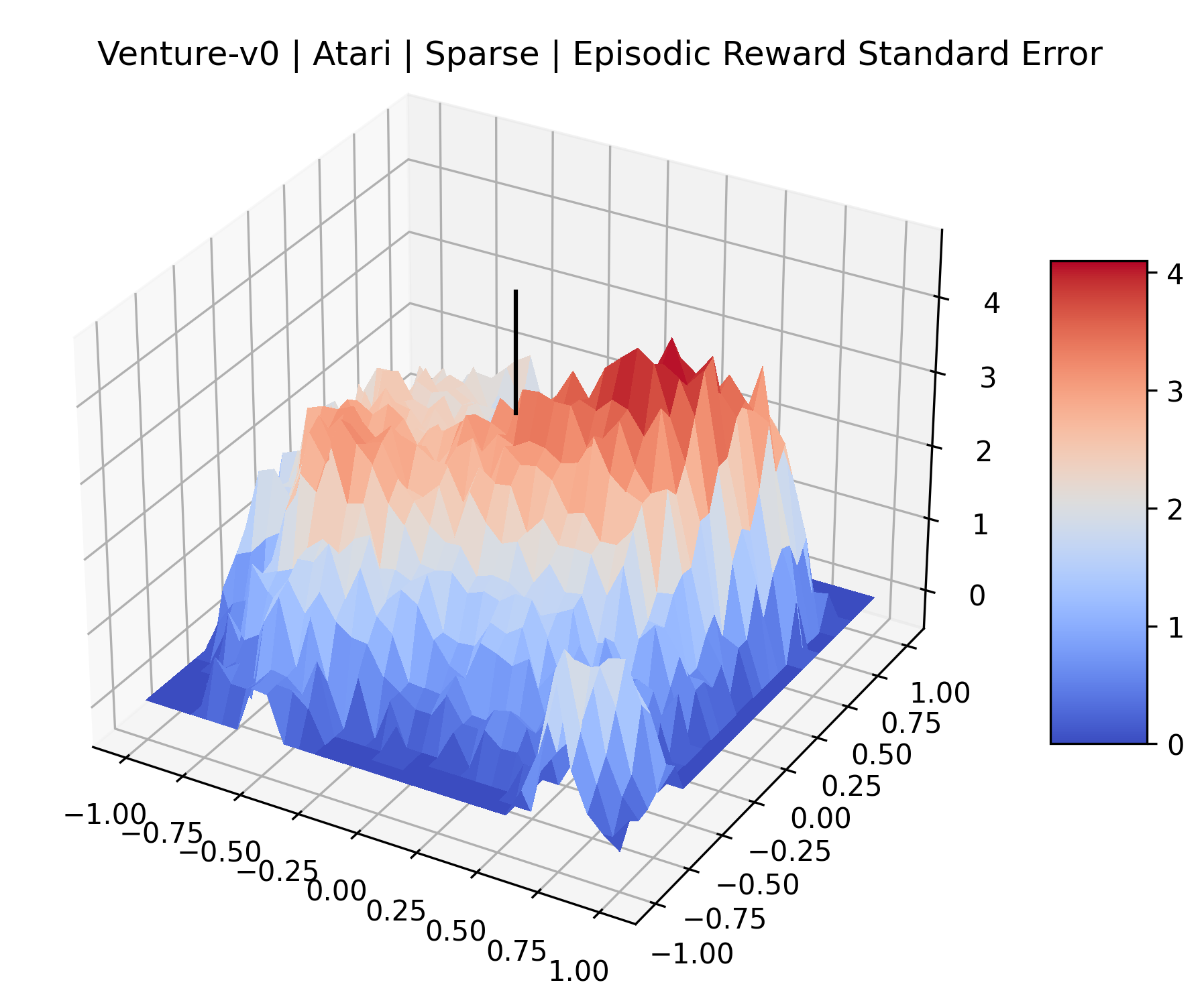} \\
\end{tabular}
\caption{Standard error surfaces for 12 Atari environments.}
\label{fig:atari_standarderror_table}
\end{figure*}
\pagebreak

\section{All Gradient Heat Maps}
\label{appendix:heat_maps}

\newcommand\heatscale{0.31\linewidth}
\subsection{Classic Control}
\begin{figure*}[!ht]
\centering
\begin{tabular}{ccc}
 \includegraphics[width=\heatscale]{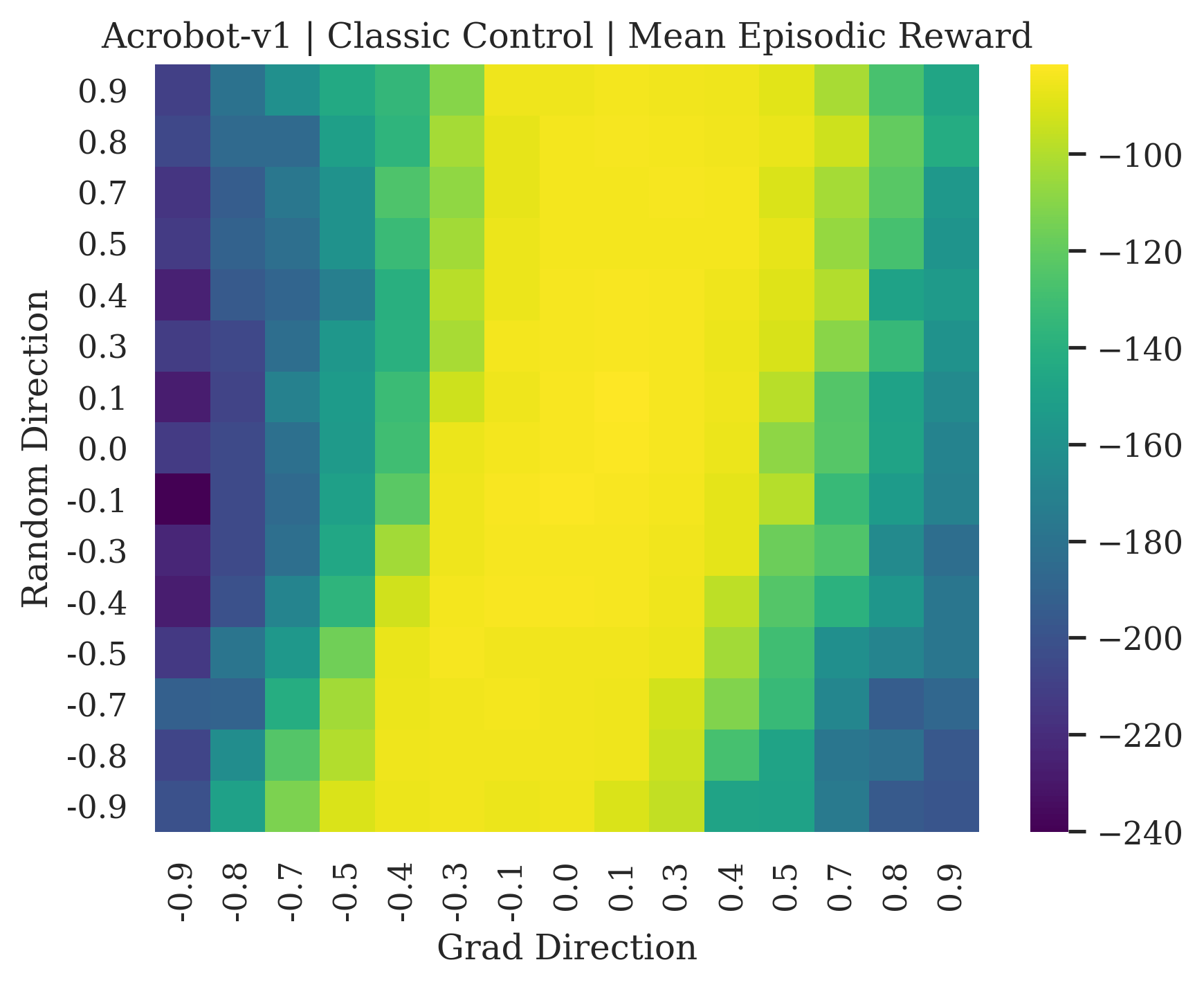} &
 \includegraphics[width=\heatscale]{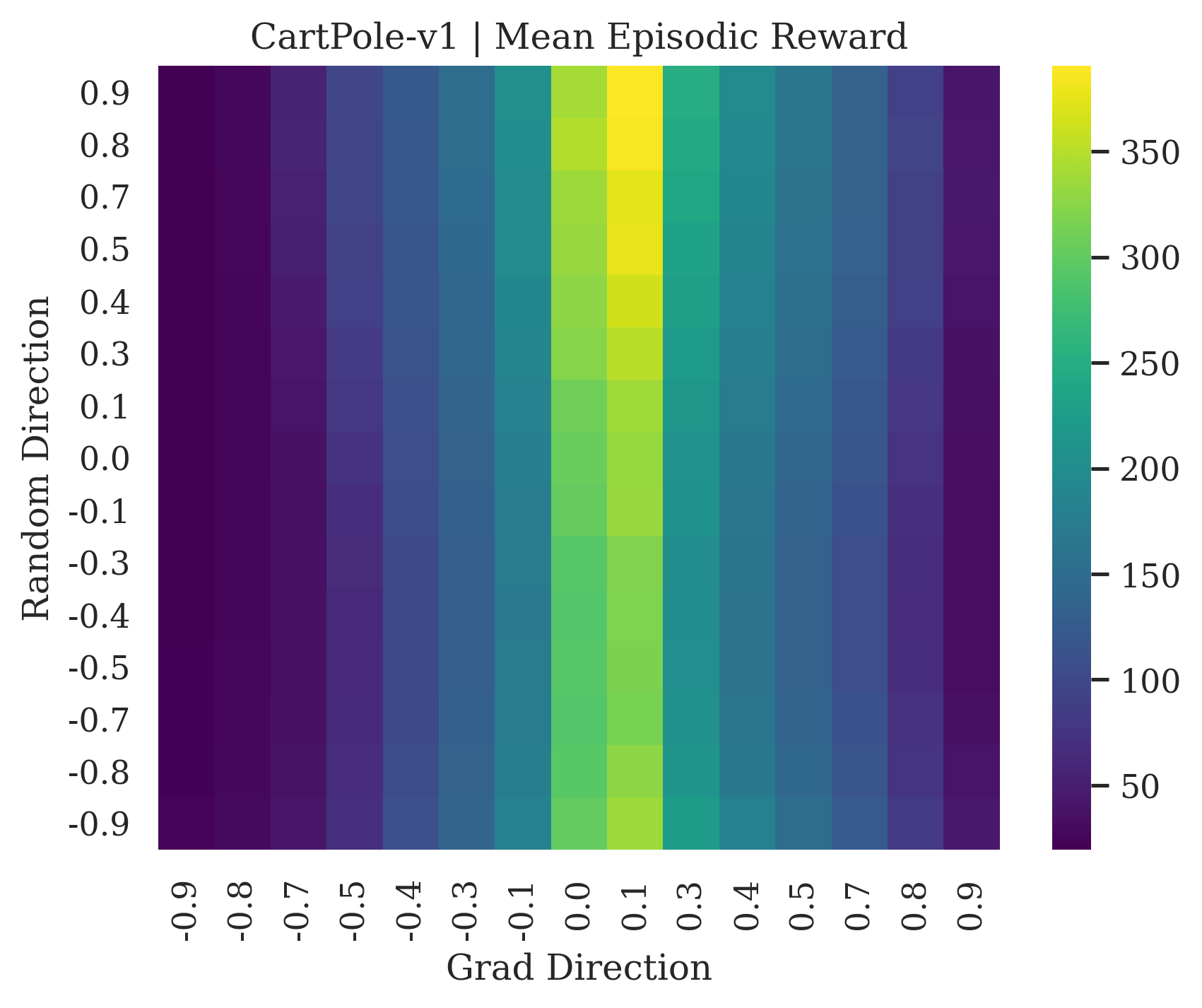} &
 \includegraphics[width=\heatscale]{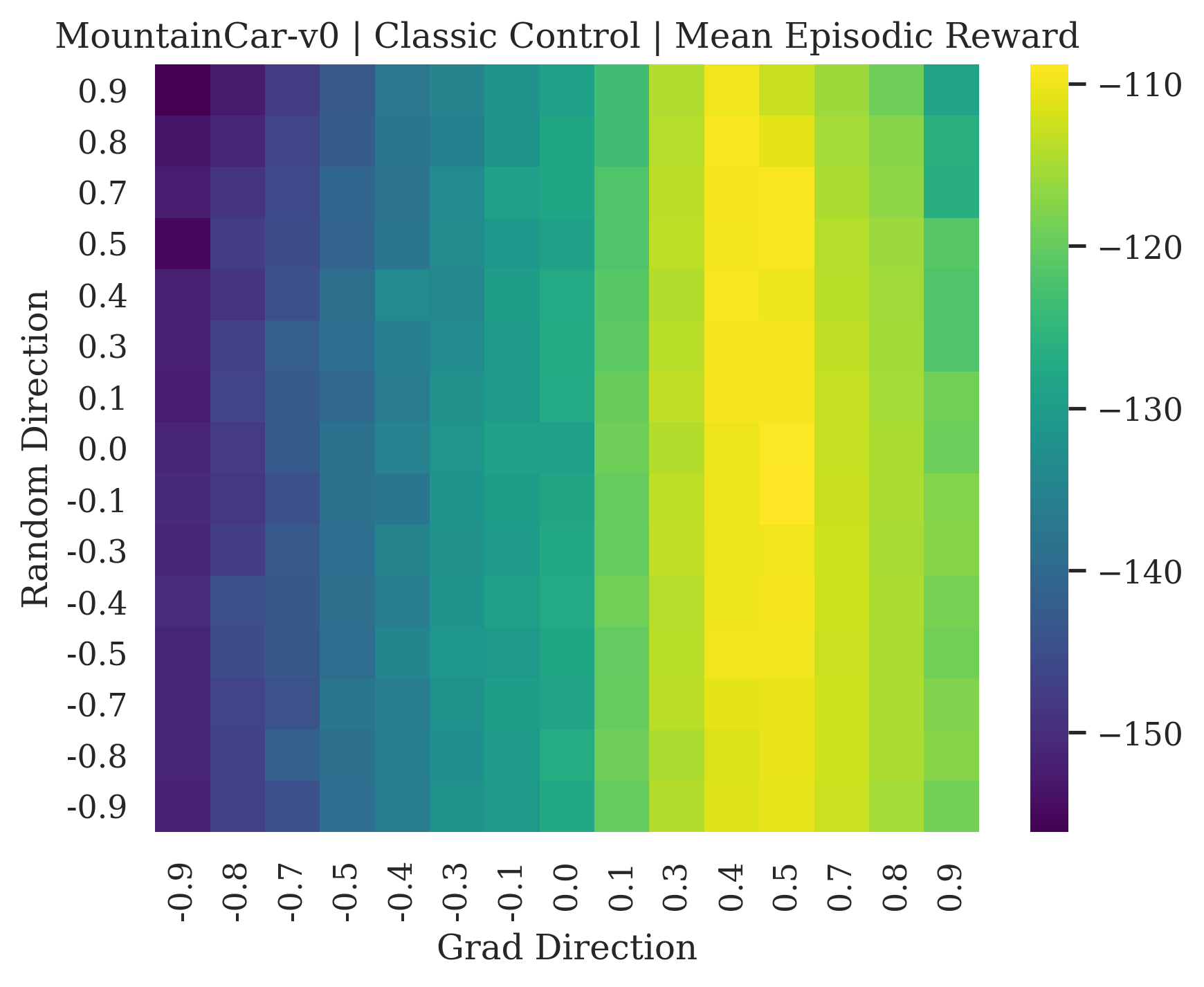} \\
\end{tabular}
\begin{tabular}{cc}
 \includegraphics[width=\heatscale]{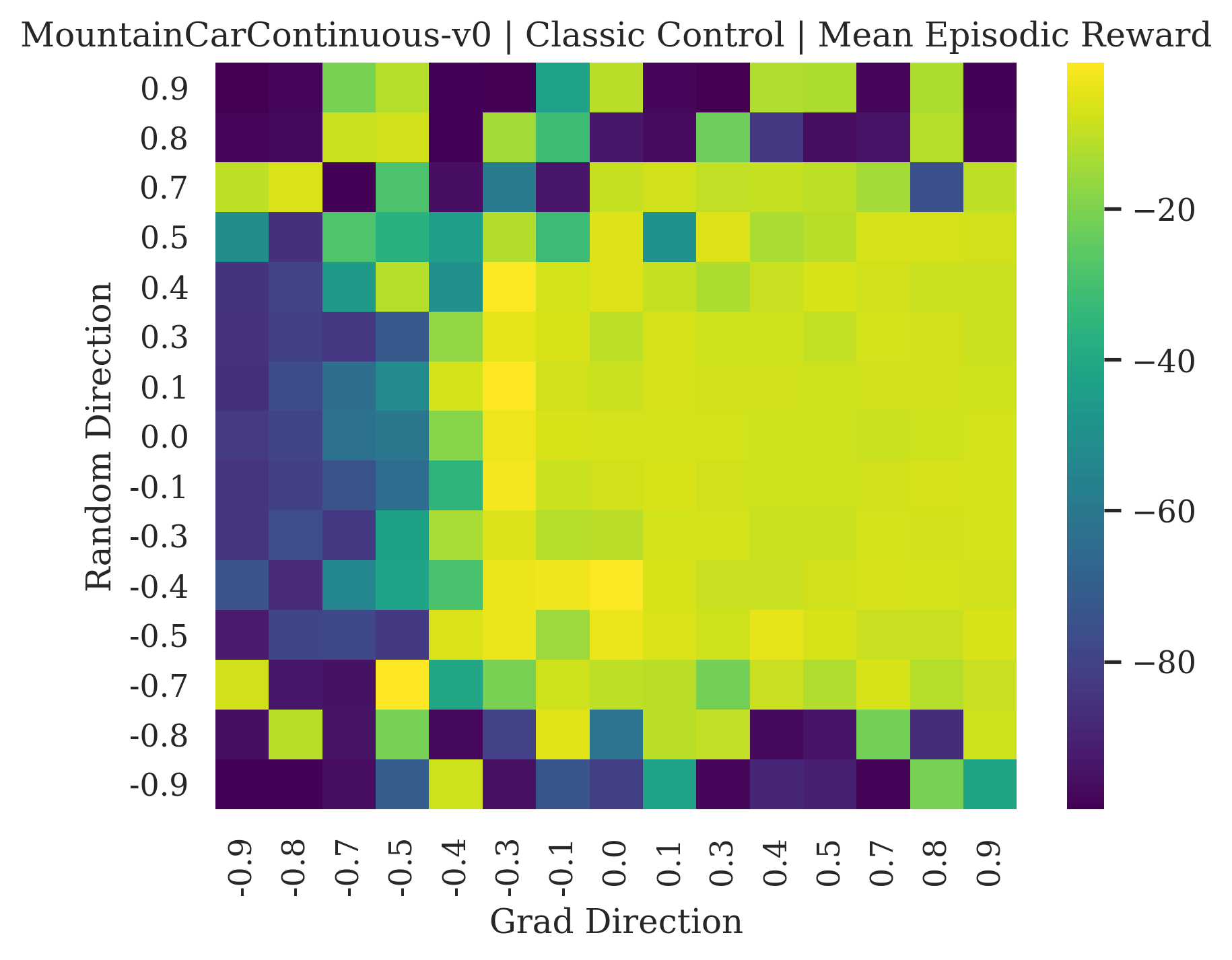} &
 \includegraphics[width=\heatscale]{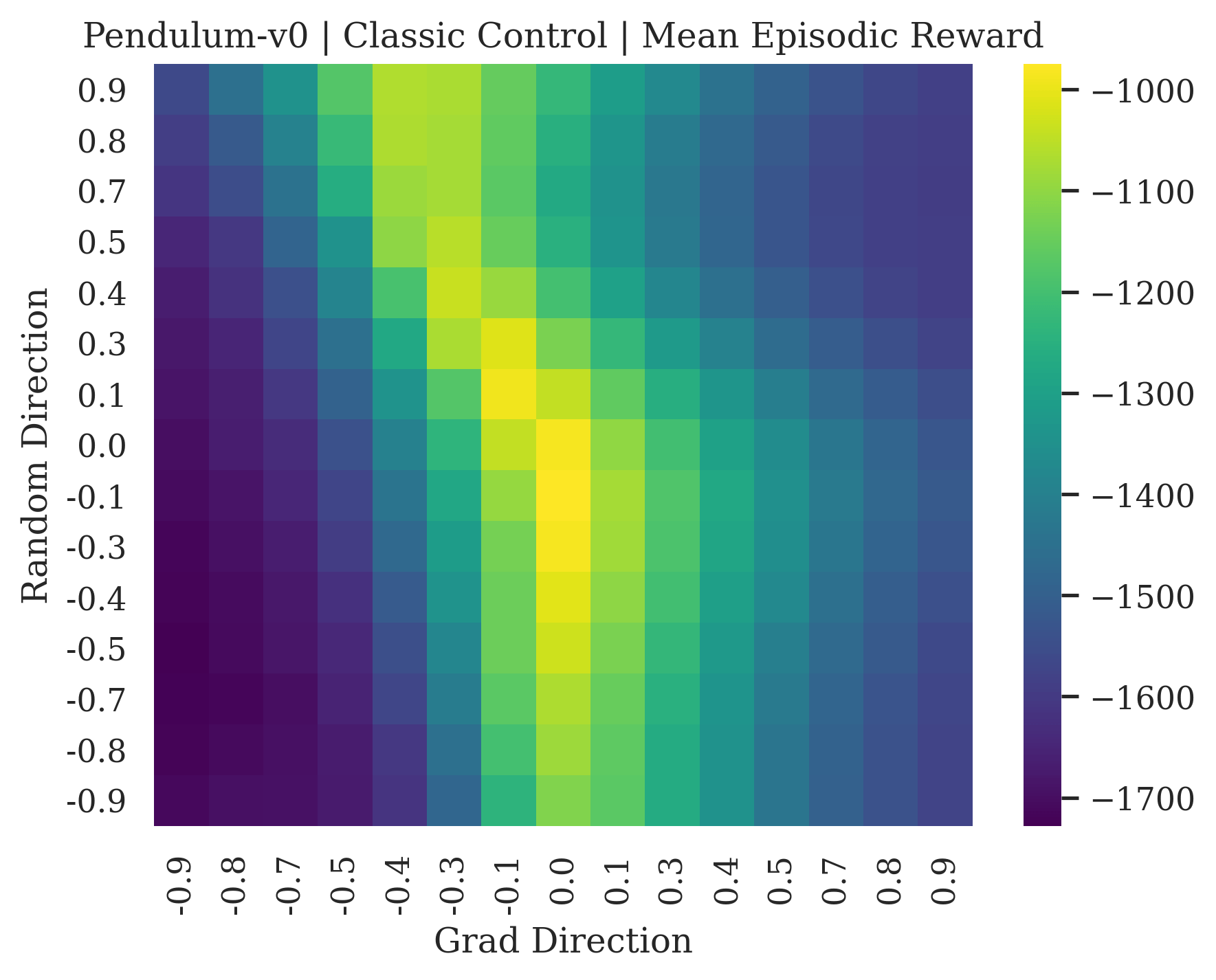} \\
\end{tabular}
\caption{Policy gradient heat maps for 5 Classic Control environments.}
\label{fig:classiccontrol_heatmap_table}
\end{figure*}
\pagebreak

\subsection{MuJoCo}
\begin{figure*}[!ht]
\centering
\begin{tabular}{ccc}
 \includegraphics[width=\heatscale]{./heat/ant_1900000_heatmap_episoderewards_2dheat.png} &
 \includegraphics[width=\heatscale]{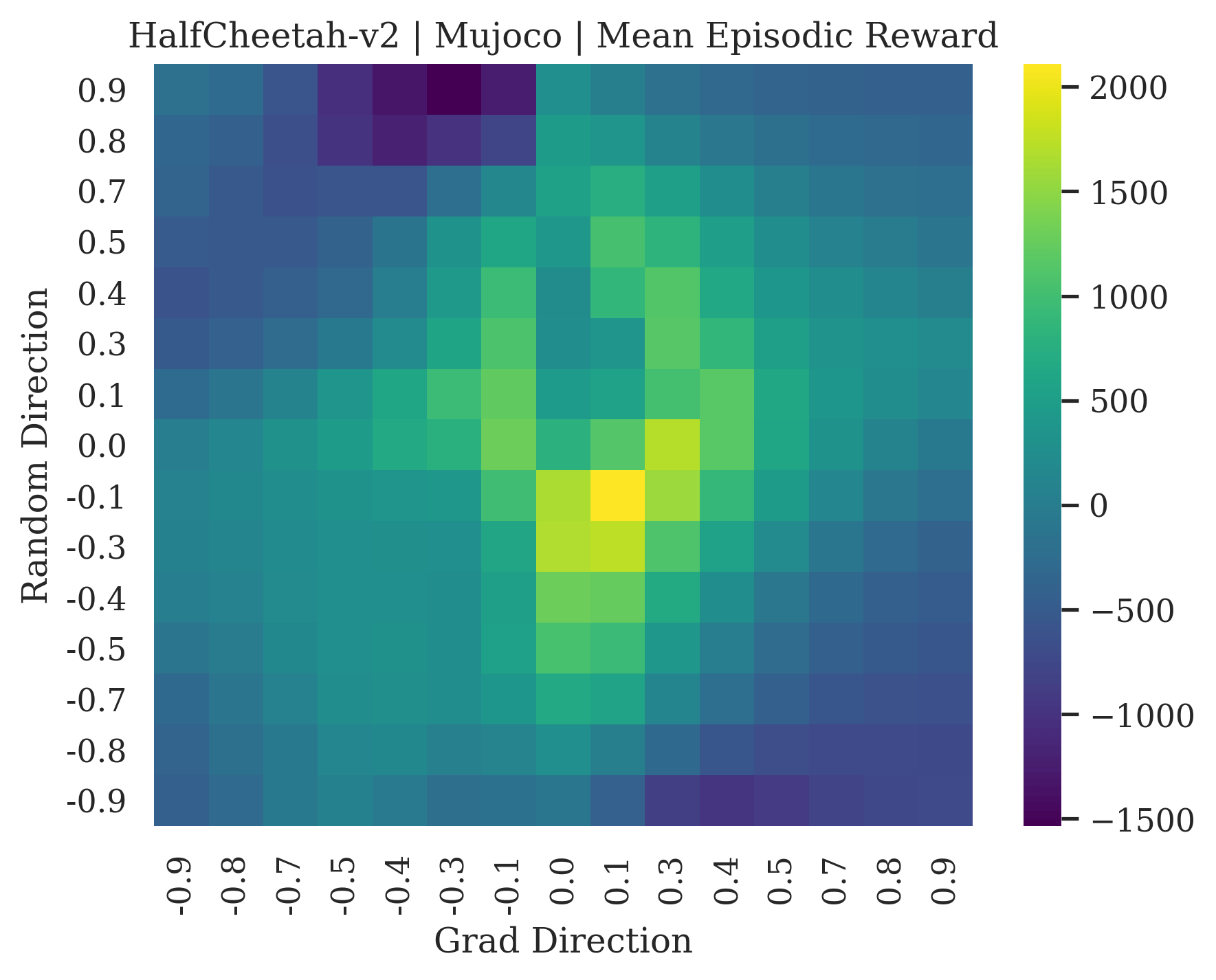} &
 \includegraphics[width=\heatscale]{./heat/hopper_0200000_heatmap_episoderewards_2dheat.png} \\
 \includegraphics[width=\heatscale]{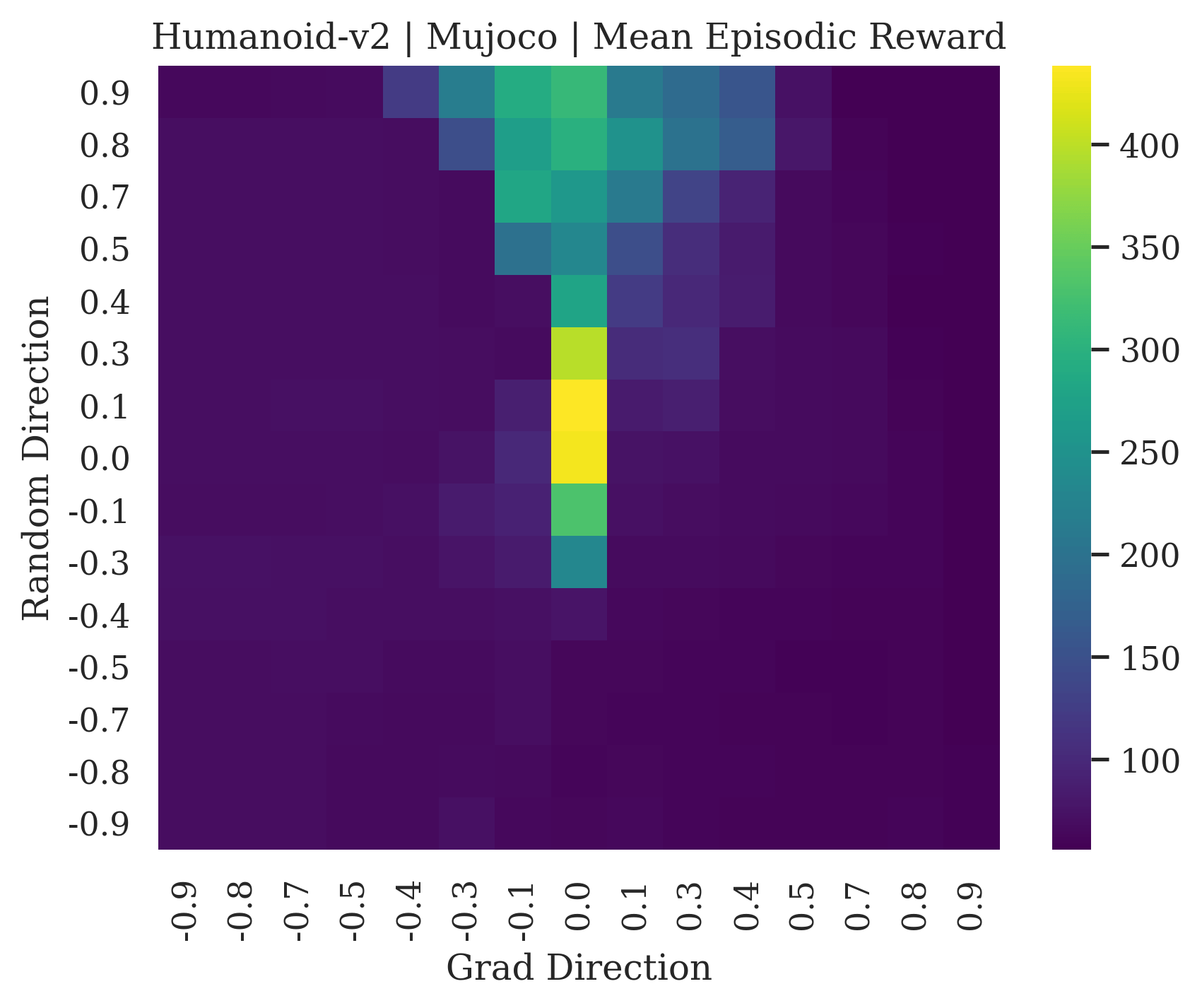} &
 \includegraphics[width=\heatscale]{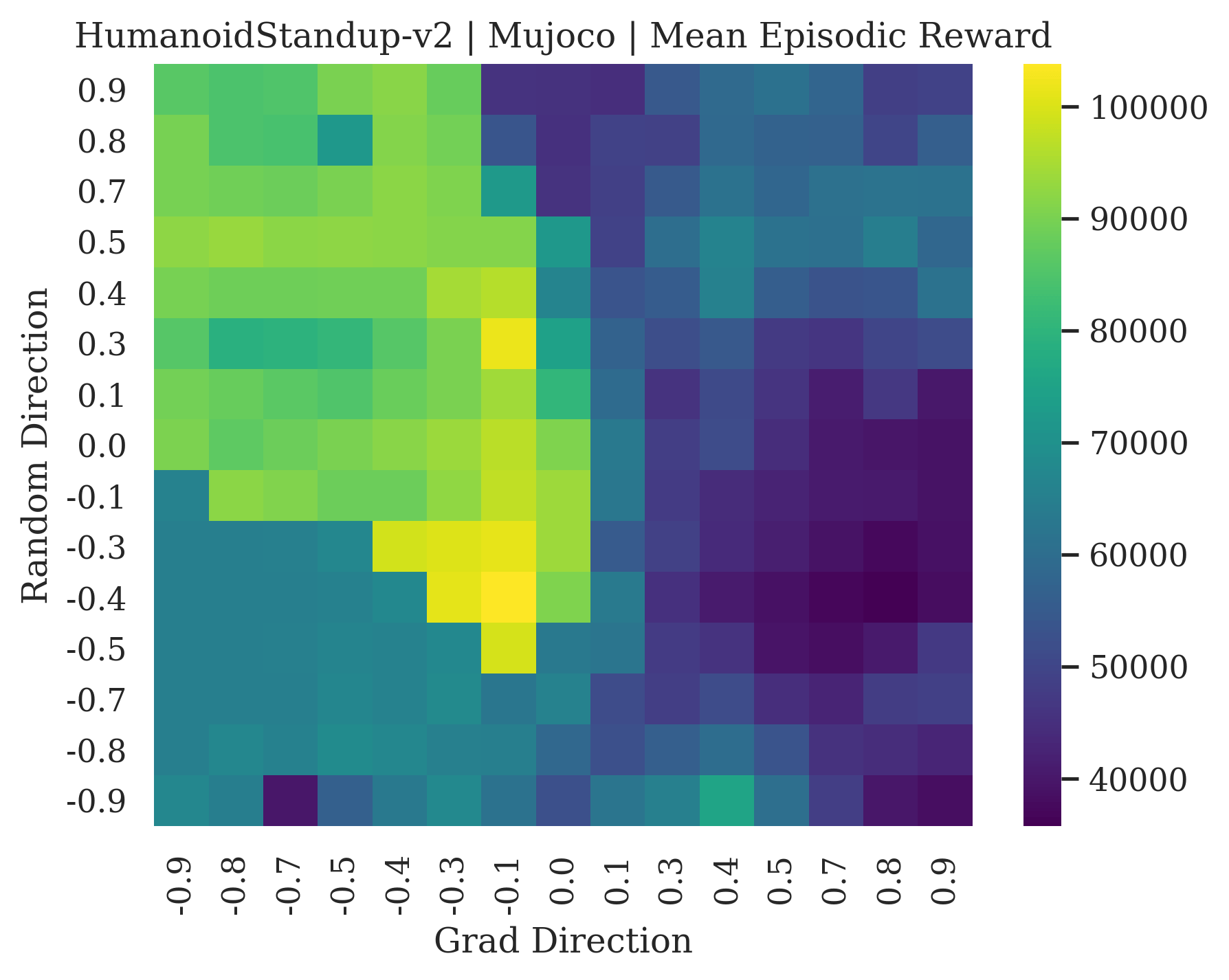} &
 \includegraphics[width=\heatscale]{./heat/inverteddoublependulum_0200000_heatmap_episoderewards_2dheat.png} \\
 \includegraphics[width=\heatscale]{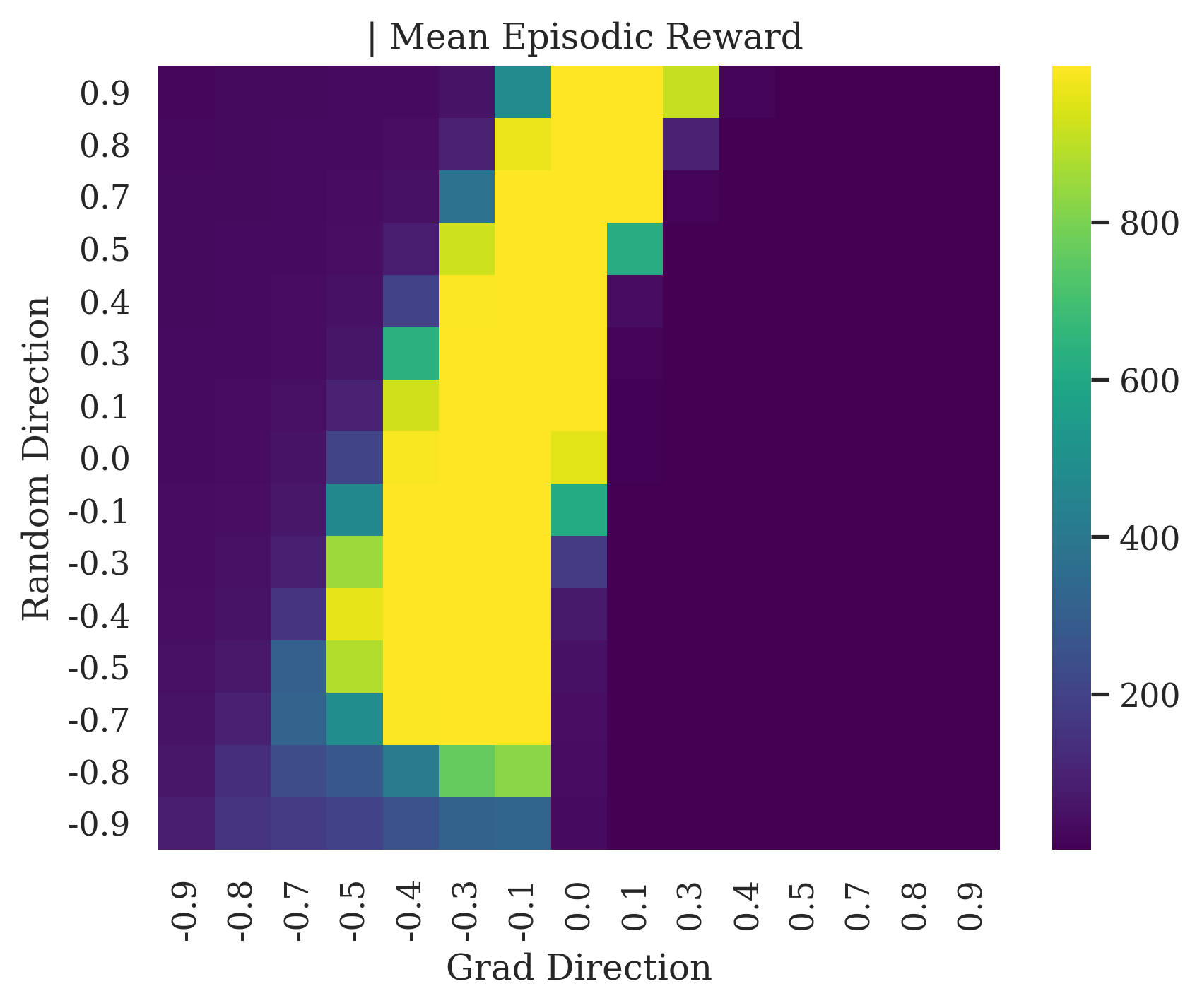} &
 \includegraphics[width=\heatscale]{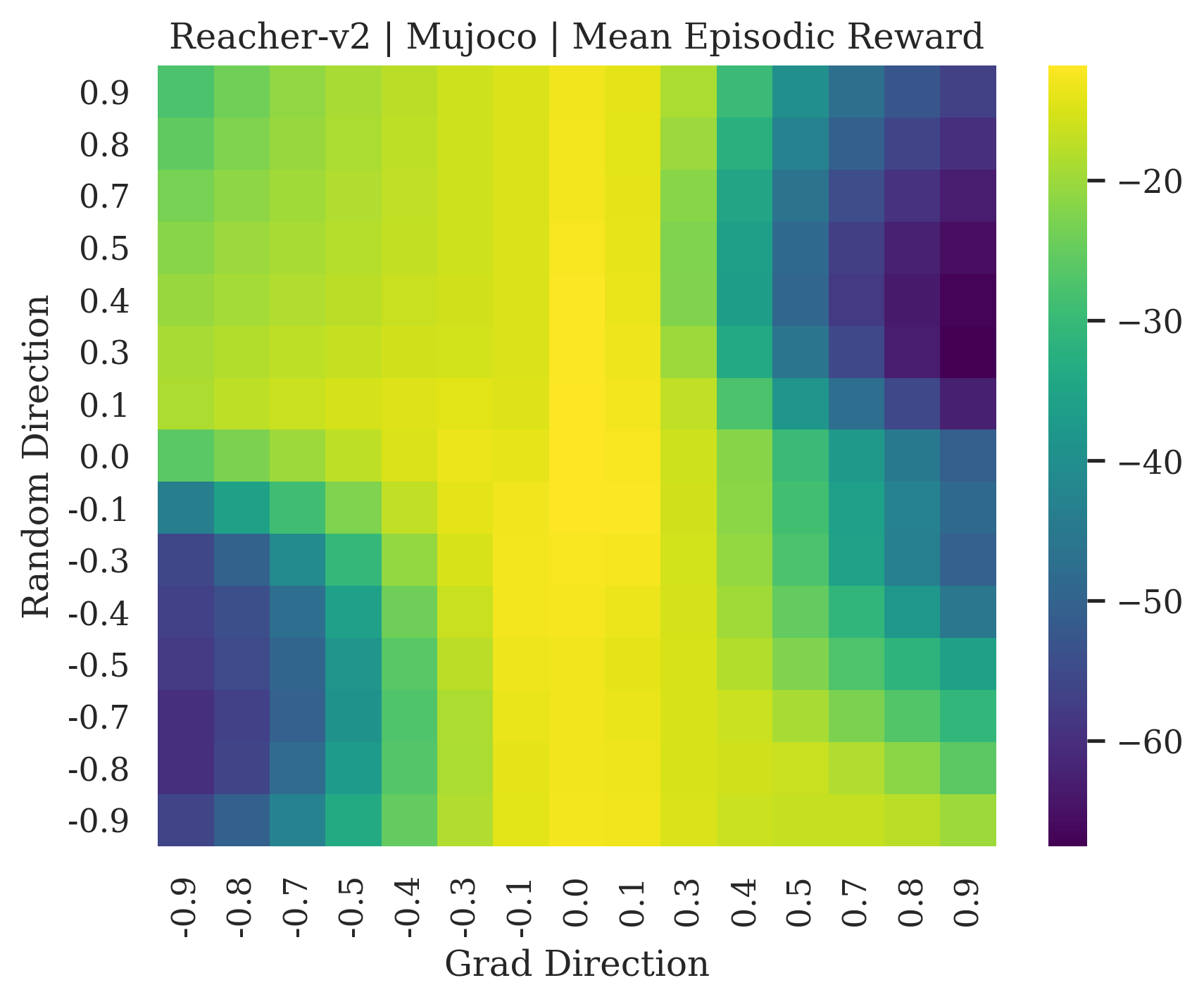} &
 \includegraphics[width=\heatscale]{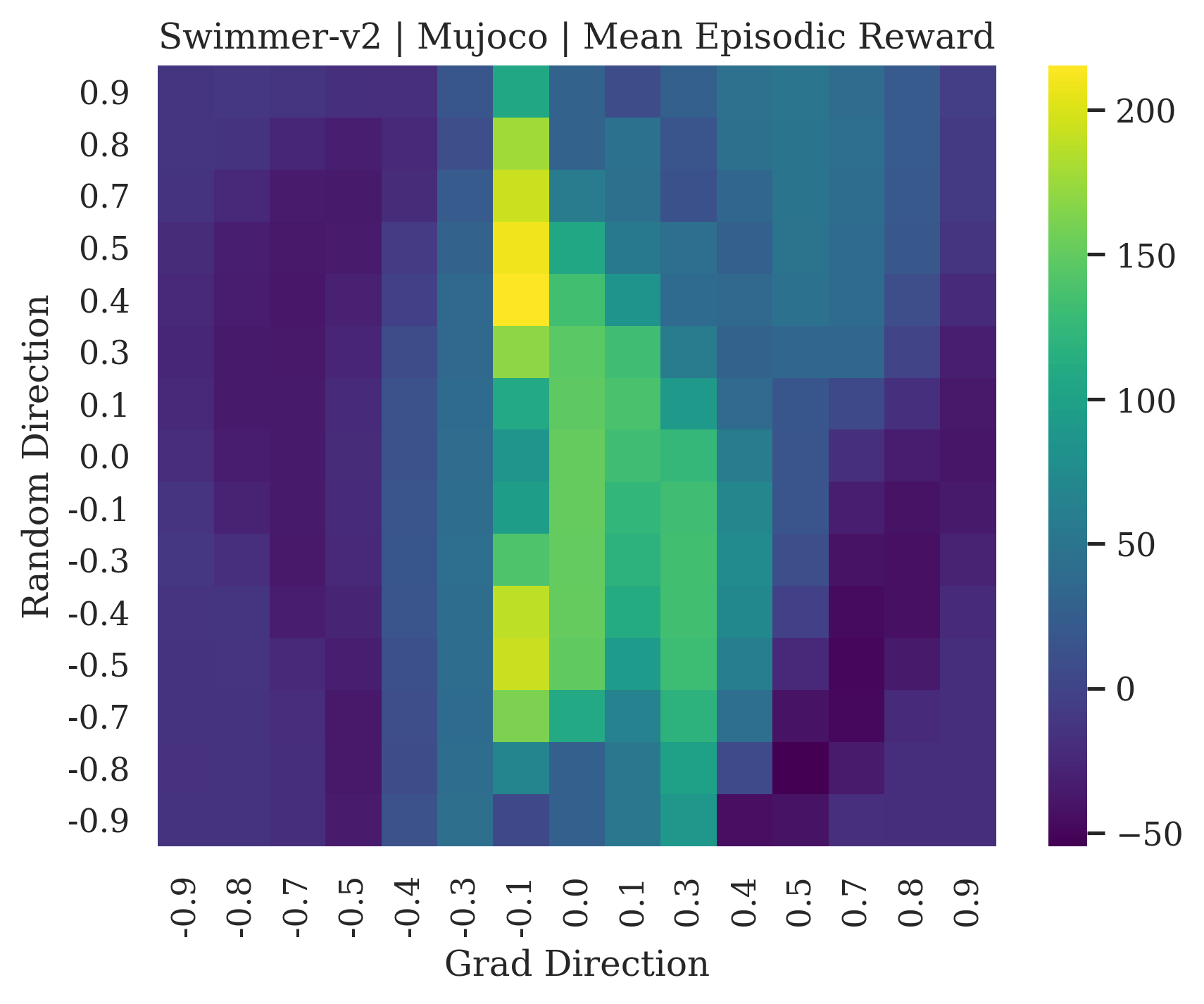} \\ 
 & \includegraphics[width=\heatscale]{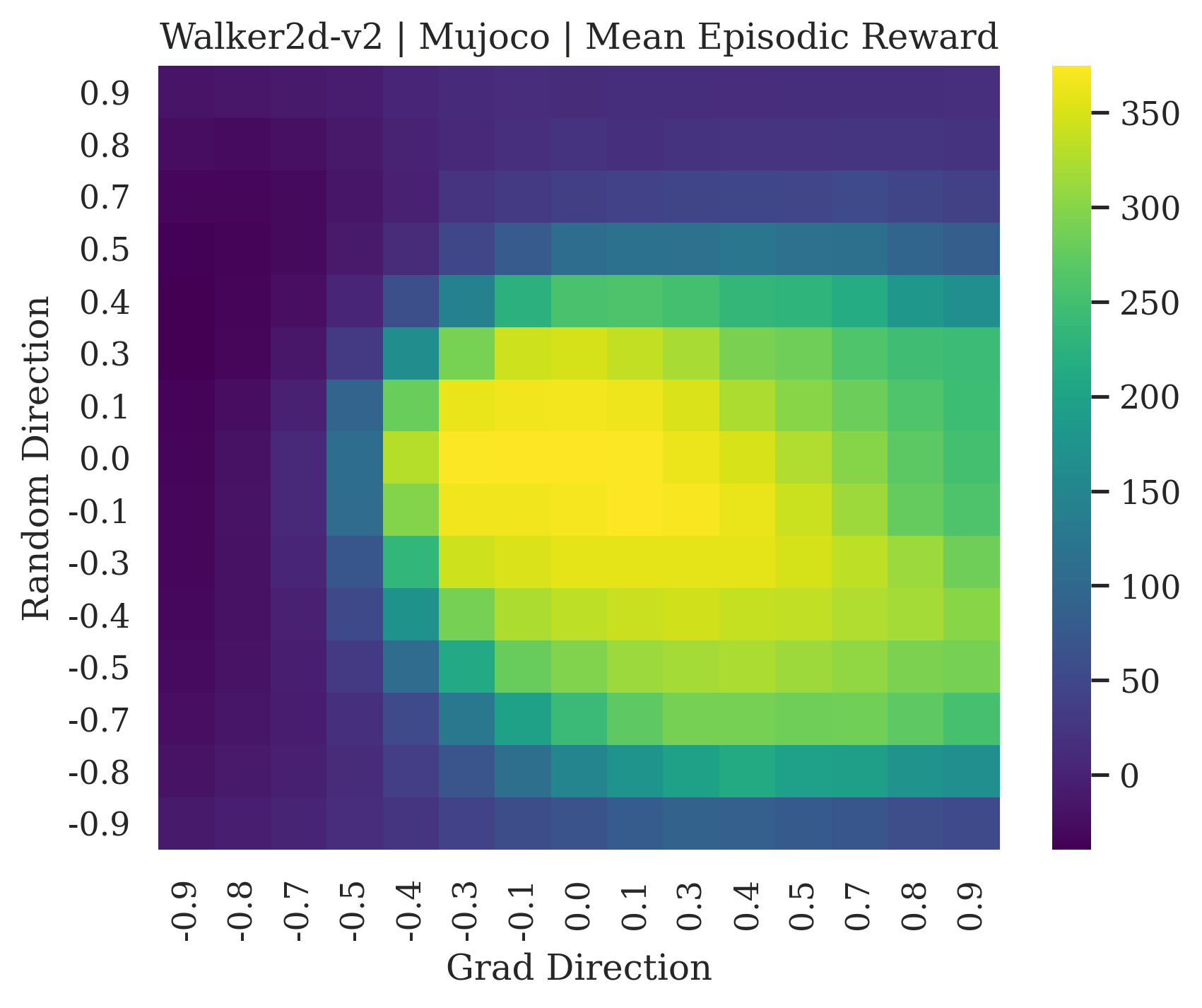} & \\
\end{tabular}
\caption{Policy gradient heat maps for 10 MuJoCo environments.}
\label{fig:MuJoCo_heatmap_table}
\end{figure*}
\pagebreak

\section{All Gradient Line Plots}
\label{appendix:line_plots}

\newcommand\linescale{0.31\linewidth}
\subsection{Classic Control}
\begin{figure*}[!ht]
\centering
\begin{tabular}{ccc}
 \includegraphics[width=\linescale]{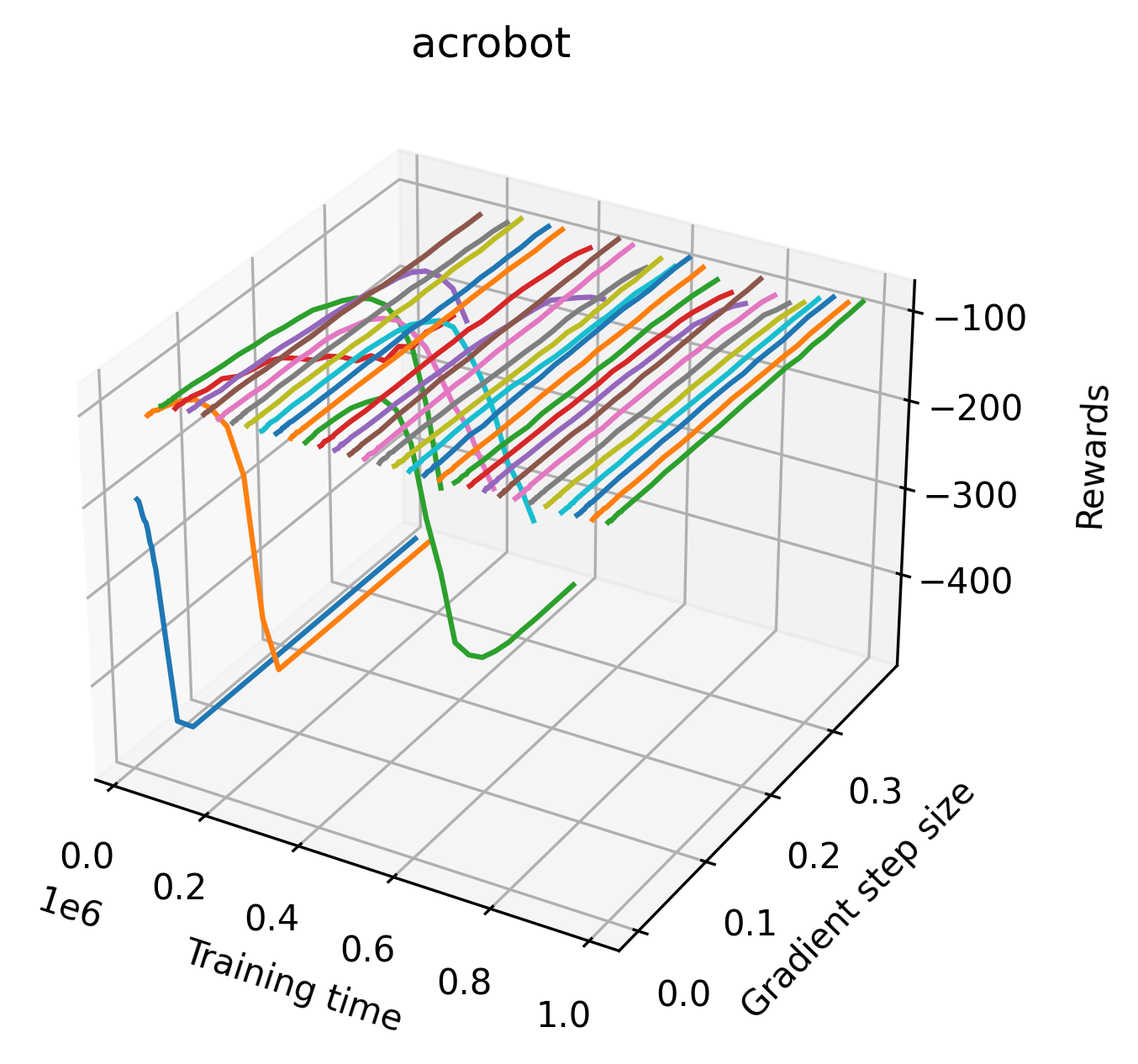} &
 \includegraphics[width=\linescale]{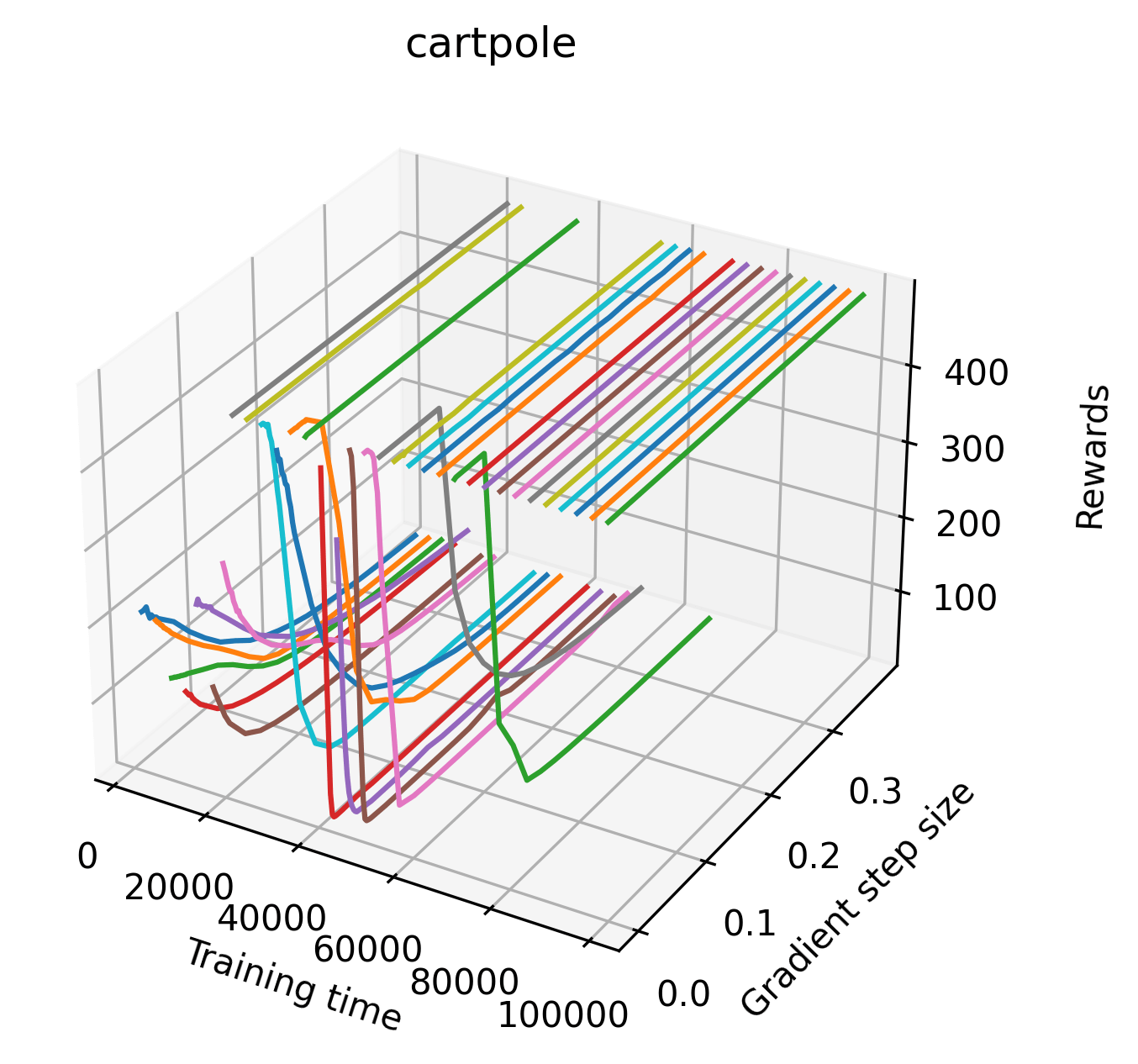} &
 \includegraphics[width=\linescale]{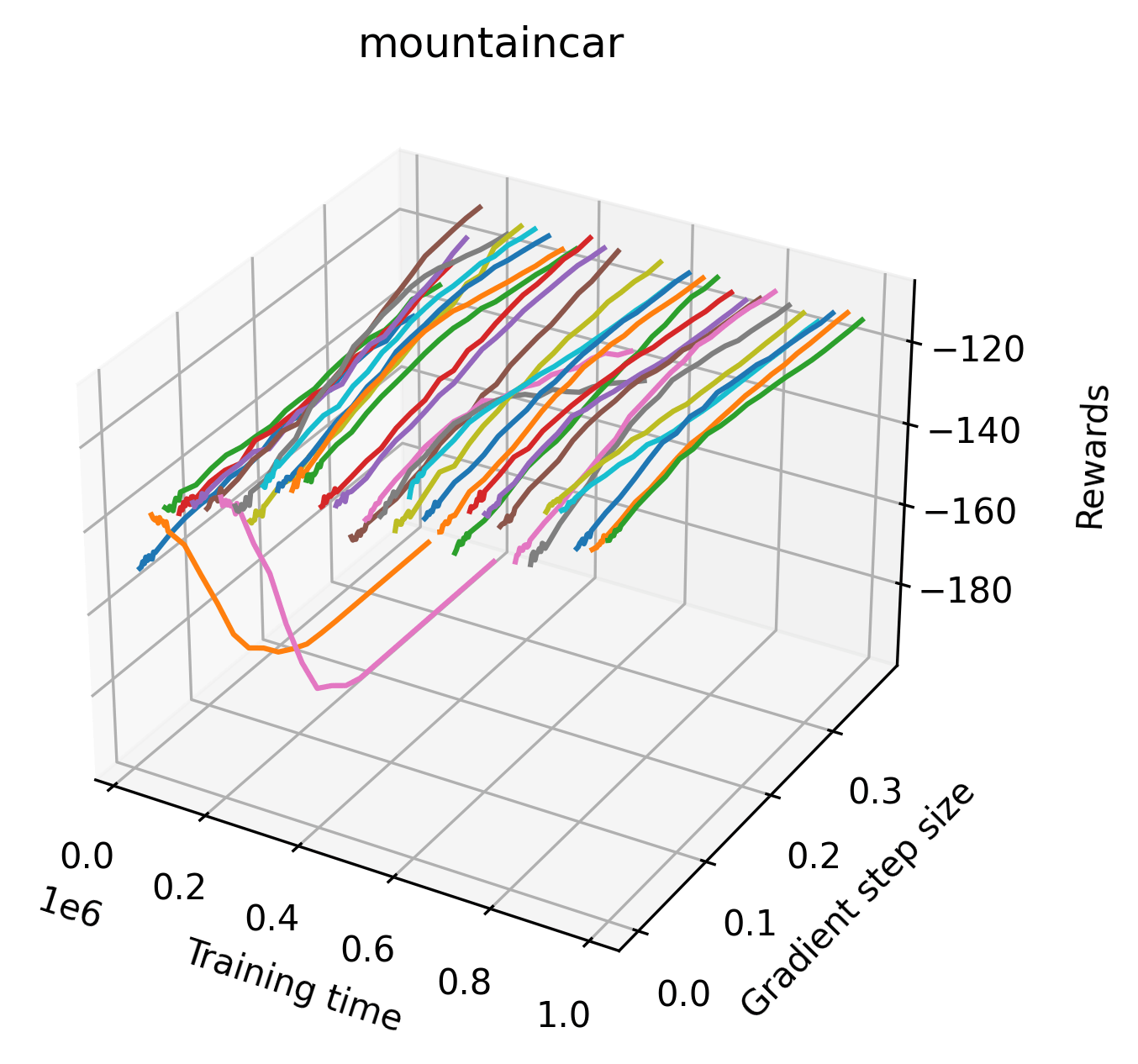} \\
\end{tabular}
\begin{tabular}{cc}
 \includegraphics[width=\linescale]{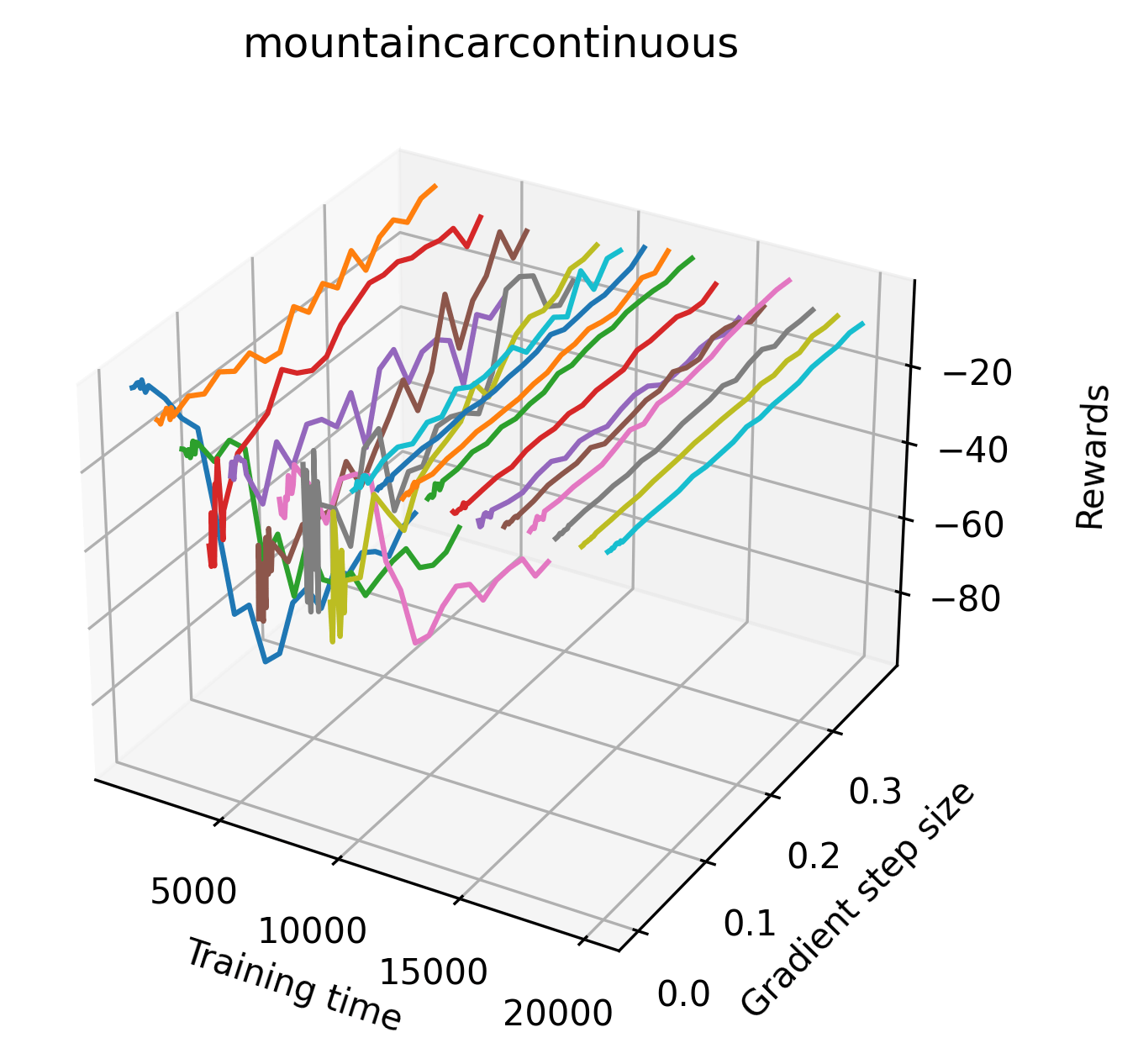} &
 \includegraphics[width=\linescale]{./line/pendulum_lines.png} \\
\end{tabular}
\caption{Policy gradient line search plots for 5 Classic Control environments.}
\label{fig:classiccontrol_lineplot_table}
\end{figure*}
\pagebreak

\subsection{MuJoCo}
\begin{figure*}[!ht]
\centering
\begin{tabular}{ccc}
 \includegraphics[width=\linescale]{./line/ant_lines.png} &
 \includegraphics[width=\linescale]{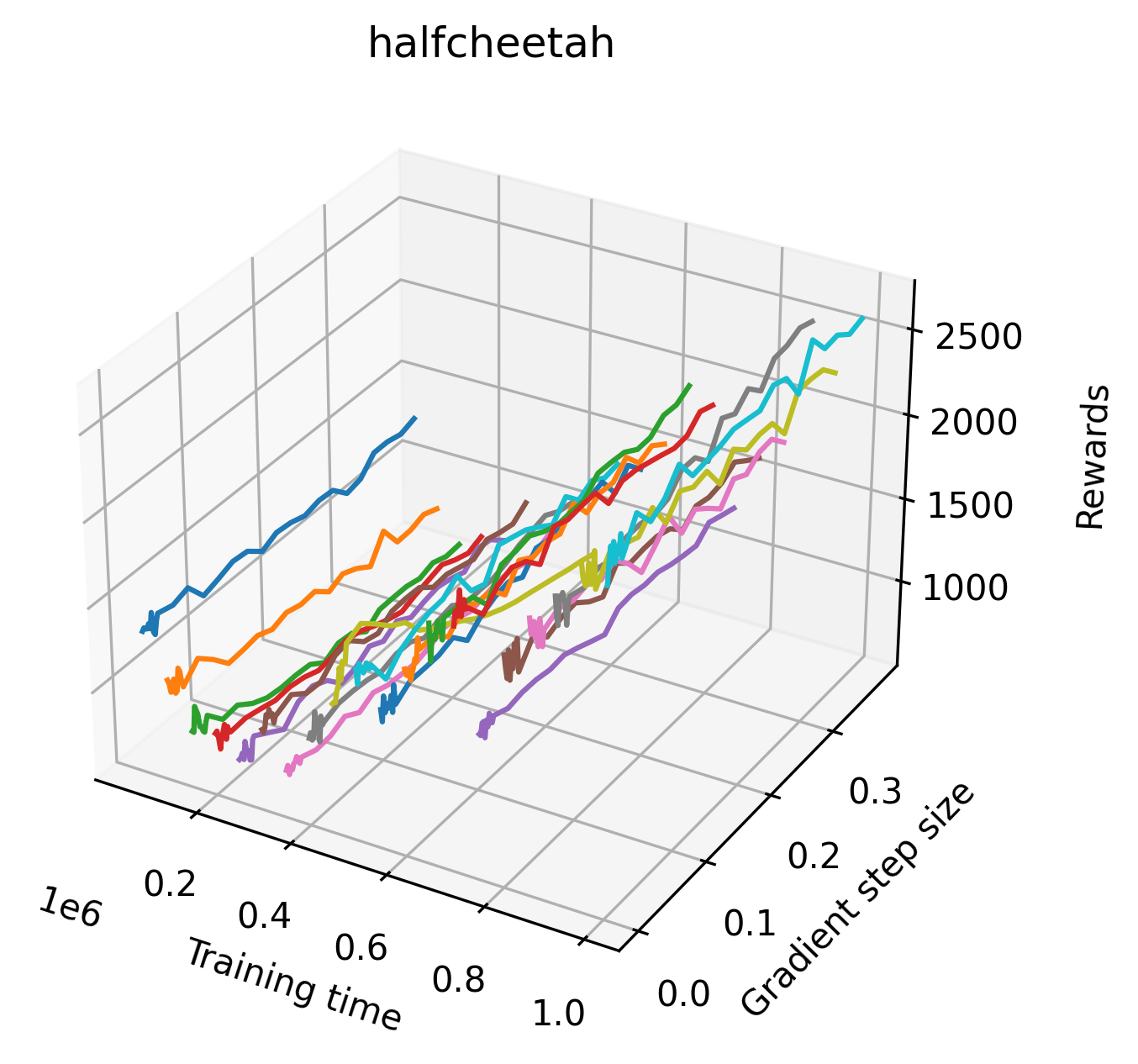} &
 \includegraphics[width=\linescale]{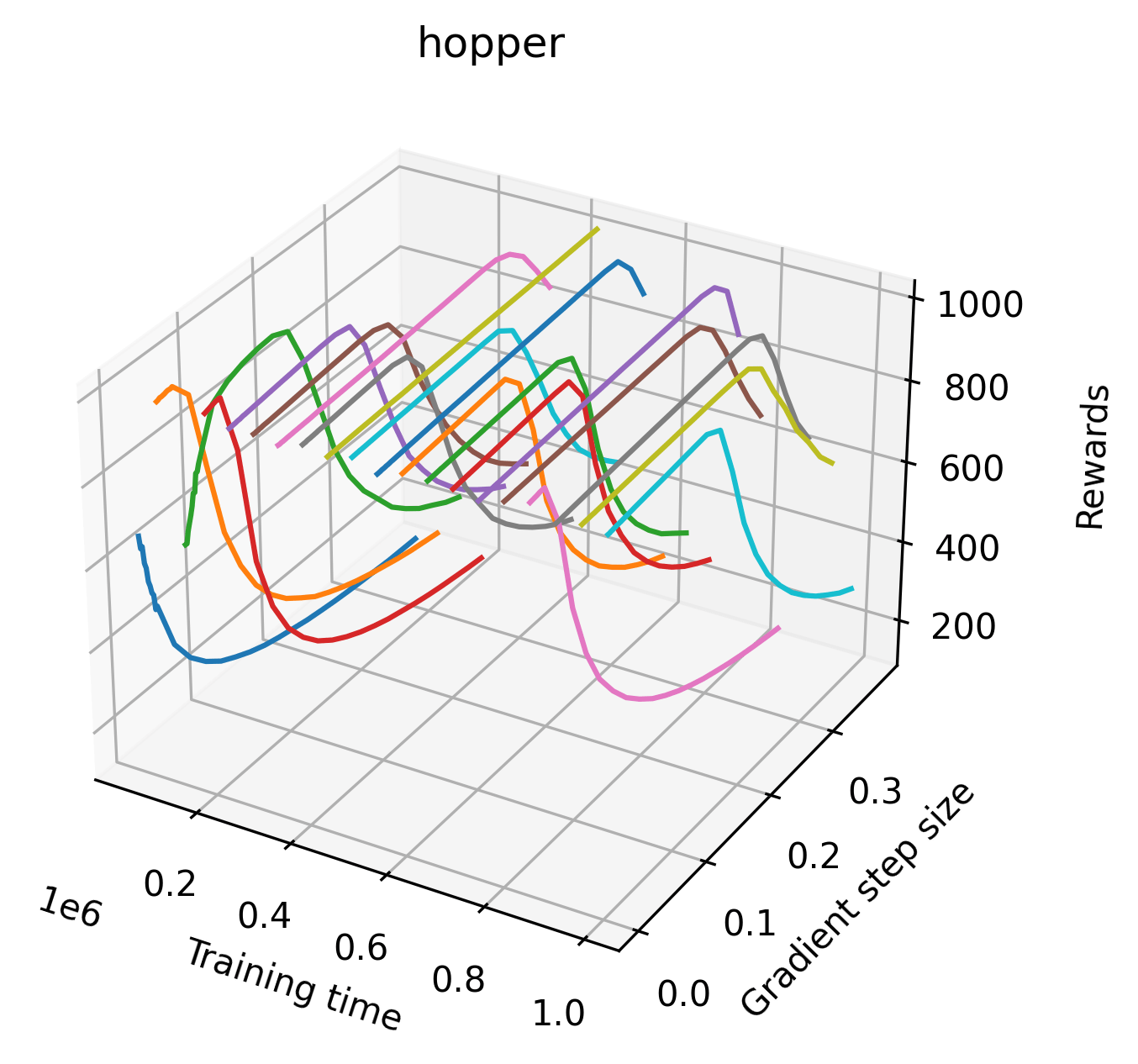} \\
 \includegraphics[width=\linescale]{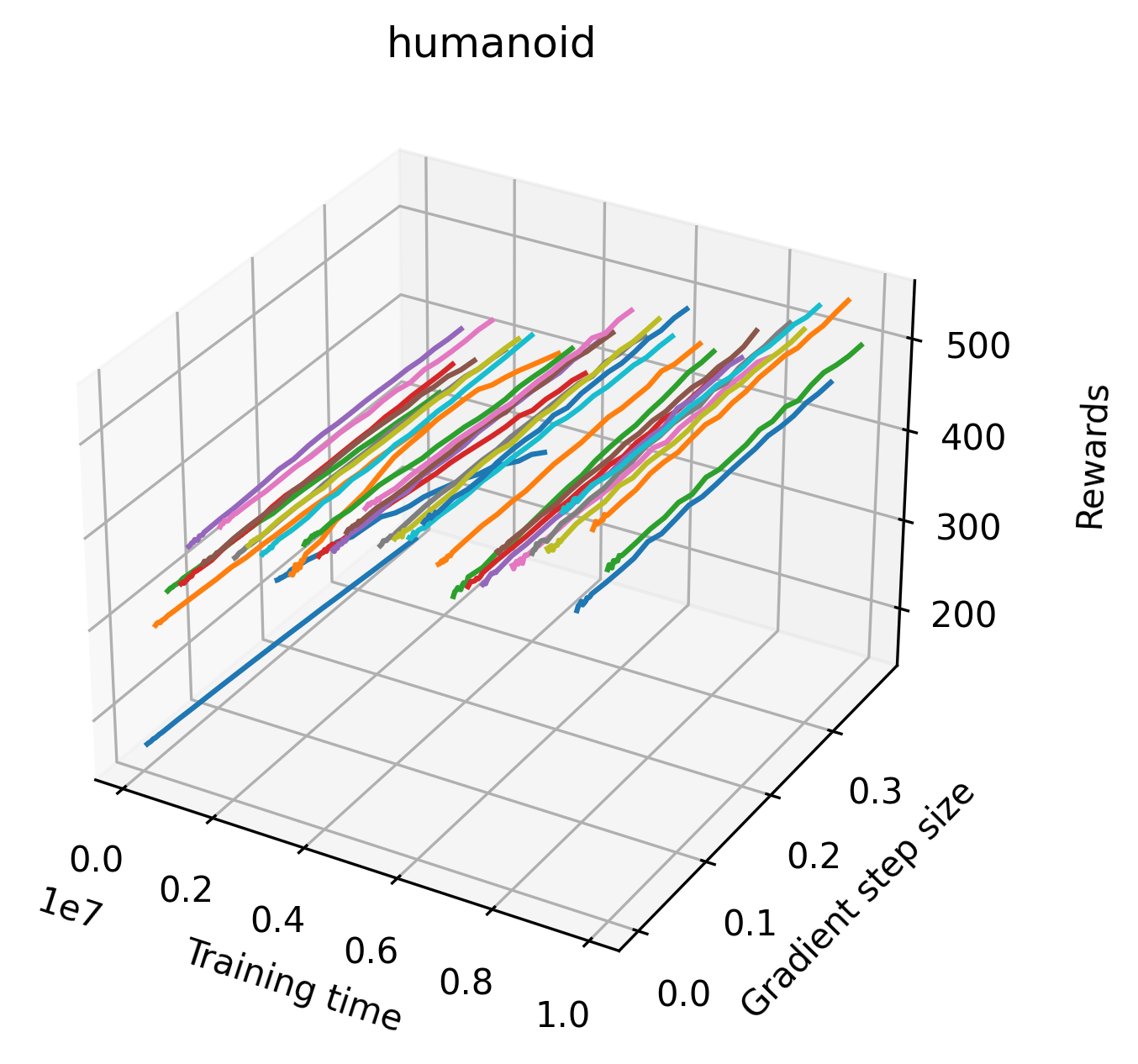} &
 \includegraphics[width=\linescale]{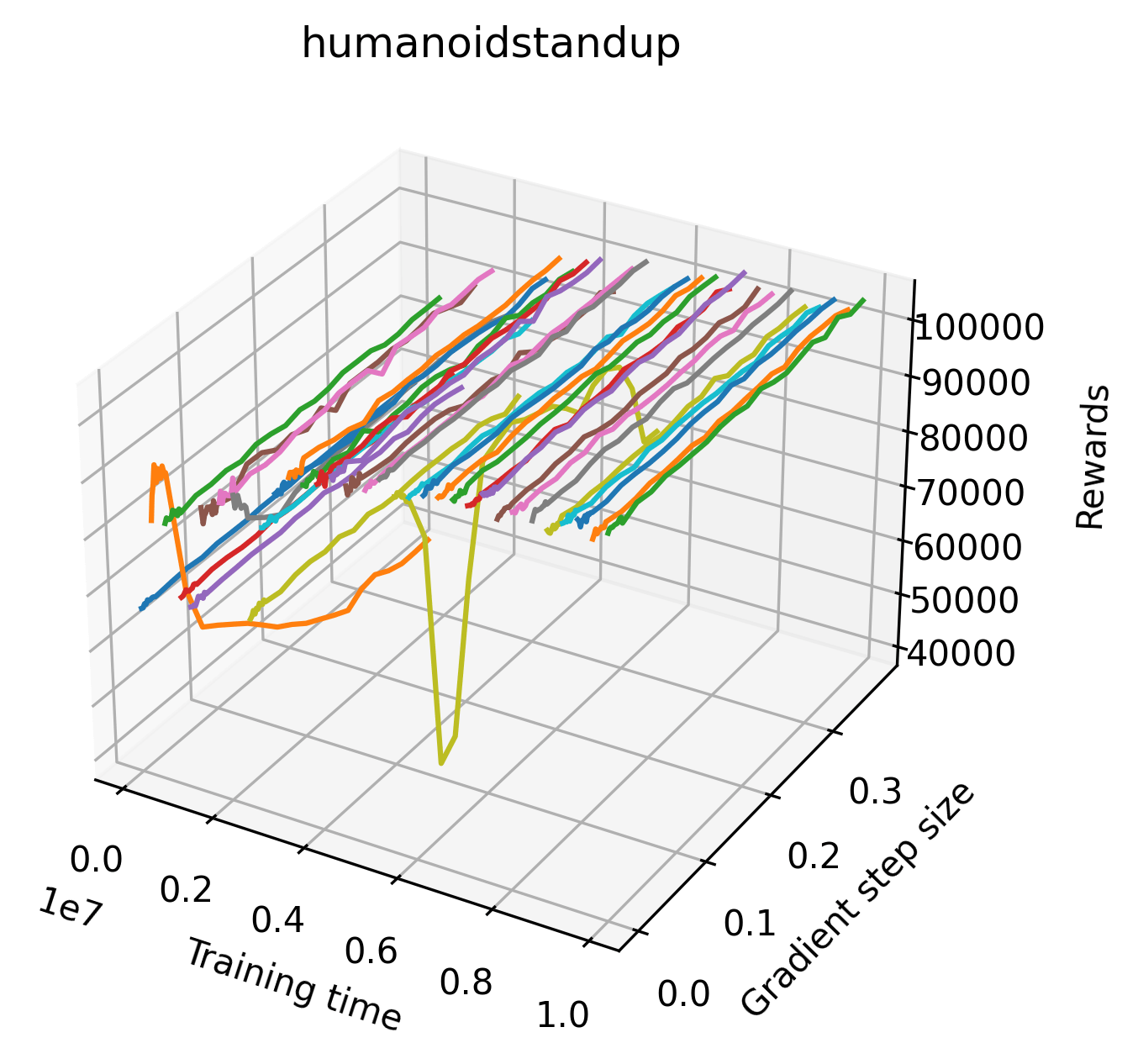} &
 \includegraphics[width=\linescale]{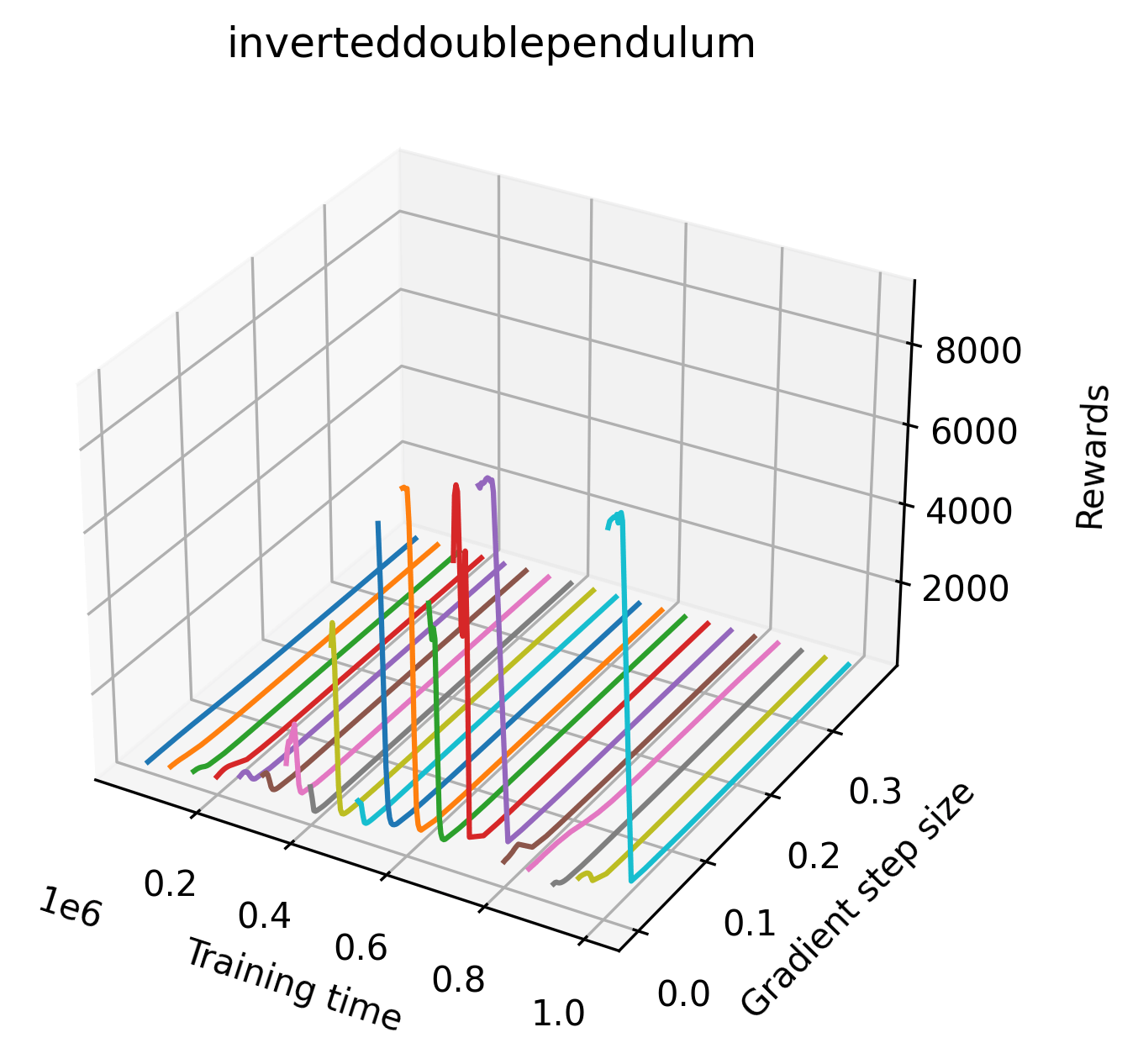} \\
 \includegraphics[width=\linescale]{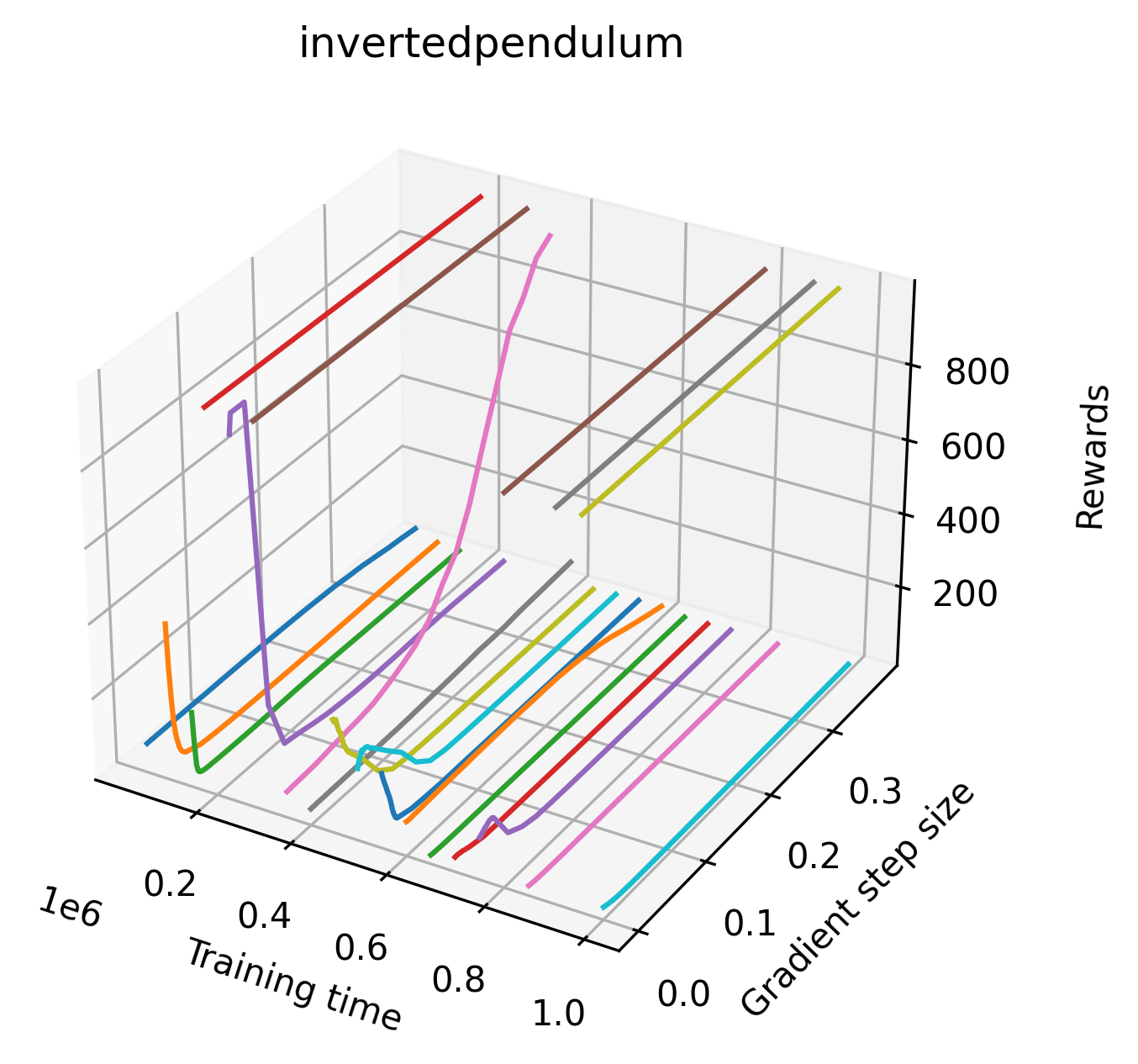} &
 \includegraphics[width=\linescale]{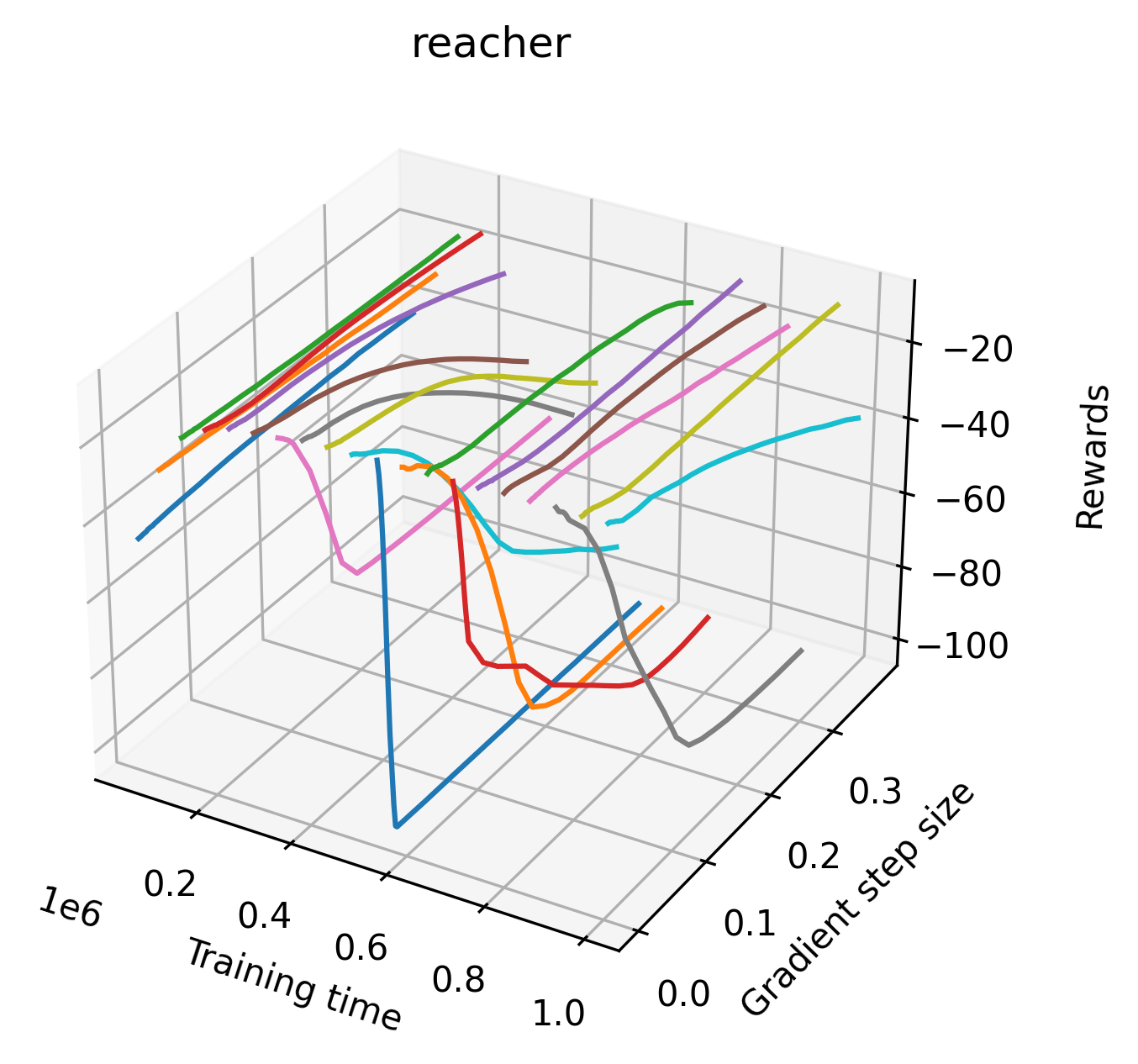} &
 \includegraphics[width=\linescale]{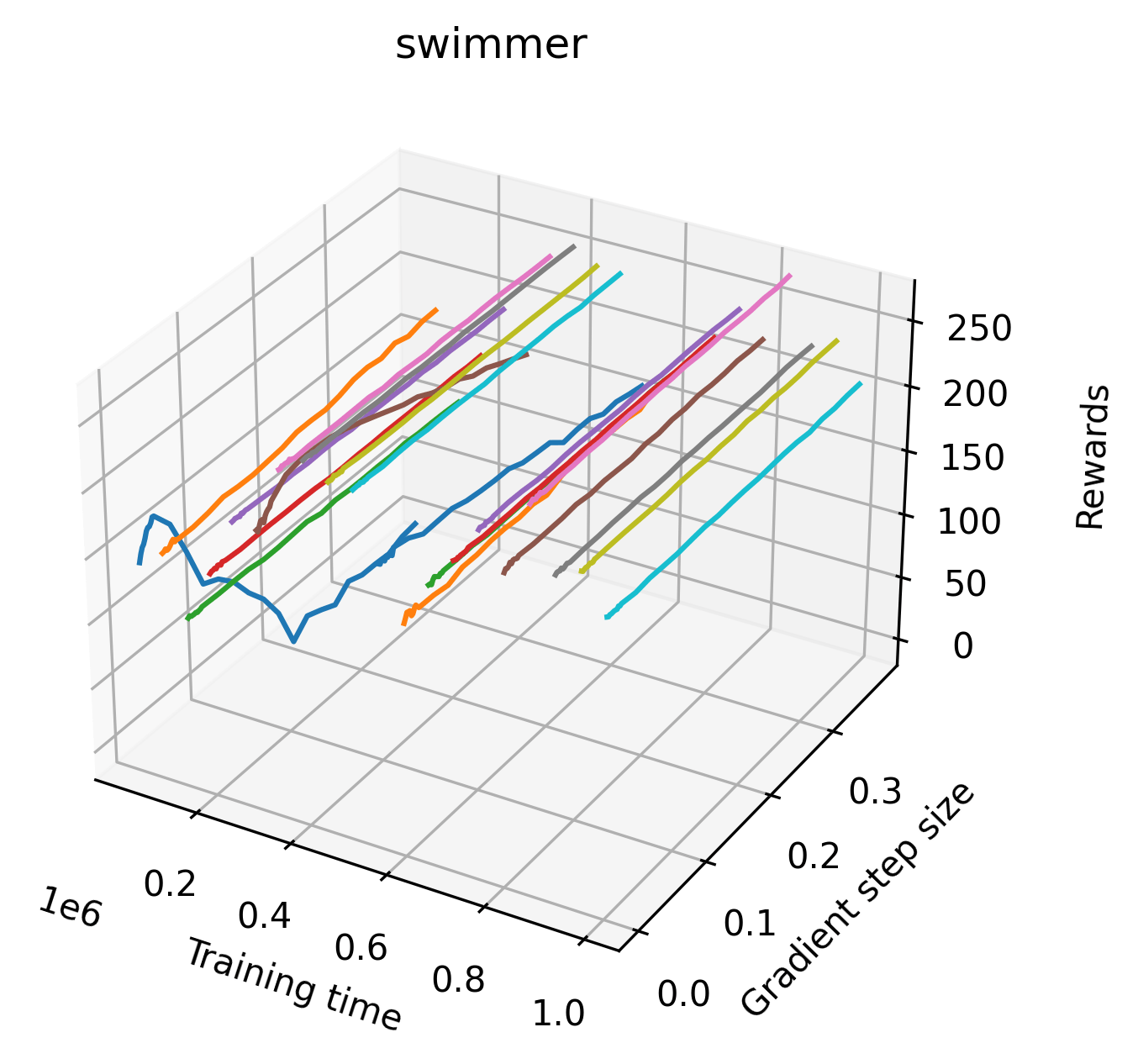} \\ 
 & \includegraphics[width=\linescale]{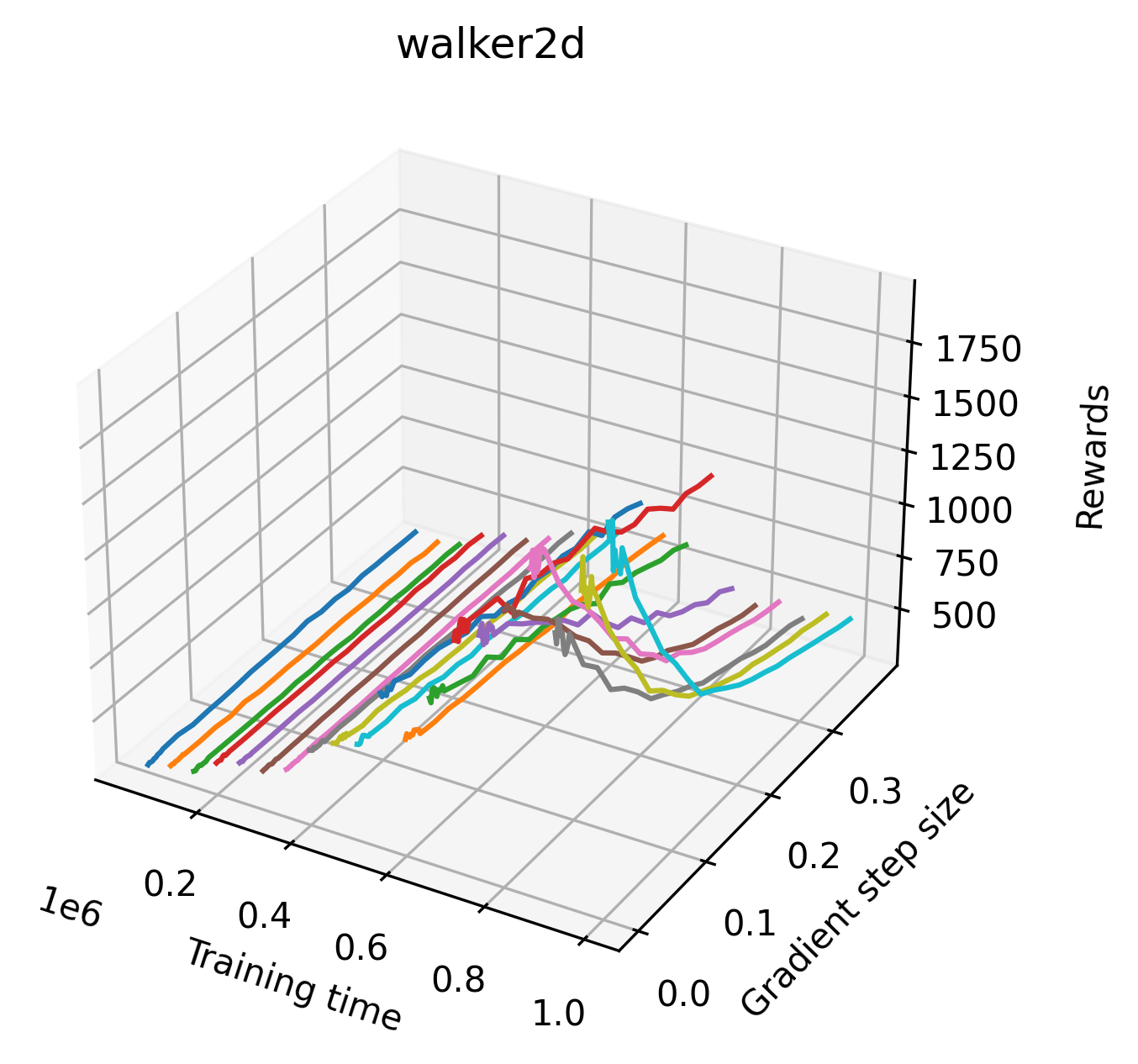} & \\
\end{tabular}
\caption{Policy gradient line search plots for 10 MuJoCo environments.}
\label{fig:MuJoCo_lineplot_table}
\end{figure*}
\pagebreak

\subsection{Atari}
\begin{figure*}[!ht]
\centering
\begin{tabular}{ccc}
 \includegraphics[width=\linescale]{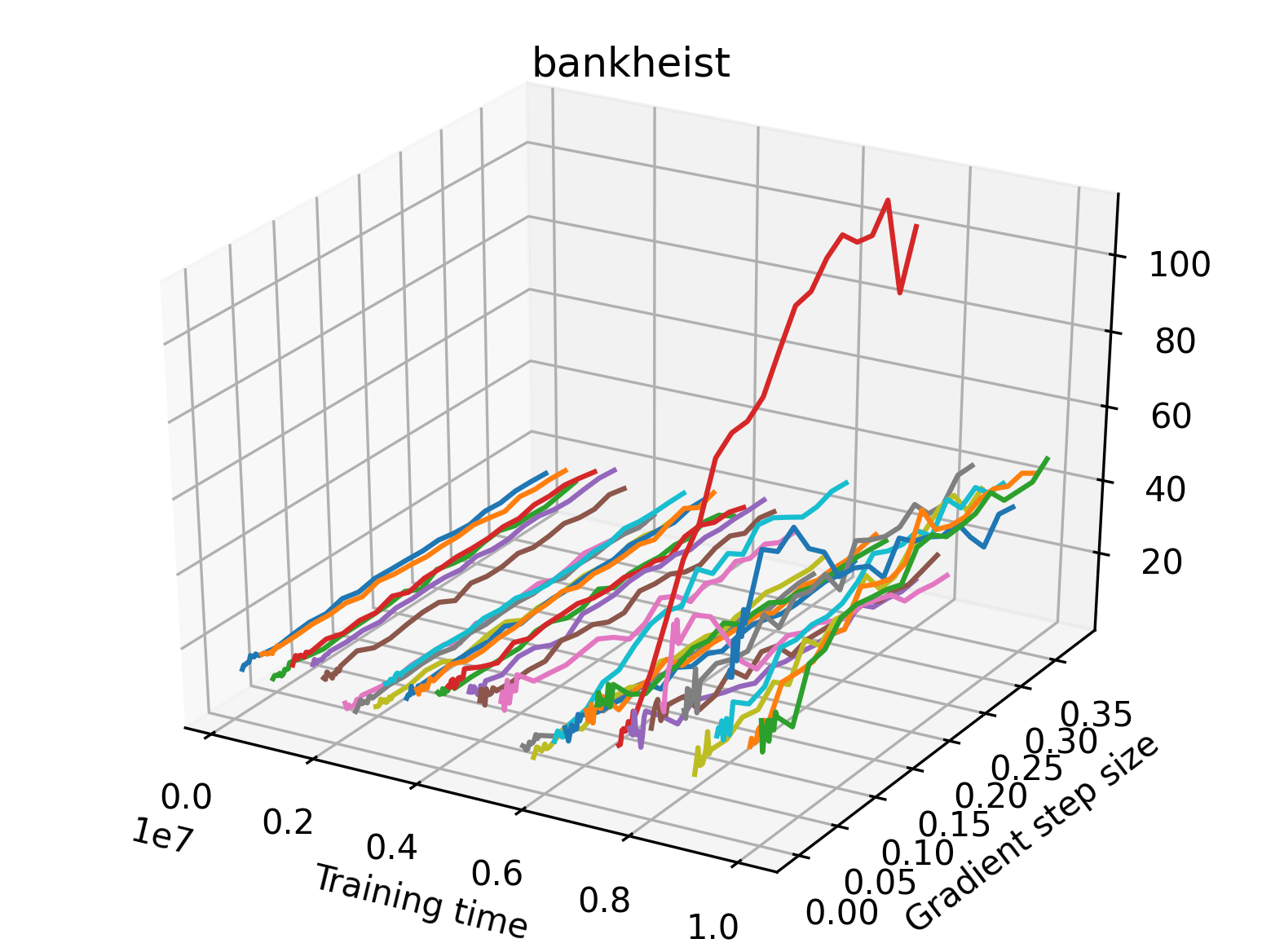} &
 \includegraphics[width=\linescale]{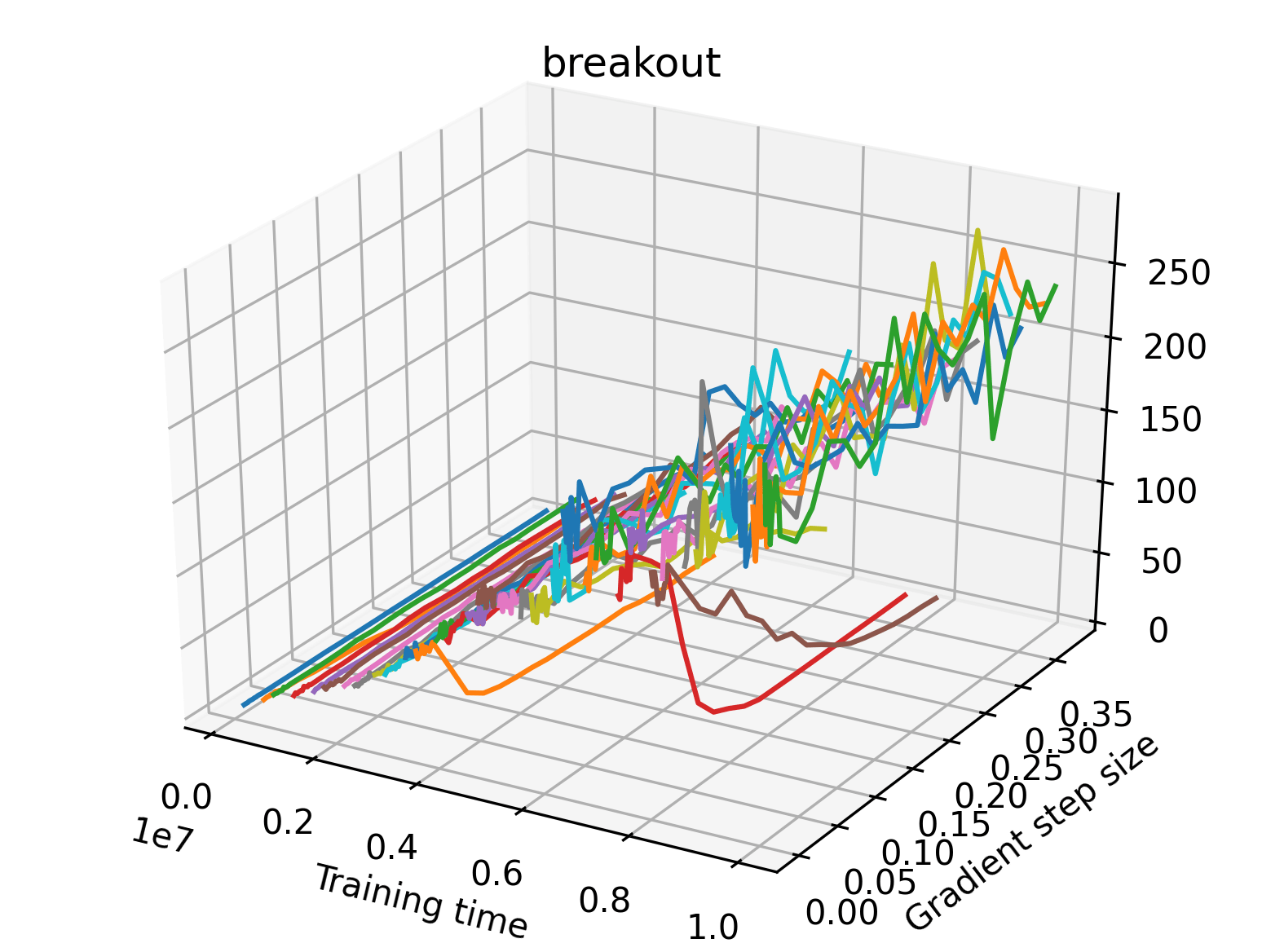} &
 \includegraphics[width=\linescale]{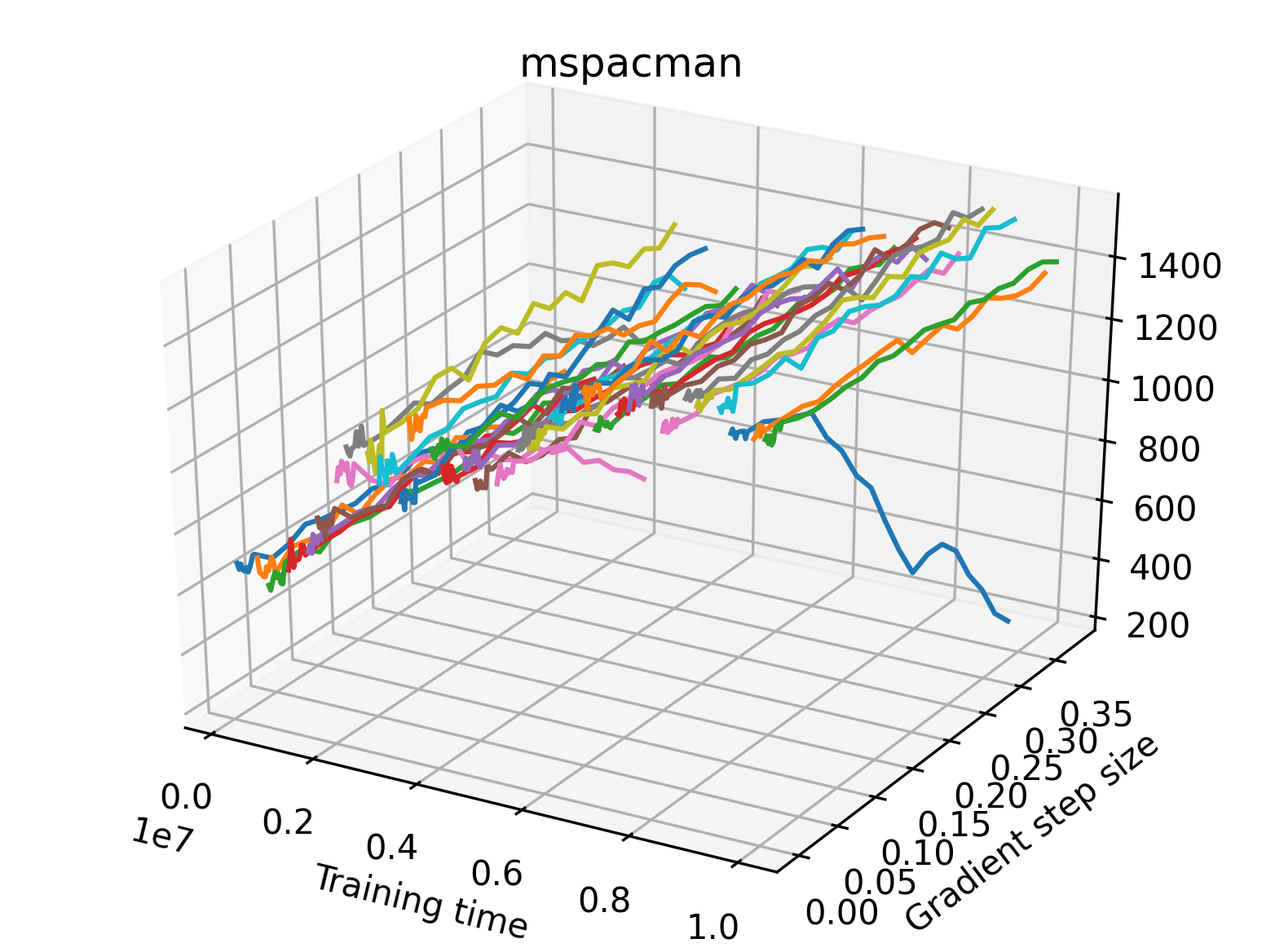} \\
 \includegraphics[width=\linescale]{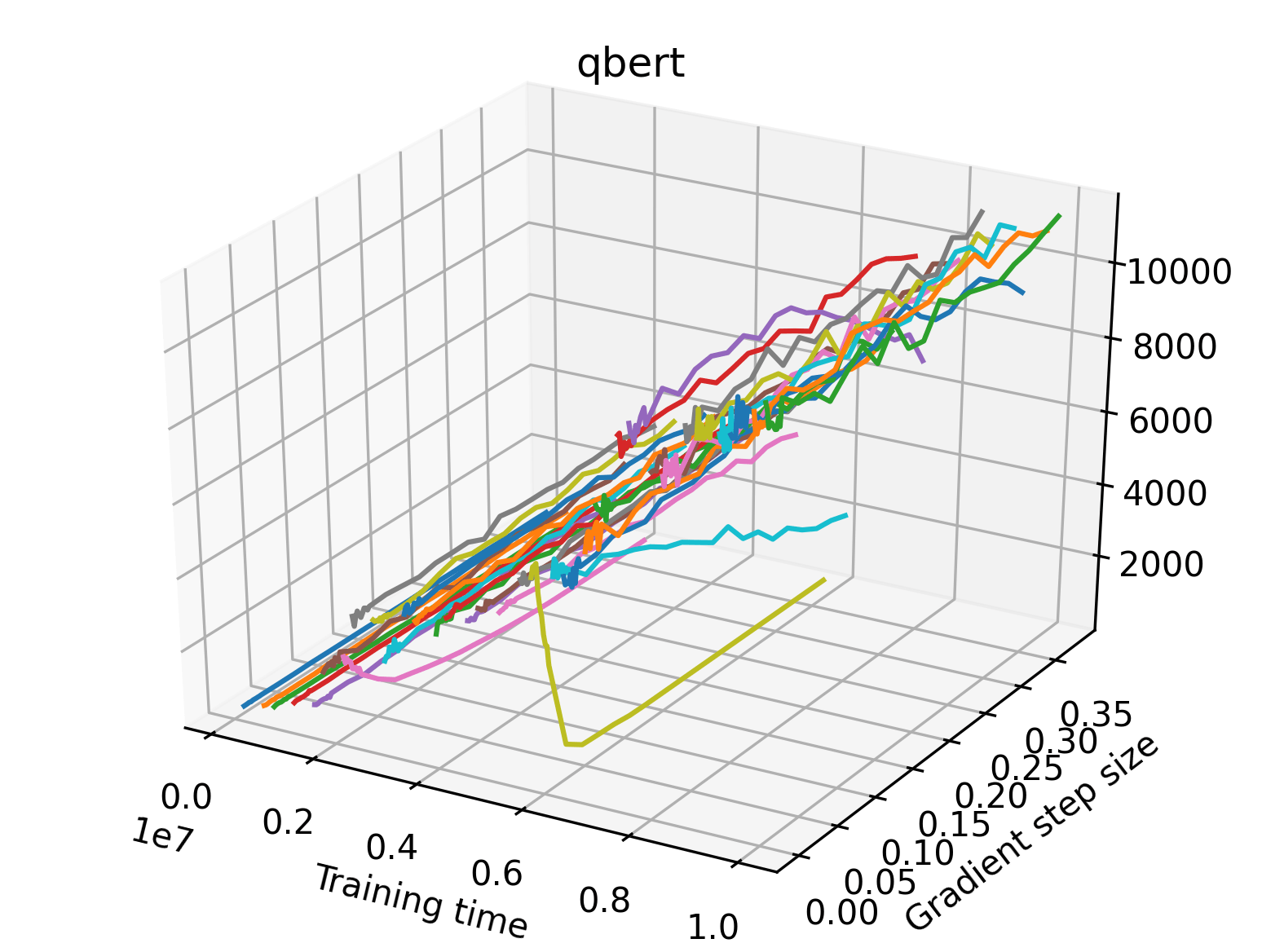} &
 \includegraphics[width=\linescale]{./line/pong_lines.png} &
 \includegraphics[width=\linescale]{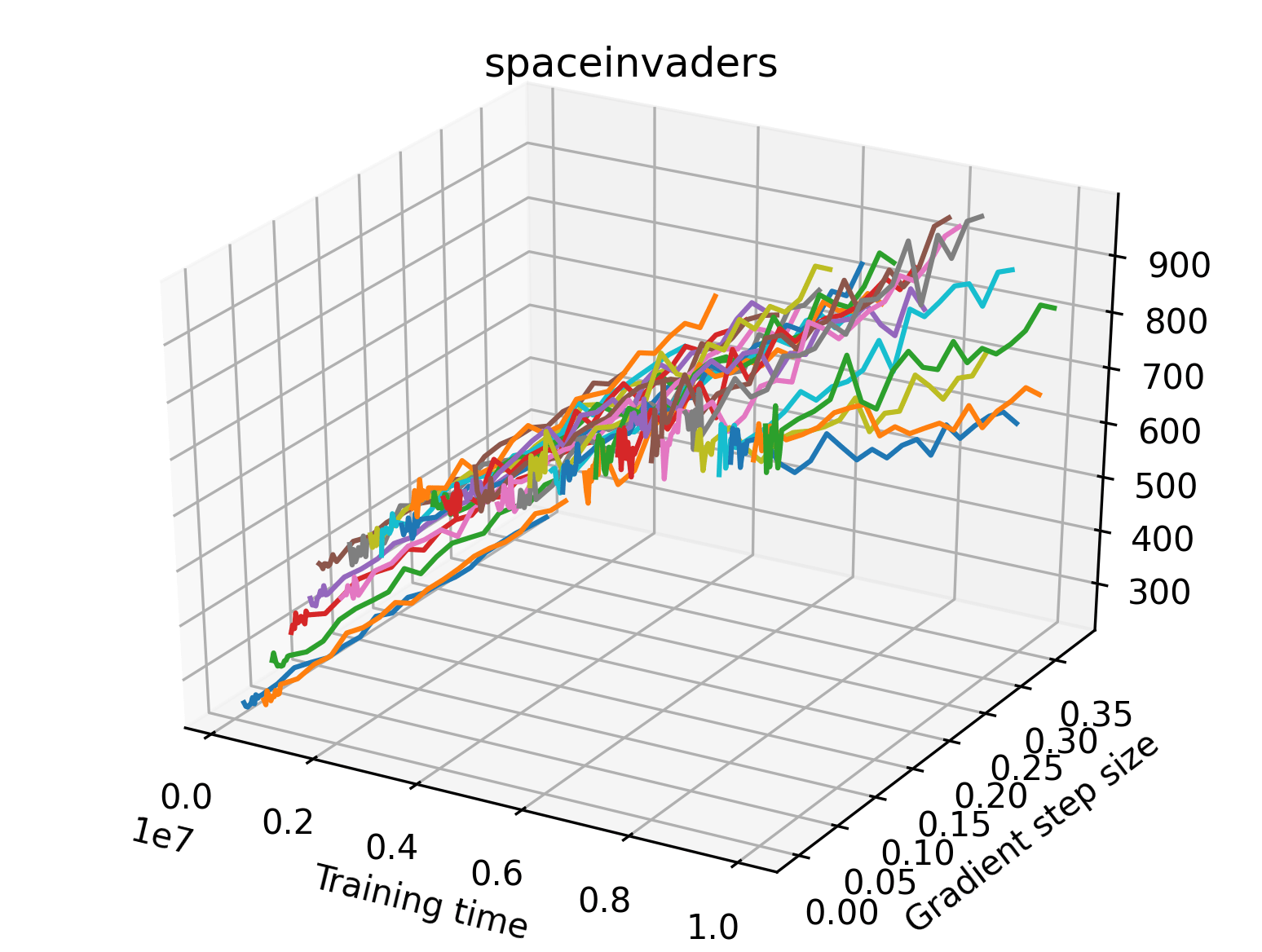} \\
 \includegraphics[width=\linescale]{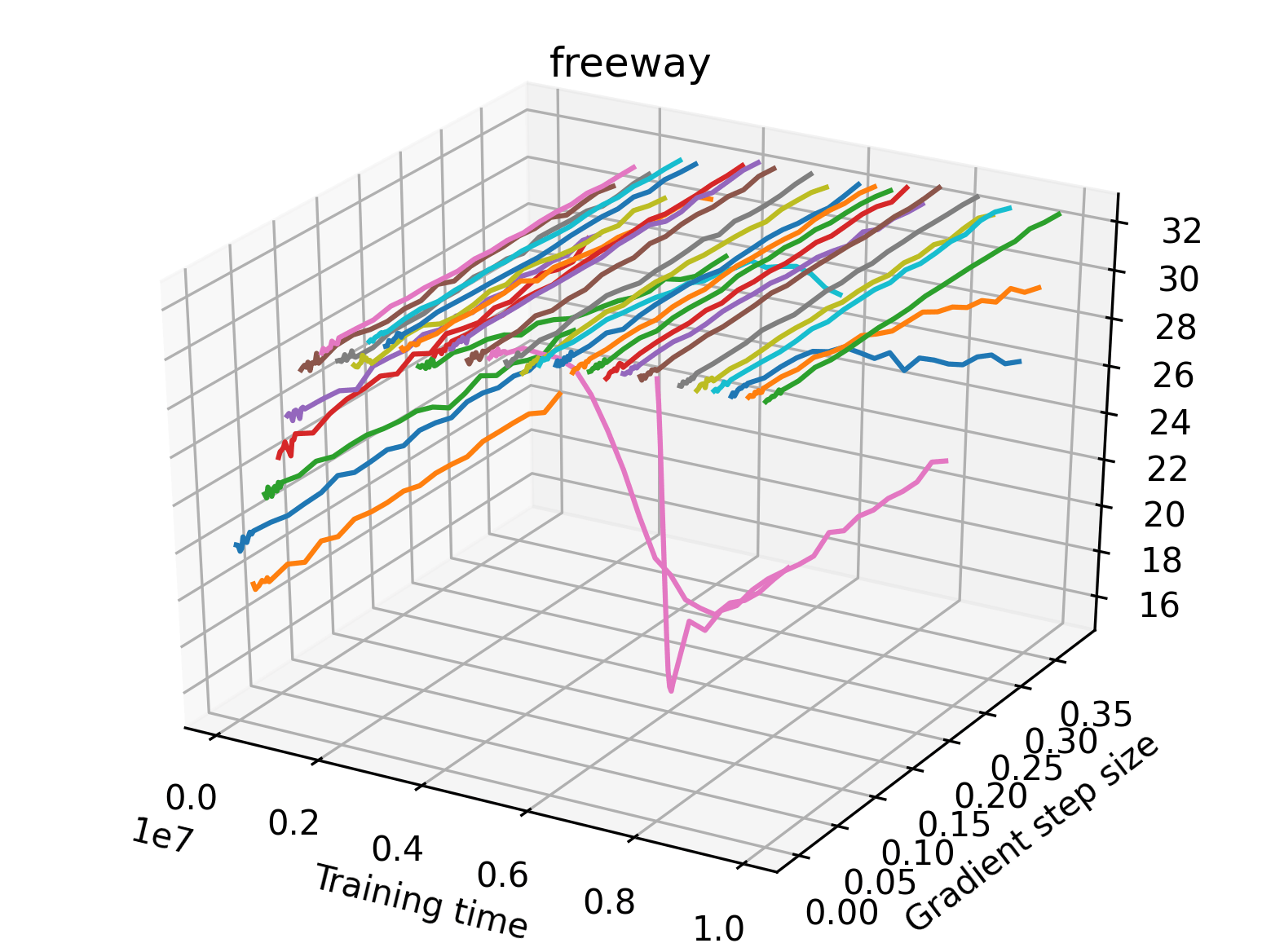} &
 \includegraphics[width=\linescale]{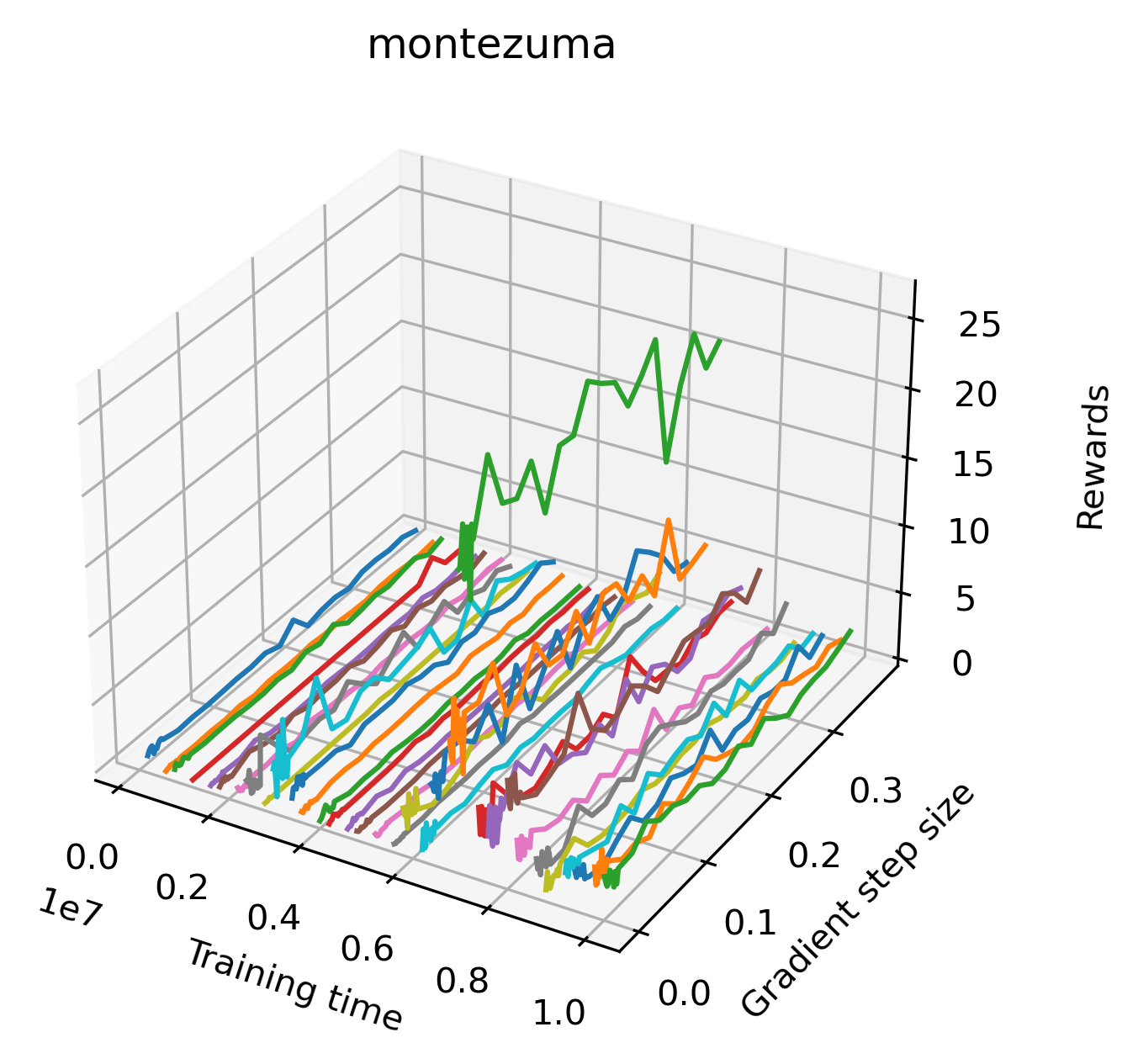} &
 \includegraphics[width=\linescale]{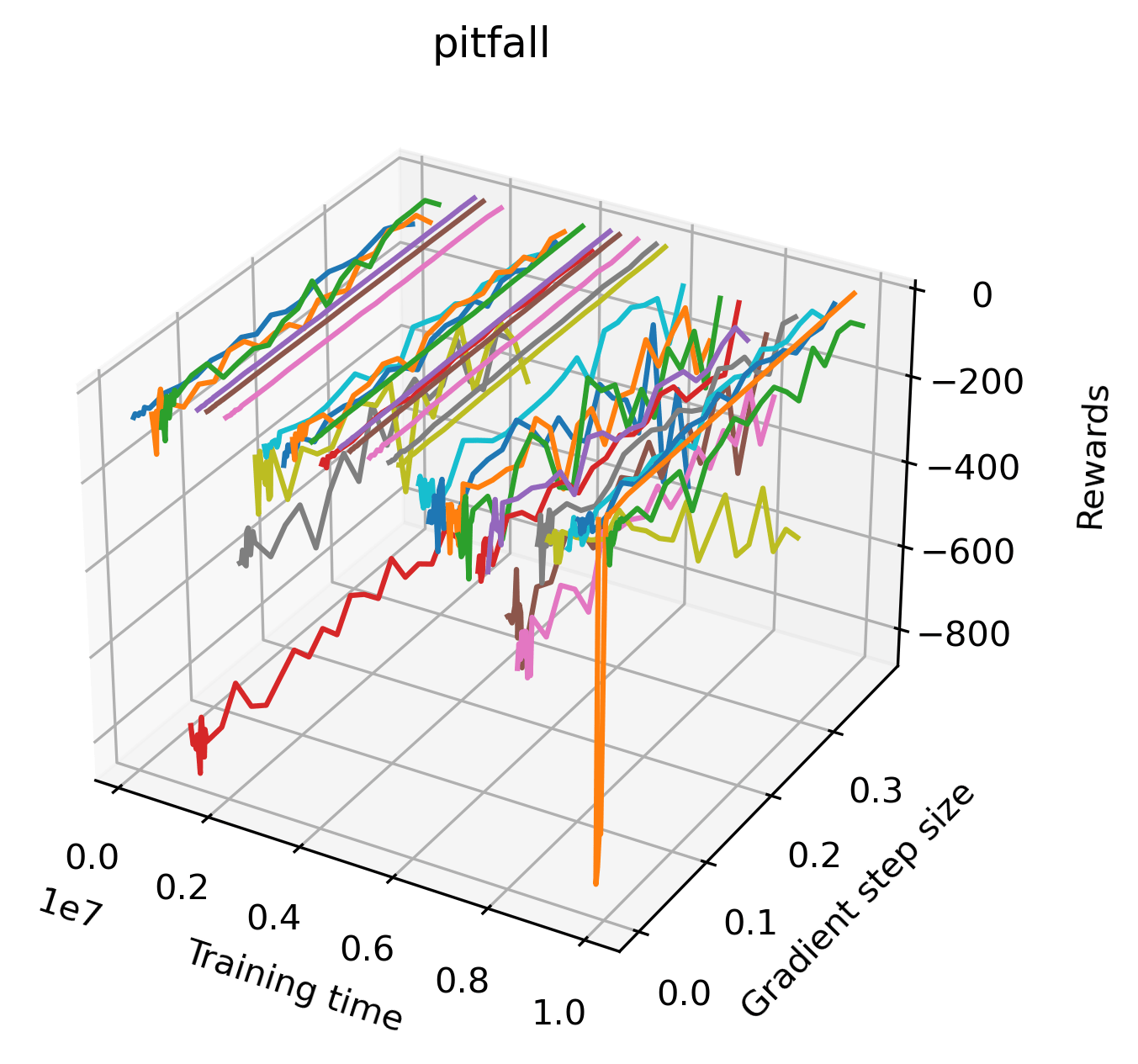} \\ 
 \includegraphics[width=\linescale]{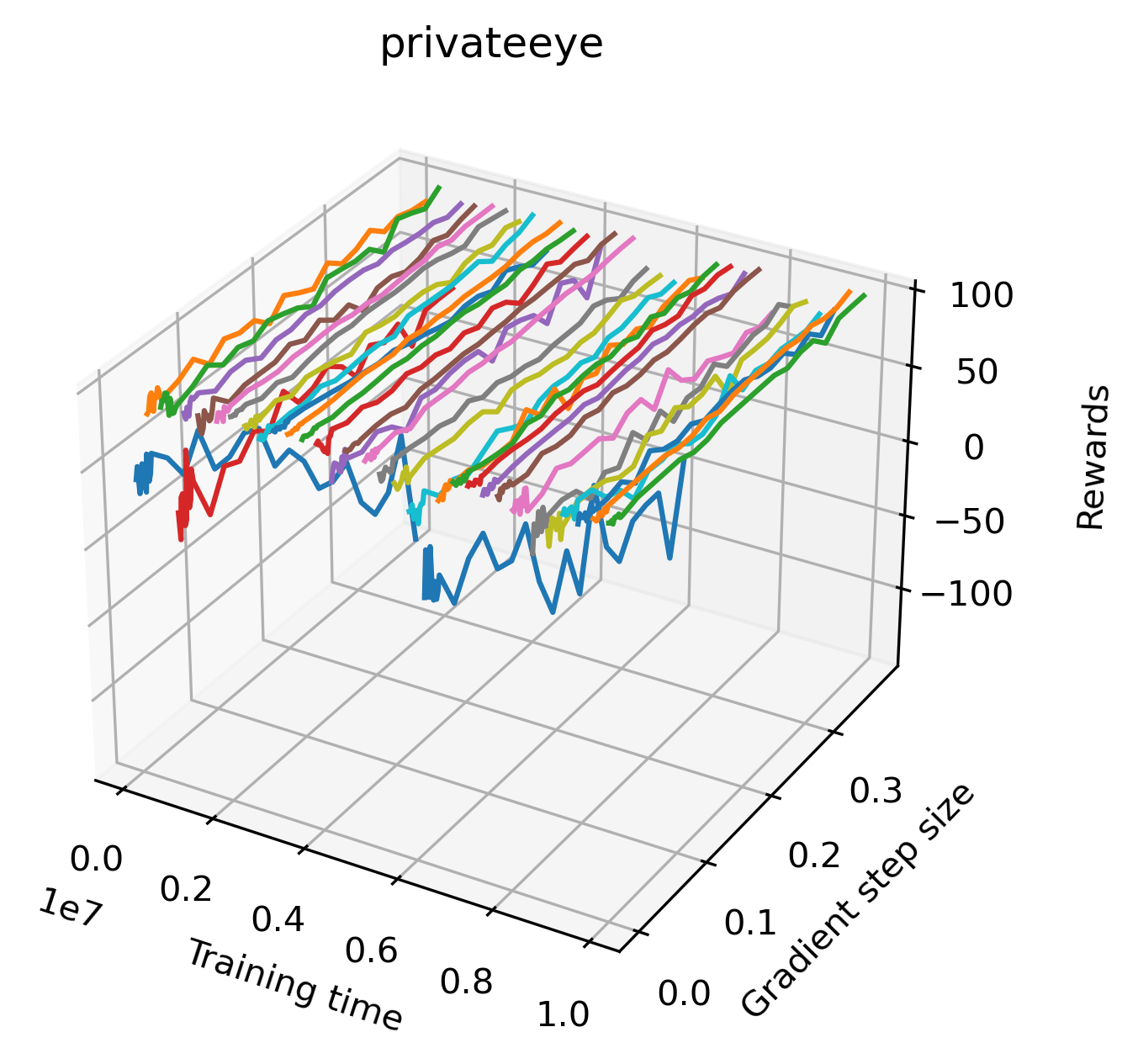} &
 \includegraphics[width=\linescale]{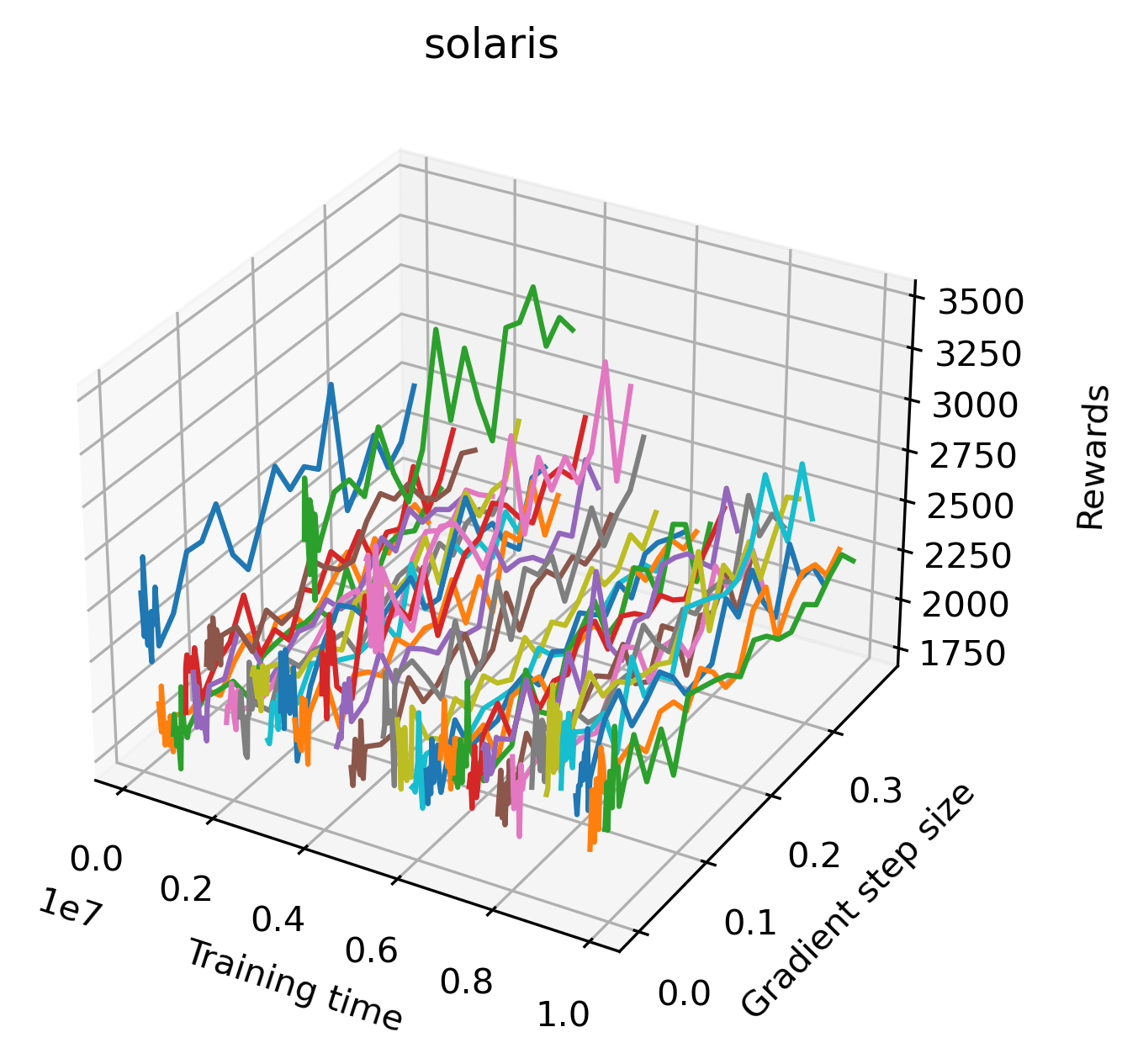} &
 \includegraphics[width=\linescale]{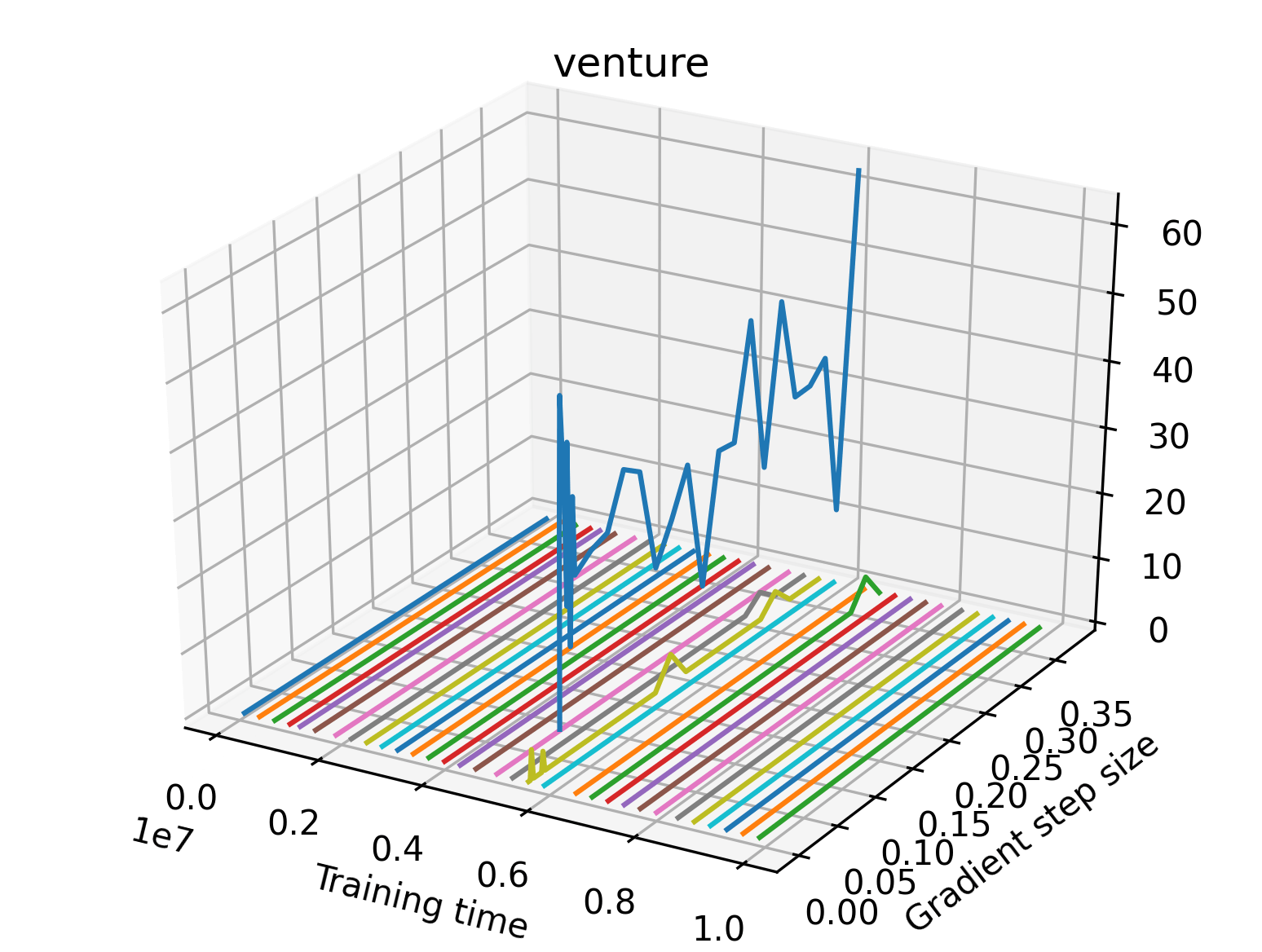} \\
\end{tabular}
\caption{Policy gradient line search plots for 12 Atari environments.}
\label{fig:atari_lineplot_table}
\end{figure*}
\pagebreak

\section{Network Architecture Experiments}
\label{appendix:network_architecture}
We produce plots of the InvertedDoublePendulum-v2, Swimmer-v2, and Walker2D-v2 environments with different network architectures to show the effect that increasing policy network depth has on a reward surface. We test actor and critic networks with 2, 4, 6, 8, 12, and 16 shared layers of 128 nodes. In each environment, as the network depth increases, either the maximizer present in the reward surface becomes sharper, or the maximum reward in the plot decreases. Often we see both occur. This is expected from the findings in \cite{li2017visualizing} that when filter-normalization is used, generalization error decreases as the sharpness of the loss landscape increases. We hope that this serves as a useful preliminary result for those studying network architectures in reinforcement learning, and leave further investigation as future work. For an investigation of loss landscapes for different RL architectures, see \citet{ota2021training}.
 
 \newcommand\architecturescale{0.31\linewidth}
\begin{figure*}[!ht]
\centering
\begin{tabular}{ccccc}
 \includegraphics[width=\architecturescale]{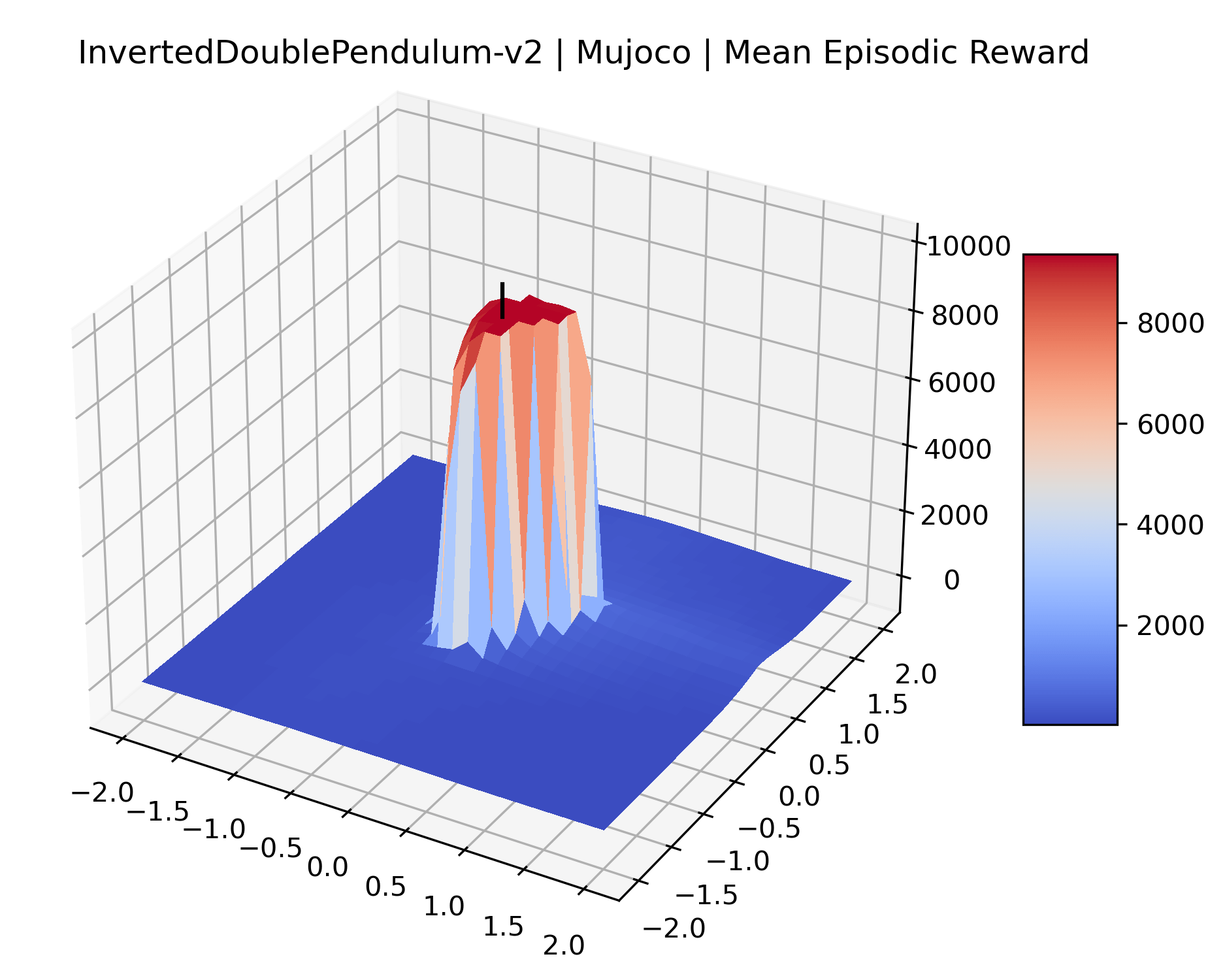} &
 \includegraphics[width=\architecturescale]{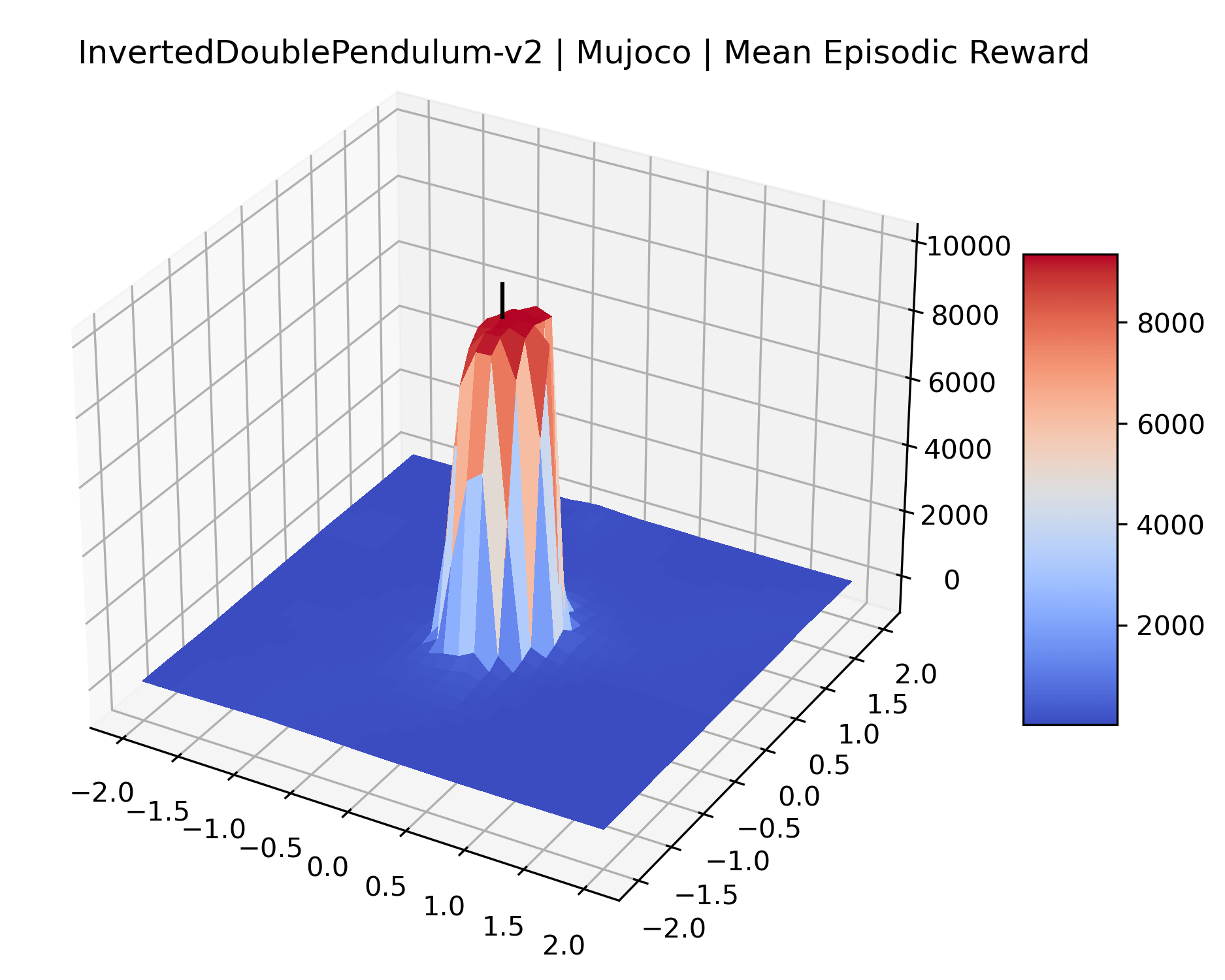} &
 \includegraphics[width=\architecturescale]{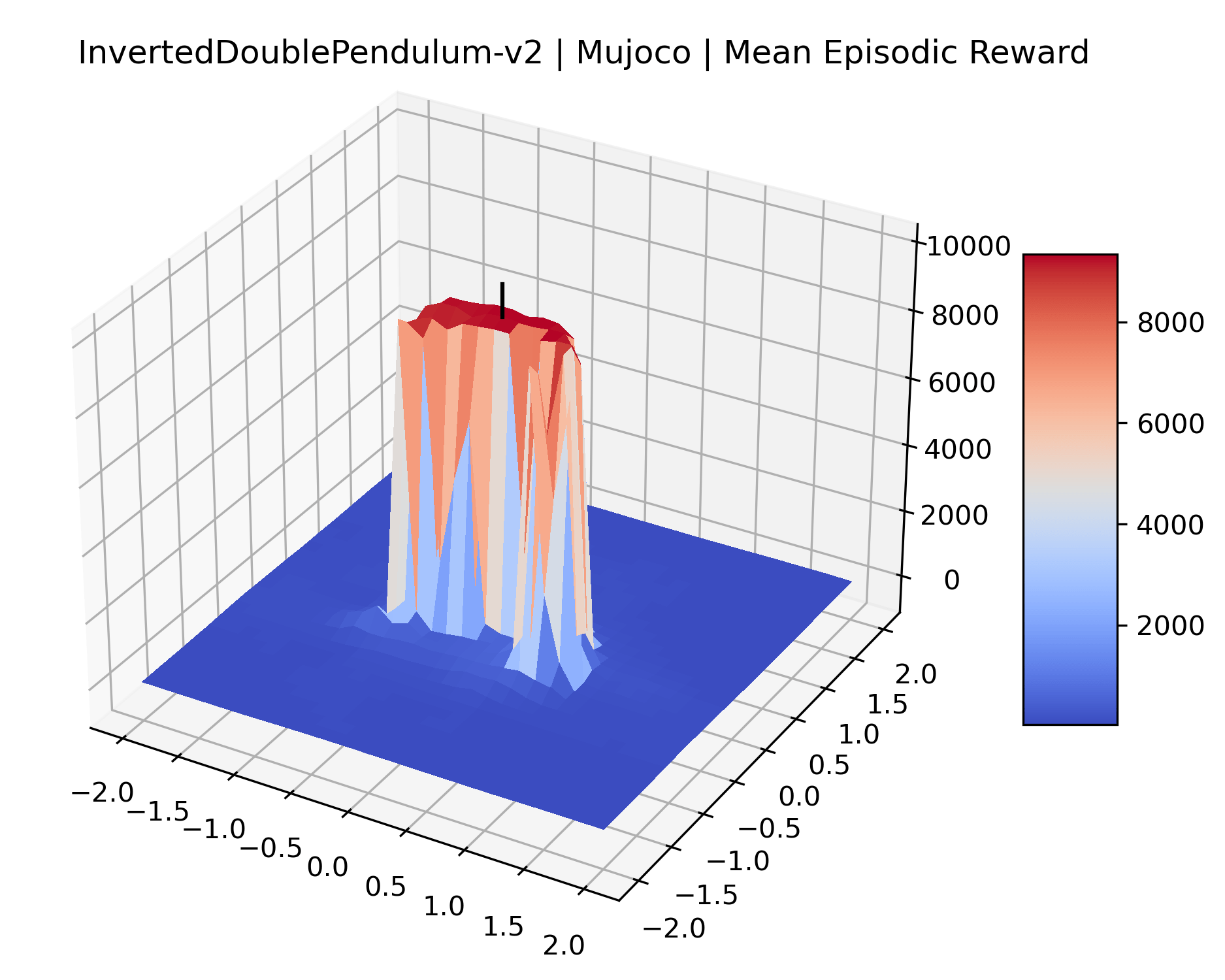} \\
 \includegraphics[width=\architecturescale]{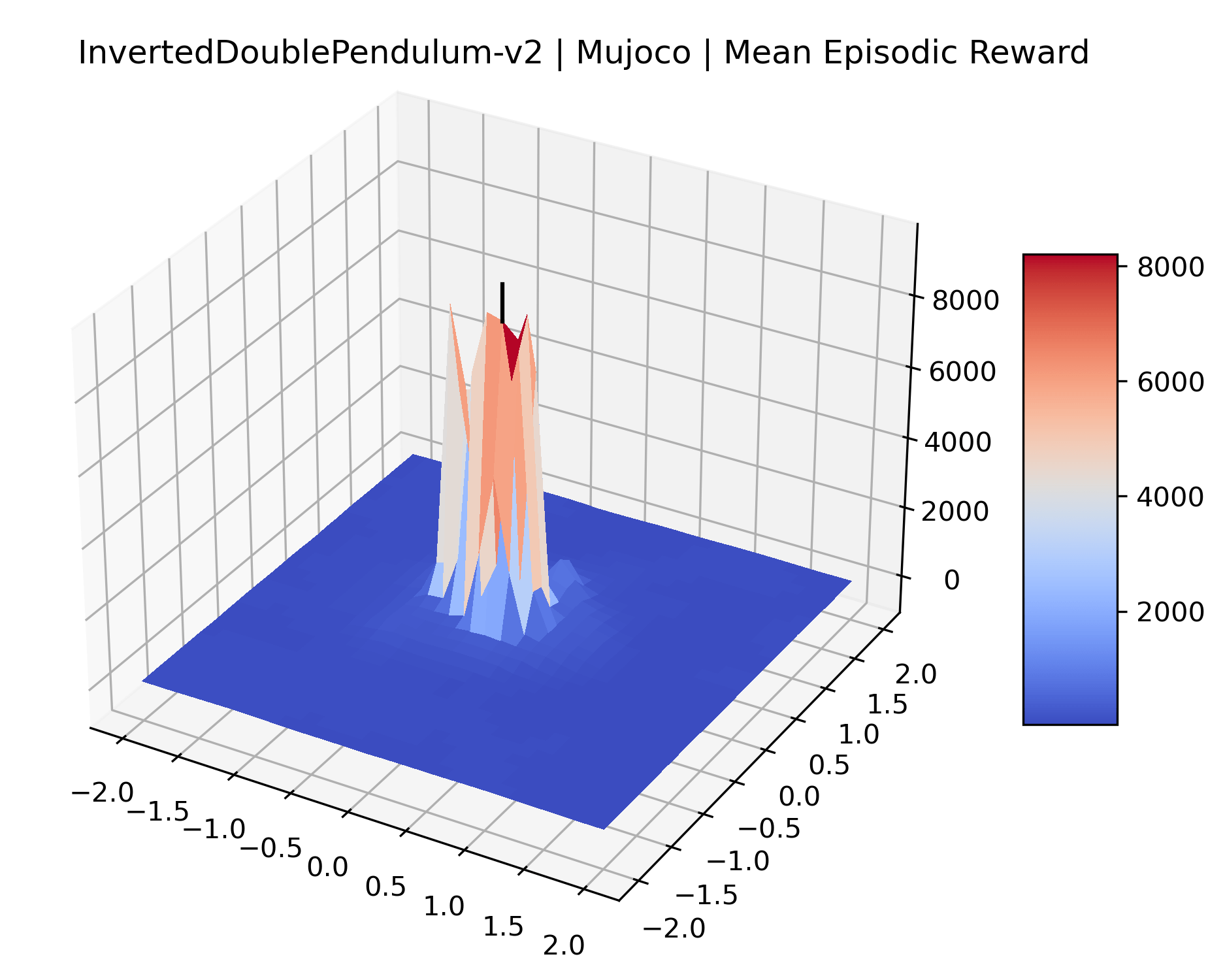} &
 \includegraphics[width=\architecturescale]{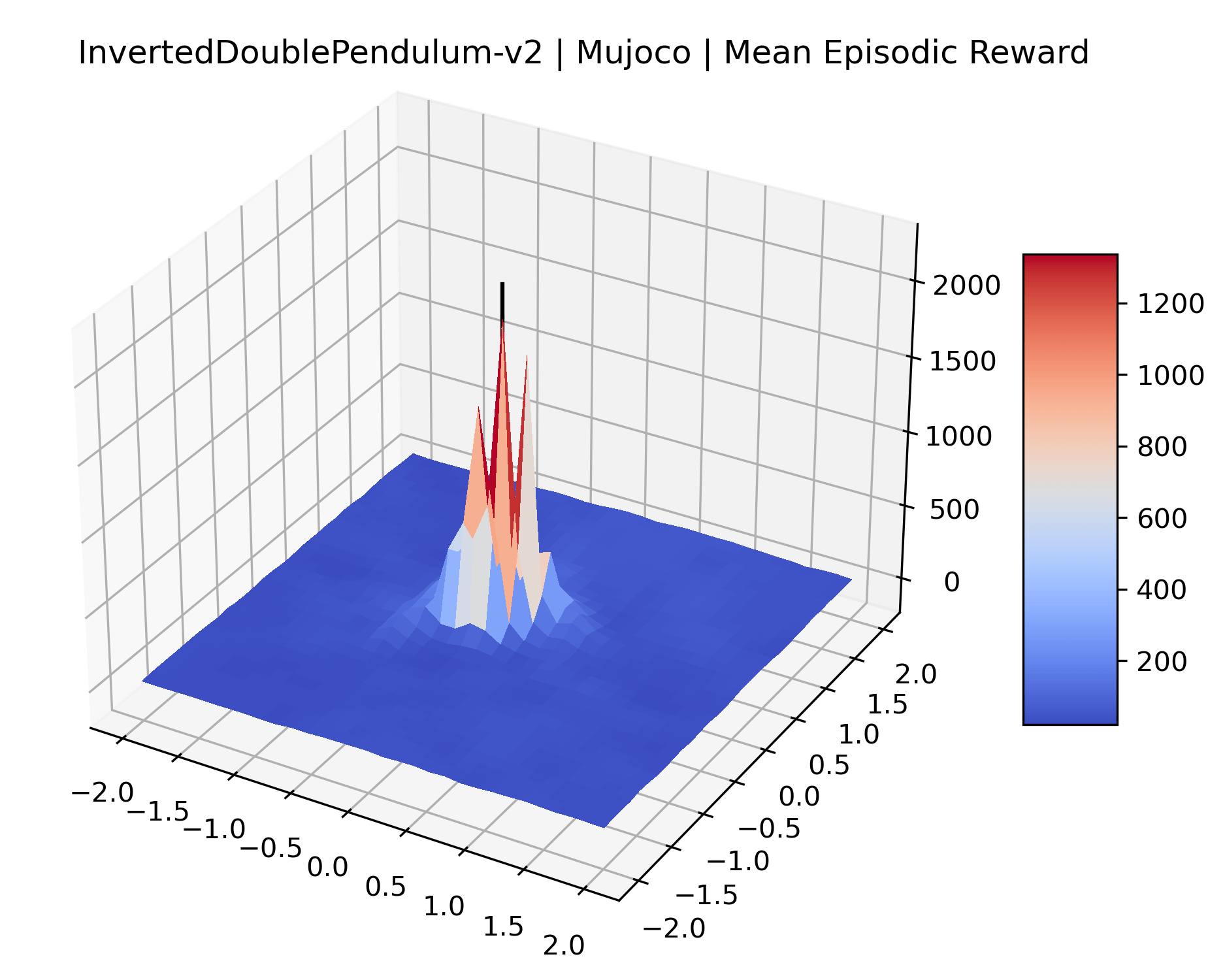} &
 \includegraphics[width=\architecturescale]{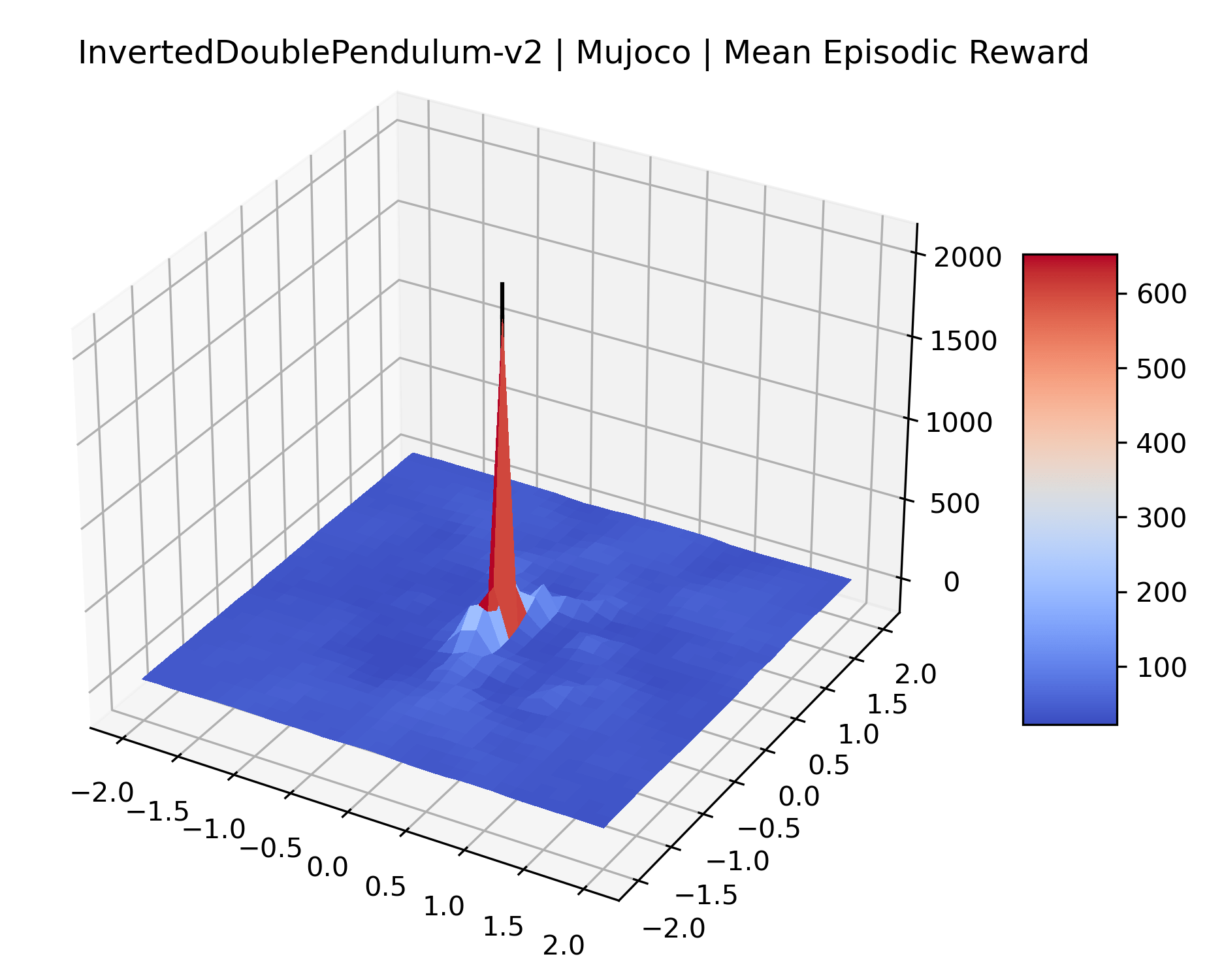} \\
 \end{tabular}
\caption{Inverted Double Pendulum environment with different number of layers in policy network. Top: 2, 4, and 6 layer networks. Bottom: 8, 12, and 16 layer networks}
\label{fig:inv_architecture_rewardsurface_table}
\end{figure*}

\begin{figure*}[!ht]
\centering
\begin{tabular}{ccccc}
 \includegraphics[width=\architecturescale]{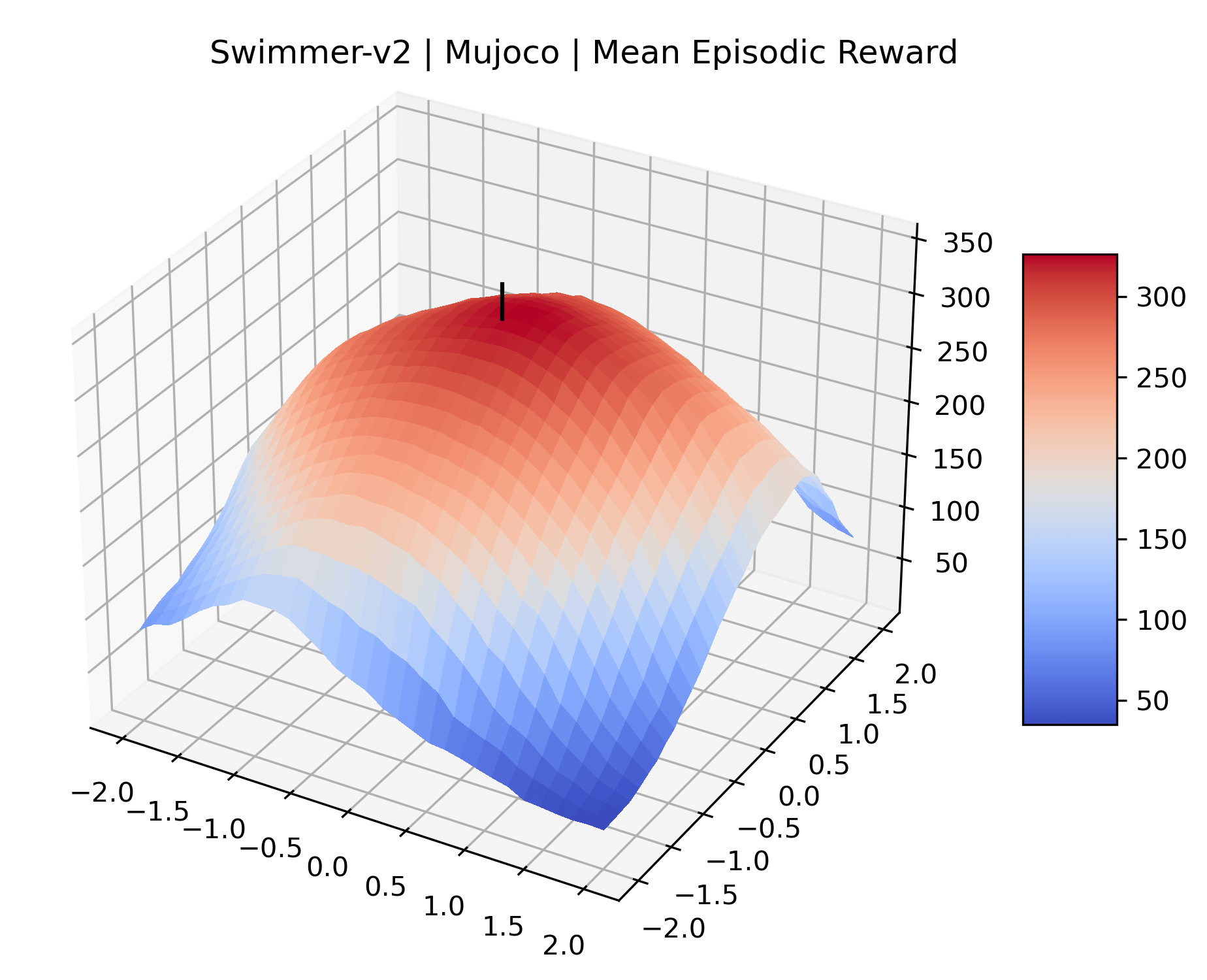} &
 \includegraphics[width=\architecturescale]{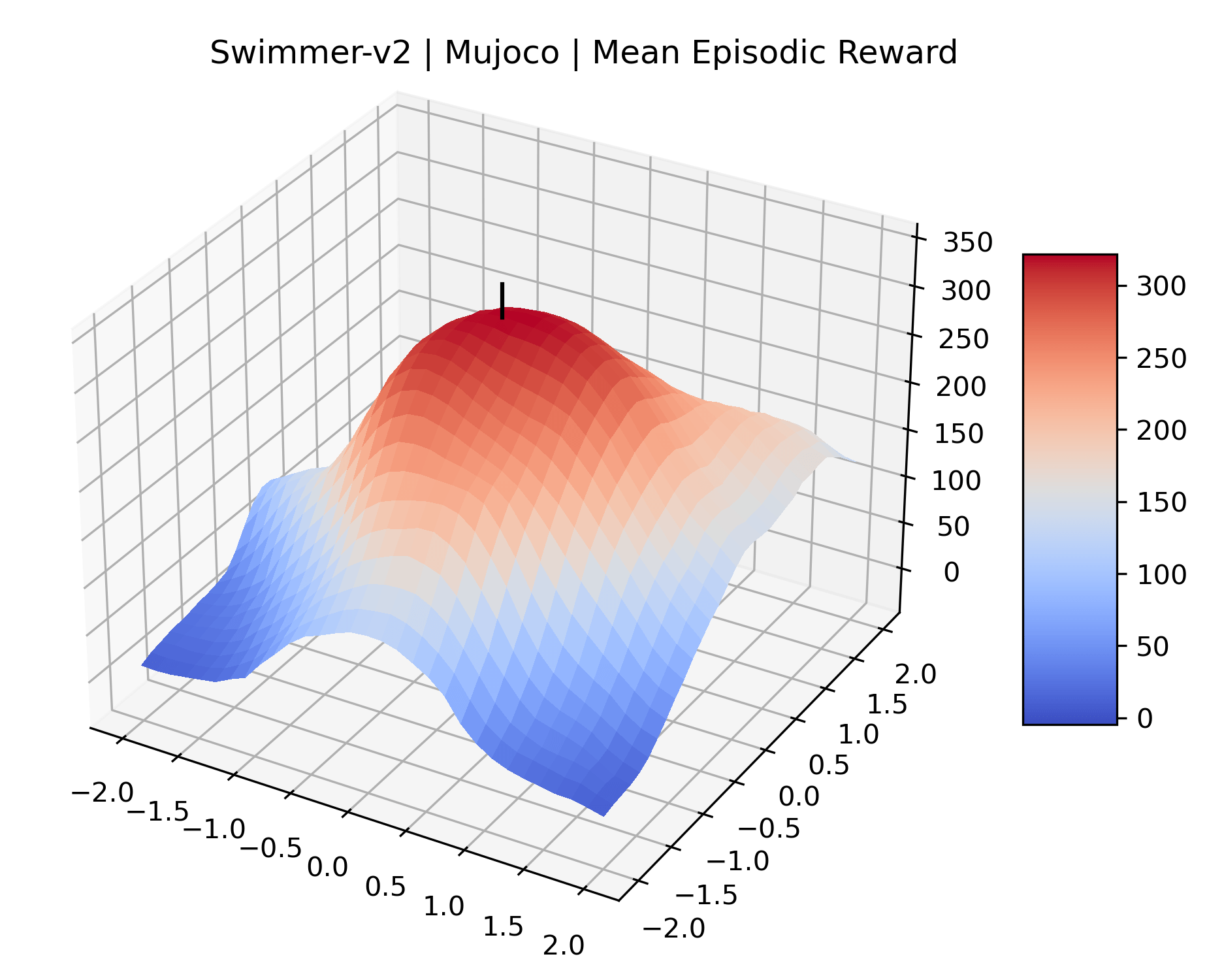} &
 \includegraphics[width=\architecturescale]{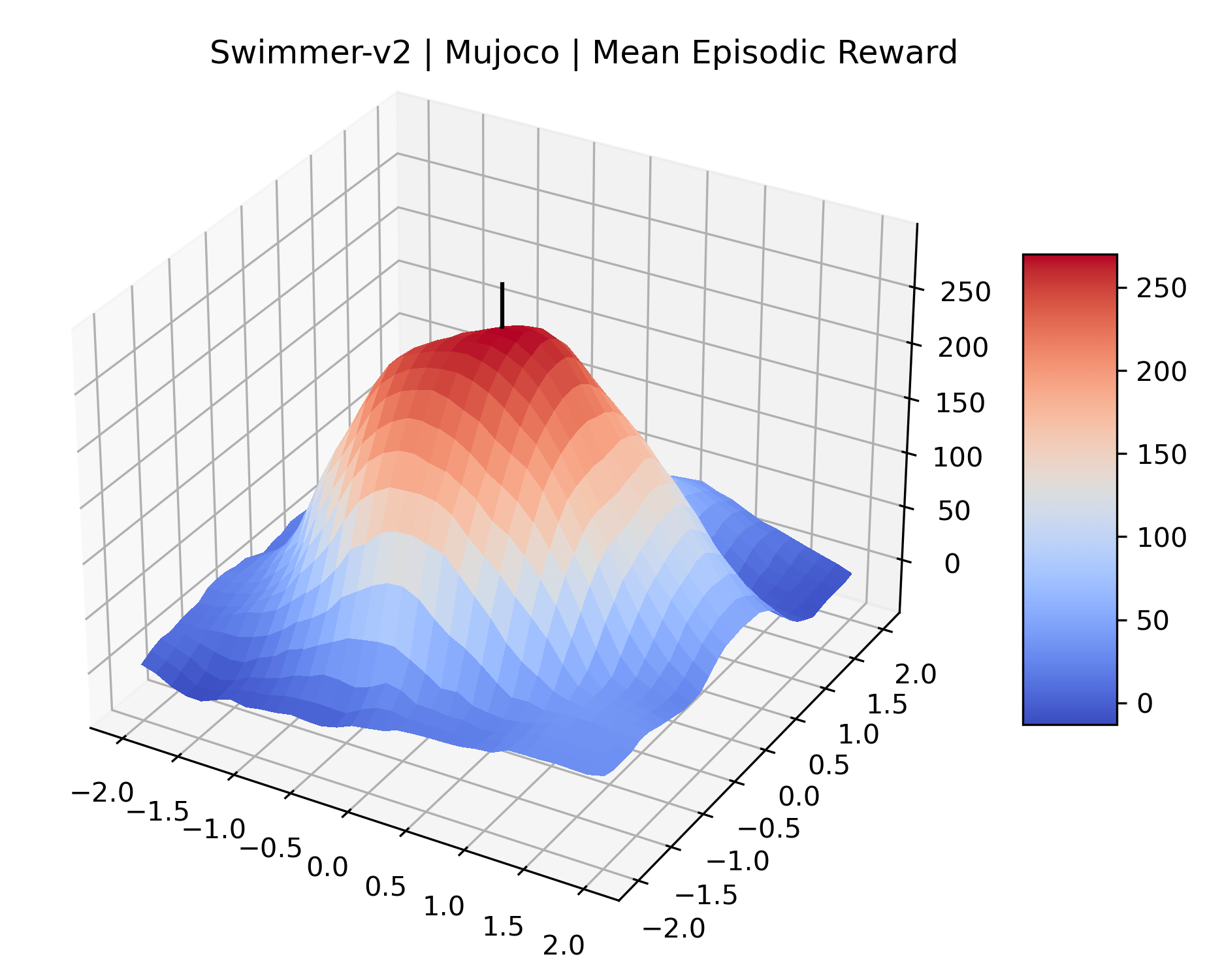} \\
 \includegraphics[width=\architecturescale]{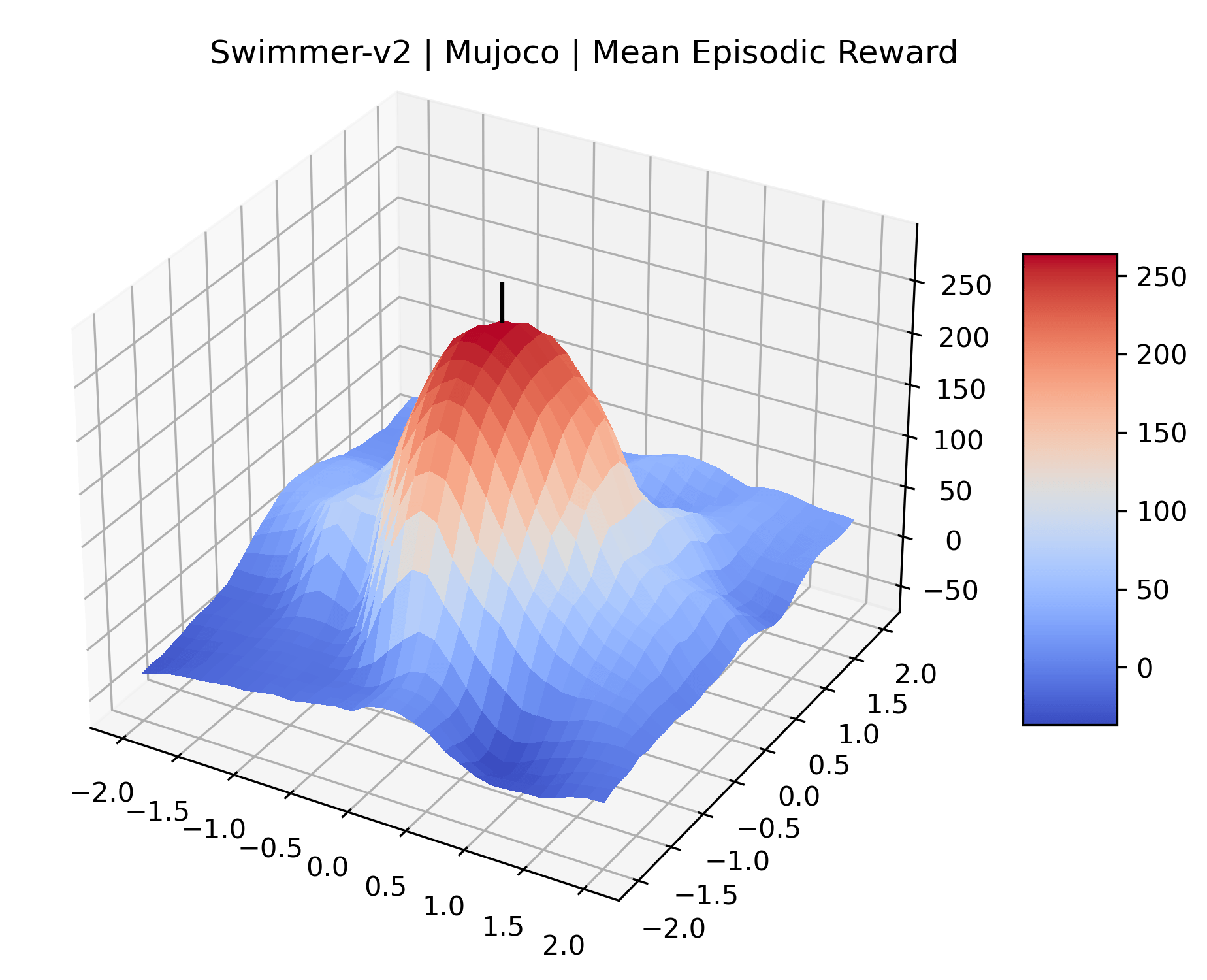} &
 \includegraphics[width=\architecturescale]{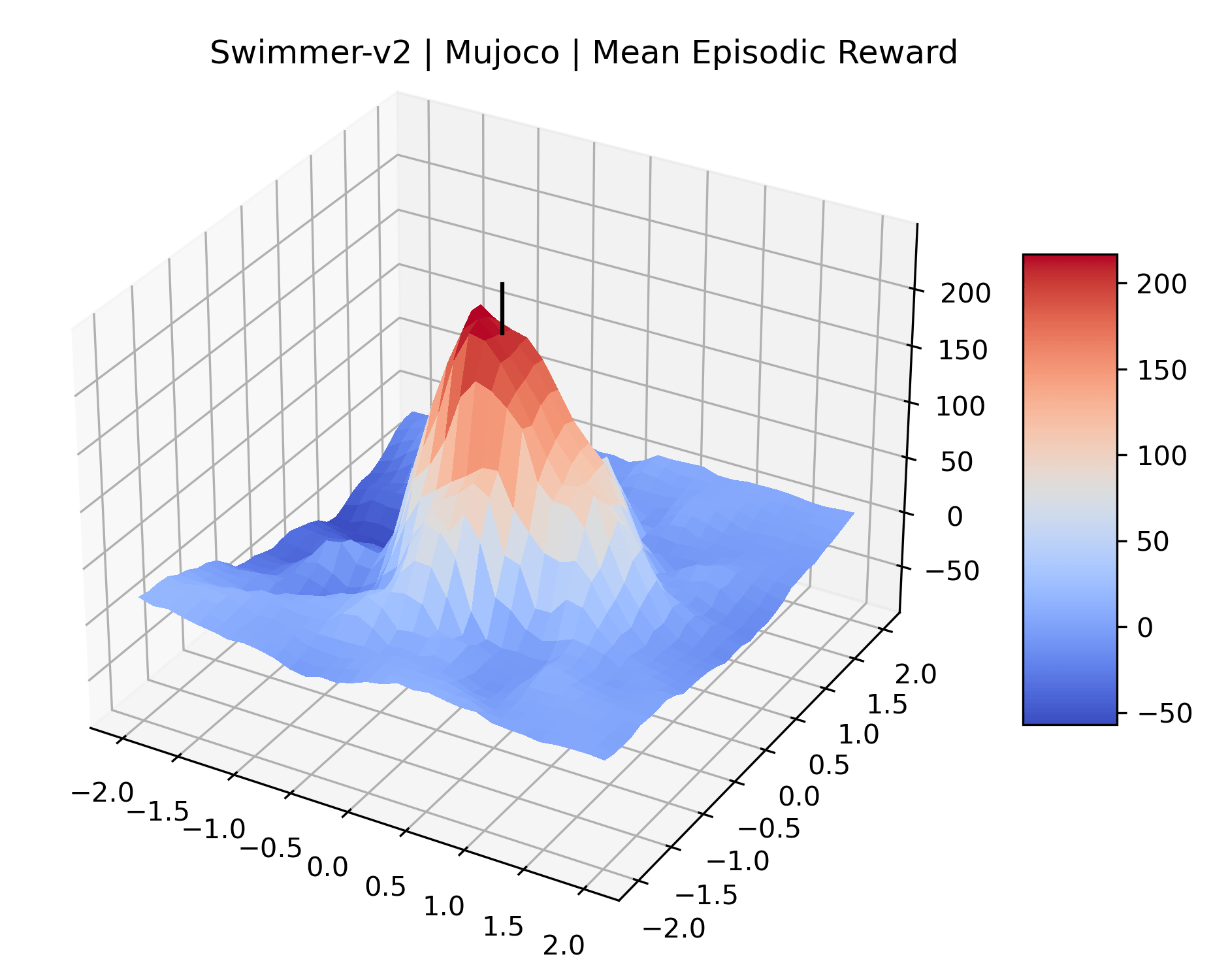} &
 \includegraphics[width=\architecturescale]{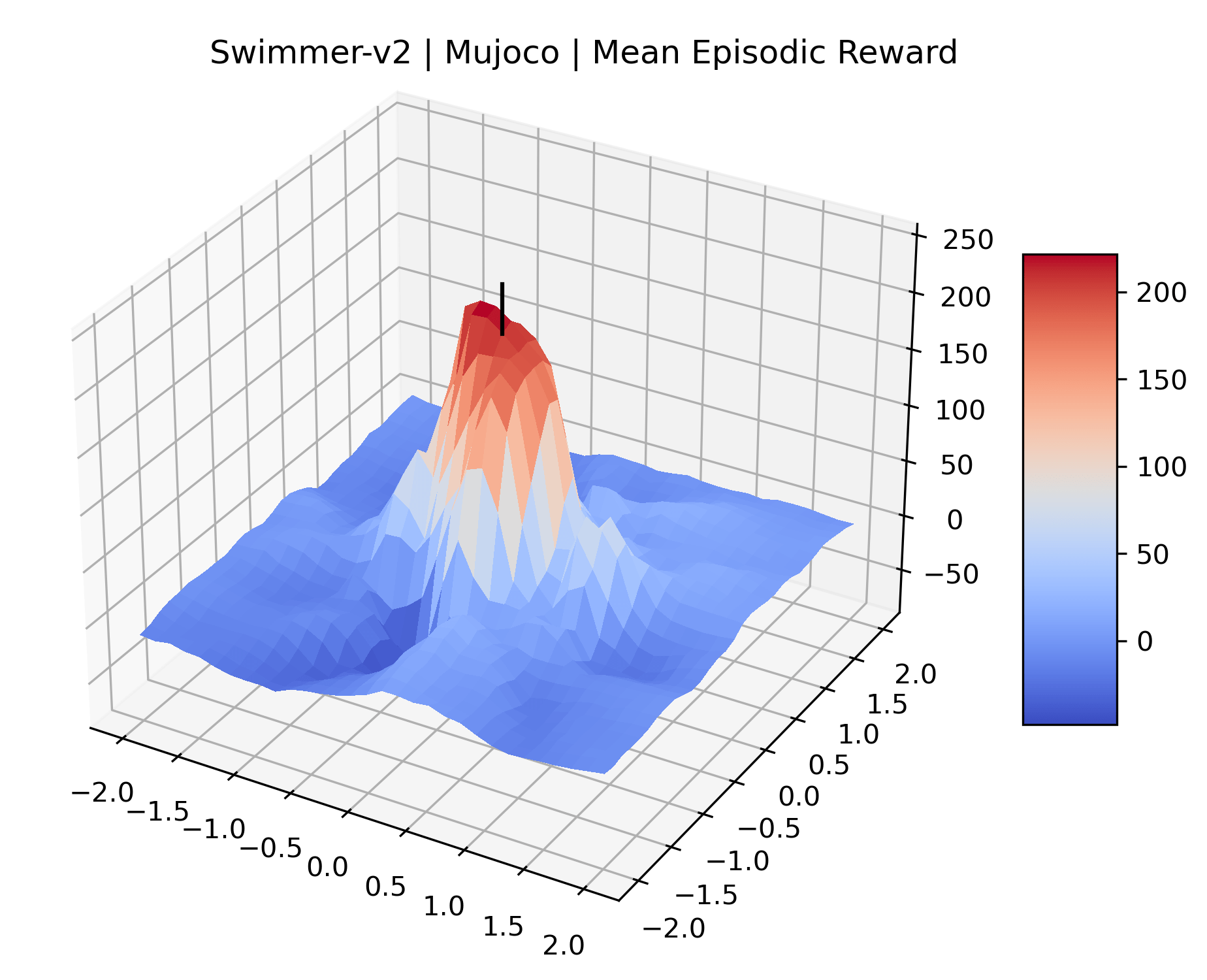} \\
 \end{tabular}
\caption{Swimmer environment with different number of layers in policy network. Top: 2, 4, and 6 layer networks. Bottom: 8, 12, and 16 layer networks}
\label{fig:swimmer_architecture_rewardsurface_table}
\end{figure*}
\pagebreak

\begin{figure*}[!ht]
\centering
\begin{tabular}{ccccc}
 \includegraphics[width=\architecturescale]{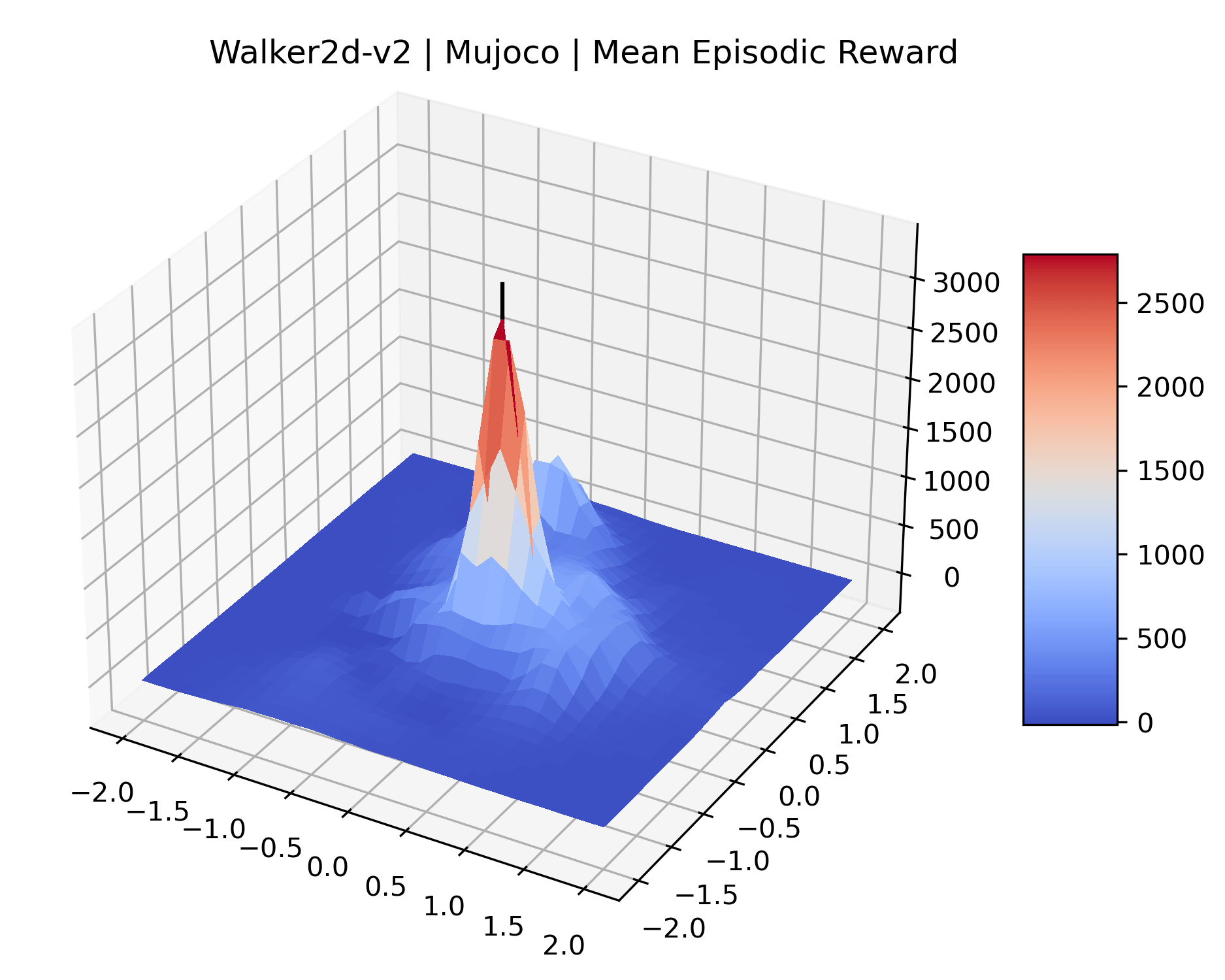} &
 \includegraphics[width=\architecturescale]{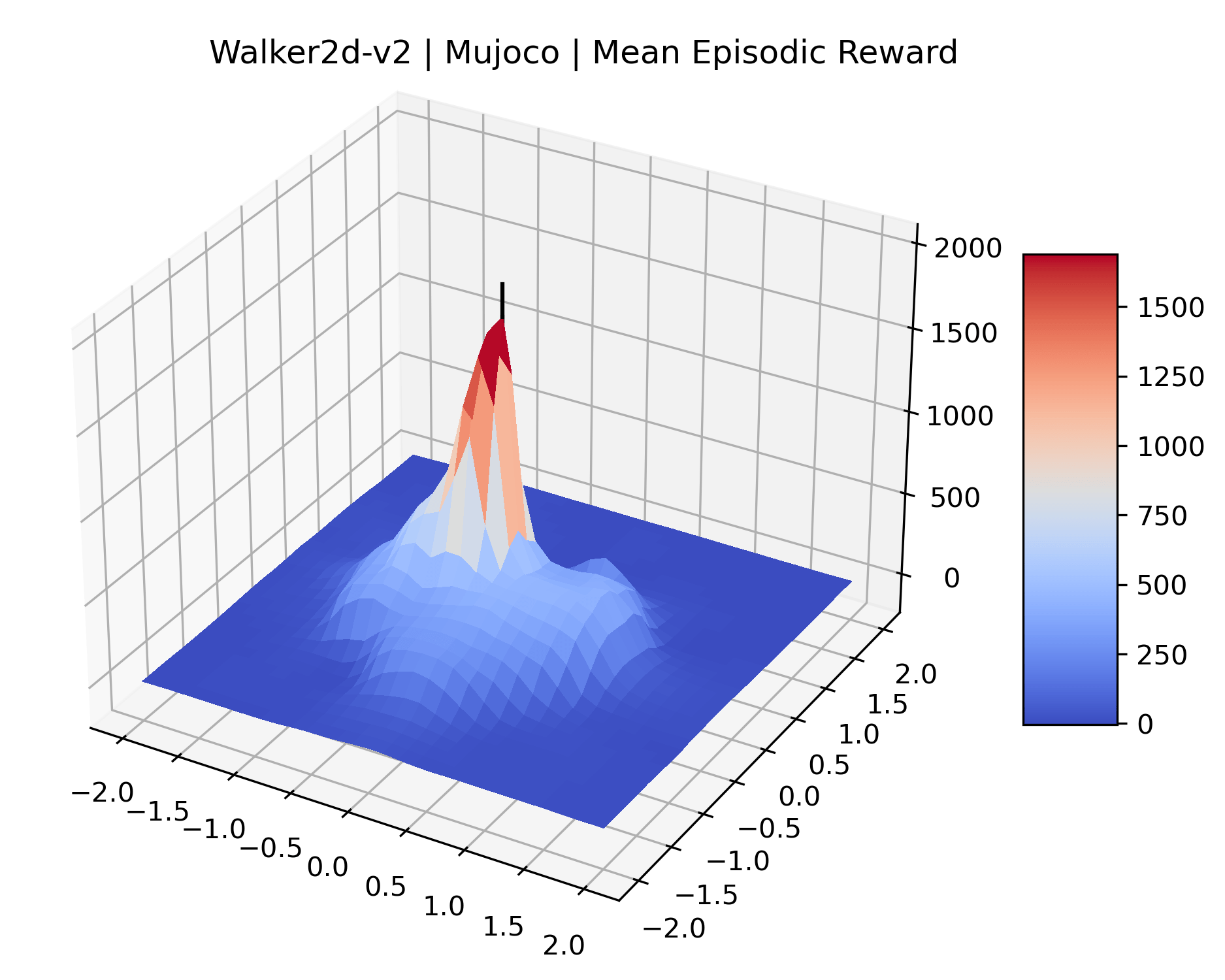} &
 \includegraphics[width=\architecturescale]{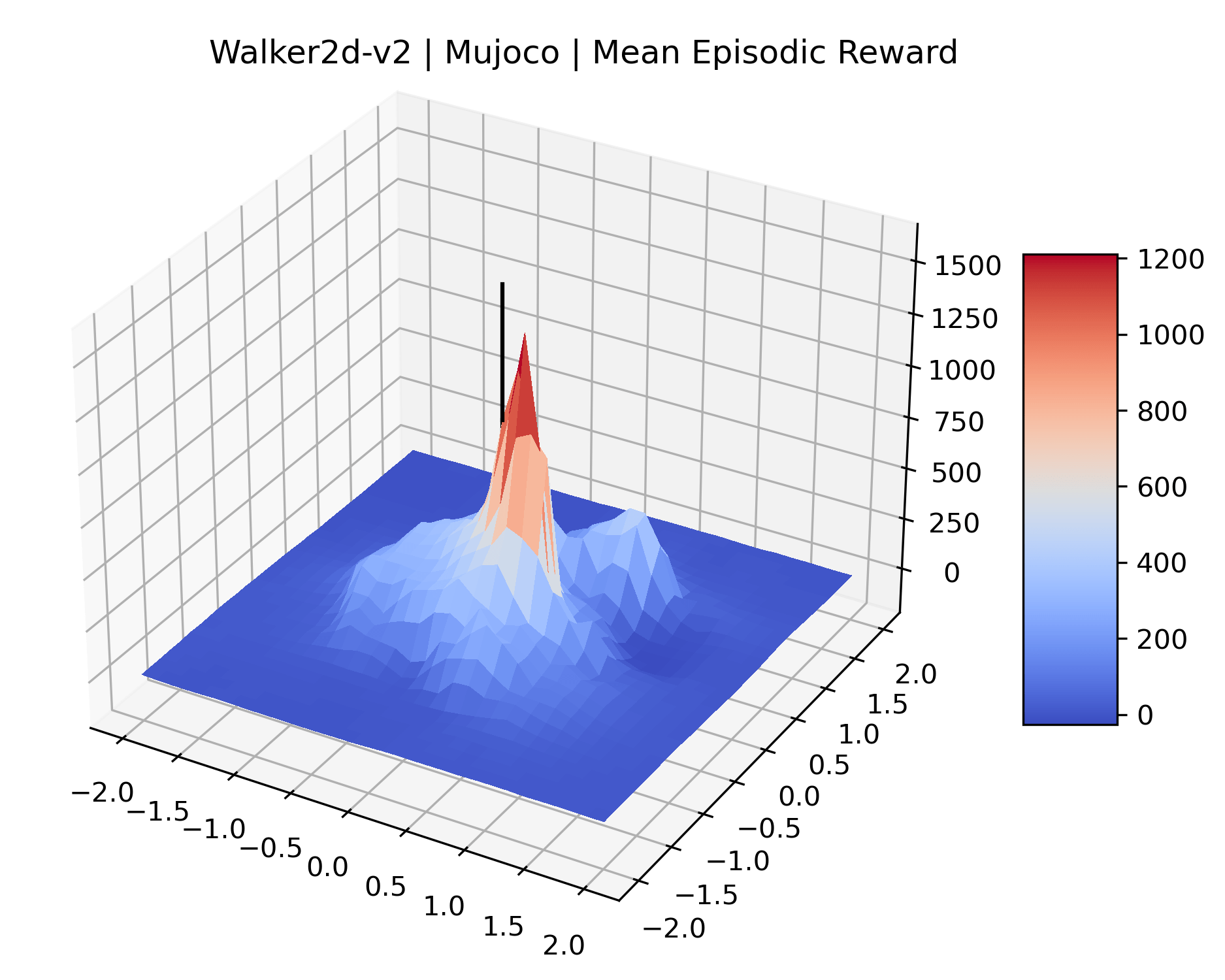} \\
 \includegraphics[width=\architecturescale]{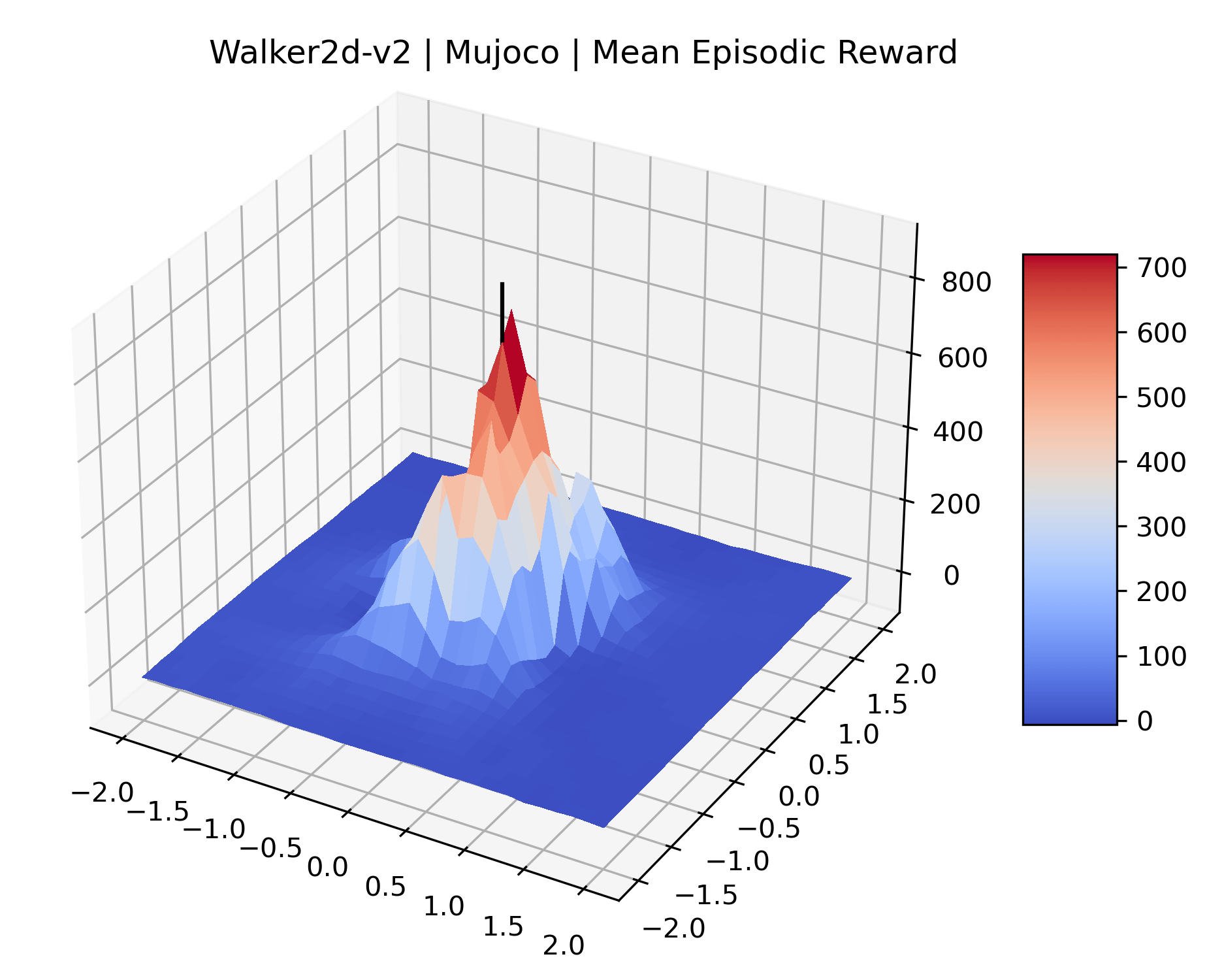} &
 \includegraphics[width=\architecturescale]{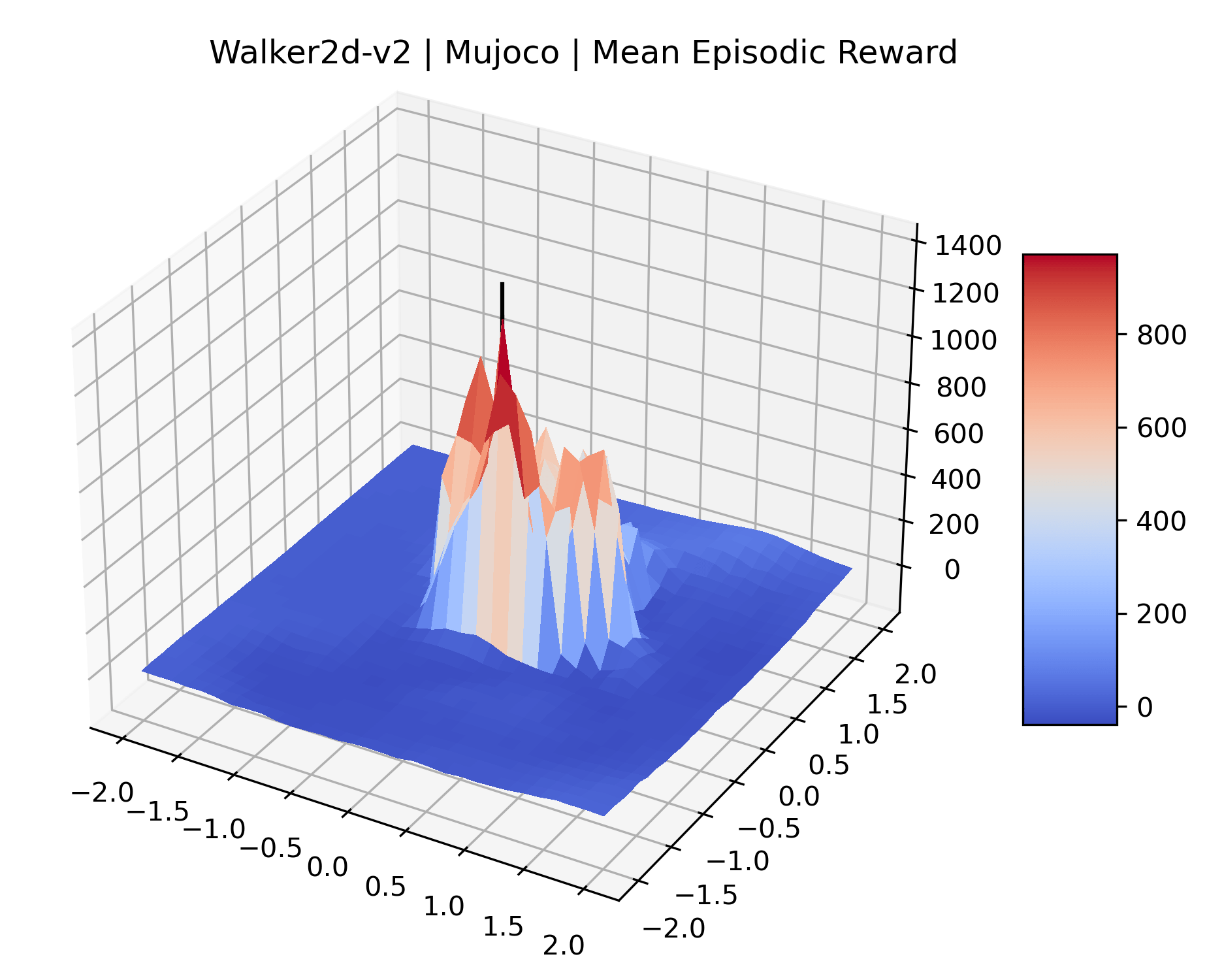} &
 \includegraphics[width=\architecturescale]{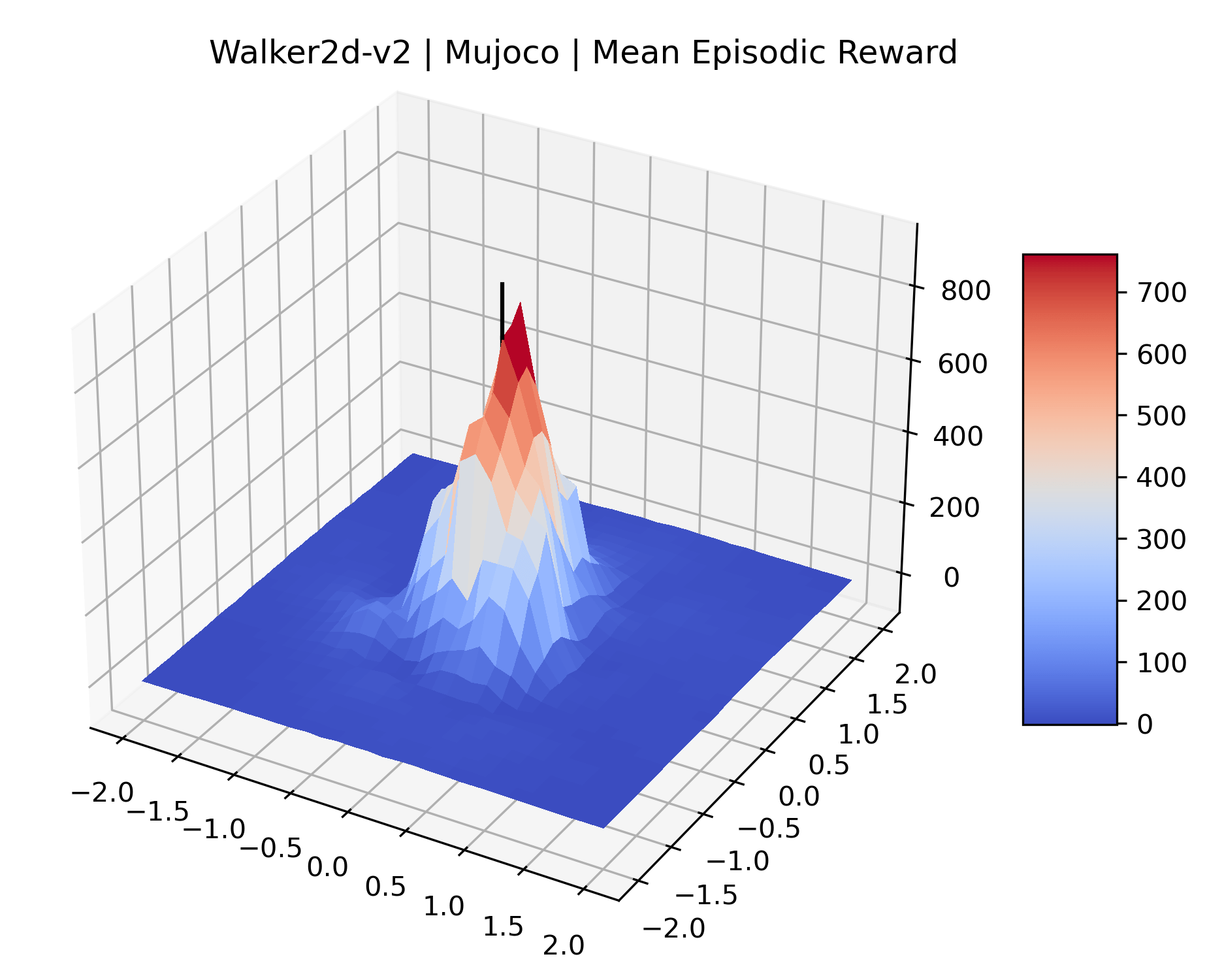} \\
 \end{tabular}
\caption{Walker2d environment with different number of layers in policy network. Top: 2, 4, and 6 layer networks. Bottom: 8, 12, and 16 layer networks}
\label{fig:walker_architecture_rewardsurface_table}
\end{figure*}
\pagebreak

\section{Cliff Checkpoint Selection}
\label{appendix:cliff-selection}
For cliffs, we select checkpoints that exhibit a decrease in returns of at least 50\% in the first, high-resolution section of the line search (that is, within a normalized gradient magnitude of 0.04). By selecting points closer to the initial, unperturbed weights, we choose cliffs that the optimization method is likely to reach. Although we do our best to evaluate the gradient line searches with enough environment steps to reduce their standard error, it is not computationally feasible to completely account for the high variance of returns in sparse reward environments. To mitigate this, the decrease in returns that identify a cliff must also be at least 25\% of the global reward range across all checkpoints in the line plot. This ensures that we select cliffs that are significant in the environment's overall reward scale, and not likely to be caused by evaluation noise. We select non-cliff checkpoints arbitrarily from the remaining checkpoints that do not meet these criteria.

\section{Limitations}
\label{appendix:limitations}
The scale of axes in our plots is a ubiquitous concern in this paper. Fortunately for the 3d reward surface plots, filter-normalized random directions have been well studied, and we find that many of the original findings from the loss landscapes paper hold. The sharpness of the minimizer or maximizer in loss landscapes and reward surfaces respectively correlate with the difficulty of the task. 

The correct scale for plots that use the policy gradient direction are less obvious. In general, a continuous, sharp change in rewards can be made to look gentle by zooming into the slope, and any continuous change in rewards can appear to occur instantly if you zoom out enough.  Gradient directions are calculated from network parameters, and therefore have their own implicit normalization, so we cannot apply filter-normalization. We choose to normalize these directions to have the same magnitude, and assume that the gradient's implicit normalization will highlight important features. While it is possible that our gradient line searches miss some cliffs, or make some checkpoints incorrectly appear as cliffs, the results of our experiments with A2C appear to confirm that more often than not, the cliffs in our line searches are real.

Our comparison of PPO and A2C on cliff checkpoints is a preliminary result and does not identify which component of PPO makes it more robust to cliffs than A2C. Previous work has highlighted that much of the improved performance of PPO can be attributed to implementation details rather than algorithmic improvements like ratio clipping \citep{Engstrom2020Implementation}. We hope that future work may narrow down the specific component in PPO that allows it to perform well on cliffs. 

We also do not use only optimal hyperparameters in our experiments comparing A2C and PPO on cliff vs. non-cliff checkpoints. This allows us to verify the existence of cliffs, but does not provide clear indication as to how impactful they are on agents with well-tuned hyperparameters. Naturally, optimized hyperparameters should not allow for the sharp drops in reward that we expect to see on the cliff checkpoints. We hope that future work will explore how often cliffs affect training in practice.


\end{document}